\documentclass[journal]{IEEEtran}
\usepackage{cite}
\usepackage[colorlinks, citecolor=green]{hyperref}
\usepackage{booktabs}
\usepackage{subfigure} 
\usepackage{graphicx}
\usepackage{float} 
\usepackage{multirow}

\usepackage{makecell}
\usepackage{textcomp}
\usepackage{array} 
\usepackage{longtable}
\usepackage{booktabs}
\usepackage{mathptmx}
\usepackage{fixltx2e}
\usepackage{caption}

\usepackage[export]{adjustbox}
\begin{document}
		%
		\title{RSAM-Seg: A SAM-based Approach with Prior Knowledge Integration for Remote Sensing Image Semantic Segmentation}
		%
		%
		%
		
		\author{Jie Zhang, Xubing Yang, Rui Jiang, Wei Shao and Li Zhang
		
		\thanks{Jie Zhang, Xubing Yang, Li Zhang are with the College of Information Science and Technology, Nanjing Forestry University, Nanjing 210037, China (e-mail:j.zhang.fn@gmail.com, xbyang, lizhang@njfu.edu.cn)
		}
		\thanks{Rui Jiang is with the College of Telecommunications and InformationEngineering, Nanjing University of Posts and Telecommunications, Nanjing 210003, China (e-mail: j ray@njupt.edu.cn).
		}
		\thanks{Wei Shao is with Shenzhen Research Institute, Nanjing University of Aeronautics and Astronautics, Guangdong 518038, China. (shaowei20022005@nuaa.edu.cn)}
		\thanks{This work is supported by National Natural Science Foundation of
			China (NSFC) under grant (No. 61802193) and The Shenzhen Science and Technology Program under Grant JCYJ20220530172403008, (Corresponding author: Wei Shao, Li
			Zhang.)}
		}

	\maketitle
	
	\begin{abstract}
	The development of high-resolution remote sensing satellites has provided great convenience for research work related to remote sensing. Segmentation and extraction of specific targets are essential tasks when facing the vast and complex remote sensing images. Recently, the introduction of Segment Anything Model (SAM) provides a universal pre-training model for image segmentation tasks. While the direct application of SAM to remote sensing image segmentation tasks does not yield satisfactory results, we propose RSAM-Seg, which stands for Remote Sensing SAM with Semantic Segmentation, as a tailored modification of SAM for the remote sensing field and eliminates the need for manual intervention to provide prompts. Adapter-Scale, a set of supplementary scaling modules, are proposed in the multi-head attention blocks of the encoder part of SAM. Furthermore,  Adapter-Feature are inserted between the Vision Transformer (ViT) blocks. These modules aim to incorporate high-frequency image information and image embedding features to generate image-informed prompts. Experiments are conducted on four distinct remote sensing scenarios, encompassing cloud detection, field monitoring, building detection and road mapping tasks  . The experimental results not only showcase the improvement over the original SAM and U-Net across cloud, buildings, fields and roads scenarios, but also highlight the capacity of RSAM-Seg to discern absent areas within the ground truth of certain datasets, affirming its potential as an auxiliary annotation method. In addition, the performance in few-shot scenarios is commendable, underscores its potential in dealing with limited datasets. Our code is available at: \url{https://chief-byte.github.io/RSAM-Seg-Site}.
	\end{abstract}
	

	%
	\IEEEpeerreviewmaketitle

	\section{Introduction}
	%
	%
	%
	\IEEEPARstart{W}{ith} the development of remote sensing satellite technology, high-resolution remote sensing images have been widely used in various fields, such as cloud detection, urban infrastructure assessment, agricultural land planning, and road condition analysis\cite{1,2,3,4,5}. Cloud detection plays a pivotal role as the initial step in the data processing pipeline for earth observation and remote sensing technologies\cite{6}. Urban infrastructure assessment leverages remote sensing capabilities to evaluate a diverse range of structures, including buildings, roads, and bridges, supporting maintenance and planning efforts\cite{7}. Furthermore, in the realm of agricultural land planning, remote sensing assumes a crucial function by monitoring crop health, scrutinizing land use patterns, and optimizing irrigation management to enhance farming practices\cite{8}. However, the satellite images often suffer from object occlusion, blurring, incomplete coverage and other issues, which pose challenges for identifying objects\cite{10}. 
	
	Thus, a multitude of methods have been proposed to address above issues, which can be generally divided into three categories: threshold-based algorithms, classical machine learning algorithms and deep learning algorithms\cite{11,12,13,14,15}. Threshold-based algorithms utilize the spectral characteristics of remote sensing data to complete semantic segmentation tasks based on prior knowledge and judgment conditions provided by experts\cite{16,17,18,19}, Li \textit{et al.} proposed an object-based approach to create a land-cover classification map. However, threshold-based methods rely on a large number of pre-designed rules and require remote sensing professionals to design and evaluate the rules, which leads to problems such as high costs, long processing times, and poor results\cite{20}. In terms of classical machine learning algorithms, Support Vector Machines (SVM) and Random Forest (RF) have gained much attention\cite{65,66,86}. Melgani \textit{et al.} assessed the potential of SVM classifiers in high-dimensional feature spaces of hyperspectral remote sensing images\cite{66}. The experimental results suggested that SVM classifiers are a viable option for classifying hyperspectral remote sensing data. However, SVM is subject to certain limitations. The choice of kernel function poses a challenge, where a smaller kernel width parameter may lead to overfitting and a larger value may cause excessive smoothing\cite{67}. RF has been successfully used to map urban buildings and land cover categories\cite{87,88}. However, RF model trained on one region is not applicable or transferable to new regions\cite{89} and depends on the number of variables used for splitting the tree nodes\cite{90}.
	
	With the rapid development of deep learning, this technology has shown great potential in addressing segmentation challenges in remote sensing\cite{21}. However, despite the accomplishments of deep learning methods in remote sensing segmentation, there are still several challenges that demand attention. One prominent challenge lies in the significant within-class variance and limited between-class variance observed in pixel values of objects of interest\cite{22}. In addition, the quality and availability of labeled data play a crucial role in this regard, particularly when dealing with small datasets or rare classes\cite{23}. Insufficient labeled data also hampers the model's ability to generalize well and accurately identify and segment objects of interest. Consequently, there is a notable lack of universality and ease of transfer across different remote sensing scenarios. Overcoming this challenge requires innovative strategies such as domain adaptation, transfer learning, or incorporating prior information that can help the model better leverage the available data and extract meaningful features that are specific to the remote sensing domain to enhance the model's ability to generalize and perform effectively in diverse remote sensing applications\cite{24,25,26,27,28}.
	
	Recently, the general-purpose vision segmentation models has brought new and more effective solutions to the field of image segmentation\cite{29,30}. These models are pre-trained on large amounts of data and can be generalized to new tasks and data distributions through the use of prompt engineering, demonstrating outstanding capabilities in few-shot and zero-shot learning\cite{28,29,30}. SAM is a new general-purpose vision segmentation model based on Natural Language Processing (NLP) developed by Meta\cite{31,32}. It focuses on promptable segmentation tasks and uses prompt engineering to adjust to various downstream segmentation tasks. SAM can automatically identify objects present in an image and immediately provide segmentation masks for any prompt by simply marking points to include or exclude objects, or by drawing bounding boxes to create segmentation\cite{31}, which is considered to be a game-changer in the field of image segmentation. Additionally, it achieves fully automated segmentation of potential objects within the images and aims to achieve effective segmentation of any object in any image, without the need for additional task-specific or dataset-specific adaptation (such as training). The segmentation accuracy of SAM on a wide range of diverse benchmark datasets show that SAM has a high generalization ability. Carefully tuned prompts could even surpass popular supervised-training models designed specifically for object segmentation tasks.
	
	Although SAM has shown promising results on open datasets, its effectiveness is often limited when applied to specific downstream tasks, particularly remote sensing segmentation tasks. This is due to the complex characteristics of the interested objects in remote sensing images, such as blurriness, occlusion, and irregular shapes, which pose challenges for segmentation algorithms. In addition, the prompt requires manual input. As a result, there is a need to develop an domain specific SAM that can better handle these challenges and improve the overall performance of remote sensing segmentation tasks without manual prompt input.
	
	To address these issues, we propose RSAM-Seg. Feature information is extracted from specific domains and inserted into the ViT blocks in the encoder to improve the performance in remote sensing field. By incorporating prior knowledge specific to remote sensing image data, such as emedding features and spectral features, the model adjusts better to the segmentation tasks of remote sensing images. To validate the effectiveness of the proposed methodology, experiments were conducted on cloud, buildings, fields and roads scenarios.
	
	The main contributions of the work were summarized as follows:
	
	1). Based on our extensive research and analysis, we have pioneered the application of SAM to object segmentation tasks in remote sensing images, and RSAM-Seg demonstrates better adaptability to remote sensing images.
	
	2). RSAM-Seg eliminates the need for manual intervention to provide prompts, thereby streamlining the workflow of SAM.
	
	3). RSAM-Seg can incorporate custom domain-specific prior information, making it adaptable to diverse tasks in the remote sensing field.
	
	4). RSAM-Seg outperforms the original SAM and U-Net across multiple scenarios such as cloud, buildings, fields and roads in the experiments. Moreover, it discerns missing areas in dataset ground truths and demonstrates few-shot capability.
	
	The rest of this article is organized as follows. Section II provides an overview of the related work in the field of remote sensing segmentation tasks. Section III presents the proposed method for the tasks. In Section IV, the datasets, experimental settings and performance metrics are described in detail. The experimental results and analysis are presented in Section V. Section VI offers a comprehensive discussion of the findings. Finally, Section VII concludes the article by summarizing the main contributions and highlighting future research directions.
	
	\section{Related Work}
	In recent years, deep learning methods have been widely applied to segmentation tasks in remote sensing\cite{44,45}. Since deep learning networks are typically trained using large datasets to learn specific features and patterns within the input data, which are then used to classify new data, they can be categorized into three types based on the availability of labeled training data: supervised learning, weakly-supervised learning, and unsupervised learning\cite{46}. 
	
	Over the past ten years of its development, supervised learning has witnessed the emergence of Convolutional Neural Network (CNN). CNN extracts local features from images through convolutional operations, and then reduces the dimensionality of the feature map through pooling operations\cite{46}. DeepLab, a CNN-based model, utilizes techniques such as dilated convolution and multi-scale pyramid pooling to improve segmentation accuracy. It performs well in the field of remote sensing image processing and has been widely used in high-resolution remote sensing image segmentation tasks\cite{48}. DeepLab V3 and DeepLab V3+ are the successors of DeepLab\cite{48,49}. DeepLab V3 applies global average pooling on the last feature map of the model. DeepLab V3+ brings about a decoder module to the DeepLab V3 to refine the boundary details. Liu \textit{et al.} proposed FieldSeg-DA based on DeepLab V3+ to automatically extract individual arable fields (IAF) from Chinese Gaofen-2 images\cite{49}. U-Net, a neural network based on CNN but utilizing the Encoder-Decoder architecture, has shown excellent performance in the field of image segmentation tasks\cite{50}. Improved networks based on the U-Net structure have gained considerable attention for their potential in remote sensing image segmentation. Sun \textit{et al.} proposed L-UNet, which replaces the partial convolution layers of U-Net with Conv-LSTM and Atrous in order to improve both the quantity and quality of the network compared to the original U-Net\cite{51}. Hou \textit{et al.} prposed C-UNet on basis of the standard U-Net, where four more modules are added for road extraction tasks and show improved performance compared to standard U-Net\cite{52}. 
	
	In the field of weakly-supervised learning, semantic segmentation with weak supervision offers a potential solution to address the challenges associated with labeling complexity in land cover classification. Weakly-supervised Semantic Segmentation (WSS) methods often rely on the utilization of Class Activation Maps (CAMs), which is a CNN trained for image-level classification to perform rough localization of object areas based on global average pooling or gradient backpropagation, have been widely used for natural images\cite{67,68,69,70}. Fu \textit{et al.} proposed WSF-Net, calculates CAMs using fused features of objects in remote sensing image especially in the water and cloud scenarios\cite{71}. Wang \textit{et al.} proposed U-CAM, which adapts CAMs for U-Net to perform cropland segmentation\cite{76}. Nyborg \textit{et al.} proposed the utilization of fix-point Generative Adversarial Network (GAN) for weakly-supervised cloud detection, referred as FCD\cite{72}. Chen \textit{et al.} utilized a WSS framework based on point labels with transfer method to accurately classify land cover with minimal human intervention\cite{74}. Wang \textit{et al.} proposed a novel RS-WSOD framework, which addresses the challenges of background noises and missing detections in remote sensing images\cite{75}. Xu \textit{et al.} proposed the Consistency-Regularized Region-Growing Network (CRGNet) for semantic segmentation of urban scenes, leveraging point-level labels\cite{77}. 
	
	In terms of unsupervised learning, unsupervised learning addresses the reliance of annotated data and domain shifts in high-resolution remote sensing imagery. Zhu \textit{et al.} proposed Memory Adapt Net (MAN), which established an adversarial learning scheme in output space to bridge the domain distribution discrepancy between the source and the target domains to perform cross-domain segmentation tasks of the high-resolution remote sensing imagery\cite{78}. Chen \textit{et al.} proposed a category-certainty attention mechanism to effectively handle unadapted regions for semantic segmentation of high-resolution satellite imagery\cite{80}. Li \textit{et al.} employed an objective function that integrated multiple weakly supervised constraints to minimize the distributional shift of data between the source and target domains to address challenges related to sensor and landscape variations in diverse geographic locations\cite{81}. Zhang \textit{et al.} proposed a stagewise domain adaptation model called RoadDA that aimed to align the features of the source and target domains through interdomain adaptation using GAN to achieve promising road segmentation on unlabeled target images.\cite{82}. Chen \textit{et al.} proposed a unsupervised domain adaptation method and a contrastive-learning based and Memory-Contracted (MCD) module for building extraction in high-resolution remote sensing imagery\cite{83}. Cai \textit{et al.} proposed the segmentation model from two opposite directions where source domain images are transformed into images featuring the style of the target domain then adapt the classifier to the target domain to improve the performance of the cross-domain semantic segmentation in urban city areas\cite{84}. 
	
	Recently, few-shot learning, as a nascent method under weakly supervised learning, has gained attention in the field of remote sensing to address the issue of limited datasets\cite{58}. Zhang \textit{et al.} first introduced the concept of few-shot learning\cite{60}. Liu Y \textit{et al.} proposed NTRENet to distinguish ambiguous regions, which is benefit for satellite images\cite{61}. Prior-knowledge based method utilizes pretraining on various other datasets to continuously accumulate learning ability and experience. Domain-specific knowledge is incorporated into the network backbone through various methods on the tasks of few-shot semantic segmentation in aerial images\cite{56,57,58}. Cheng \textit{et al.} proposed SPNet to tackle interclass similarity issues in remote sensing scenes during few-shot segmentation by considering the validity of prototypes\cite{62}. Li \textit{et al.} proposed SCL-MLNet to boost few-shot classification in remote sensing scenarios through the fusion of multi-scale spatial features and integration of self-supervised contrastive learning methods\cite{63}. Liu \textit{et al.} enforce the tunable parameters focusing on the explicit individual image and achieved high performances on domin-specific tasks\cite{57}. By leveraging the acquired general knowledge, the model can achieve fast learning with only a small amount of labeled data.
	

	\section{Method}
	\begin{figure*} 
	\centering 
	\includegraphics[width=1.0\textwidth]{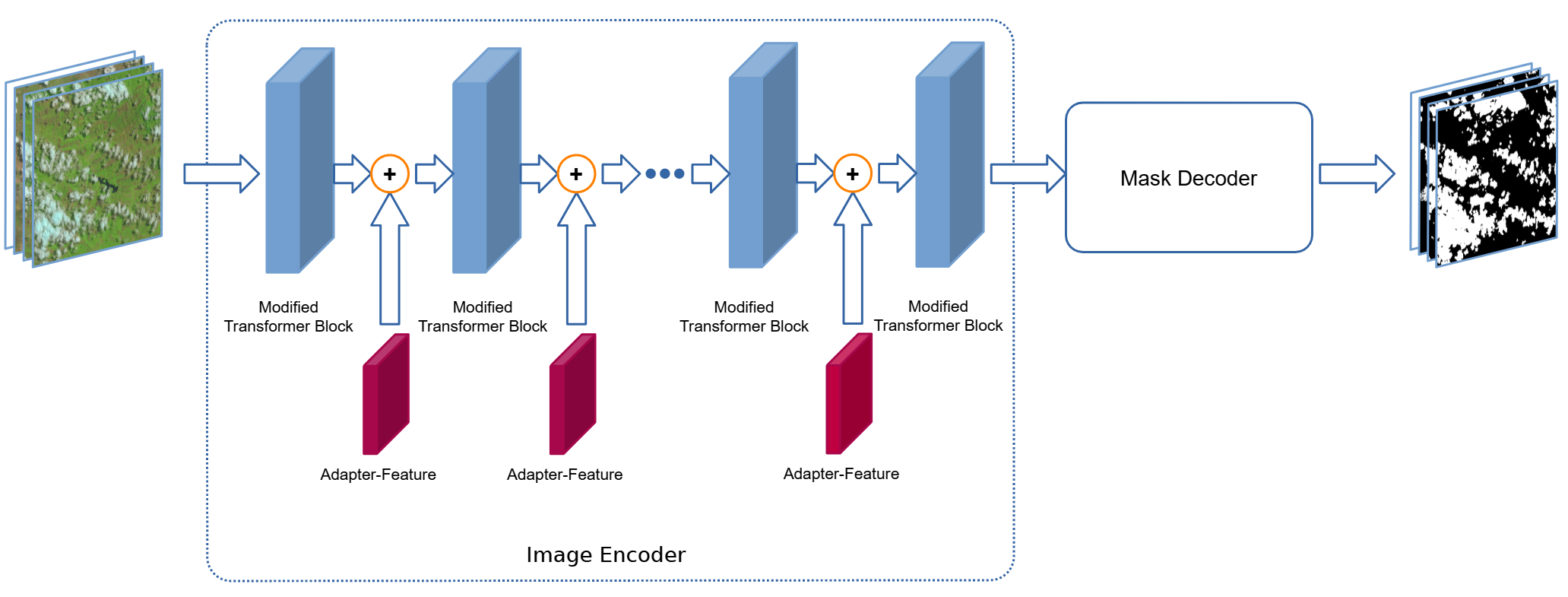} 
	\caption{The structure of RSAM-Seg. Adapter-Feature are inserted between modified ViT blocks while maintaining the mask decoder identical to the original SAM.}
	\label{fig.RSAMS}
	\end{figure*}
	\subsection{RSAM-Seg architecture}
	RSAM-Seg uses SAM as the backbone while retaining most of the structure of the decoder part. RSAM-Seg extracts features from remote sensing images without the necessity of human-provided prompts. To obtain more task-related information, the original encoder and decoder part of the model are modified. This adaptation enables better performance on remote sensing related tasks. To be specific, the ViT blocks of the encoder are modified by incorporating Adapter-Scale inside, and embedding Adapter-Feature between ViT layers to extract image information. We assume $P^{i}$ refers to the prompts that generated from the extracted features of the image.
	\begin{figure}[!htb] 
		\centering
		\includegraphics[width=0.35\textwidth]{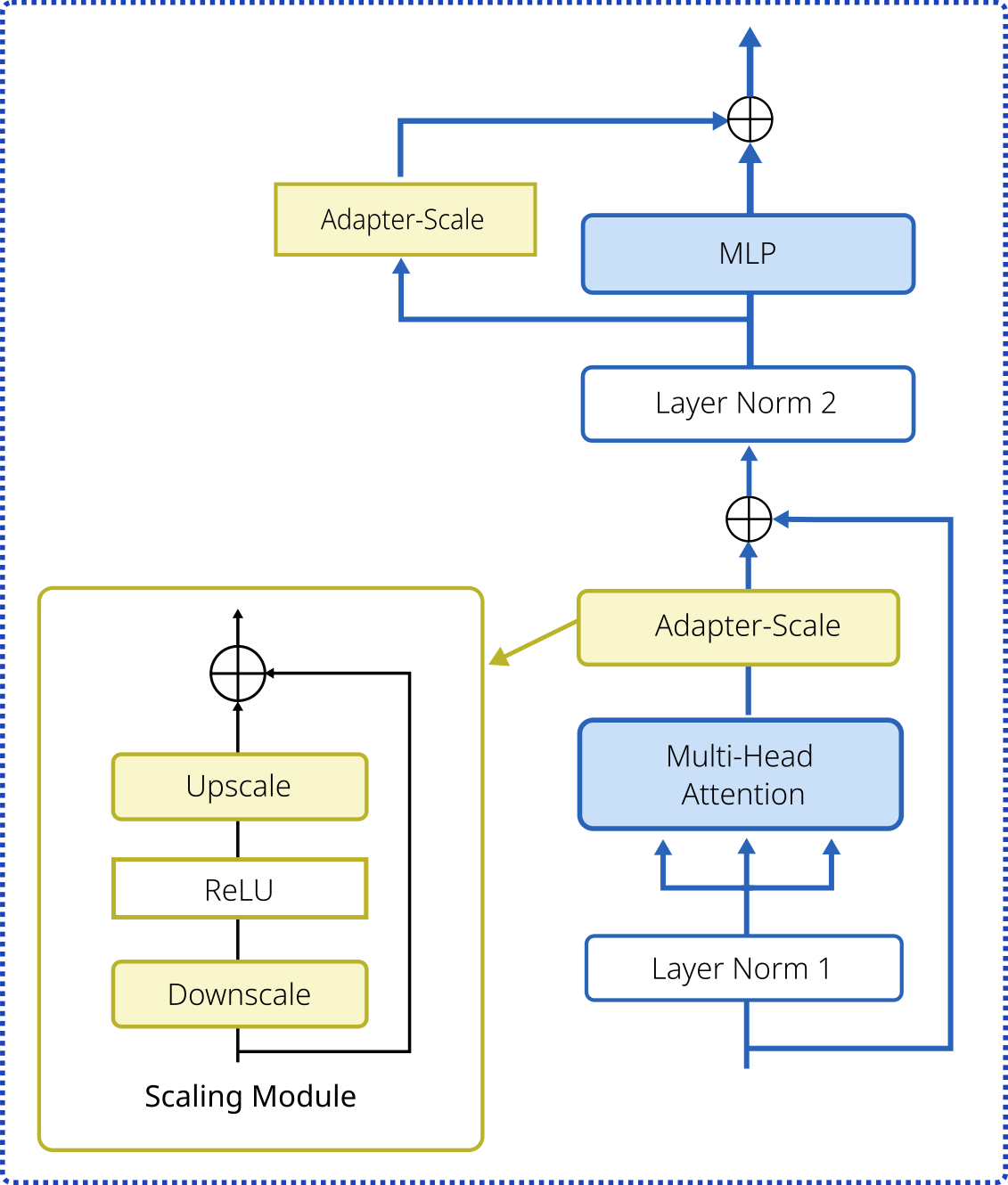} 
		\caption{The structure of the modified transformer block and Adapter-Scale in the encoder of RSAM-Seg.}
		\label{fig.RSAM1}
	\end{figure}
		\begin{equation}
		P^{i}=\mathrm{MLP}_{\mathit{up}}\left(\mathrm{GELU}\left(\mathrm{MLP}_\mathit{{tune }}^{i}\left(\mathrm{F}_{\mathrm{pe}}+\mathrm{F}_{\mathrm{hfc}}\right)\right)\right)
	\end{equation}
	Where $i$ denotes each individual adapter between ViT layers. $\mathrm{F}_{\mathrm{pe}}$ and $\mathrm{F}_{\mathrm{hfc}}$ stand for embedding features and High-Frequency Components (HFC) features. The mask decoder remains unchanged with no given prompt inputs and is fine-tuned using a pre-trained model. The architecture is shown in Figure \ref{fig.RSAMS}.
	\subsection{Adapter details}
	\subsubsection{Adapter-Scale}
	In the encoder, Adapter-Scale consists of three parts: Downscale, ReLu, and Upscale. The Downscale part uses a single Multi-Layer Perceptron (MLP) layer to reduce the dimensionality of the embedding. After applying the ReLu activation function, the embedding is restored to its original dimensionality using another MLP layer in the Upscale part. Two Adapter-Scale modules are inserted into each ViT block. The first is before the multi-head attention blocks and residual connections. The second is within the residual structure of the MLP. Additionally, a scale factor of 0.5 is applied to each adapter. The structure of ViT blocks is shown in Figure \ref{fig.RSAM1}. 
	\subsubsection{Adapter-Feature}
\begin{figure}[!htb] 
		\centering
		\includegraphics[width=0.35\textwidth]{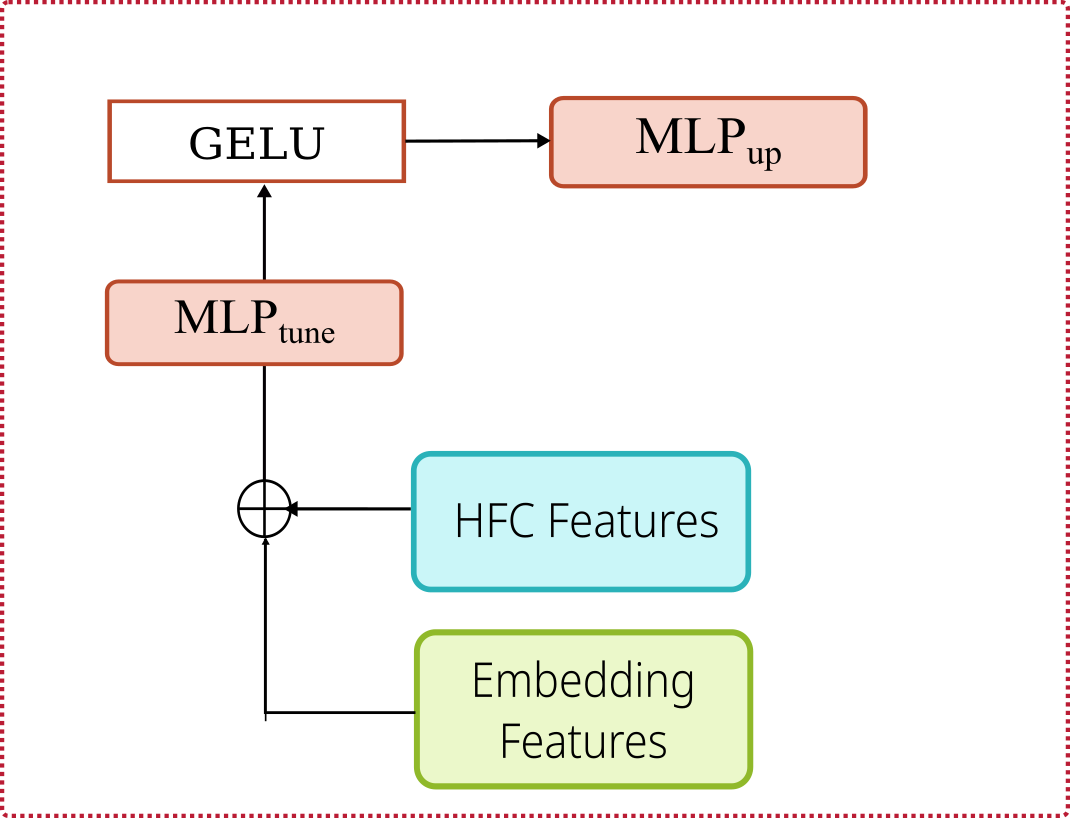} 
		\caption{The structure of the Adapter-Feature between the ViT blocks in the encoder of RSAM-Seg.}
		\label{fig.RSAM2}
	\end{figure}

	Between the ViT layers, Adapter-Feature consists of two MLPs. The first is the $\mathrm{MLP}_{\mathit{tune}}$, which extracts features from remote sensing images to serve as prompts. The second MLP, $\mathrm{MLP}_{\mathit{up}}$, is used to adjust the feature dimension to input into the ViT layer. The Adapter-Feature structure is shown in Figure \ref{fig.RSAM2}.
	
	In our work, both embedding features and high-frequency components features are tuned. In the part of embedding features, a linear layer with a scale factor is used to change the original embedding dimension. The structure of Adapter-Feature is shown in Figure \ref{fig.RSAM2}. 
	
	In the part of HFC features, the HFC of the images are extracted and then inputted as prompts into the encoder. For an image $I$ with dimensions of $H \times W$, high-frequency and low-frequency information can be extracted through Fast Fourier Transform (FFT) and inverse transforms. The high-frequency information of the image is of particular interest to us. $\mathtt{fft}$ and $\mathtt{ifft}$ are used to represent the Fast Fourier Transform and its inverse transform, respectively. The frequency components extracted from image $I$ can be expressed by $\mathit{f}=\mathtt{ftt}(I)$. Image $I$ can also be restored through $\mathtt{ifft}$ by $\mathit{I}=\mathtt{iftt}(\mathit{f})$. To avoid the loss of information at the edges, a mask is used to selectively filter the high-frequency components,which can be done by shifting the low-frequency coefficients to the center of the image($\frac{H}{2}$,$\frac{W}{2}$). The mask is generated with a mask ratio $\tau$.

	\begin{equation}
	\mathbf{M}_{h}^{i, j}(\tau) = \left\{\begin{array}{ll}
		0, & \frac{4\left|\left(i-\frac{H}{2}\right)\left(j-\frac{W}{2}\right)\right|}{H W} \leq \tau \\
		1, & \textbf{otherwise}
	\end{array}\right.
	\end{equation}
	where the symbol $\tau$ represents the proportion of the masked region. The HFC feature 	can obtain by$:$
	\begin{equation}
		I_{h f c}=\mathtt{ifft}\left(f \mathbf{M}_{h}(\tau)\right)
	\end{equation}
	
	\begin{figure*}[htbp]
		\centering
		\subfigure{
			\rotatebox{90}{\scriptsize{Clouds}}
		\begin{minipage}[b]{0.07\textwidth}
			\centering
			\includegraphics[width=\textwidth]{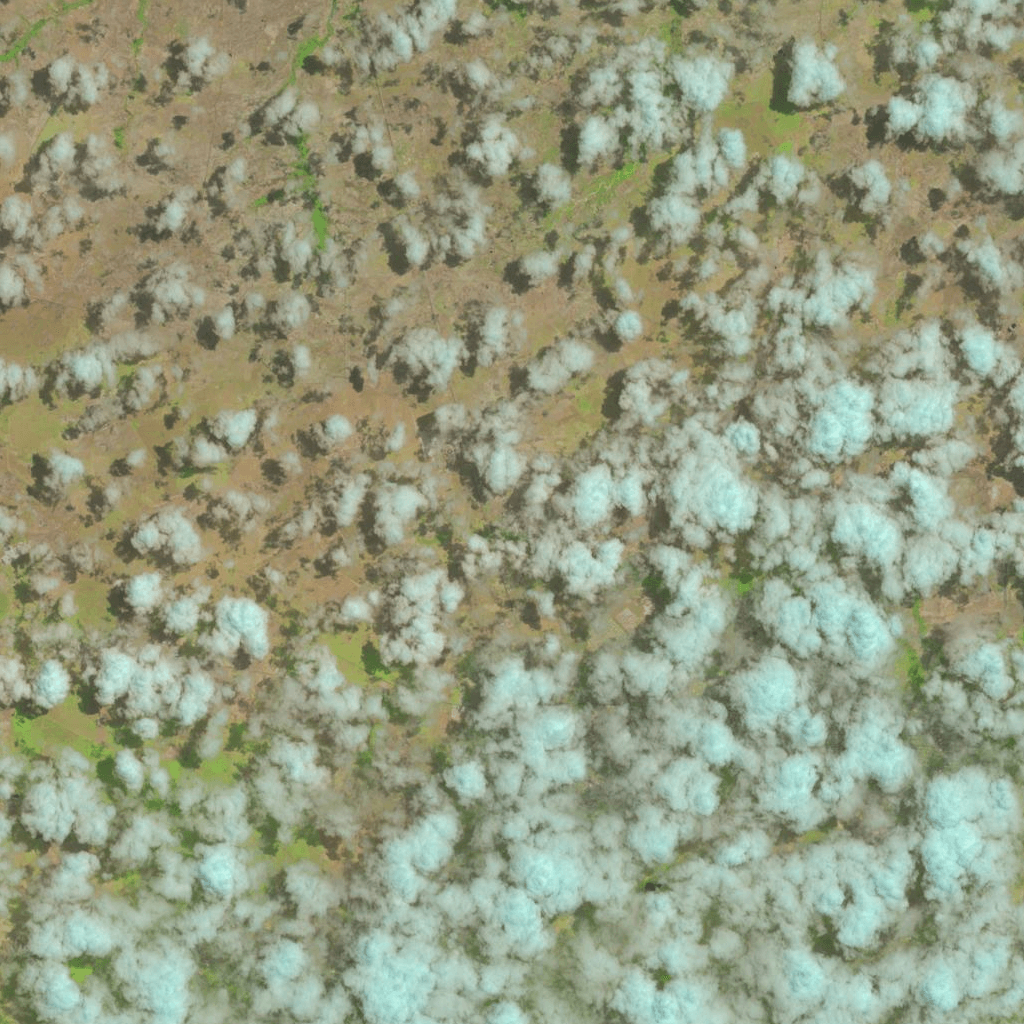}
		\end{minipage}}
		\begin{minipage}[b]{0.07\textwidth}
			\centering
			\includegraphics[width=\textwidth]{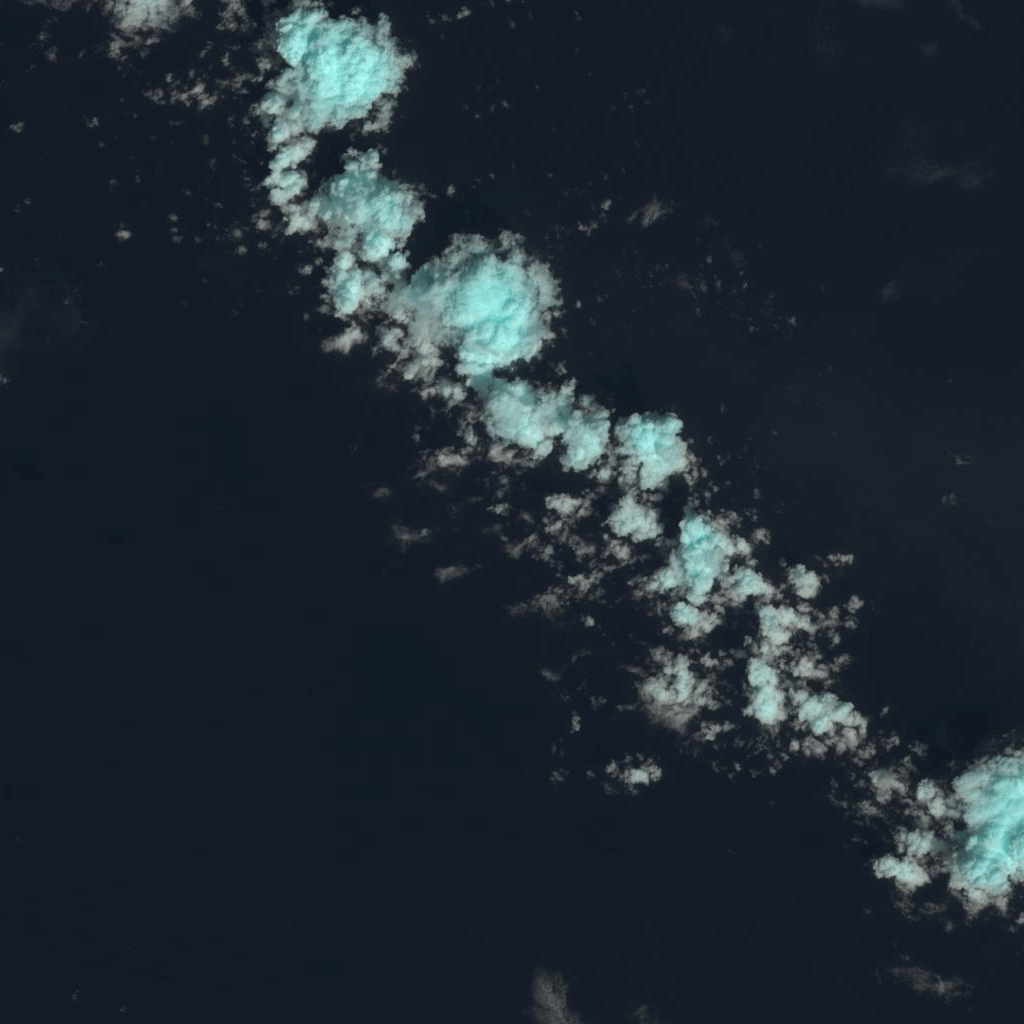}
		\end{minipage}
		\begin{minipage}[b]{0.07\textwidth}
			\centering
			\includegraphics[width=\textwidth]{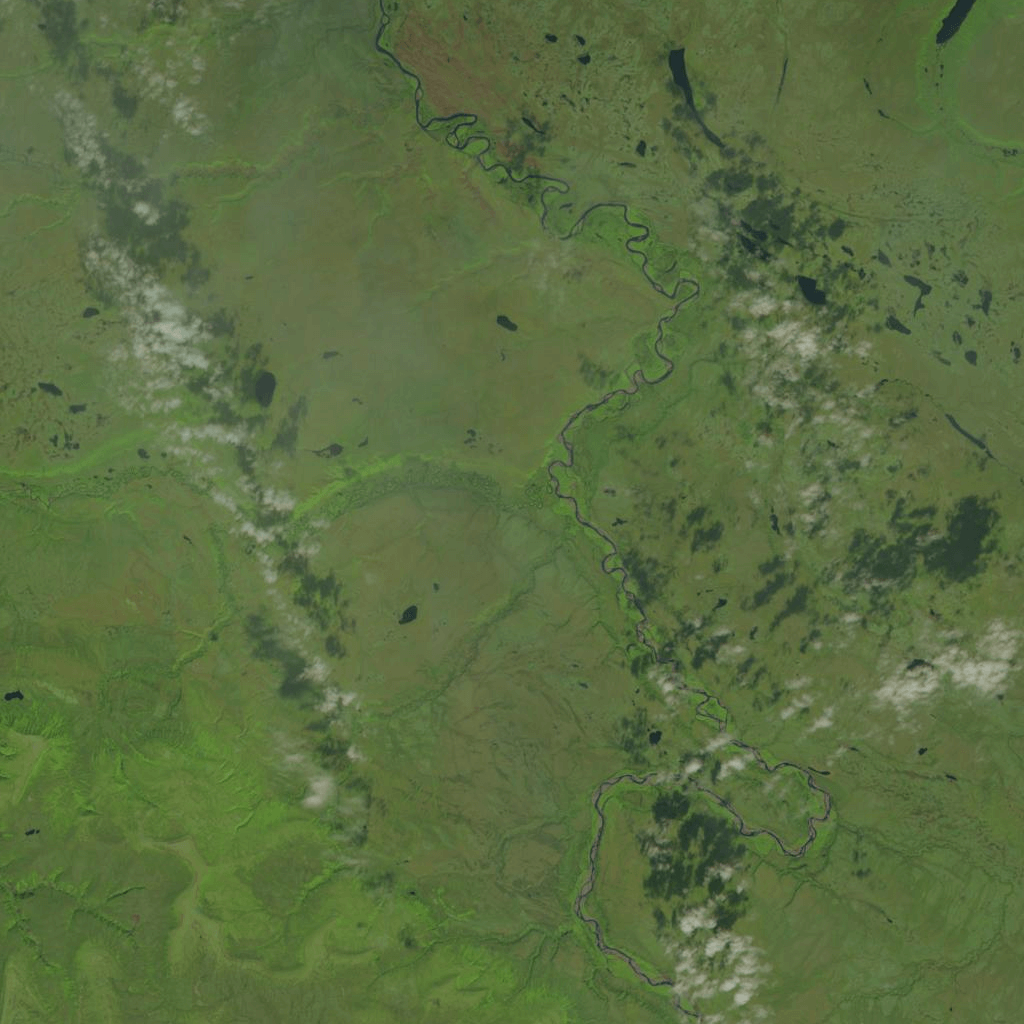}
		\end{minipage}
		\begin{minipage}[b]{0.07\textwidth}
			\centering
			\includegraphics[width=\textwidth]{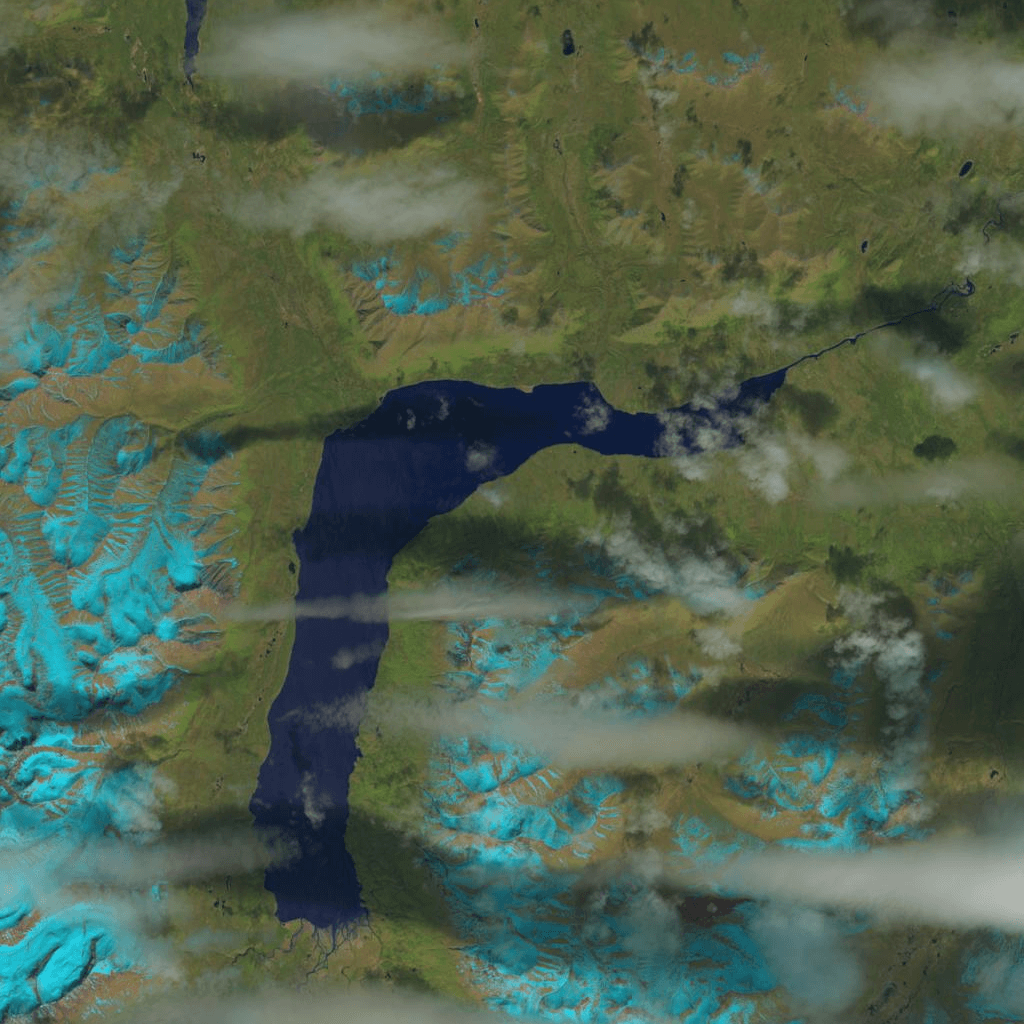}
		\end{minipage}
		\begin{minipage}[b]{0.07\textwidth}
			\centering
			\includegraphics[width=\textwidth]{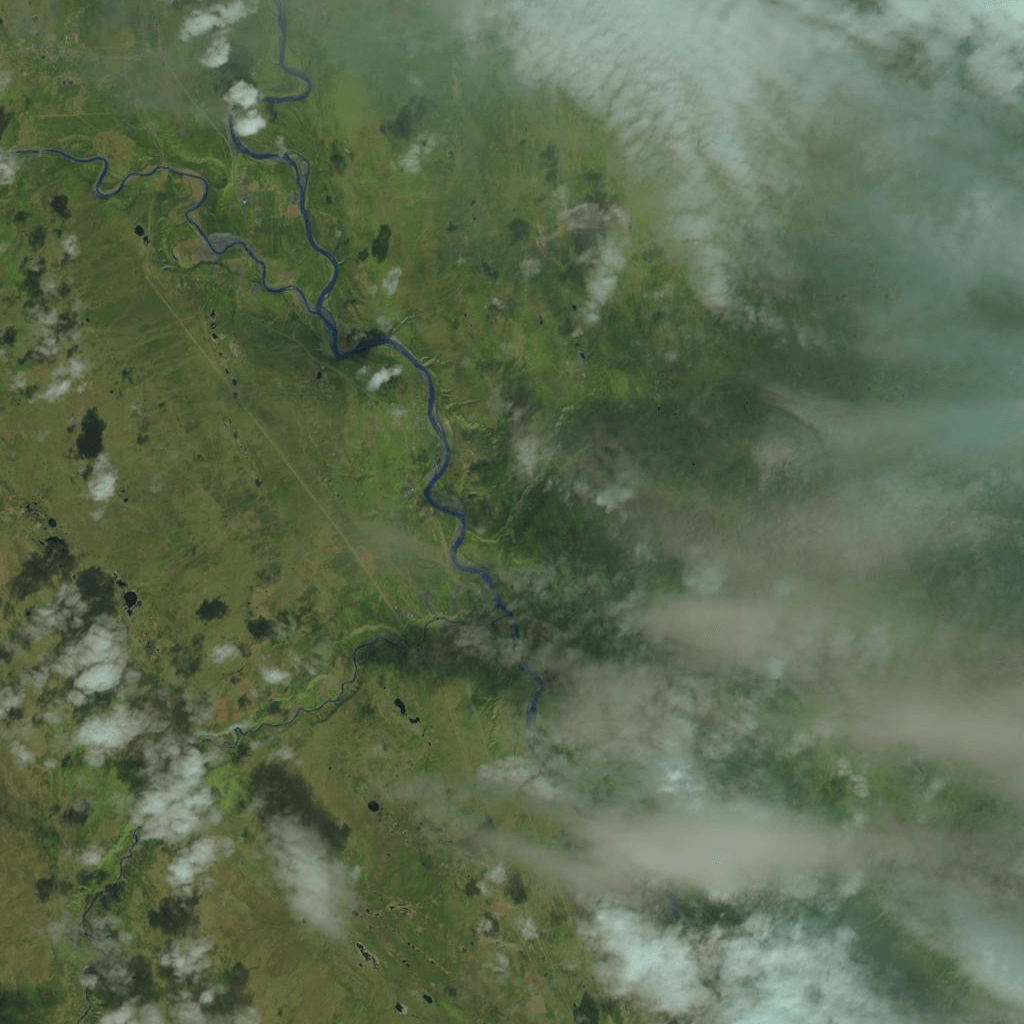}
		\end{minipage}
		\begin{minipage}[b]{0.07\textwidth}
			\centering
			\includegraphics[width=\textwidth]{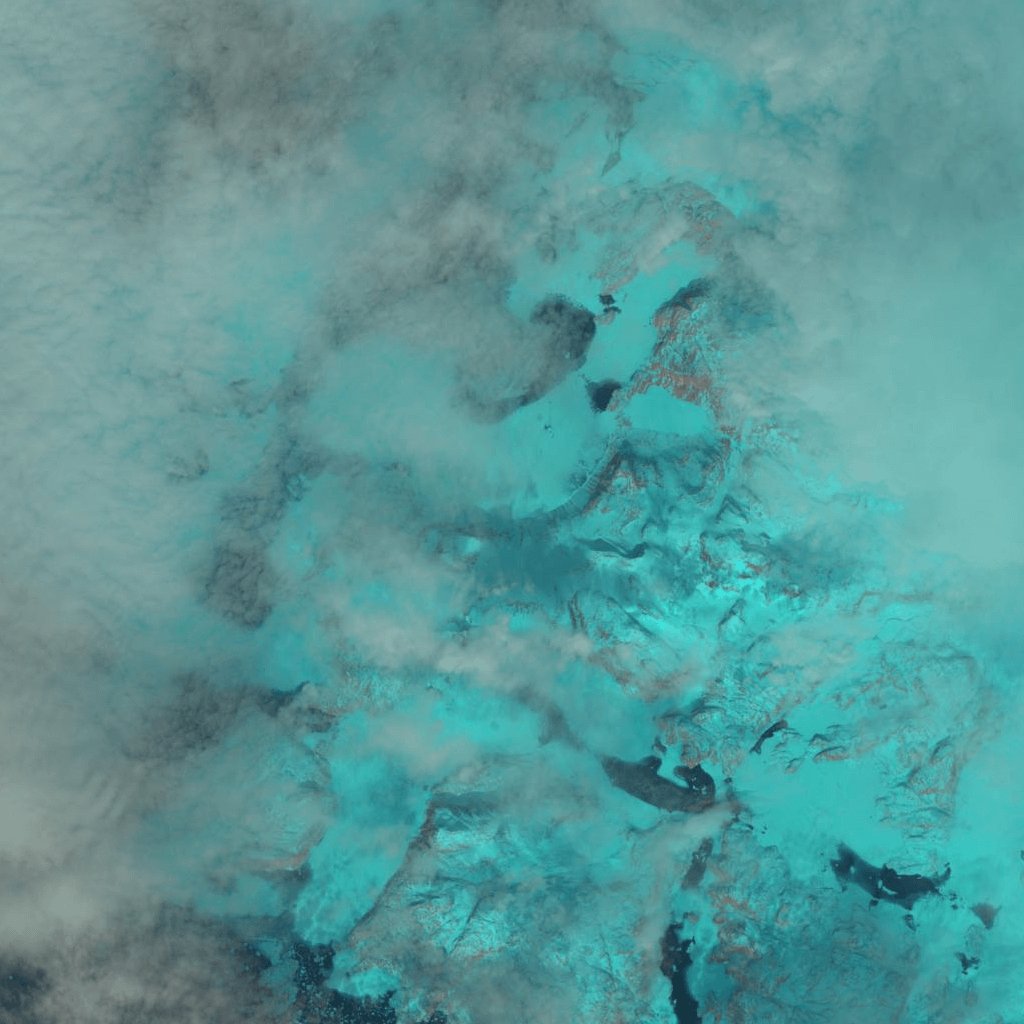}
		\end{minipage}
		\subfigure{
			\rotatebox{90}{\scriptsize{Buildings}}
		\begin{minipage}[b]{0.07\textwidth}
			\centering
			\includegraphics[width=\textwidth]{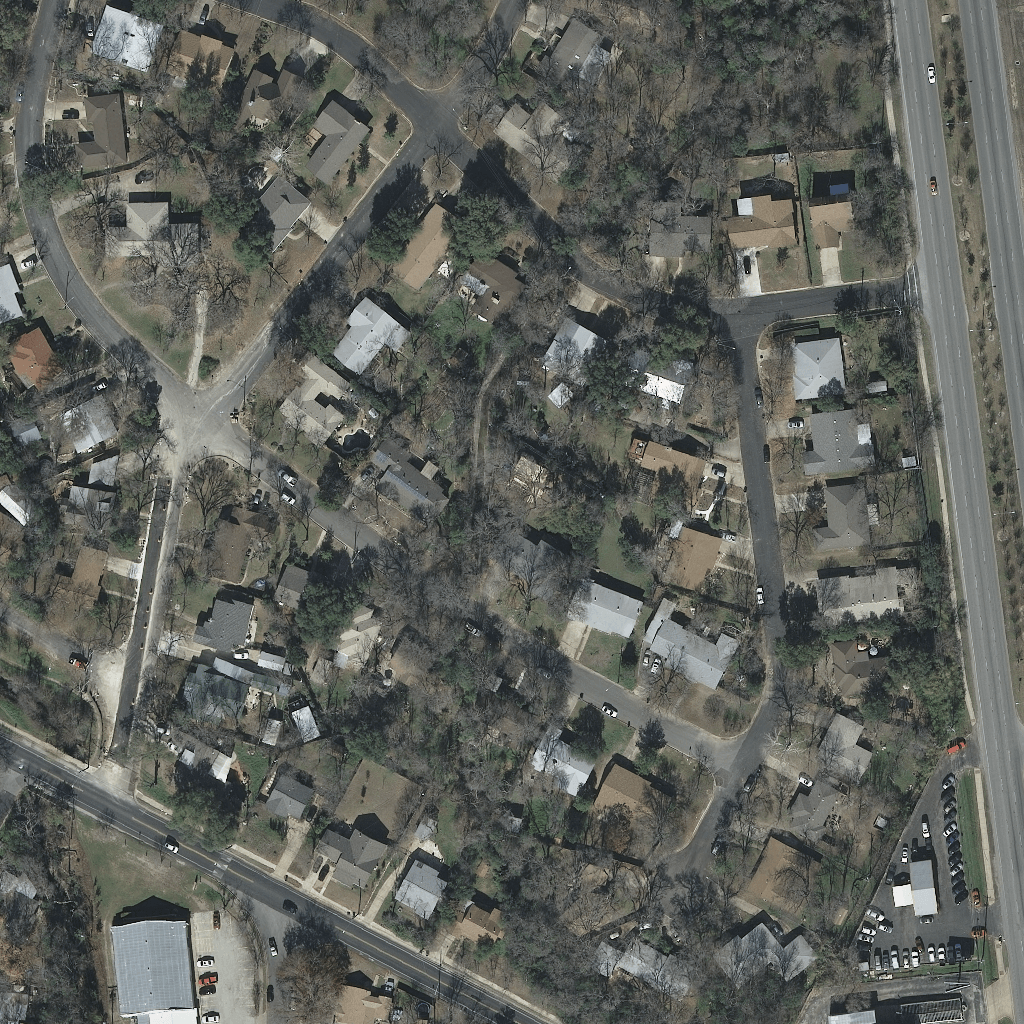}
		\end{minipage}}
		\begin{minipage}[b]{0.07\textwidth}
			\centering
			\includegraphics[width=\textwidth]{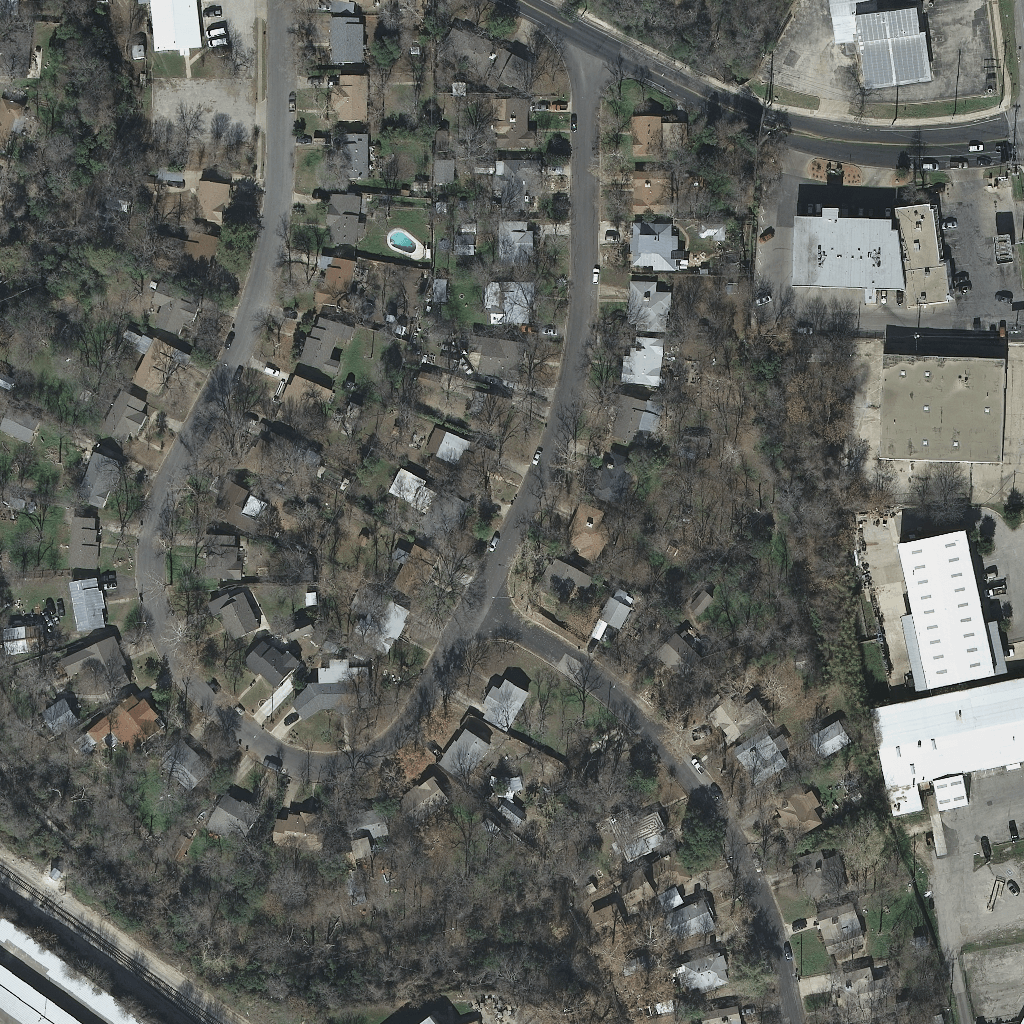}
		\end{minipage}
		\begin{minipage}[b]{0.07\textwidth}
			\centering
			\includegraphics[width=\textwidth]{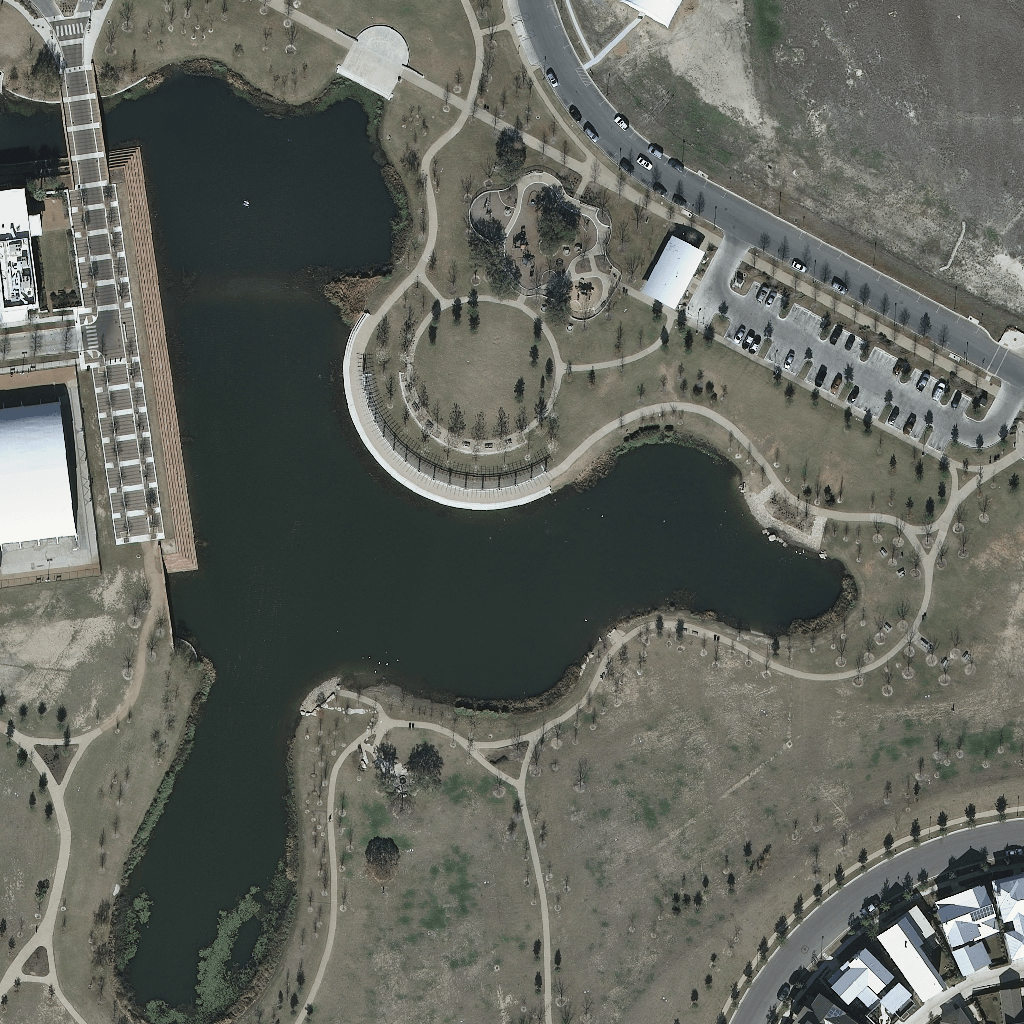}
		\end{minipage}
		\begin{minipage}[b]{0.07\textwidth}
			\centering
			\includegraphics[width=\textwidth]{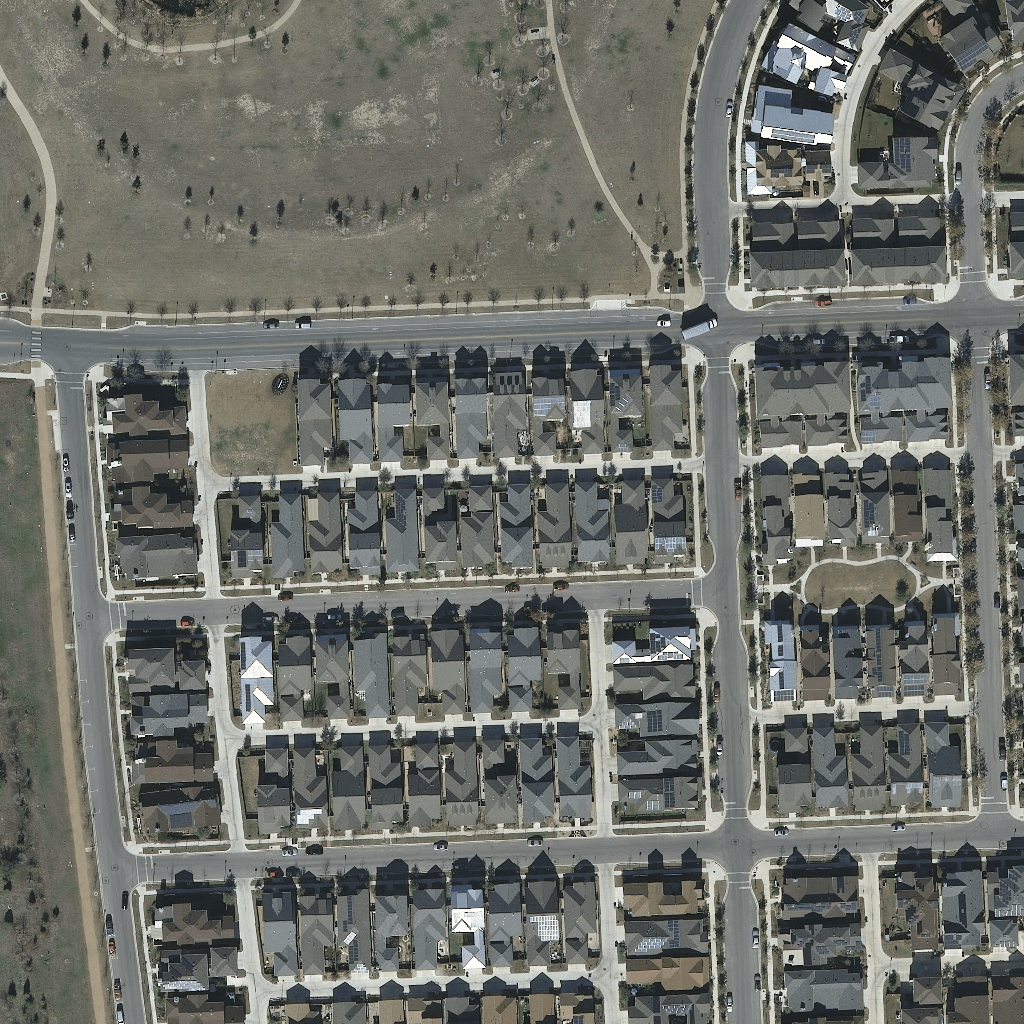}
		\end{minipage}
		\begin{minipage}[b]{0.07\textwidth}
			\centering
			\includegraphics[width=\textwidth]{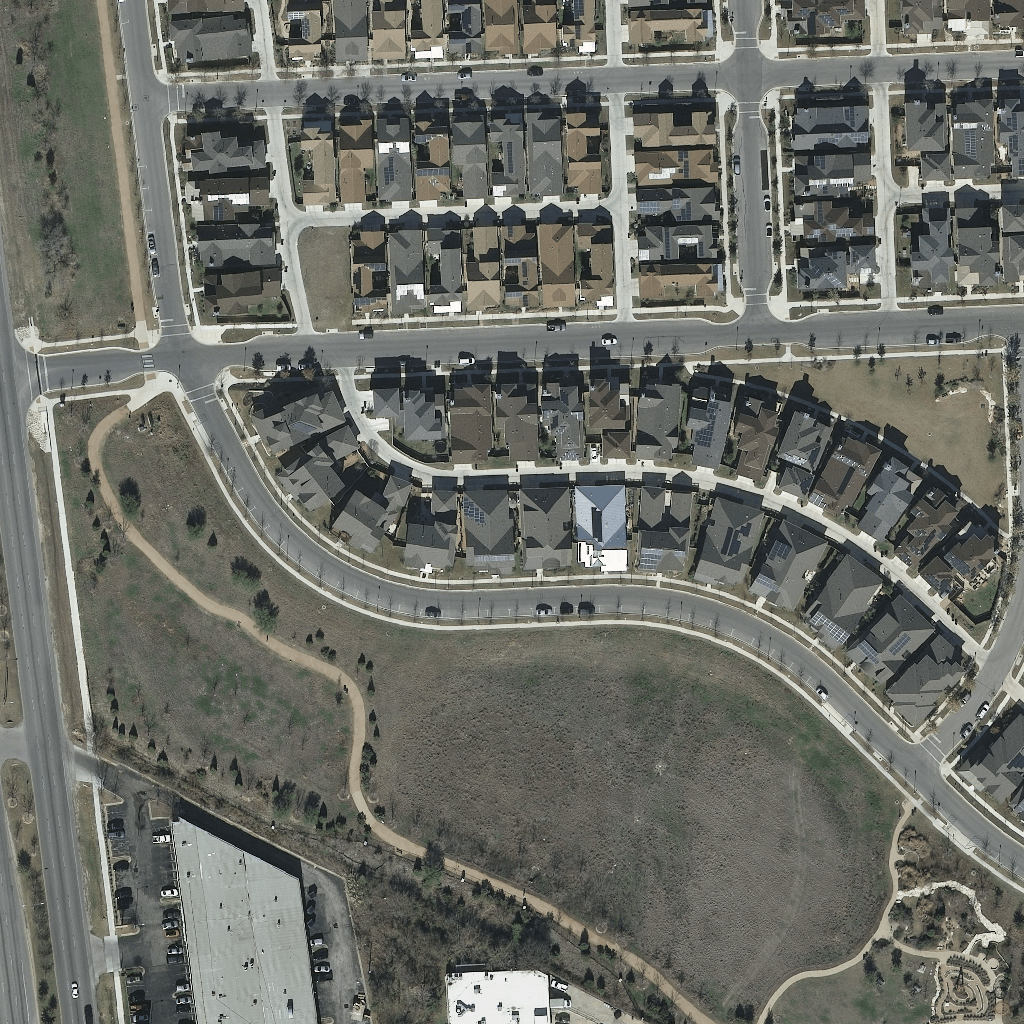}
		\end{minipage}
		\begin{minipage}[b]{0.07\textwidth}
			\centering
			\includegraphics[width=\textwidth]{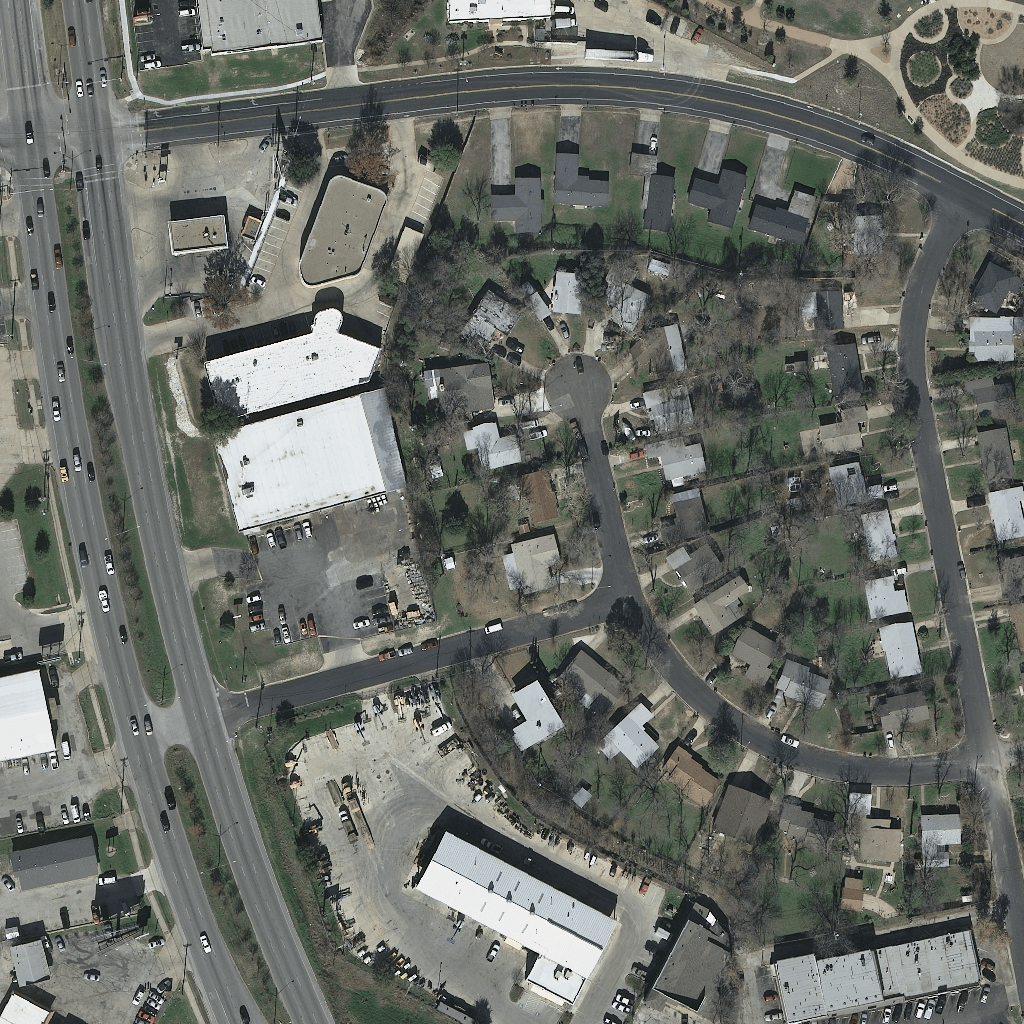}
		\end{minipage}\\[1pt]
		\subfigure{
			\rotatebox{90}{\scriptsize{GT}}
		\begin{minipage}[b]{0.07\textwidth}
			\centering
			\includegraphics[width=\textwidth]{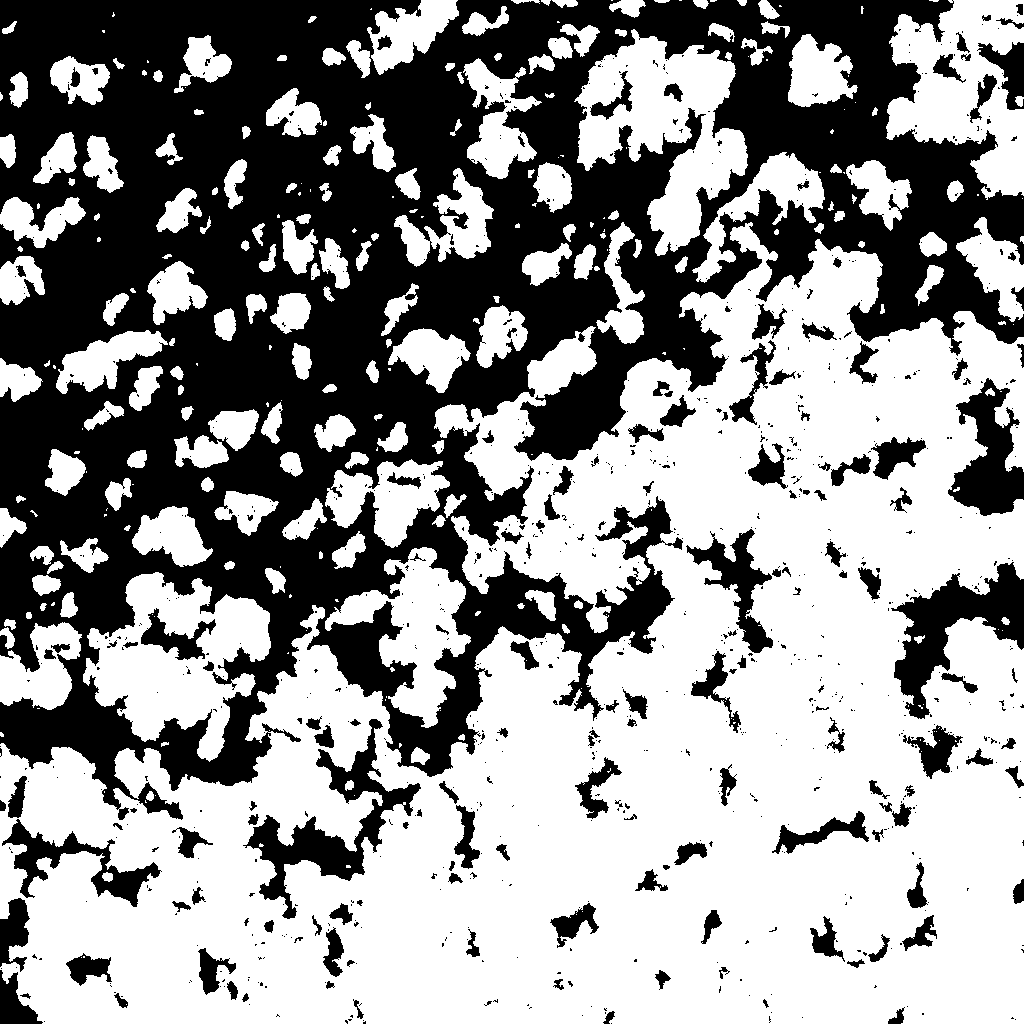}
		\end{minipage}}
		\begin{minipage}[b]{0.07\textwidth}
			\centering
			\includegraphics[width=\textwidth]{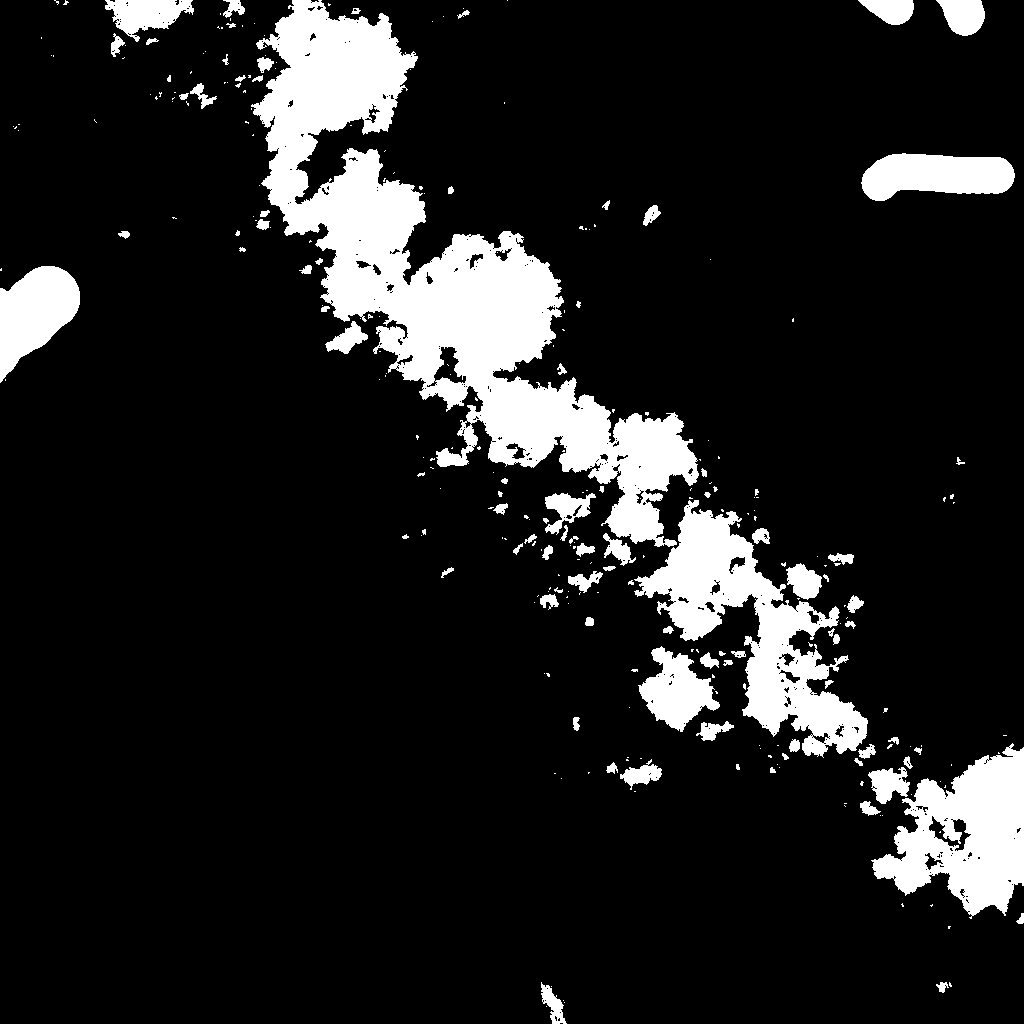}
		\end{minipage}
		\begin{minipage}[b]{0.07\textwidth}
			\centering
			\includegraphics[width=\textwidth]{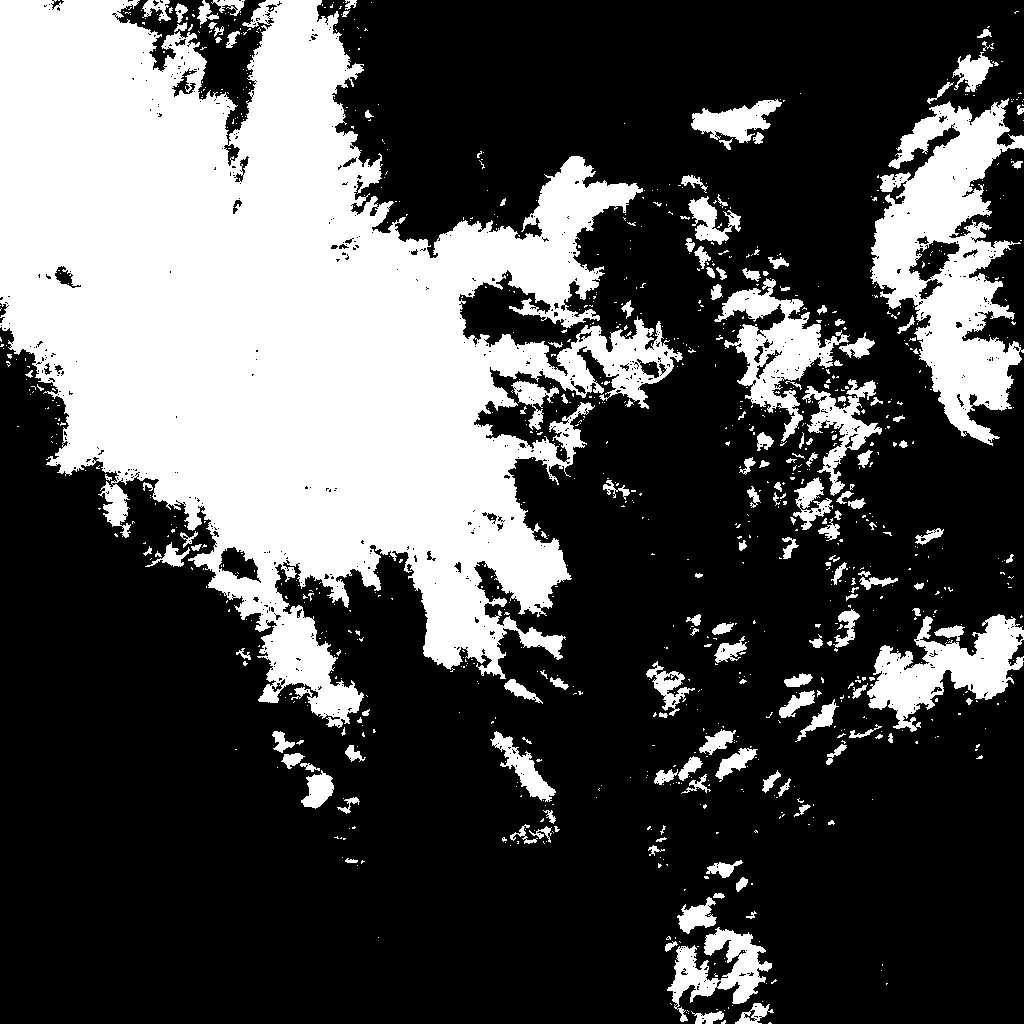}
		\end{minipage}
		\begin{minipage}[b]{0.07\textwidth}
			\centering
			\includegraphics[width=\textwidth]{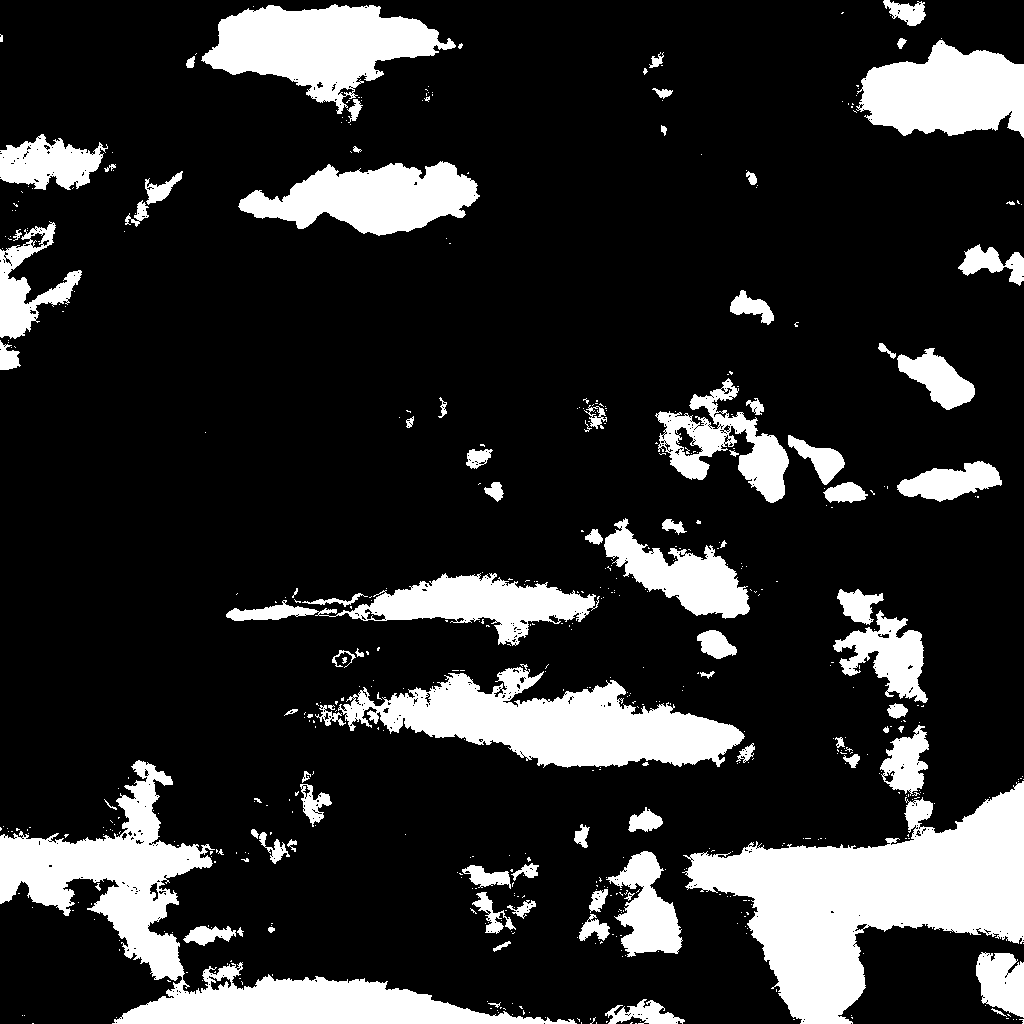}
		\end{minipage}
		\begin{minipage}[b]{0.07\textwidth}
			\centering
			\includegraphics[width=\textwidth]{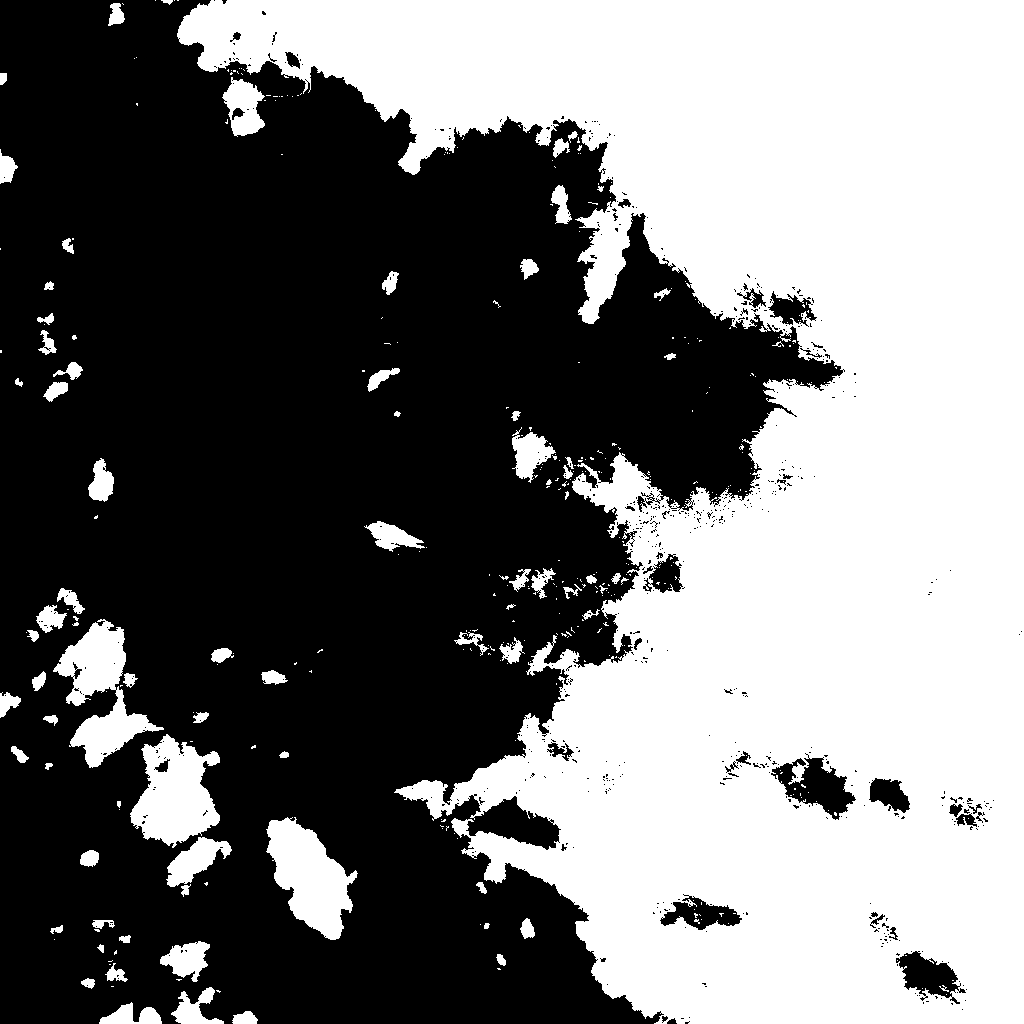}
		\end{minipage}
		\begin{minipage}[b]{0.07\textwidth}
			\centering
			\includegraphics[width=\textwidth]{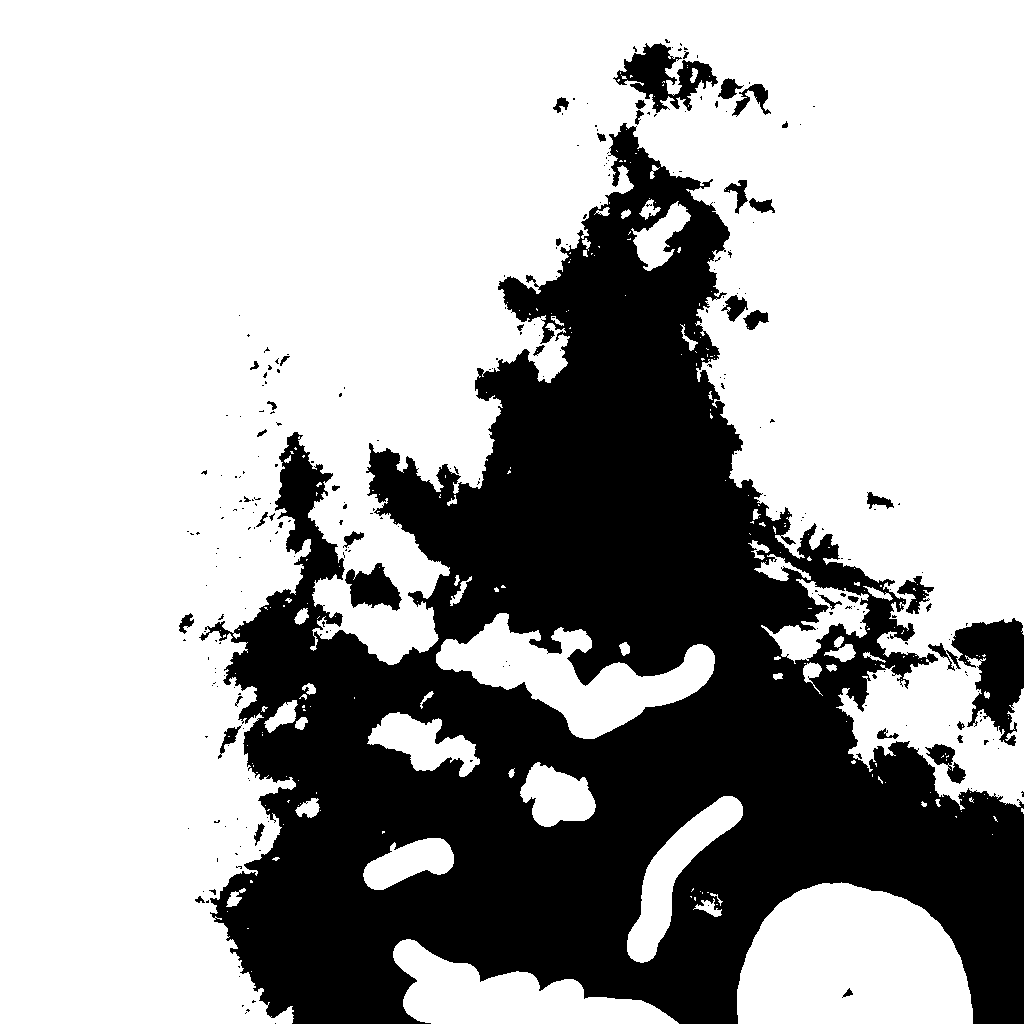}
		\end{minipage}
		\subfigure{
		\rotatebox{90}{\scriptsize{\scriptsize{GT}}}
		\begin{minipage}[b]{0.07\textwidth}
			\centering
			\includegraphics[width=\textwidth]{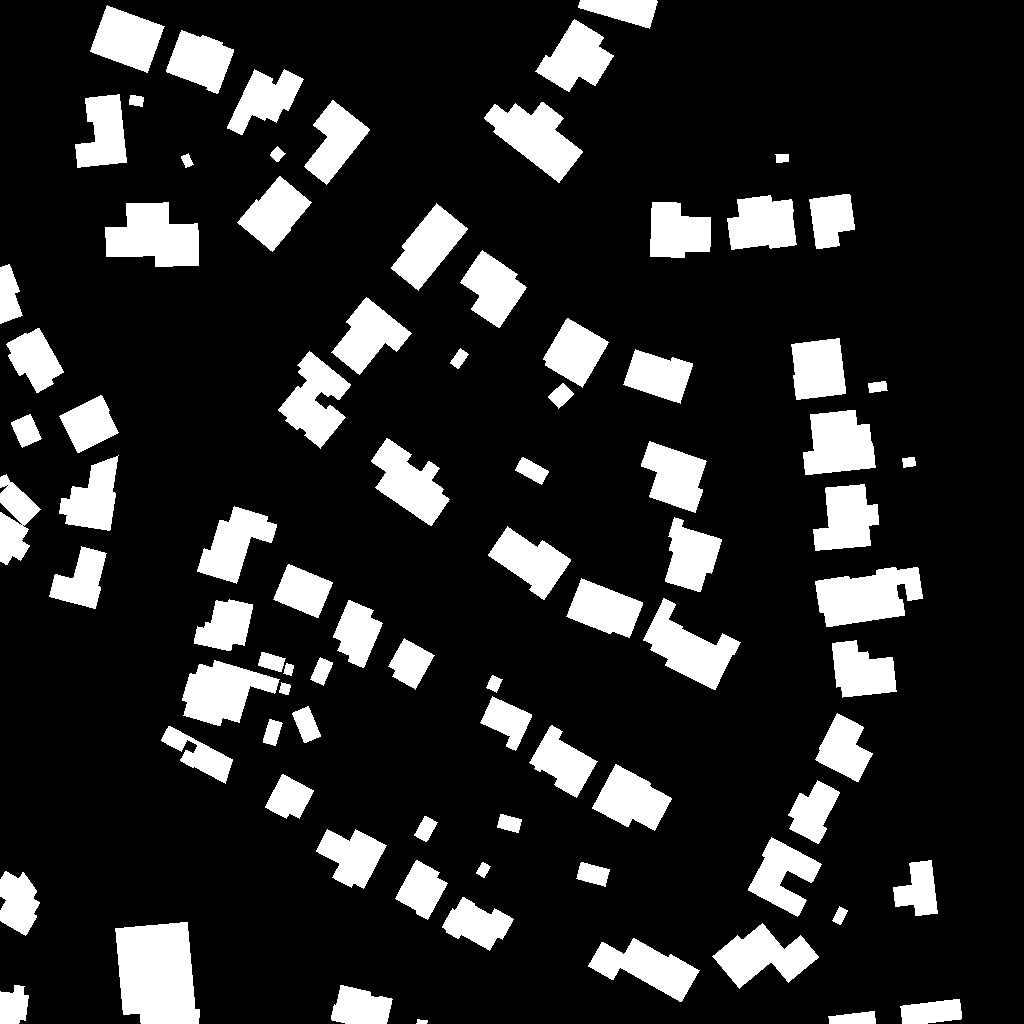}
		\end{minipage}}
		\begin{minipage}[b]{0.07\textwidth}
			\centering
			\includegraphics[width=\textwidth]{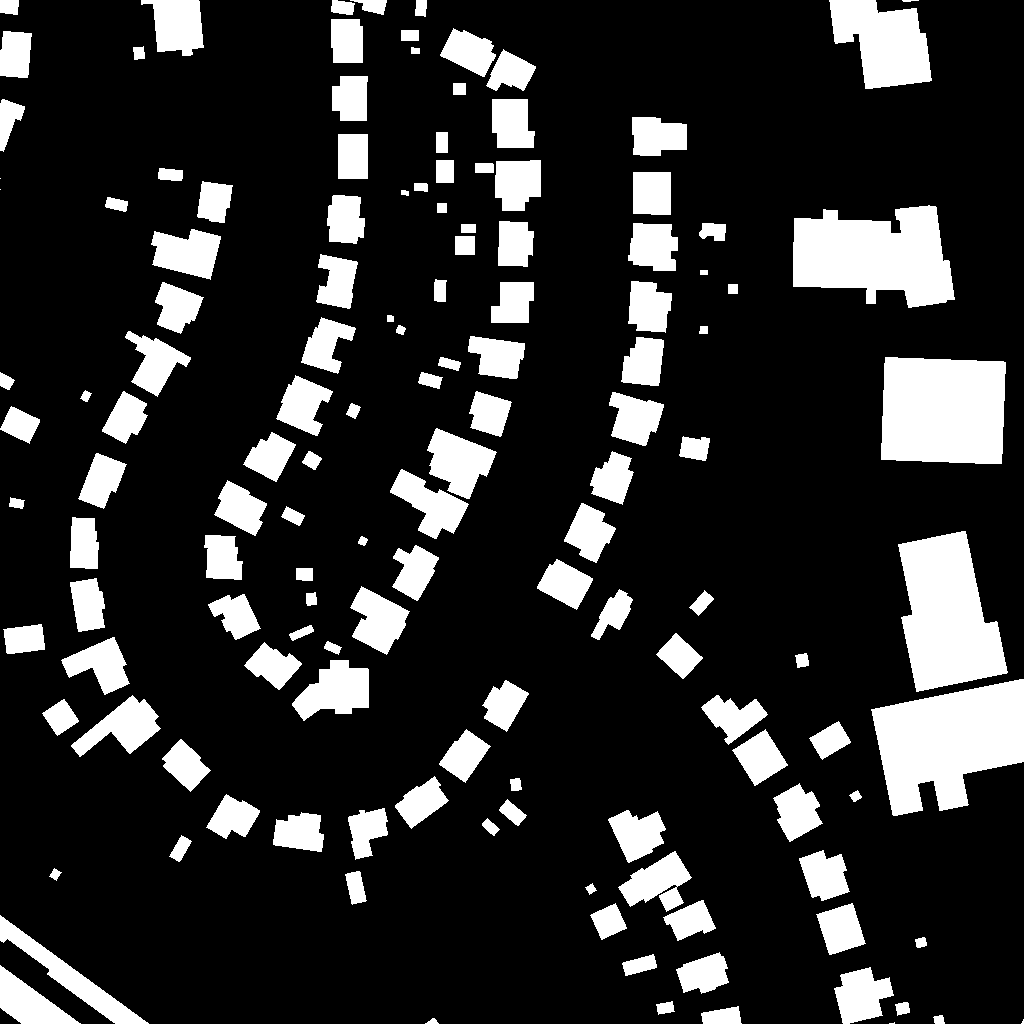}
		\end{minipage}
		\begin{minipage}[b]{0.07\textwidth}
			\centering
			\includegraphics[width=\textwidth]{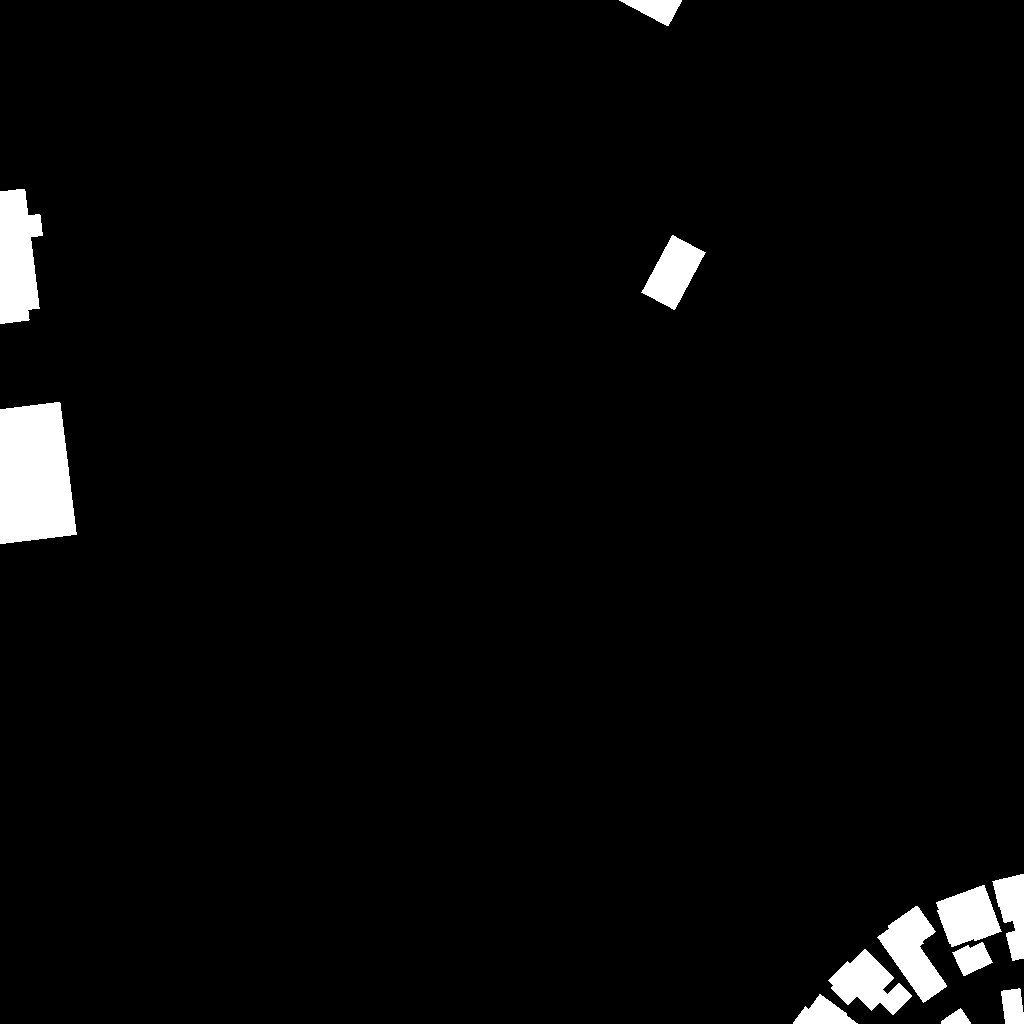}
		\end{minipage}
		\begin{minipage}[b]{0.07\textwidth}
			\centering
			\includegraphics[width=\textwidth]{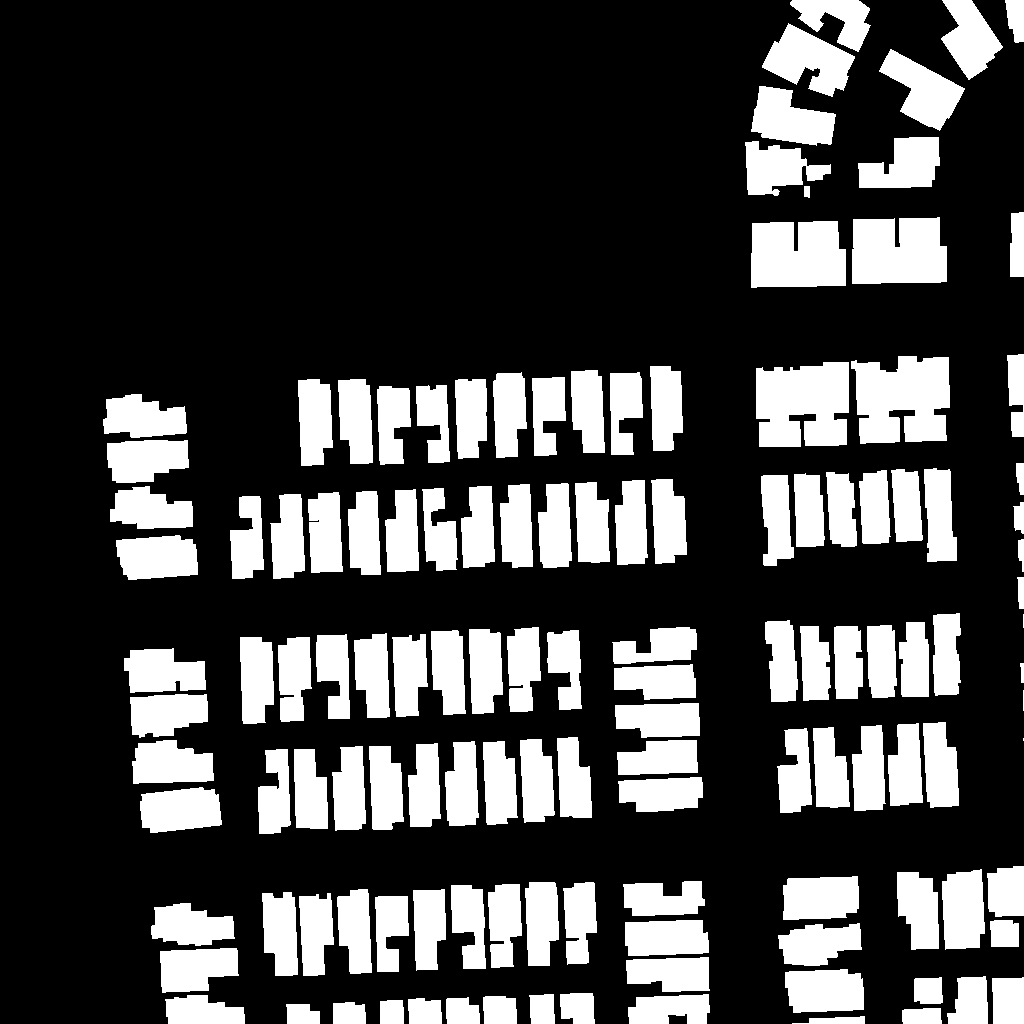}
		\end{minipage}
		\begin{minipage}[b]{0.07\textwidth}
			\centering
			\includegraphics[width=\textwidth]{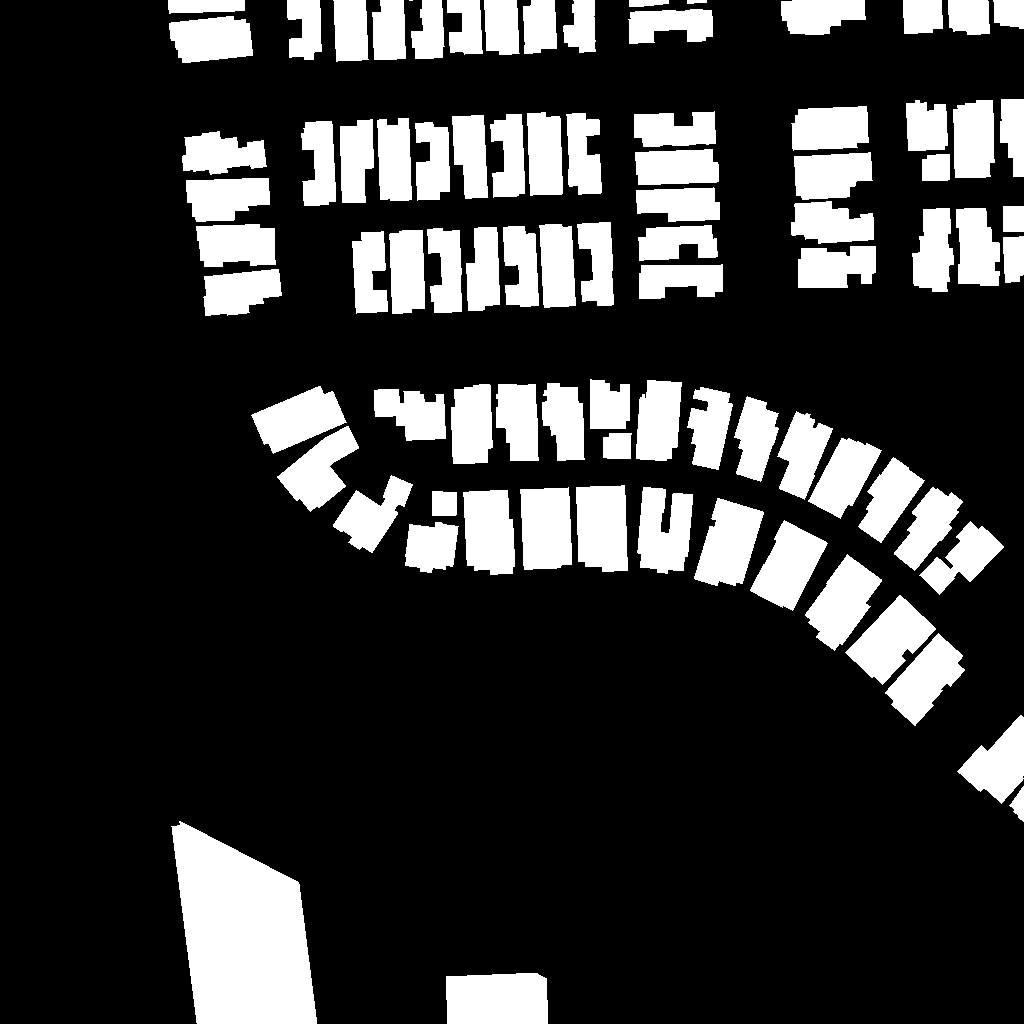}
		\end{minipage}
		\begin{minipage}[b]{0.07\textwidth}
			\centering
			\includegraphics[width=\textwidth]{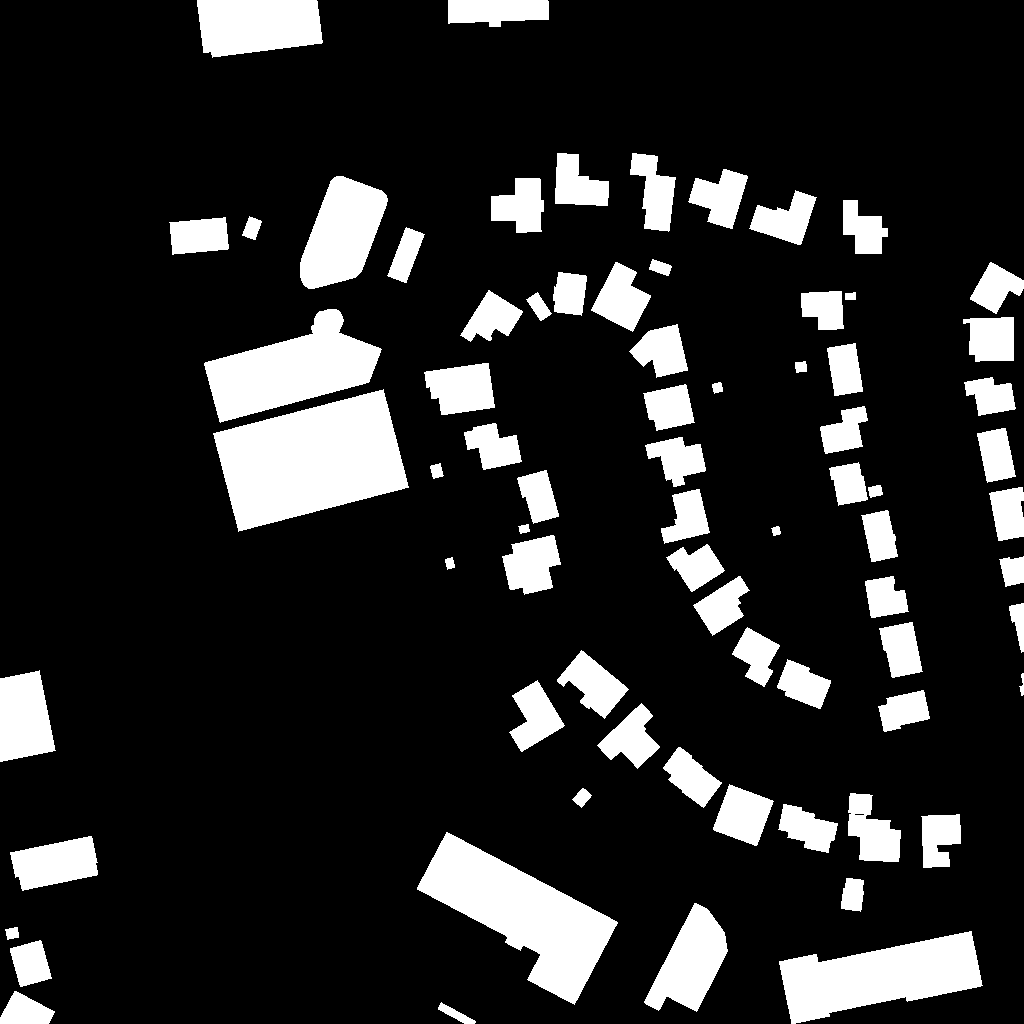}
		\end{minipage}\\[2pt]
		
		\subfigure{
			\rotatebox{90}{\scriptsize{Fields}}
			\begin{minipage}[b]{0.07\textwidth}
				\centering
				\includegraphics[width=\textwidth]{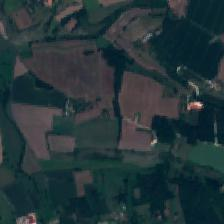}
		\end{minipage}}
		\begin{minipage}[b]{0.07\textwidth}
			\centering
			\includegraphics[width=\textwidth]{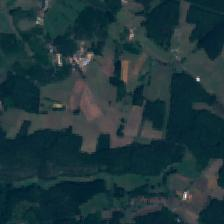}
		\end{minipage}
		\begin{minipage}[b]{0.07\textwidth}
			\centering
			\includegraphics[width=\textwidth]{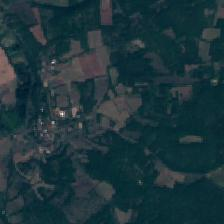}
		\end{minipage}
		\begin{minipage}[b]{0.07\textwidth}
			\centering
			\includegraphics[width=\textwidth]{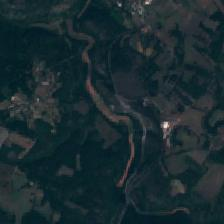}
		\end{minipage}
		\begin{minipage}[b]{0.07\textwidth}
			\centering
			\includegraphics[width=\textwidth]{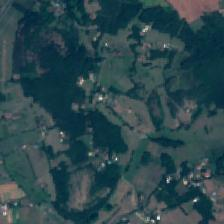}
		\end{minipage}
		\begin{minipage}[b]{0.07\textwidth}
			\centering
			\includegraphics[width=\textwidth]{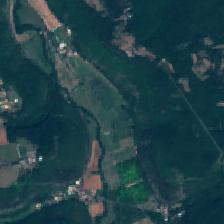}
		\end{minipage}
		\subfigure{
			\rotatebox{90}{\scriptsize{Roads}}
			\begin{minipage}[b]{0.07\textwidth}
				\centering
				\includegraphics[width=\textwidth]{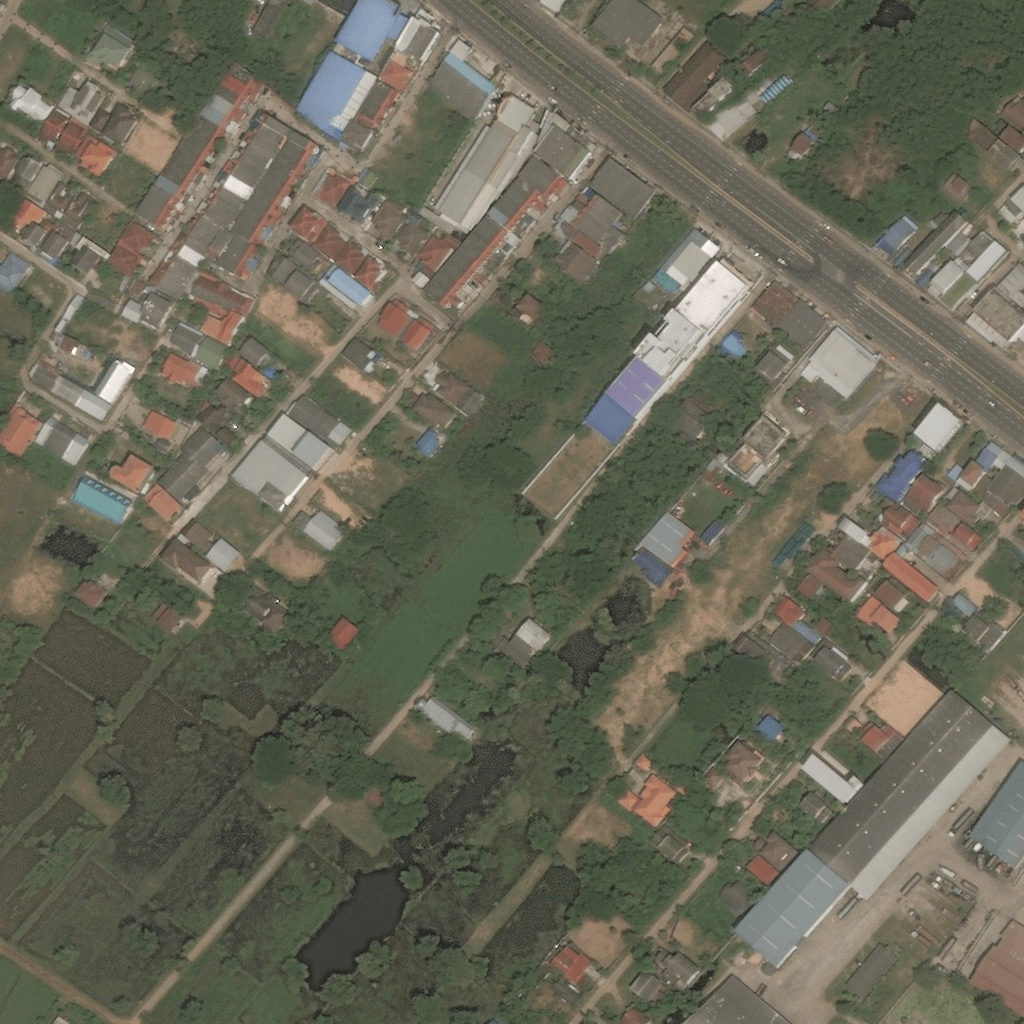}
		\end{minipage}}
		\begin{minipage}[b]{0.07\textwidth}
			\centering
			\includegraphics[width=\textwidth]{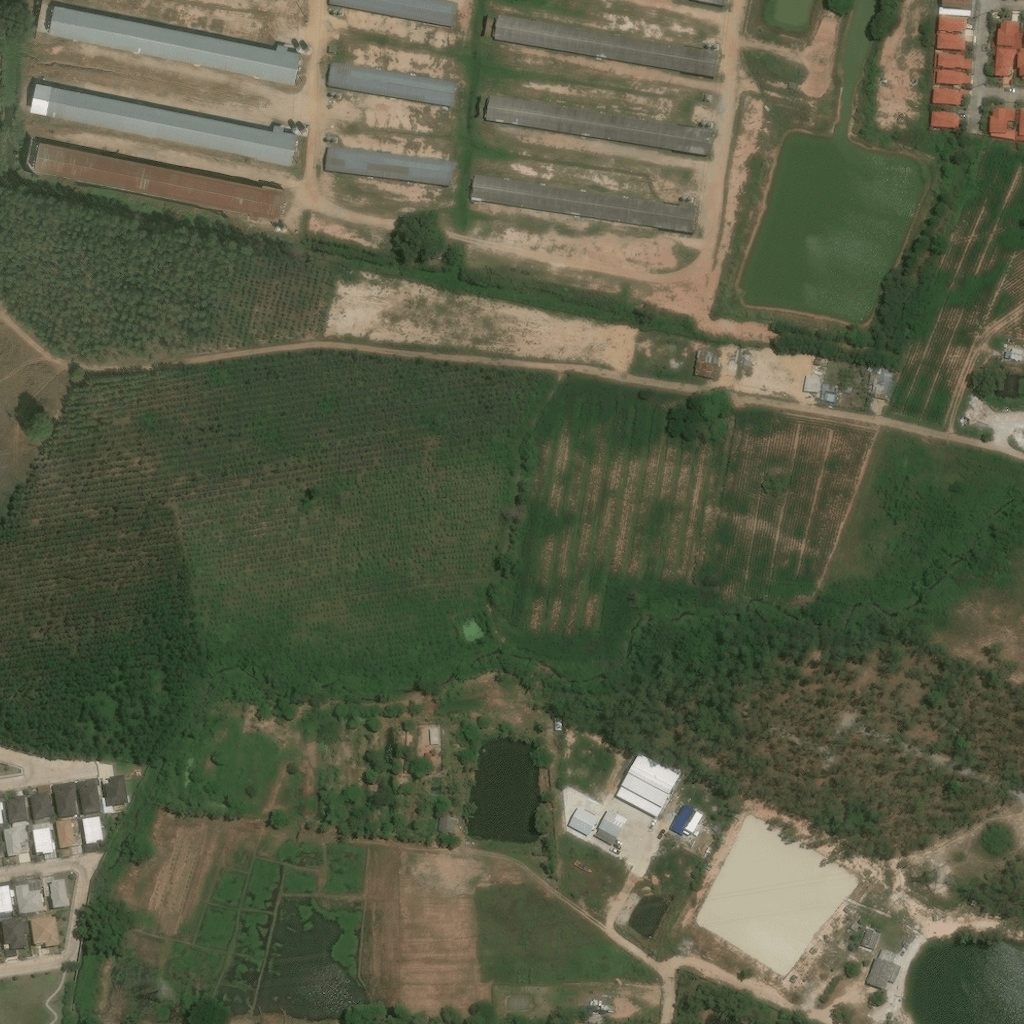}
		\end{minipage}
		\begin{minipage}[b]{0.07\textwidth}
			\centering
			\includegraphics[width=\textwidth]{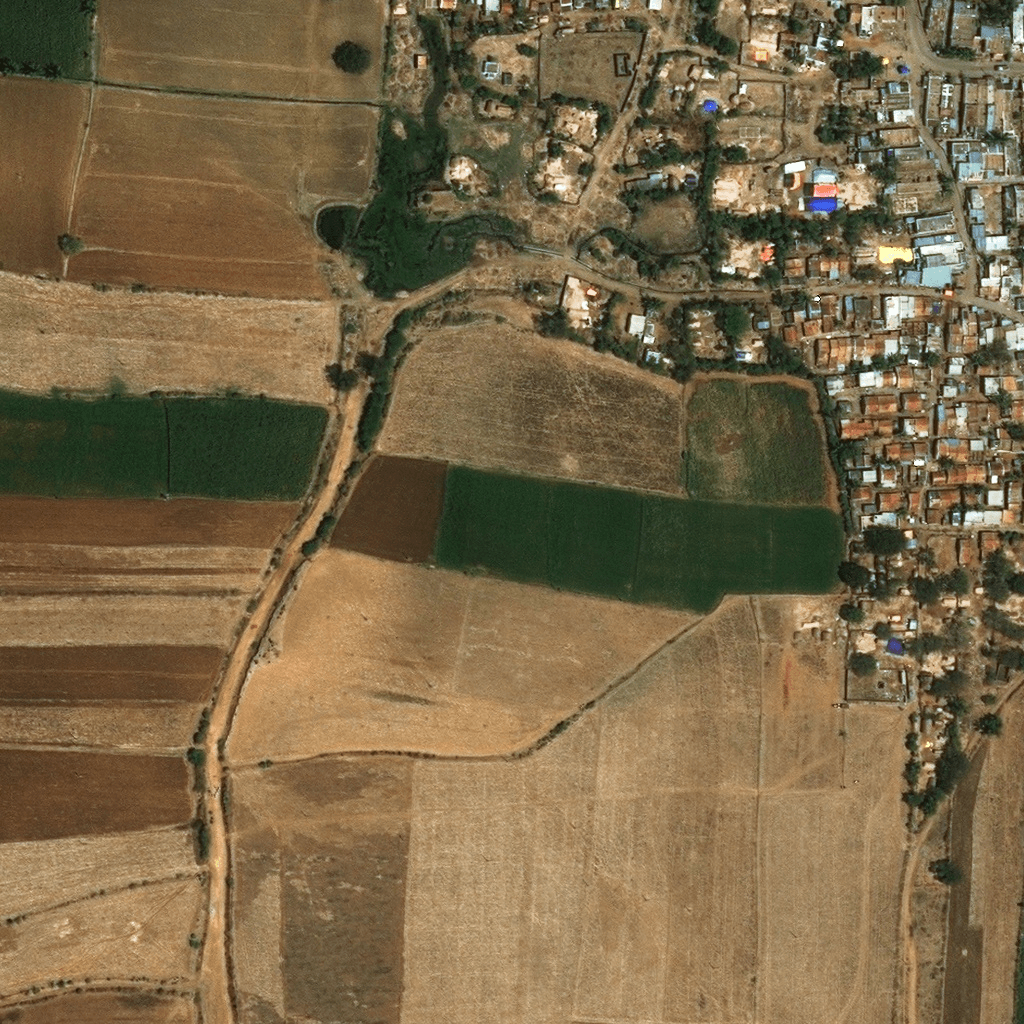}
		\end{minipage}
		\begin{minipage}[b]{0.07\textwidth}
			\centering
			\includegraphics[width=\textwidth]{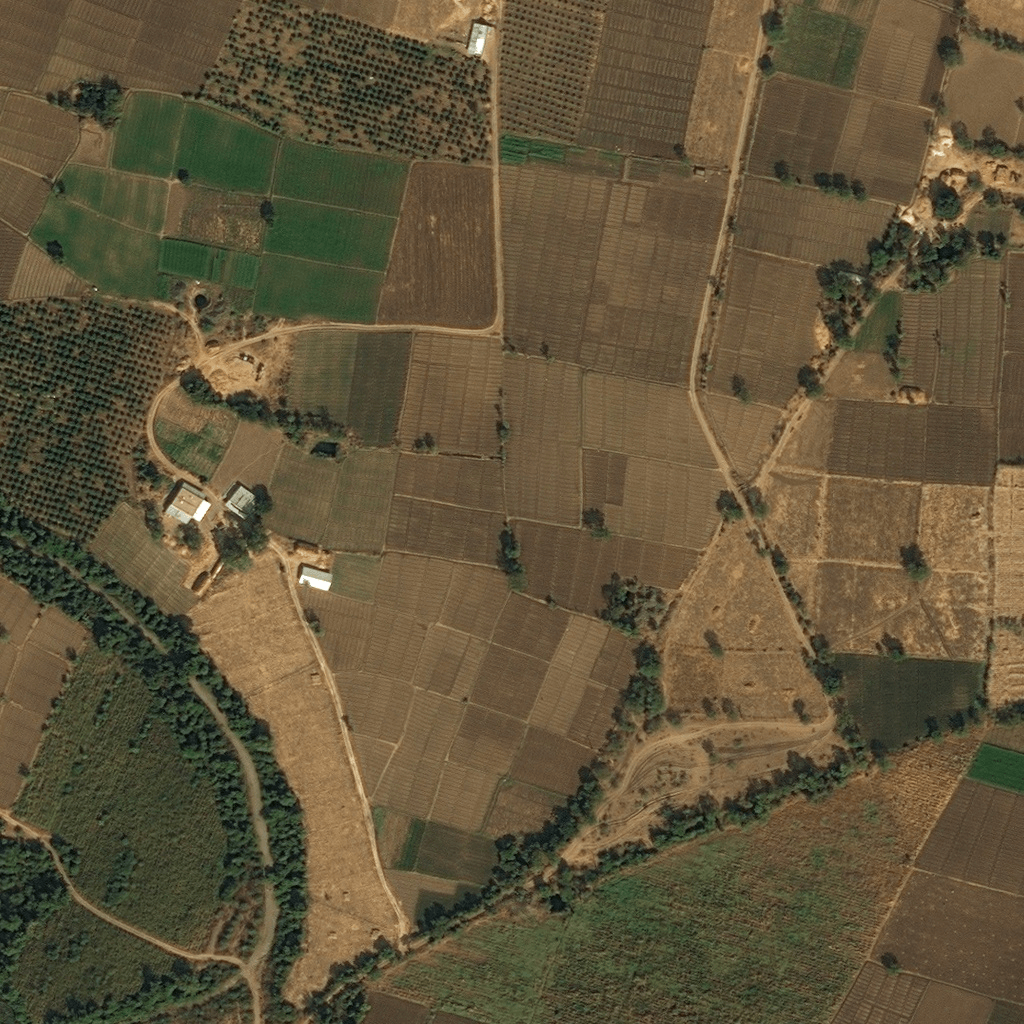}
		\end{minipage}
		\begin{minipage}[b]{0.07\textwidth}
			\centering
			\includegraphics[width=\textwidth]{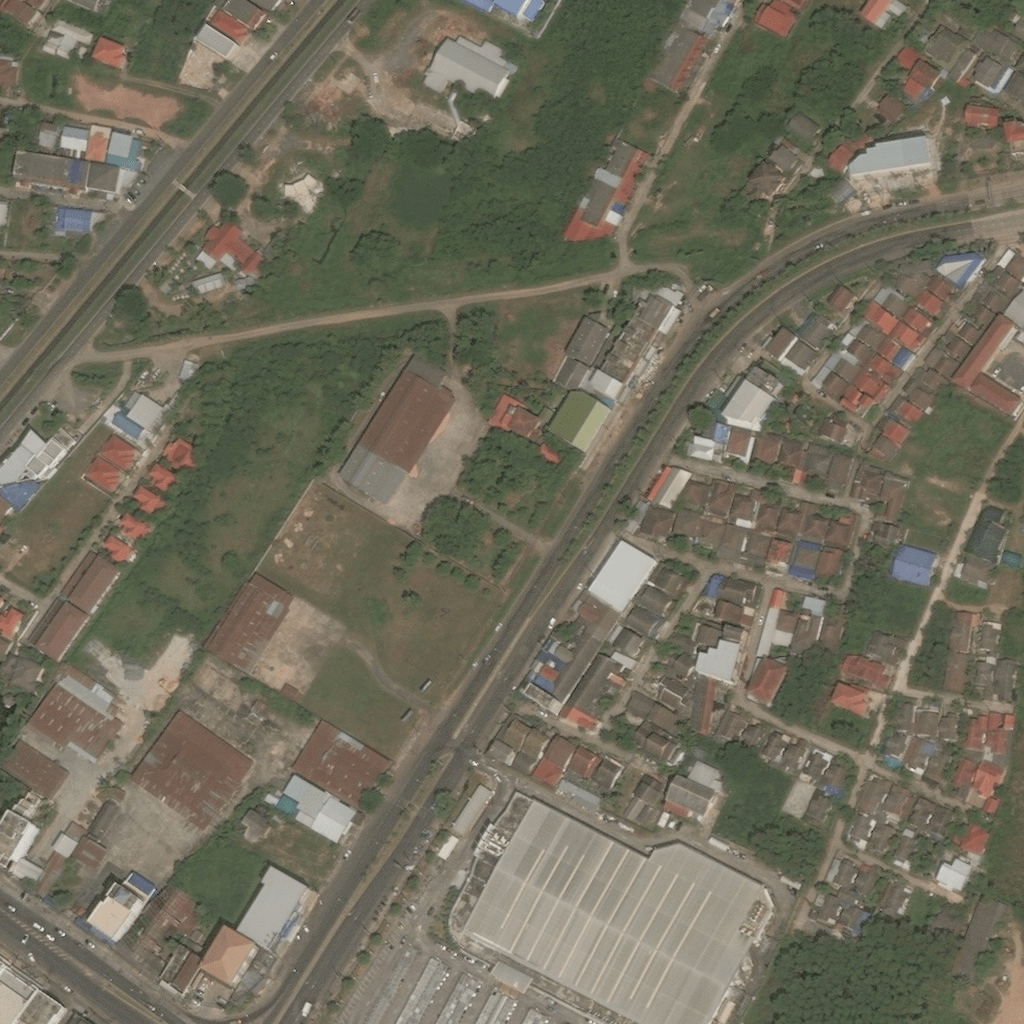}
		\end{minipage}
		\begin{minipage}[b]{0.07\textwidth}
			\centering
			\includegraphics[width=\textwidth]{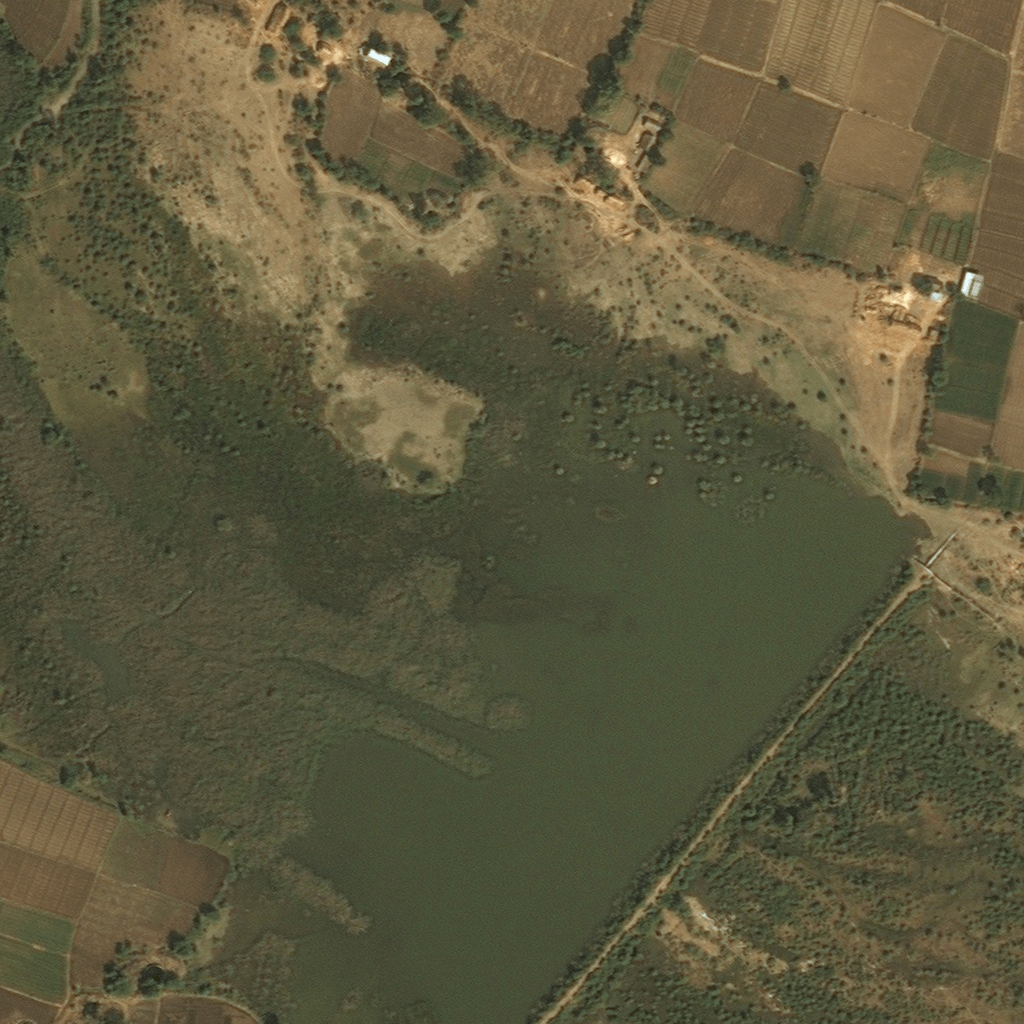}
		\end{minipage}\\[1pt]
		\subfigure{
			\rotatebox{90}{\scriptsize{GT}}
			\begin{minipage}[b]{0.07\textwidth}
				\centering
				\includegraphics[width=\textwidth]{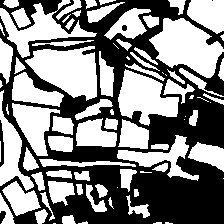}
		\end{minipage}}
		\begin{minipage}[b]{0.07\textwidth}
			\centering
			\includegraphics[width=\textwidth]{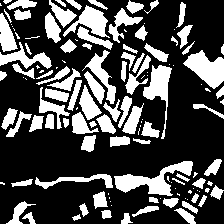}
		\end{minipage}
		\begin{minipage}[b]{0.07\textwidth}
			\centering
			\includegraphics[width=\textwidth]{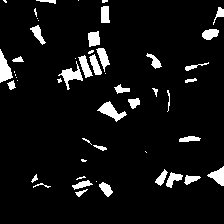}
		\end{minipage}
		\begin{minipage}[b]{0.07\textwidth}
			\centering
			\includegraphics[width=\textwidth]{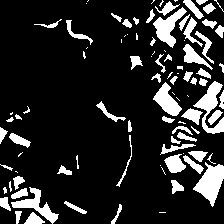}
		\end{minipage}
		\begin{minipage}[b]{0.07\textwidth}
			\centering
			\includegraphics[width=\textwidth]{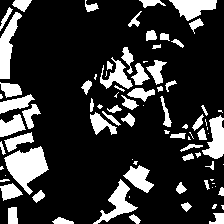}
		\end{minipage}
		\begin{minipage}[b]{0.07\textwidth}
			\centering
			\includegraphics[width=\textwidth]{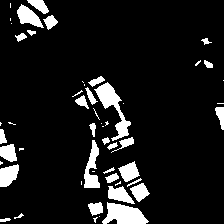}
		\end{minipage}
		\subfigure{
			\rotatebox{90}{\scriptsize{\scriptsize{GT}}}
			\begin{minipage}[b]{0.07\textwidth}
				\centering
				\includegraphics[width=\textwidth]{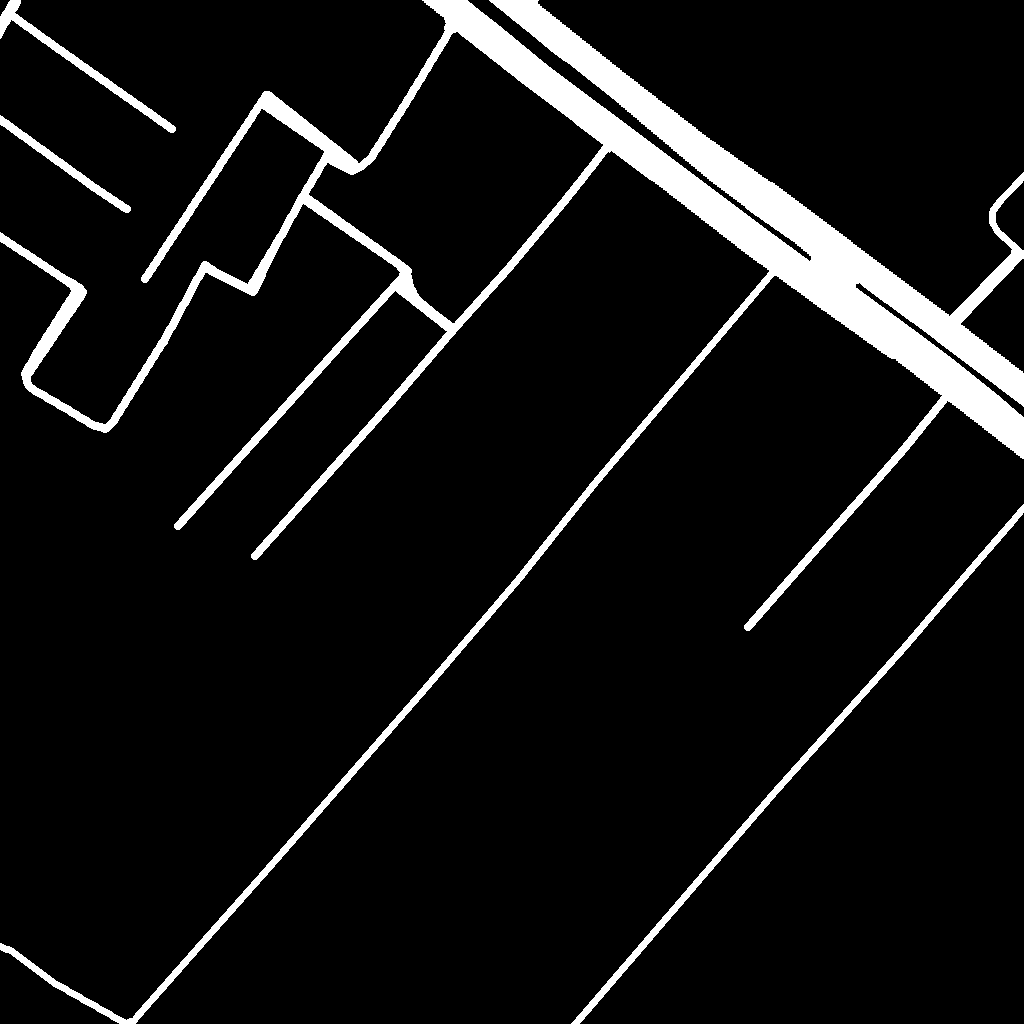}
		\end{minipage}}
		\begin{minipage}[b]{0.07\textwidth}
			\centering
			\includegraphics[width=\textwidth]{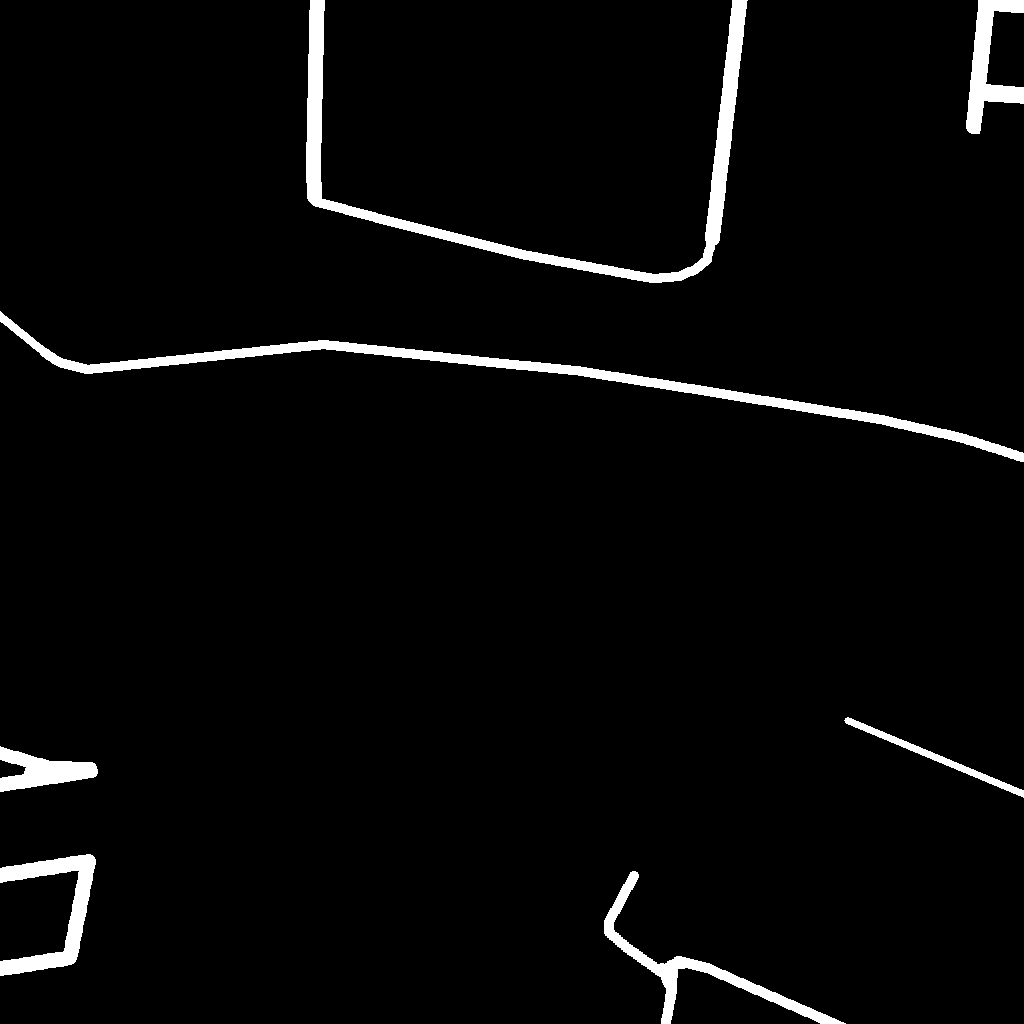}
		\end{minipage}
		\begin{minipage}[b]{0.07\textwidth}
			\centering
			\includegraphics[width=\textwidth]{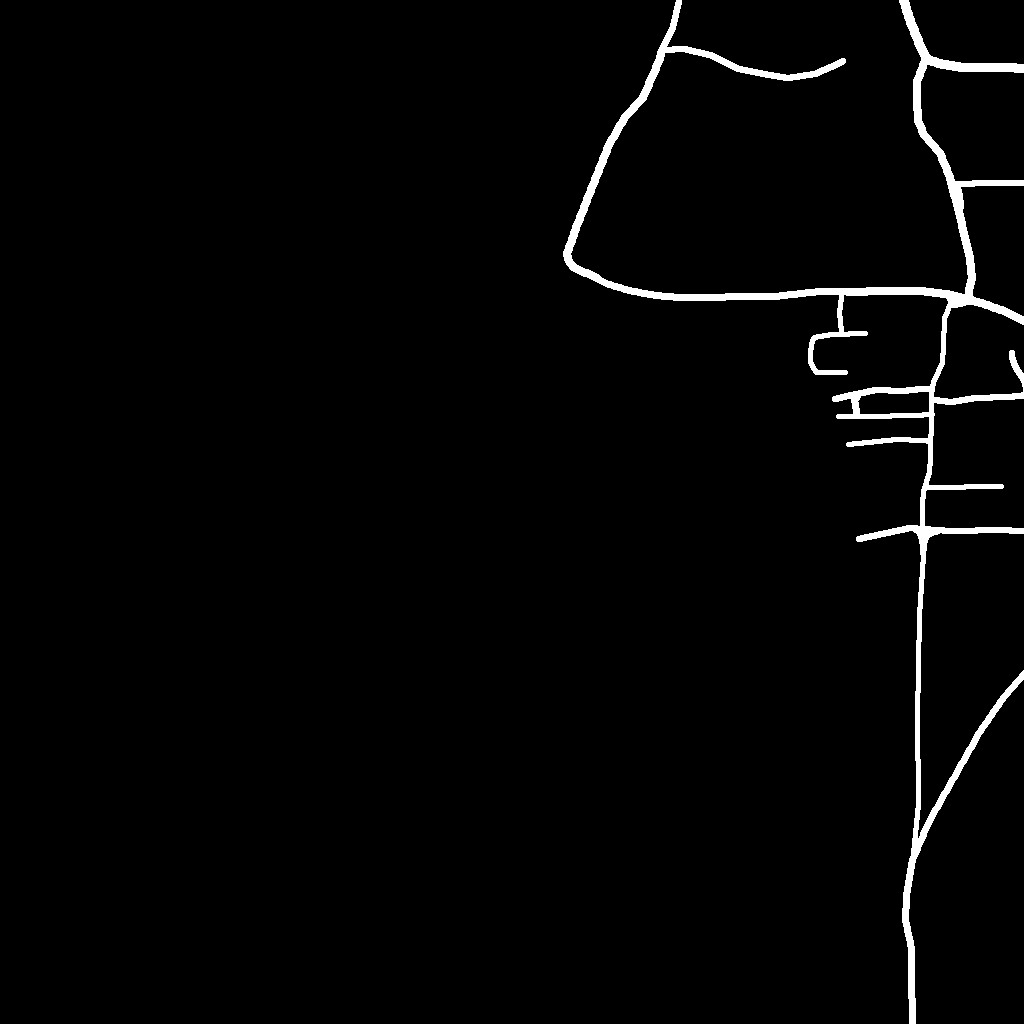}
		\end{minipage}
		\begin{minipage}[b]{0.07\textwidth}
			\centering
			\includegraphics[width=\textwidth]{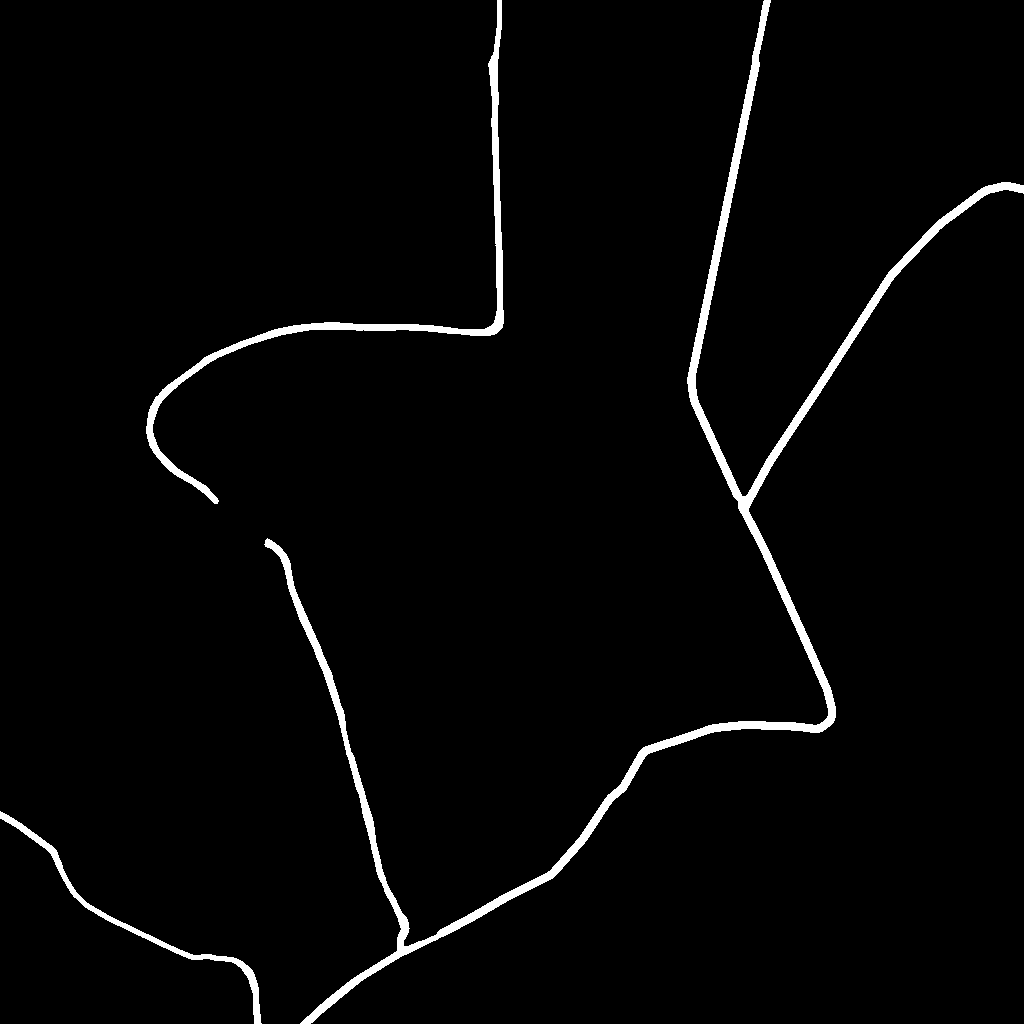}
		\end{minipage}
		\begin{minipage}[b]{0.07\textwidth}
			\centering
			\includegraphics[width=\textwidth]{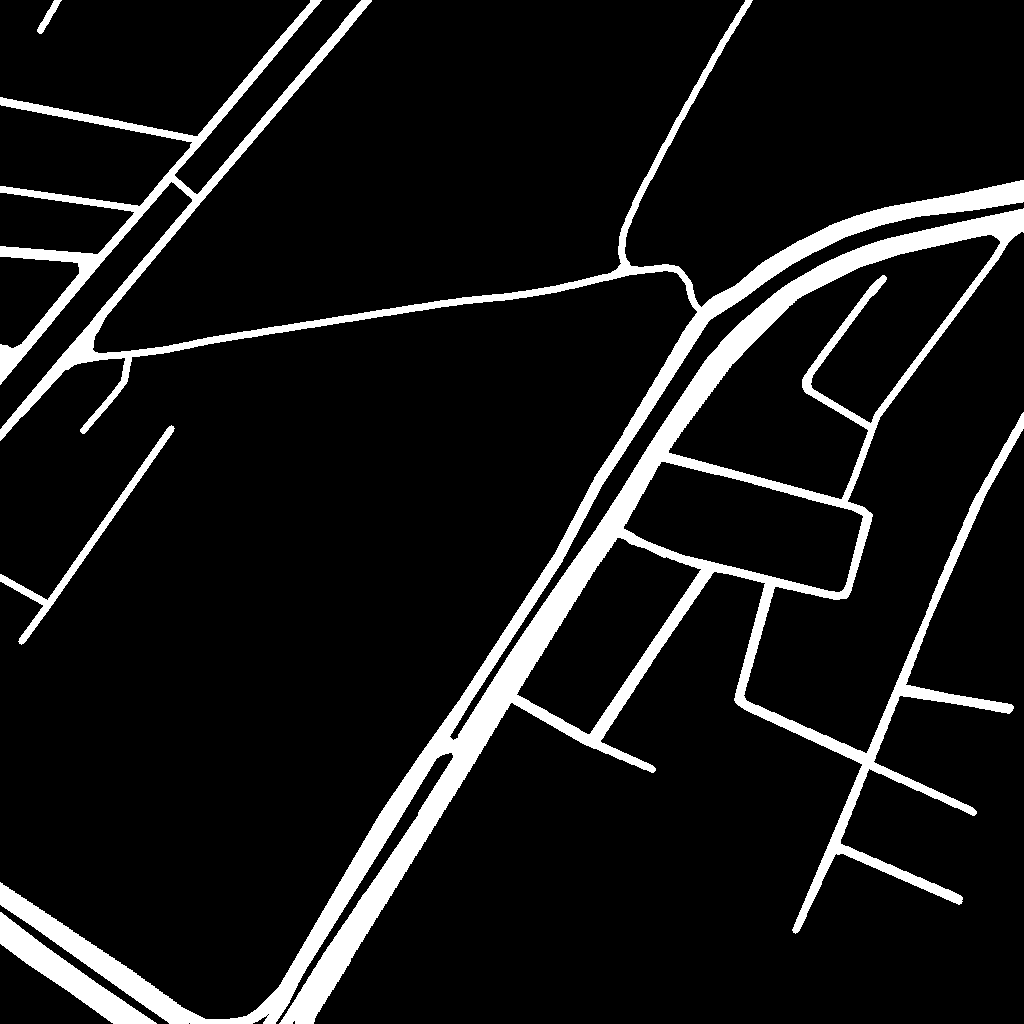}
		\end{minipage}
		\begin{minipage}[b]{0.07\textwidth}
			\centering
			\includegraphics[width=\textwidth]{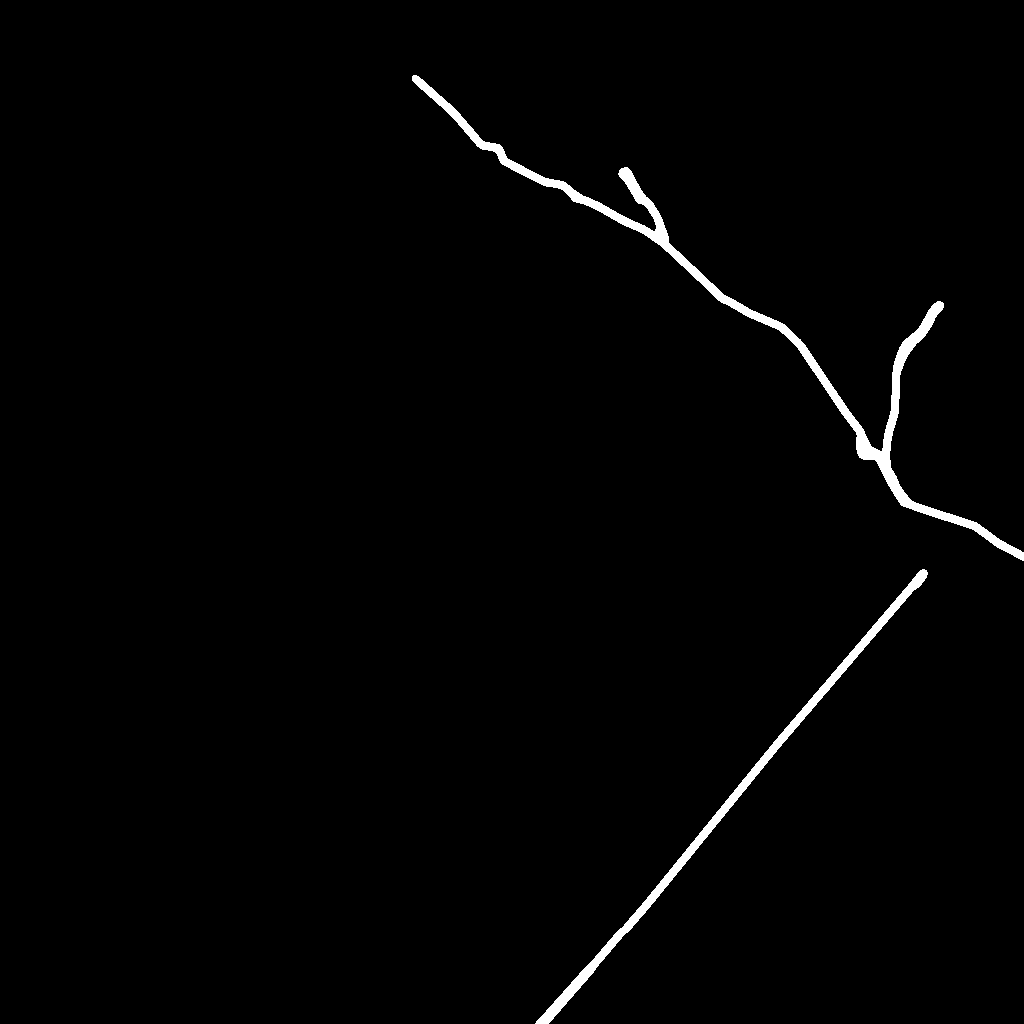}
		\end{minipage}
		\caption{The images in different datasets depict various scenes, including clouds, buildings, fields and roads. The image above shows a remote sensing image, with the corresponding mask displayed below. GT represents ground truth.}
		\label{fig:18pics}
		\end{figure*}
		
	\section{Experiment Settings}
	
		\begin{table}[!htb]\normalsize
		\caption{The detailed information of various datasets.}
		\resizebox{\columnwidth}{!}{%
			\begin{tabular}{|c|c|c|c|c|c|}
				\toprule[2pt]
				\textbf{Scenarios} & \textbf{Region}           & \textbf{Dataset} & \textbf{Total Area ${km}^2$}     & \textbf{Resolution (m)} \\ \hline
				Building  & \begin{tabular}[c]{@{}c@{}}Austin\\ Chicago\\ Kitsap CountyWest TyrolVienna\end{tabular}        & Inria  & 81   & 0.3   \\ \hline
				Cloud & -           &38-Cloud & 2,188,800       & 30  \\ \hline
				Field  & France         &Sentinel-2 & 785       & 6	0  \\ \hline
				Road & \begin{tabular}[c]{@{}c@{}}Thailand\\ Indonesia\\ India\end{tabular} &DeepGlobe & 2,220        & 0.50  \\ 	\bottomrule[2pt]
			
			\end{tabular}%
				}
		\label{tab.datasetinfo}
		
	\end{table}

	\subsection{Datasets}
	Since buildings, clouds, roads and fields are typical scenes in the realm of remote sensing, we select these four scenes to assess the performances of RSAM-Seg. Inria Aerial Image Labeling dataset, 38-Cloud dataset, DeepGlobe Roads dataset and Field Delineation dataset are utilized to respectively evaluate the building, cloud, field and road scenes \cite{inriadataset,38clouddataset,deepglobe18dataset,aung2020farmdataset}. The detailed information of four datasets are listed in Table \ref{tab.datasetinfo}. And Figure \ref{fig:18pics} shows orignal images and ground truth masks of each dataset.
	
	$\mathbf{Inria: }$ The Inria dataset has the coverage of 810 km² (405 km² for training and 405 km² for testing) and aerial orthorectified color imagery with a spatial resolution of 0.3 m. The ground truth data is separated for two semantic classes: building and not building. The original resolution of each image is 5000 × 5000 and then cropped to 1024 × 1024. Finally the building dataset contains 2380 labled patches and 500 unlabled patches as training and testing sets.
	 
	$\mathbf{38-Cloud: }$ The 38-Cloud dataset is a public satellite cloud image dataset collected by the Landsat-8 satellite which includes 9 spectral bands. In this study, three commonly used bands are selected - Band 2 (blue), Band 3 (green), and Band 4 (red) - to compose a three-channel RGB image. The average cloud coverage of the Landsat-8 dataset is 51.6\%. The original resolution of each image is 5000 × 5000 and then cropped to 1024 × 1024 patches. Finally the dataset contains 660 labled and 166 unlabeled patches as training and testing sets.
	
	$\mathbf{Sentinel-2:}$ The Sentinel-2 field dataset has the resolution ranging from 10 to 60 meters in the visible, near infrared (VNIR), and short-wave infrared (SWIR) spectral zones, including 13 spectral channels. In this study, agricultural field scenes from France are selected. The dataset contains 1566 labled patches and 400 unlabeled patches as training and testing sets, each image is cropped to 224 × 224.
	
	$\mathbf{DeepGlobe:}$ The dataset contains 6226 RGB images with resolution of 1024 × 1024, which covers images captured over Thailand, Indonesia and India. The satellite imagery mainly covers regions contained roads. In this study, owing to constraints in hardware capacity, a subset of 2500 patches is selected as the training set and 500 patches as the testing set.

	\subsection{Implementation details}
	In the experiment, ViT-L/14 version of SAM is utilized as the network backbone and trained all datasets using the AdamW optimizer. Additionally, cosine decay is applied to the learning rate and Binary Cross-Entropy (BCE) is used as the loss function. The network is trained for 60 epochs on all datasets. RSAM-Seg is implemented in PyTorch, an NVIDIA A40 (80GB) GPU are used for all experiments.
	\subsection{Performance metrics}
	Seven metrics are employed, including Jaccard index, precision, recall, specificity, F1 score, overall accuracy, and mIoU (mean Intersection over Union), to evaluate the performance of RSAM-Seg on different datasets, the baseline method chosen is U-Net, and the SAM operates in point mode in conjunction with the evaluation process. It is crucial to note that point mode encompasses two distinct variations: center(+) and center(-), representing the center point being labeled as the positive and negative class, respectively. In evaluation, center point coordinates of each image are inputted as prompts into SAM and examined its performance under both positive and negative center point labeling scenarios.
	\section{Results}
	In this section, we evaluate performance of RSAM-Seg compared to U-Net and original SAM on four datasets. 
	\begin{table*}[!htb]
		\centering
		\caption{sons across multiple datasets in terms of Jaccard index, precision, recall, specificity, F1 score, overall accuracy, and mean intersection over union (mIoU).}
		\begin{tabular}{|c|c|c|c|c|c|c|c|c|c|}
			\Xhline{1.2pt}
			\textbf{Scenario}   & \textbf{Method} & \textbf{Resolution}   & \textbf{Jaccard} & \textbf{Precision} & \textbf{Recall} & \textbf{Specificity} & \textbf{F1 score} & \textbf{Overall accuracy} & \textbf{mIoU} \\ \hline
			\Xhline{1.2pt}
			\multirow{4}{*}{Cloud} & U-Net & \multirow{4}{*}{1024*1024} & 0.6461 & $\mathbf{0.8994}$ & 0.6740 & $\mathbf{0.8994}$ & 0.7467 & 0.9021 & 0.7262 \\ \cline{2-2} \cline{4-10} 
			& SAM(center +) &    & 0.1940 & 0.4107  & 0.3929 & 0.4107  & 0.2502 & 0.4838  & 0.2666 \\ \cline{2-2} \cline{4-10} 
			& SAM(center -) &    & 0.0637 & 0.3652 & 0.0801 & 0.3652 & 0.0956 & 0.6221 & 0.3246 \\ \cline{2-2} \cline{4-10} 
			& RSAM-Seg  &    & $\mathbf{0.731}$ & 0.8301 & $\mathbf{0.8396}$ & 0.8301 & $\mathbf{0.8152}$ & $\mathbf{0.9197}$ & $\mathbf{0.7646}$ \\ \hline
			\multirow{4}{*}{Field} & U-Net  & \multirow{4}{*}{224*224} & 0.5011 & 0.6963 & 0.6484 & 0.6963 & 0.6391 & 0.7392 & 0.5545 \\ \cline{2-2} \cline{4-10} 
			& SAM(center +) &    & 0.1798 & 0.5323 & 0.3399 & 0.5323 & 0.2627 & 0.513 & 0.2866 \\ \cline{2-2} \cline{4-10} 
			& SAM(center -) &    & 0.065 & 0.5442 & 0.0751 & 0.5442 & 0.1176 & 0.5502 & 0.2989 \\ \cline{2-2} \cline{4-10} 
			& RSAM-Seg  &    & $\mathbf{0.6346}$ & $\mathbf{0.76}$ & $\mathbf{0.7818}$ & $\mathbf{0.76}$ & $\mathbf{0.7592}$ & $\mathbf{0.8201}$ & $\mathbf{0.6634}$ \\ \hline
			\multirow{4}{*}{Building} & U-Net  & \multirow{4}{*}{1024*1024} & 0.5047 & 0.7151 & 0.6318 & 0.7151 & 0.6496 & 0.9125 & 0.7017 \\ \cline{2-2} \cline{4-10} 
			& SAM(center +) &    & 0.0046 & 0.1522 & 0.0062 & 0.1522 & 0.0083 & 0.807 & 0.4057 \\ \cline{2-2} \cline{4-10} 
			& SAM(center -) &    & 0.0067 & 0.2146 & 0.0072 & 0.2146 & 0.0132 & 0.8433 & 0.4249 \\ \cline{2-2} \cline{4-10} 
			& RSAM-Seg  &    & $\mathbf{0.7353}$ & $\mathbf{0.839}$ & $\mathbf{0.836}$ & $\mathbf{0.839}$ & $\mathbf{0.8337}$ & $\mathbf{0.9583}$ & $\mathbf{0.8424}$ \\ \hline
			\multirow{4}{*}{Road} & U-Net  & \multirow{4}{*}{1024*1024} & 0.5286 & 0.6673 & 0.7276 & 0.6673 & 0.6774 & 0.974 & 0.7506 \\ \cline{2-2} \cline{4-10} 
			& SAM(center +) &    & 0.0068 & 0.0244  & 0.0354 & 0.0244  & 0.0112 & 0.8706  & 0.4383 \\ \cline{2-2} \cline{4-10} 
			& SAM(center -) &    & 0.0031 & 0.0216 & 0.005 & 0.0216 & 0.0061 & 0.9257 & 0.4644 \\ \cline{2-2} \cline{4-10} 
			& RSAM-Seg  &    & $\mathbf{0.6195}$ & $\mathbf{0.7332}$ & $\mathbf{0.8104}$ & $\mathbf{0.7332}$ & $\mathbf{0.7548}$ & $\mathbf{0.9785}$ & $\mathbf{0.7982}$ \\ \hline
			\Xhline{1.2pt}
		\end{tabular}
		\label{tab.RSAMRES}
	\end{table*}	
	\subsection{Results in various scenarios}
	\subsubsection{Results in the cloud scenario}
\begin{figure*}[!htb]
	\centering
	\begin{minipage}[b]{0.10\textwidth}
		\centering
		\includegraphics[width=\textwidth]{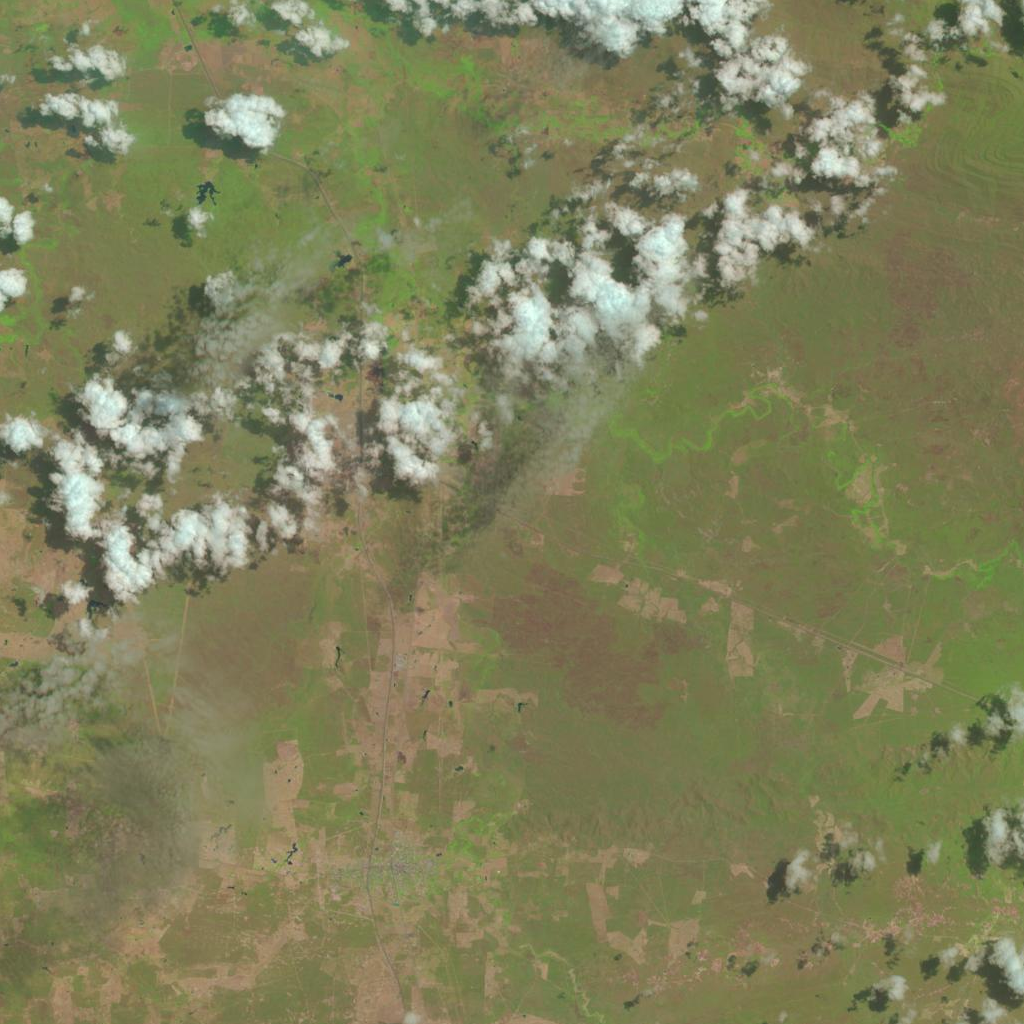}
	\end{minipage}
	\begin{minipage}[b]{0.10\textwidth}
		\centering
		\includegraphics[width=\textwidth]{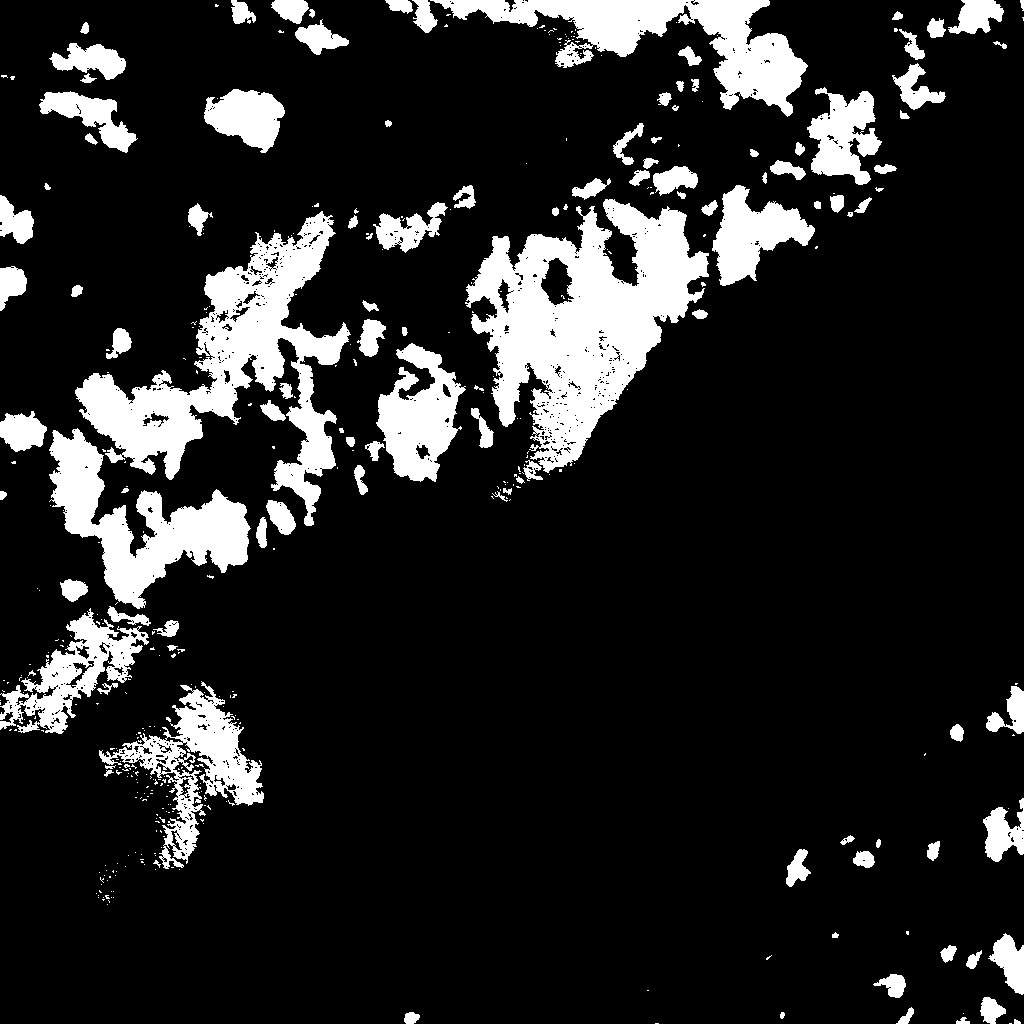}
	\end{minipage}
	\begin{minipage}[b]{0.10\textwidth}
		\centering
		\includegraphics[width=\textwidth]{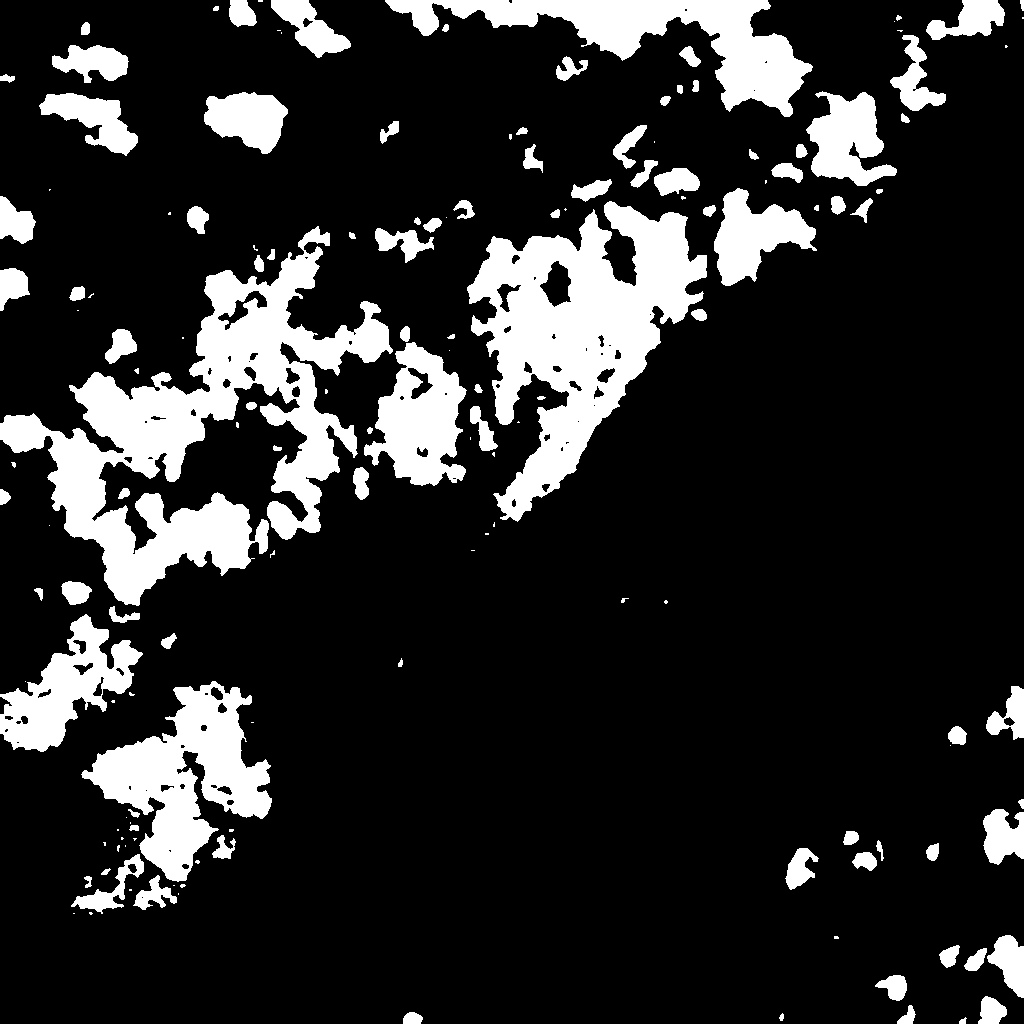}
	\end{minipage}
	\begin{minipage}[b]{0.10\textwidth}
		\centering
		\includegraphics[width=\textwidth]{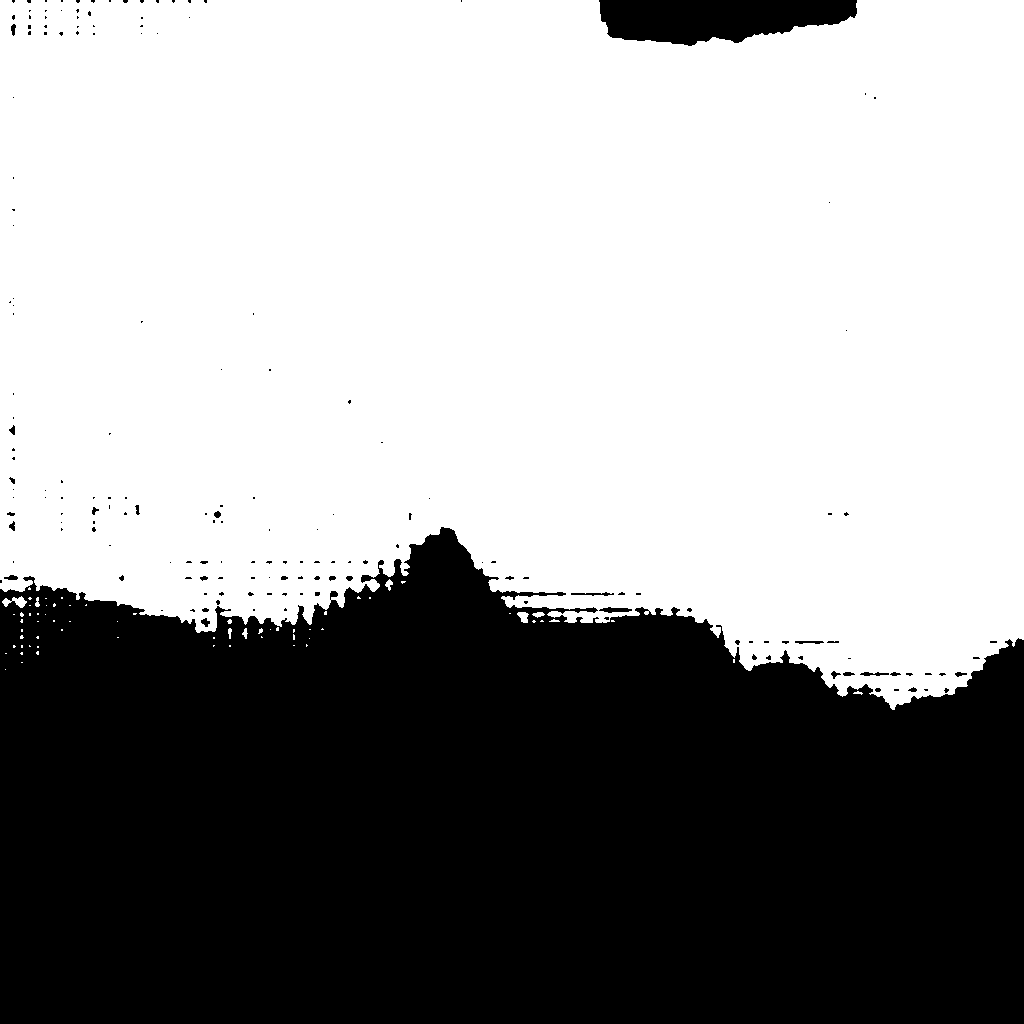}
	\end{minipage}
	\begin{minipage}[b]{0.10\textwidth}
		\centering
		\includegraphics[width=\textwidth]{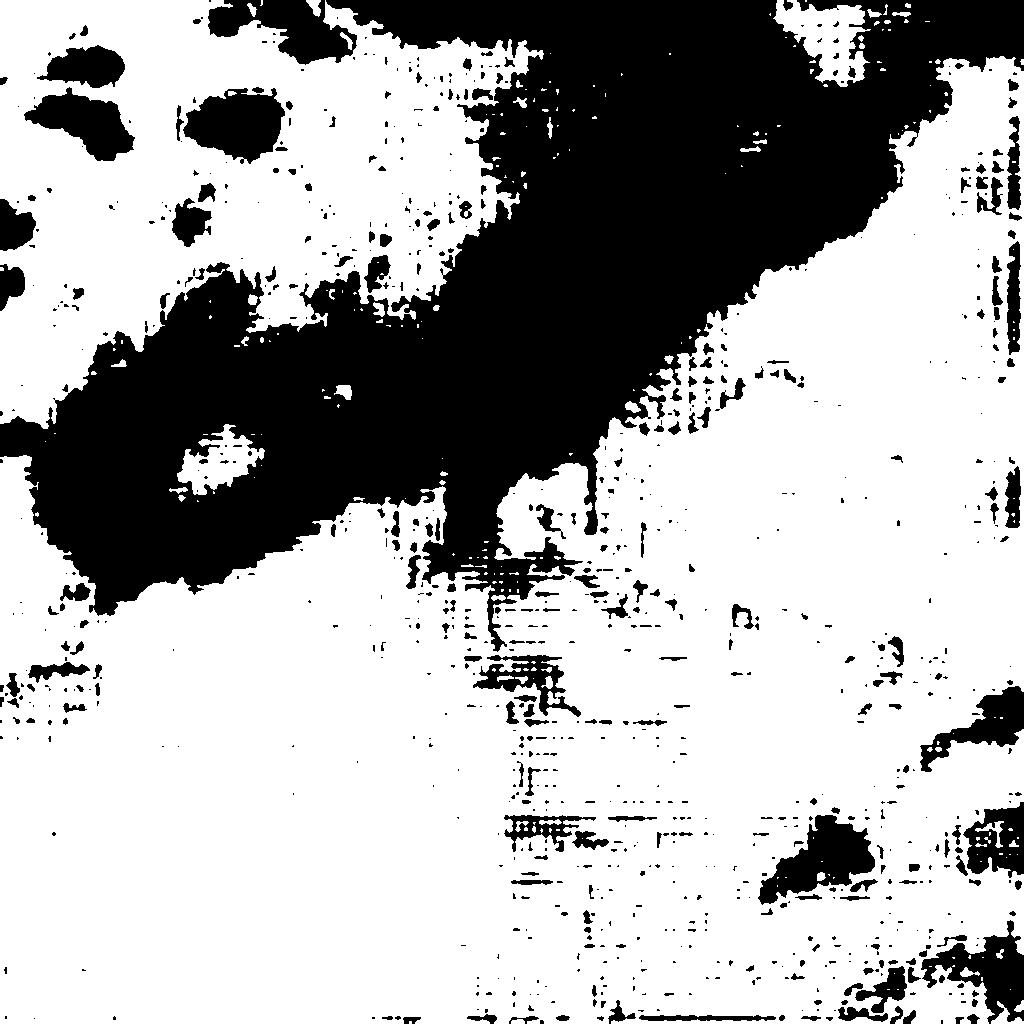}
	\end{minipage}
	\begin{minipage}[b]{0.10\textwidth}
		\centering
		\includegraphics[width=\textwidth]{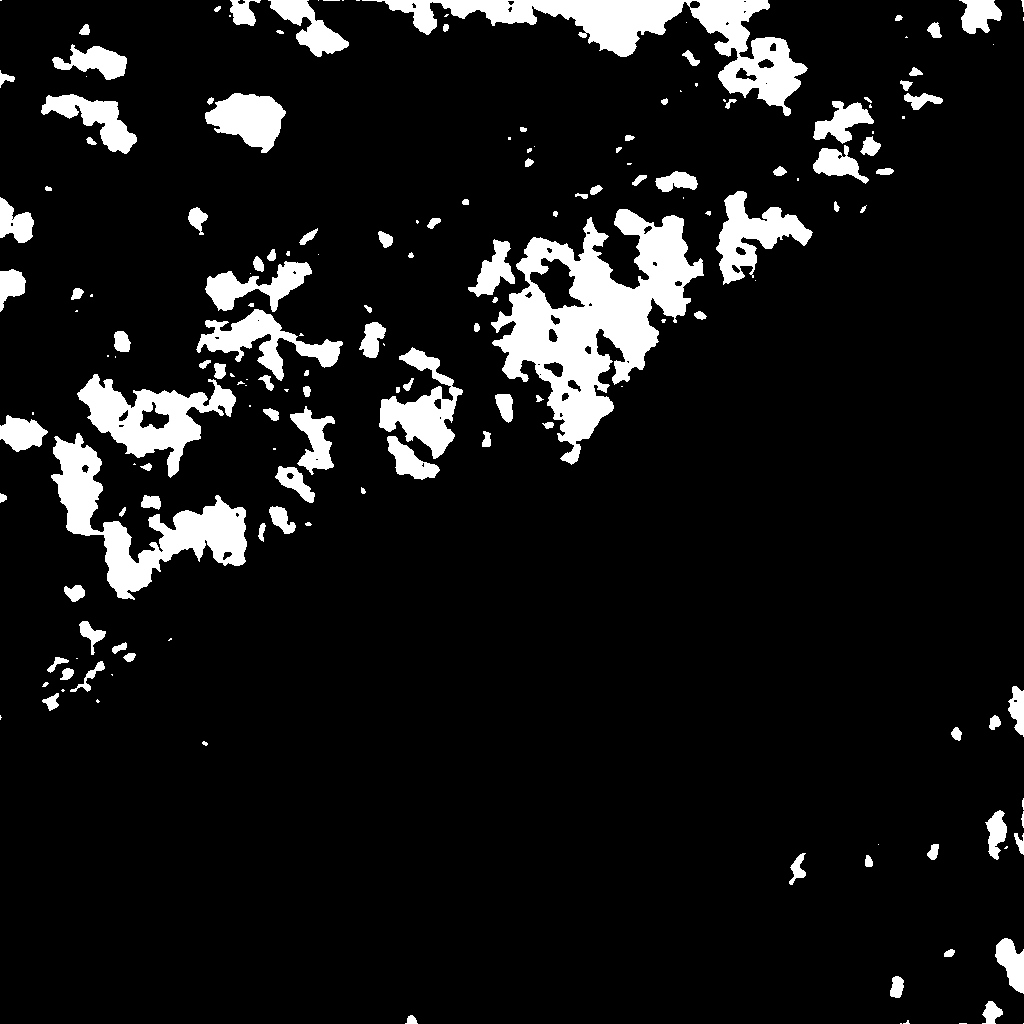}
	\end{minipage}\\[3pt]
	\begin{minipage}[b]{0.10\textwidth}
		\centering
		\includegraphics[width=\textwidth]{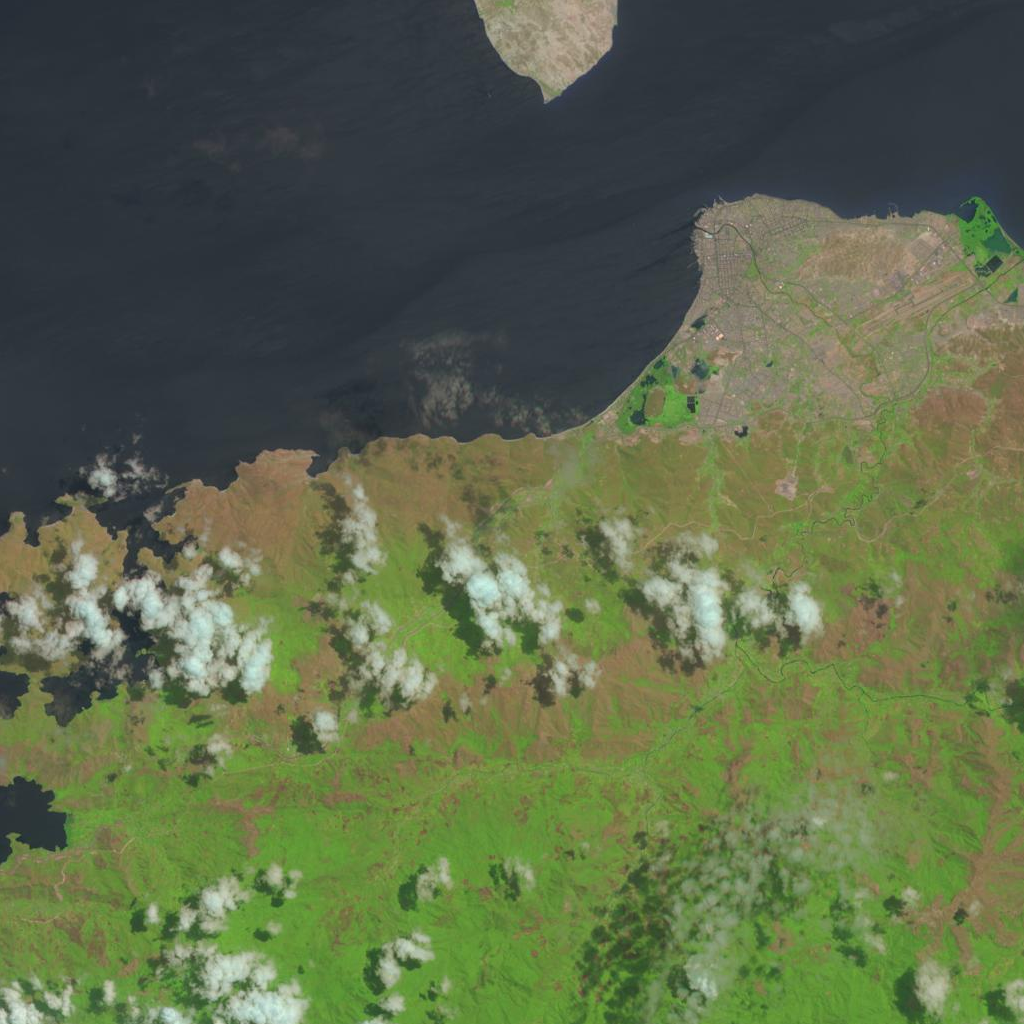}
	\end{minipage}
	\begin{minipage}[b]{0.10\textwidth}
		\centering
		\includegraphics[width=\textwidth]{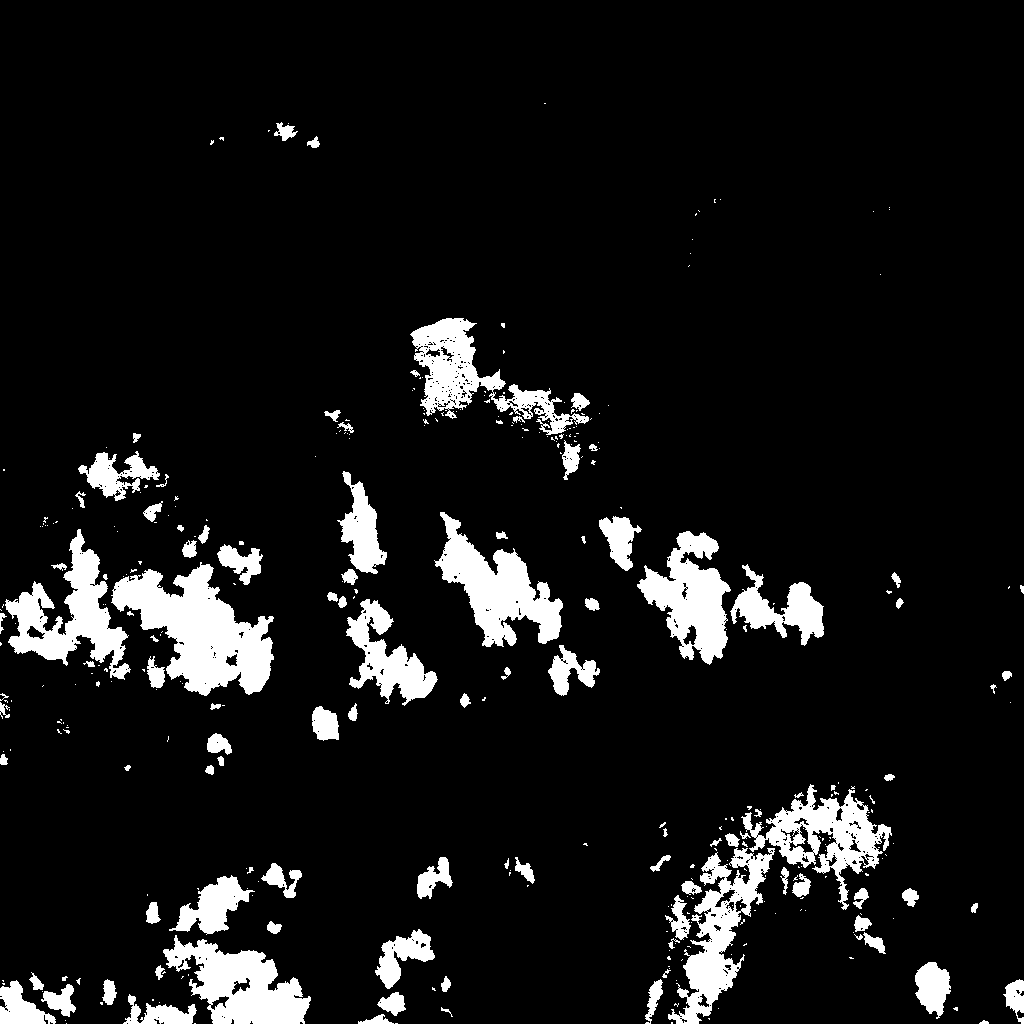}
	\end{minipage}
	\begin{minipage}[b]{0.10\textwidth}
		\centering
		\includegraphics[width=\textwidth]{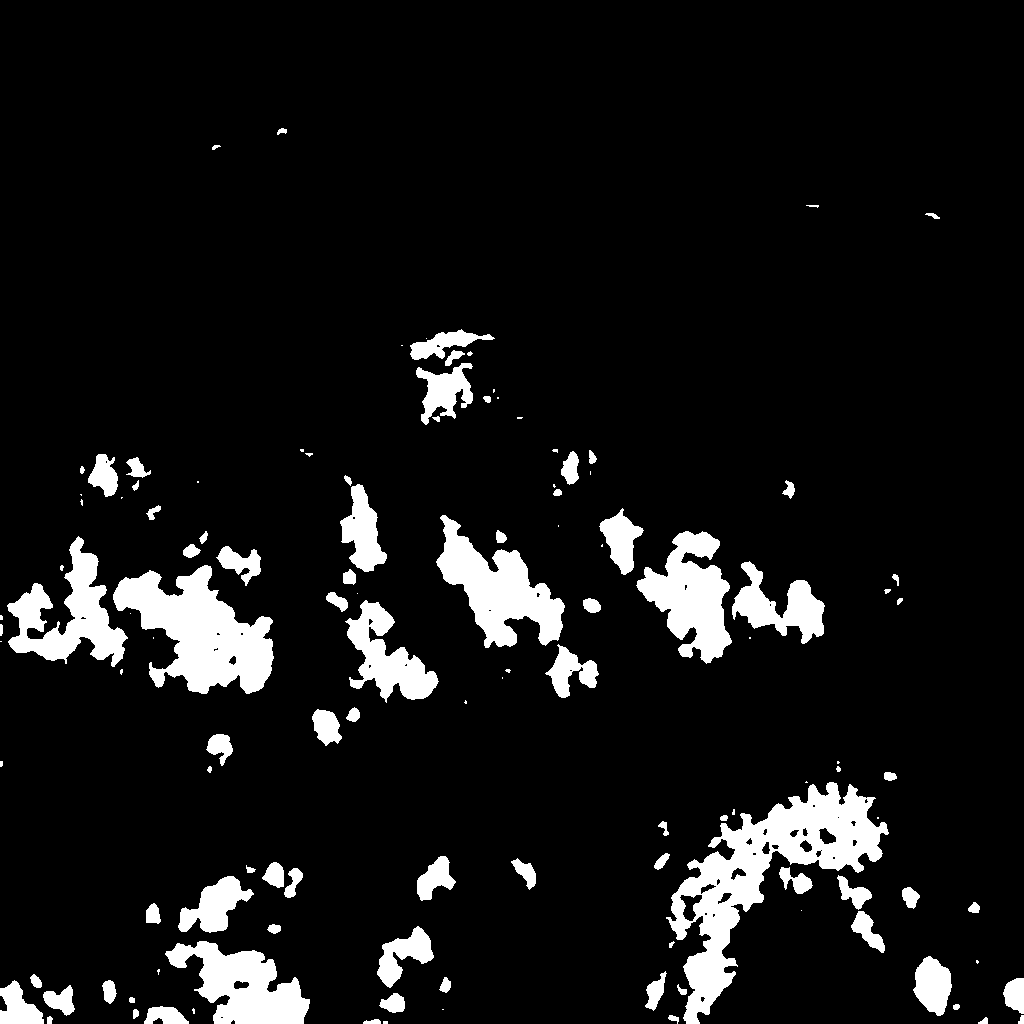}
	\end{minipage}
	\begin{minipage}[b]{0.10\textwidth}
		\centering
		\includegraphics[width=\textwidth]{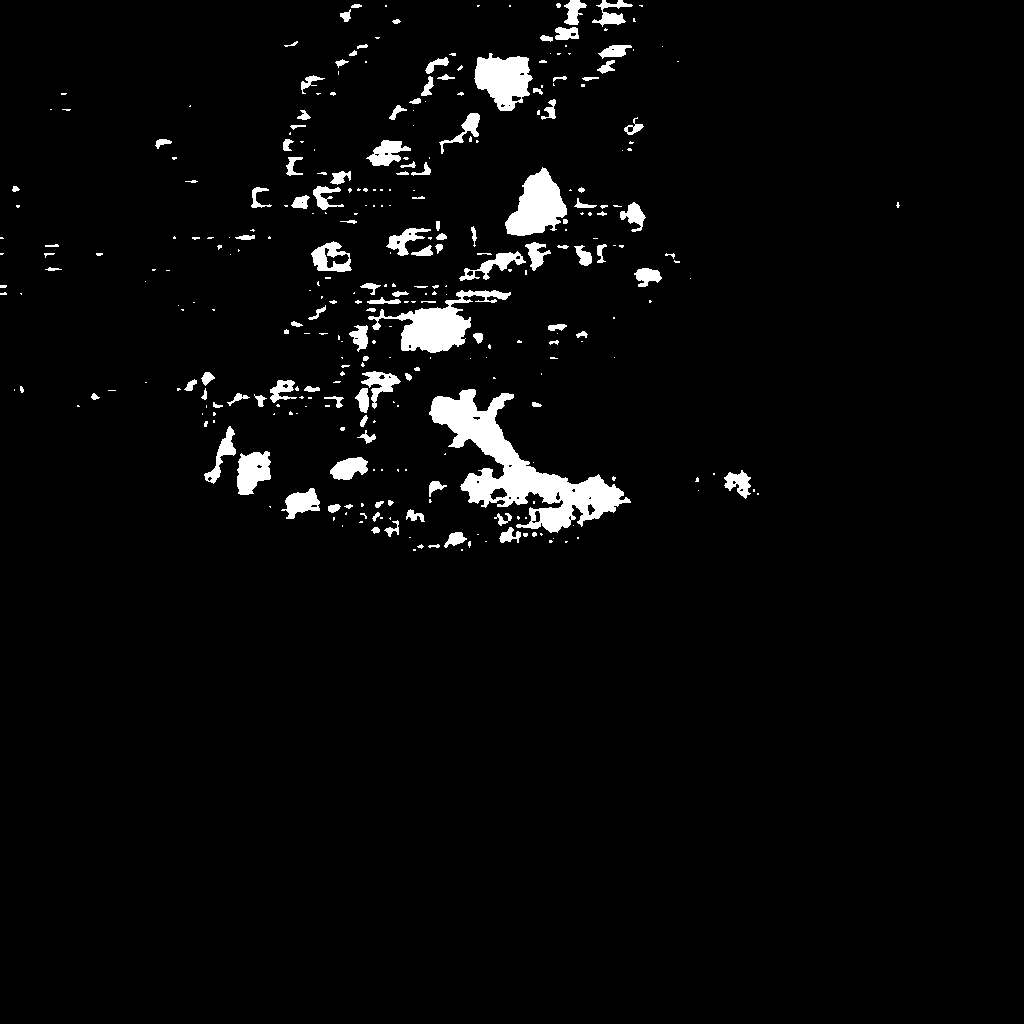}
	\end{minipage}
	\begin{minipage}[b]{0.10\textwidth}
		\centering
		\includegraphics[width=\textwidth]{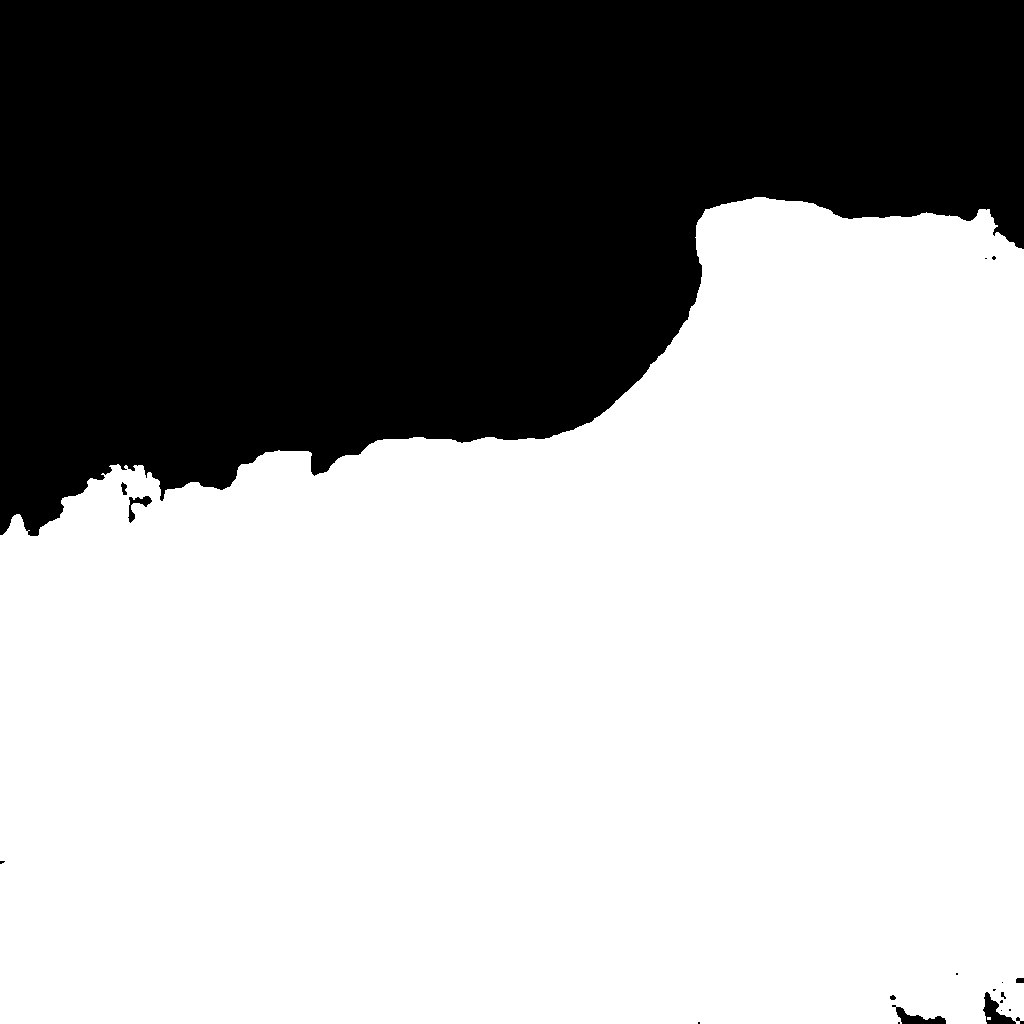}
	\end{minipage}
	\begin{minipage}[b]{0.10\textwidth}
		\centering
		\includegraphics[width=\textwidth]{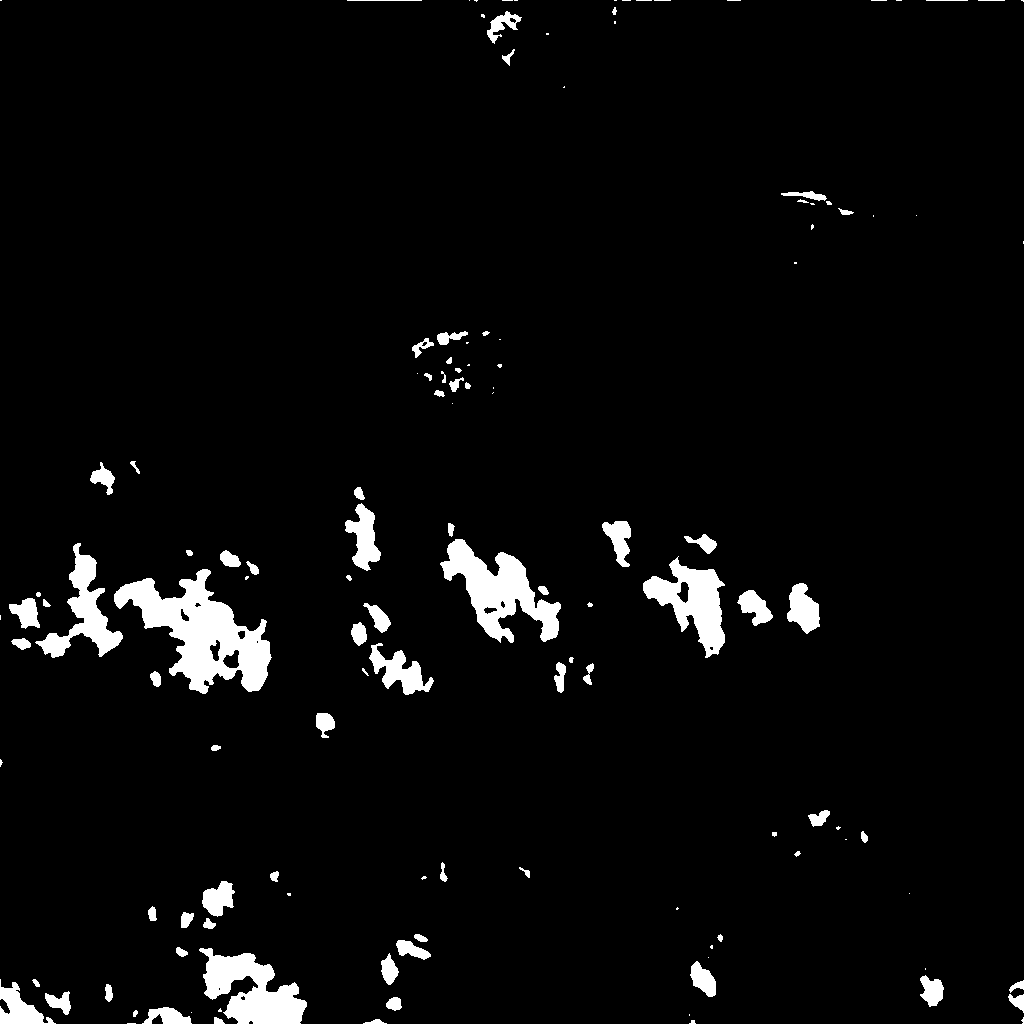}
	\end{minipage}\\[3pt]
	\begin{minipage}[b]{0.10\textwidth}
		\centering
		\includegraphics[width=\textwidth]{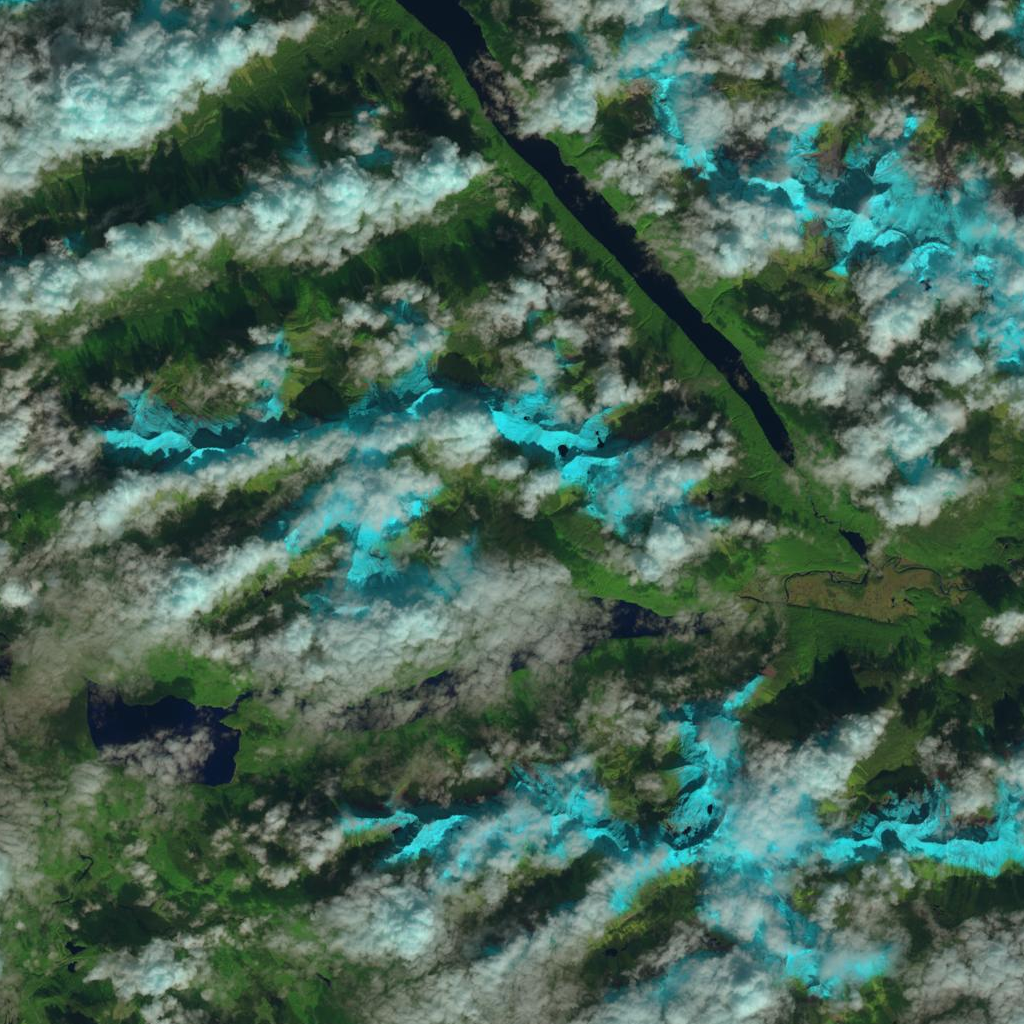}
		\scriptsize{Original Image}
	\end{minipage}
	\begin{minipage}[b]{0.10\textwidth}
		\centering
		\includegraphics[width=\textwidth]{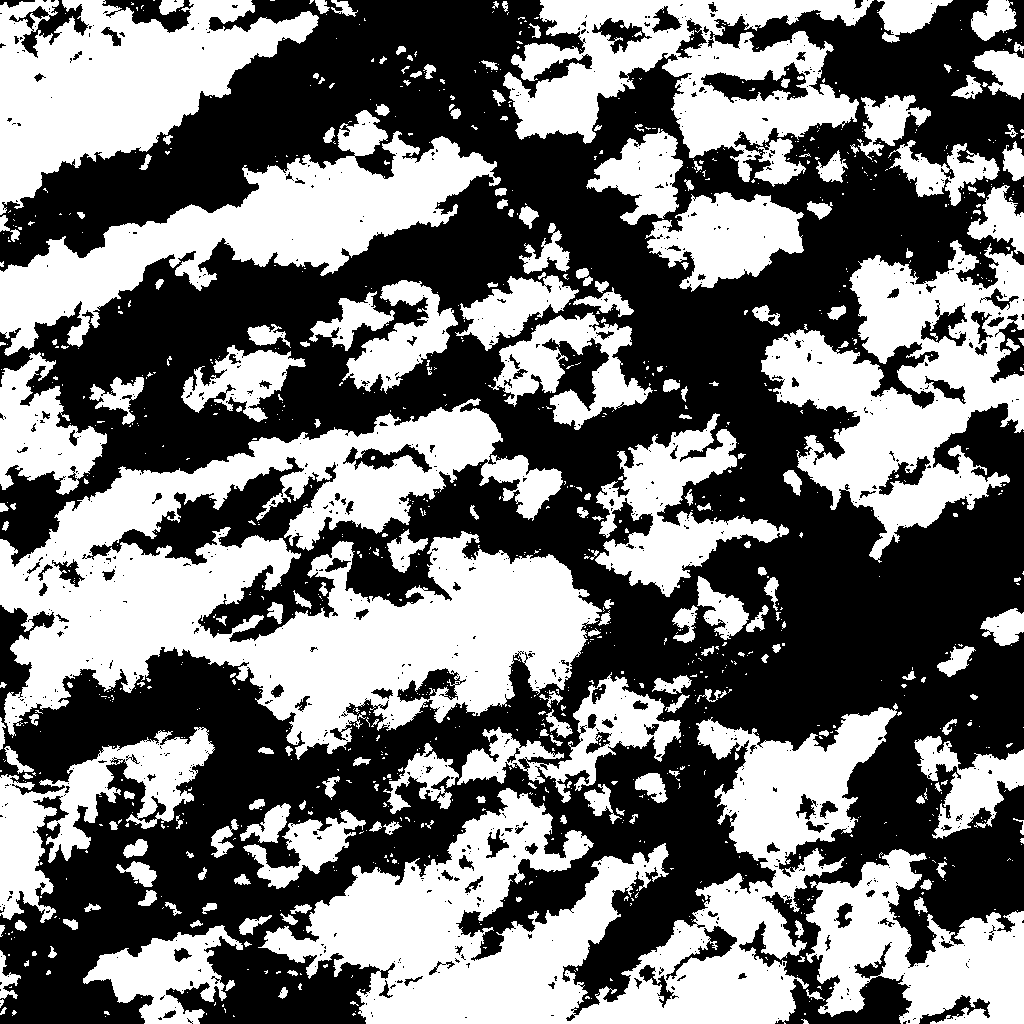}
		\scriptsize{Ground Truth}
	\end{minipage}
	\begin{minipage}[b]{0.10\textwidth}
		\centering
		\includegraphics[width=\textwidth]{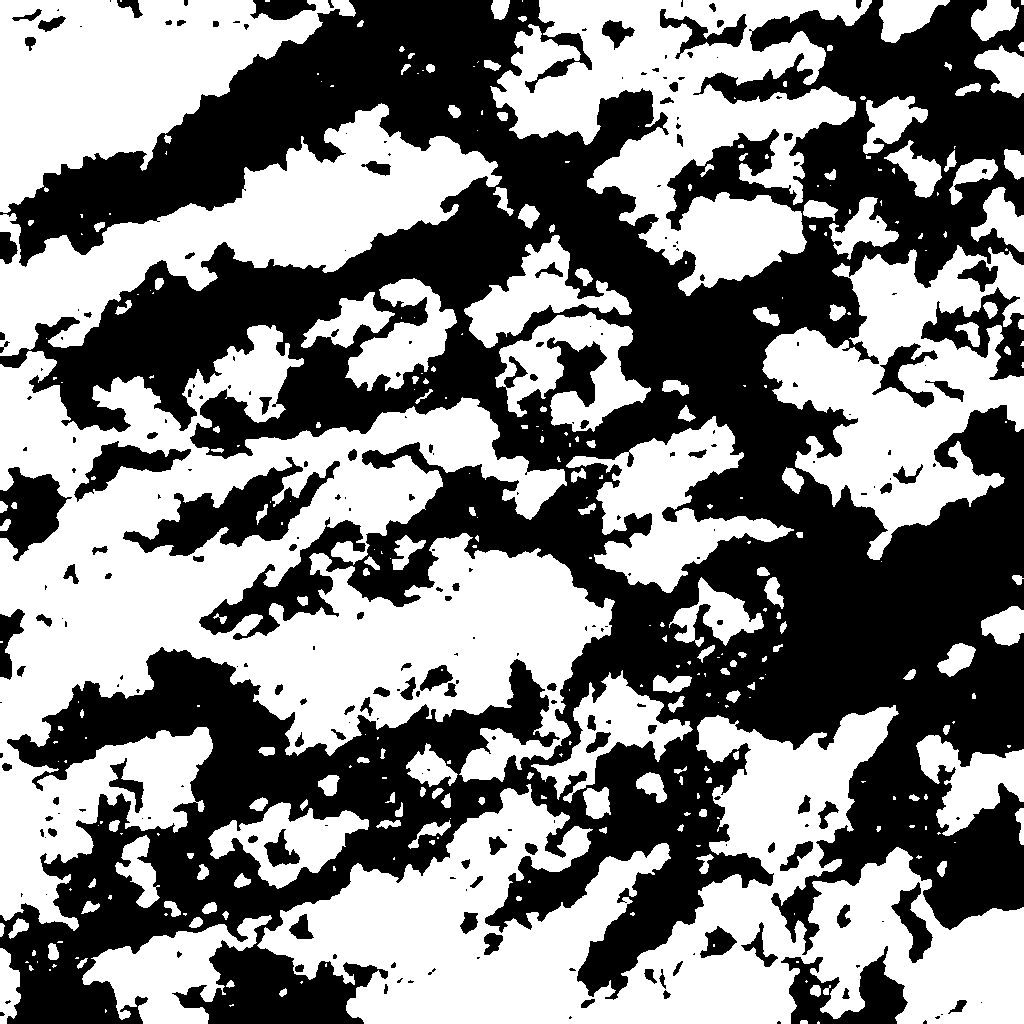}
		\scriptsize{RSAM-Seg}
	\end{minipage}
	\begin{minipage}[b]{0.10\textwidth}
		\centering
		\includegraphics[width=\textwidth]{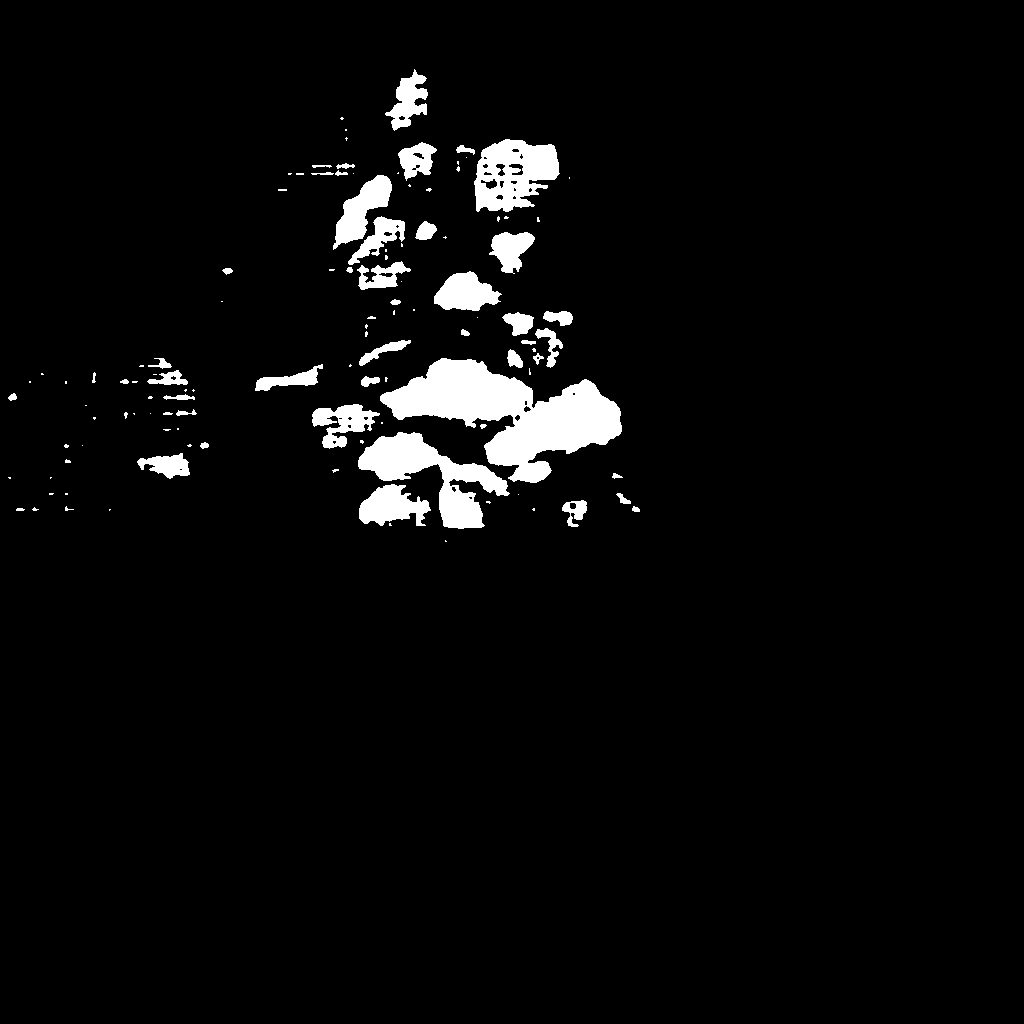}
		\scriptsize{SAM (center -)}
	\end{minipage}
	\begin{minipage}[b]{0.10\textwidth}
		\centering
		\includegraphics[width=\textwidth]{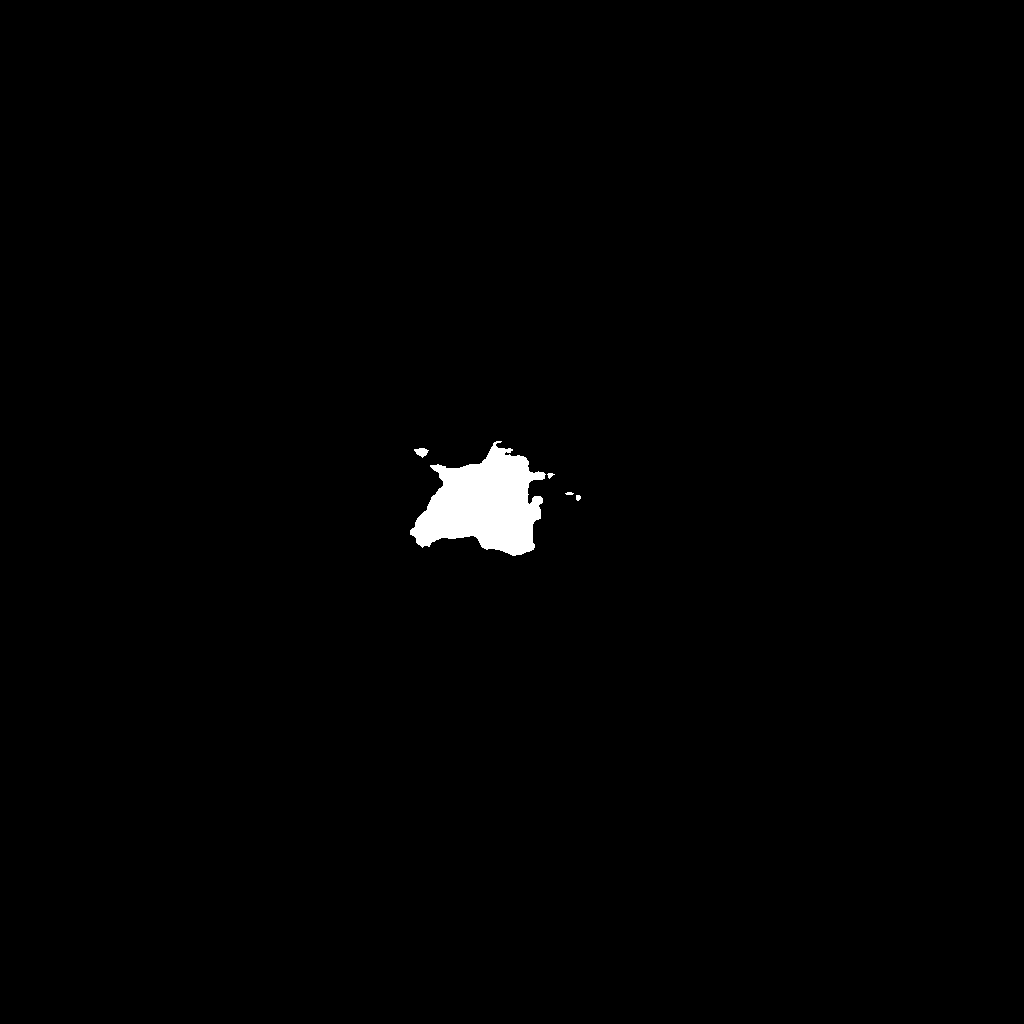}
		\scriptsize{SAM (center +)}
	\end{minipage}
	\begin{minipage}[b]{0.10\textwidth}
		\centering
		\includegraphics[width=\textwidth]{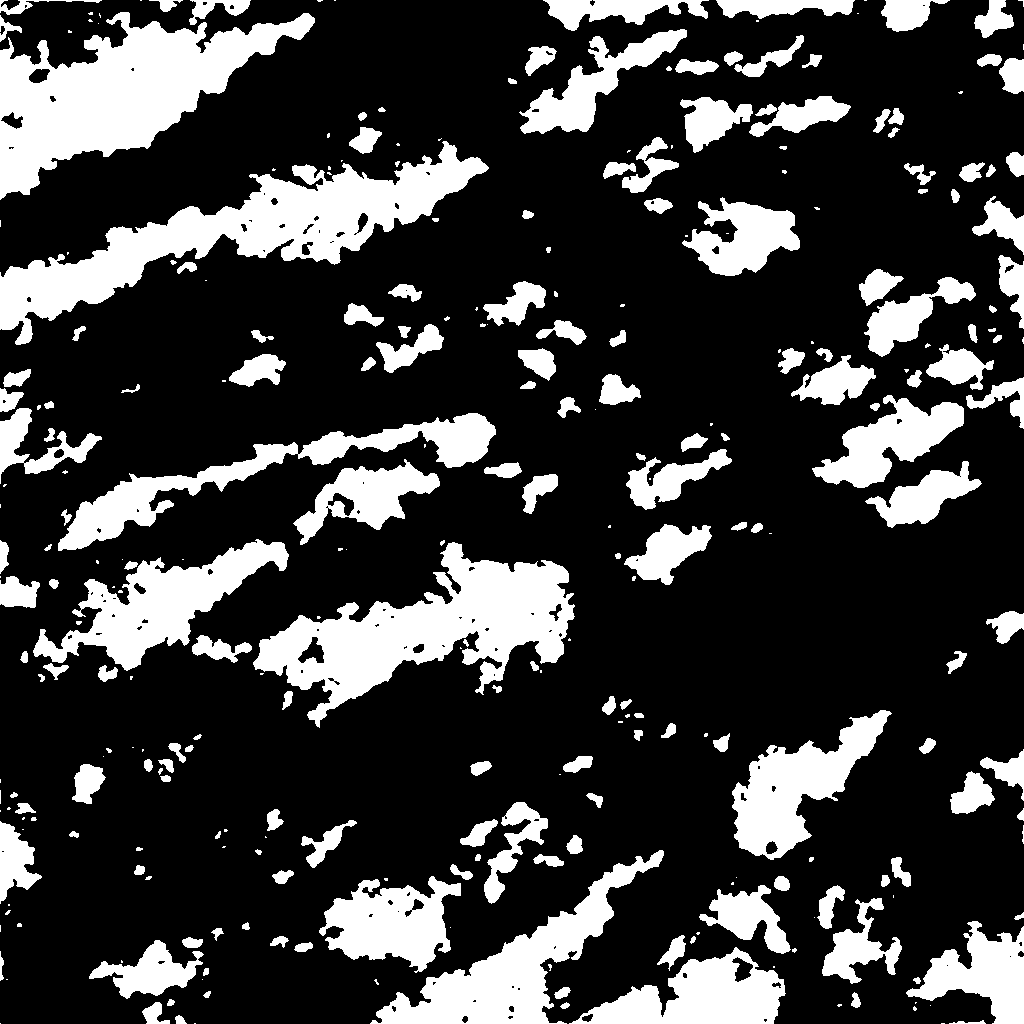}
		\scriptsize{U-Net}
	\end{minipage}
	\caption{Comparison of cloud segmentation results on 38-Cloud dataset with RSAM-Seg, SAM and U-Net.}
	\label{fig:CloudQuality}
	\end{figure*}

	The quantitative results of various methods in the cloud scenario are summarized in Table \ref{tab.RSAMRES} and the visualization results are listed in Figure \ref{fig:CloudQuality}. 
	
	From the "Cloud" scenario of Table \ref{tab.RSAMRES}, RSAM-Seg significantly exceeds the basic SAM in all metrics. Compared to U-Net, RSAM-Seg exhibit superior performance in four comprehensive metrics, Jaccard, F1 score, overall accuracy and mIoU. RSAM-Seg demonstrates an average superiority over SAM under both modes by 36.7\%.
	
	From the images in the first row of Figure \ref{fig:CloudQuality}, it is evident that RSAM-Seg performs well specially when distinguishing thin cloud segments in the bottom-left. In contrast, SAM struggles to accurately identify cloud formations, resulting in large cloud areas being grouped into a single category. U-Net is able to segment thick clouds more accurately, but struggles with the segmentation of thin clouds. This indicates that RSAM-Seg is better suited for handling thin cloud scenarios.
	
	\subsubsection{Results in the field scenario} 
\begin{figure*}[!htb]
	\centering
	\begin{minipage}[b]{0.10\textwidth}
		\centering
		\includegraphics[width=\textwidth]{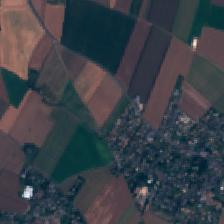}
	\end{minipage}
	\begin{minipage}[b]{0.10\textwidth}
		\centering
		\includegraphics[width=\textwidth]{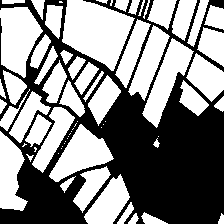}
	\end{minipage}
	\begin{minipage}[b]{0.10\textwidth}
		\centering
		\includegraphics[width=\textwidth]{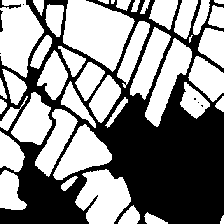}
	\end{minipage}
	\begin{minipage}[b]{0.10\textwidth}
		\centering
		\includegraphics[width=\textwidth]{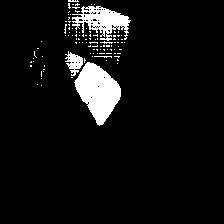}
	\end{minipage}
	\begin{minipage}[b]{0.10\textwidth}
		\centering
		\includegraphics[width=\textwidth]{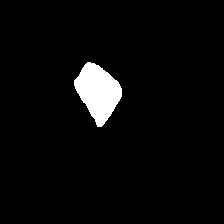}
	\end{minipage}
	\begin{minipage}[b]{0.10\textwidth}
		\centering
		\includegraphics[width=\textwidth]{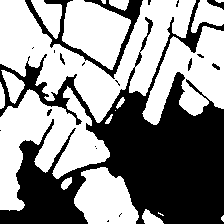}
	\end{minipage}\\[3pt]
	\begin{minipage}[b]{0.10\textwidth}
		\centering
		\includegraphics[width=\textwidth]{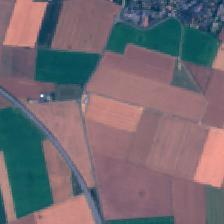}
	\end{minipage}
	\begin{minipage}[b]{0.10\textwidth}
		\centering
		\includegraphics[width=\textwidth]{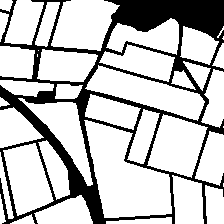}
	\end{minipage}
	\begin{minipage}[b]{0.10\textwidth}
		\centering
		\includegraphics[width=\textwidth]{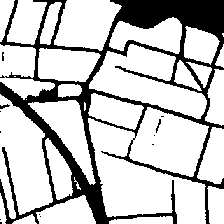}
	\end{minipage}
	\begin{minipage}[b]{0.10\textwidth}
		\centering
		\includegraphics[width=\textwidth]{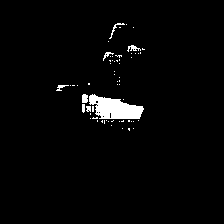}
	\end{minipage}
	\begin{minipage}[b]{0.10\textwidth}
		\centering
		\includegraphics[width=\textwidth]{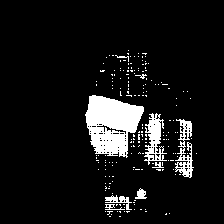}
	\end{minipage}
	\begin{minipage}[b]{0.10\textwidth}
		\centering
		\includegraphics[width=\textwidth]{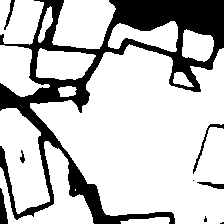}
	\end{minipage}\\[3pt]
	\begin{minipage}[b]{0.10\textwidth}
		\centering
		\includegraphics[width=\textwidth]{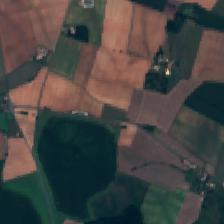}
		\scriptsize{Original Image}
	\end{minipage}
	\begin{minipage}[b]{0.10\textwidth}
		\centering
		\includegraphics[width=\textwidth]{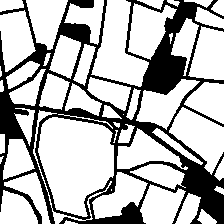}
		\scriptsize{Ground Truth}
	\end{minipage}
	\begin{minipage}[b]{0.10\textwidth}
		\centering
		\includegraphics[width=\textwidth]{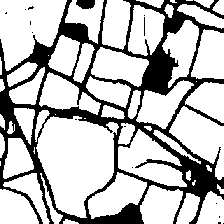}
		\scriptsize{RSAM-Seg}
	\end{minipage}
	\begin{minipage}[b]{0.10\textwidth}
		\centering
		\includegraphics[width=\textwidth]{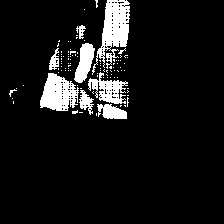}
		\scriptsize{SAM (center -)}
	\end{minipage}
	\begin{minipage}[b]{0.10\textwidth}
		\centering
		\includegraphics[width=\textwidth]{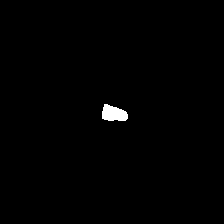}
		\scriptsize{SAM (center +)}
	\end{minipage}
	\begin{minipage}[b]{0.10\textwidth}
		\centering
		\includegraphics[width=\textwidth]{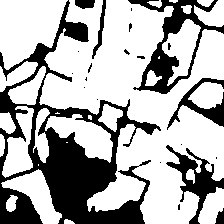}
		\scriptsize{U-Net}
	\end{minipage}
	\caption{Comparison of field segmentation results on Sentinel-2 dataset with RSAM-Seg, SAM and U-Net.}
	\label{fig:FieldQuality}
	\end{figure*}

	The quantitative results in the field scenario are summarized in Table \ref{tab.RSAMRES} and the visualization results are listed in Figure \ref{fig:FieldQuality}. 
	
	The "Field" row of Table \ref{tab.RSAMRES} reveals a enhancement across all metrics compared to the original SAM version. Specially, overall accuracy increased by 28.5\%, and F1 score improved by 56\%. Moreover, RSAM-Seg surpasses the baseline by 18\% and 10\% respectively.
	
	It can be observed in the third row image of Figure \ref{fig:FieldQuality} that RSAM-Seg performs well in distinguishing both regular and irregular field scenes. SAM performs poorly in segmenting agricultural fields in densely populated areas and only identifies the agricultural field surrounding the given point, without taking the overall layout of the fields into account. U-Net struggles to accurately identify roads and other features separating agricultural fields. This suggests that RSAM-Seg is well-suited for handling complex and heterogeneous landscapes.
	
	\subsubsection{Results in the building scenario} 
\begin{figure*}[!htb]
	\centering
	\begin{minipage}[b]{0.10\textwidth}
		\centering
		\includegraphics[width=\textwidth]{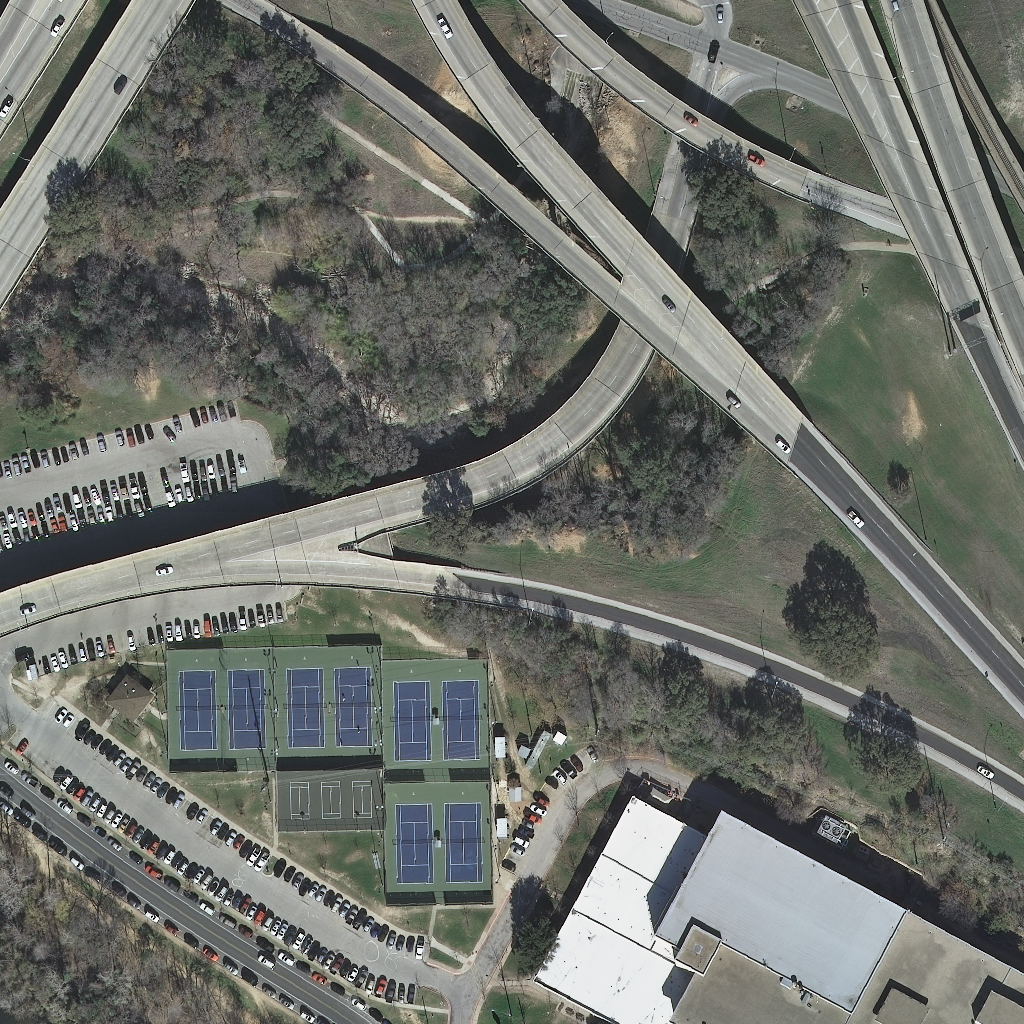}
	\end{minipage}
	\begin{minipage}[b]{0.10\textwidth}
		\centering
		\includegraphics[width=\textwidth]{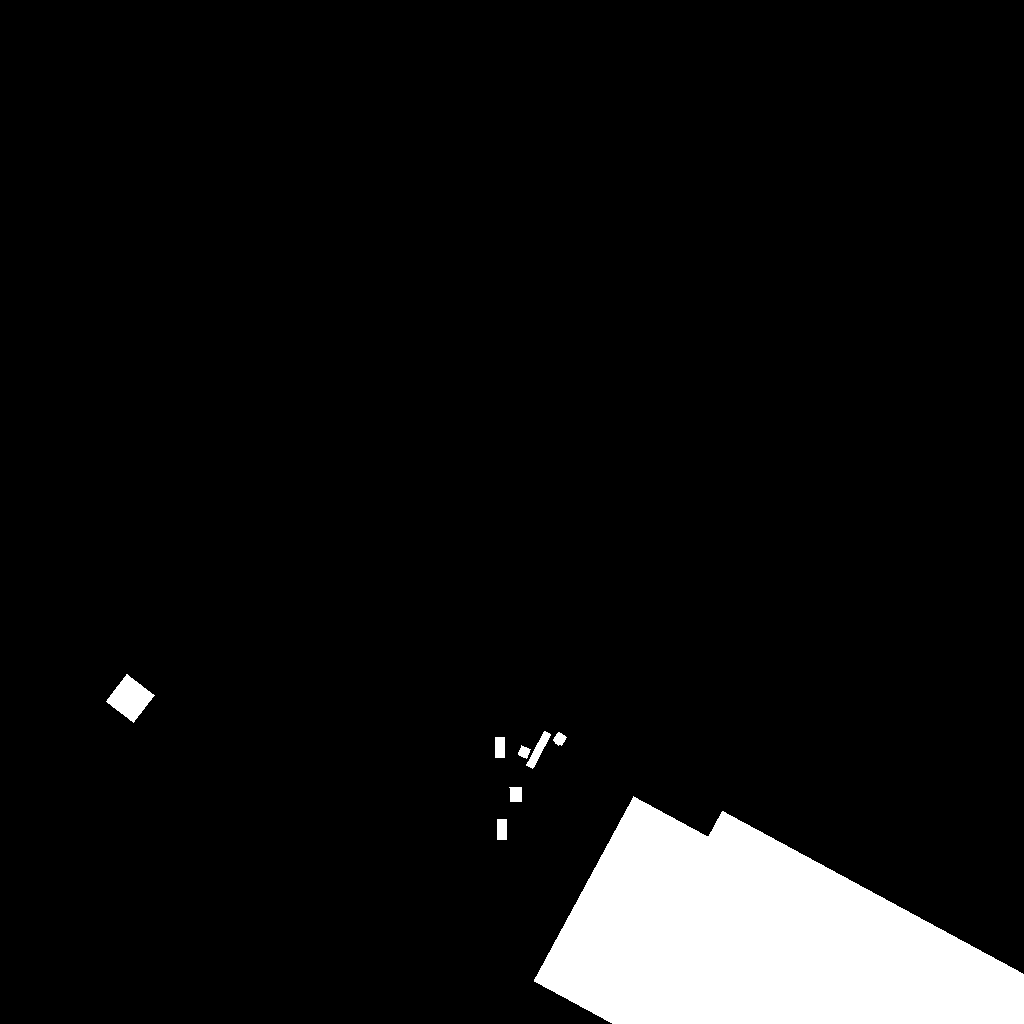}
	\end{minipage}
	\begin{minipage}[b]{0.10\textwidth}
		\centering
		\includegraphics[width=\textwidth]{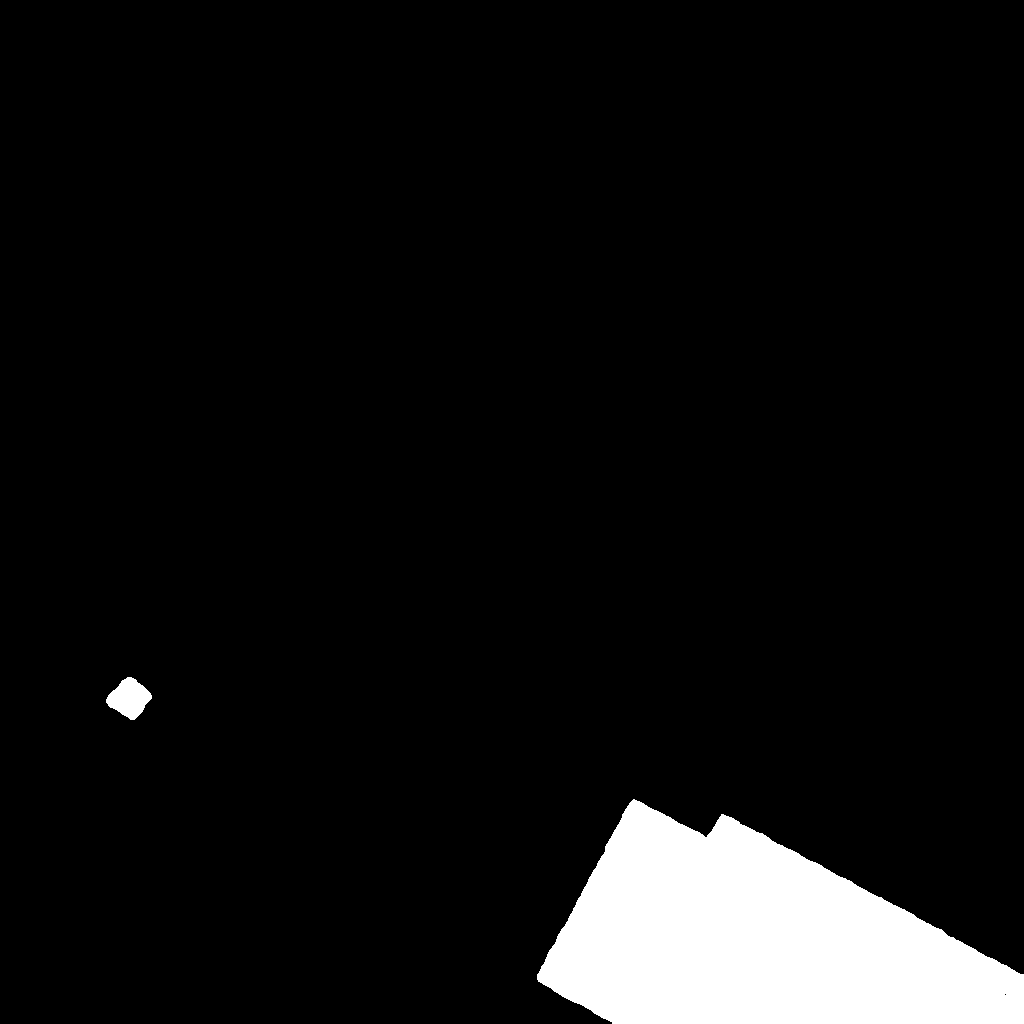}
	\end{minipage}
	\begin{minipage}[b]{0.10\textwidth}
		\centering
		\includegraphics[width=\textwidth]{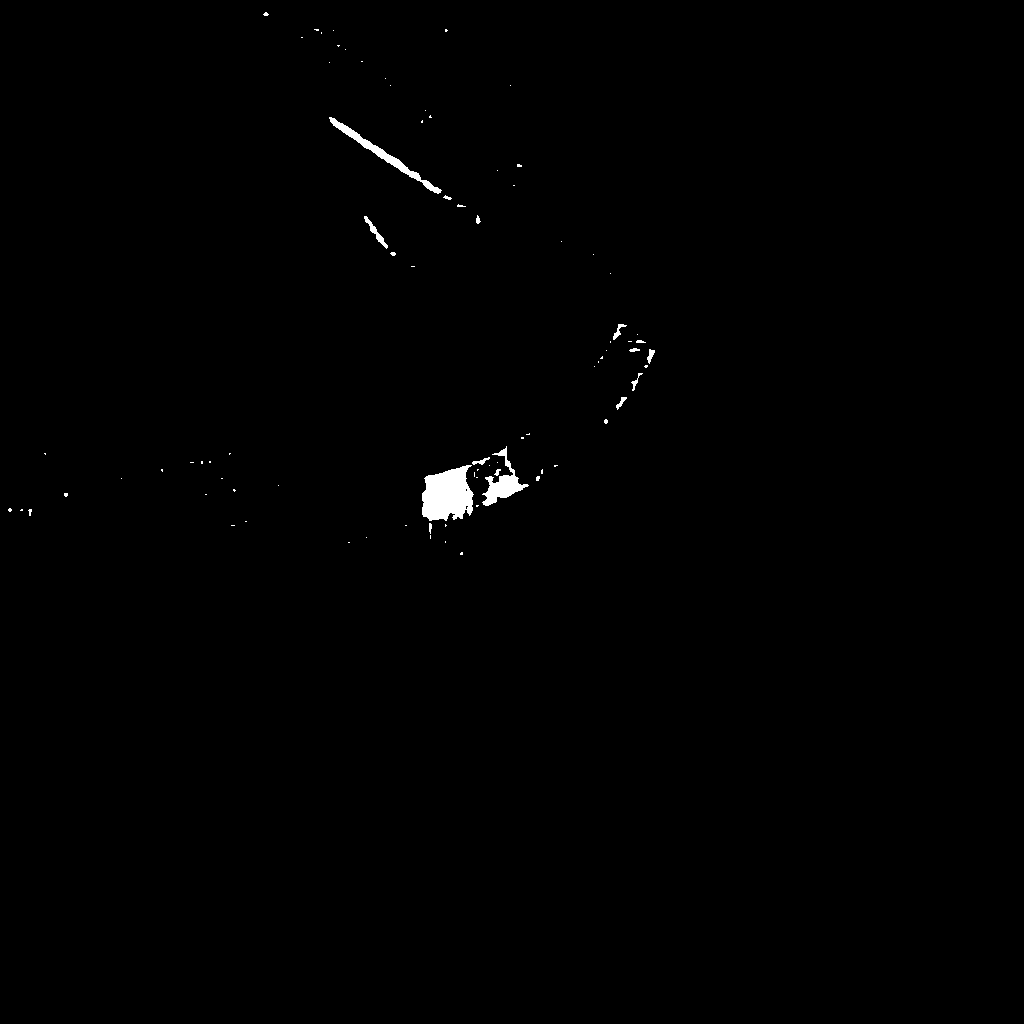}
	\end{minipage}
	\begin{minipage}[b]{0.10\textwidth}
		\centering
		\includegraphics[width=\textwidth]{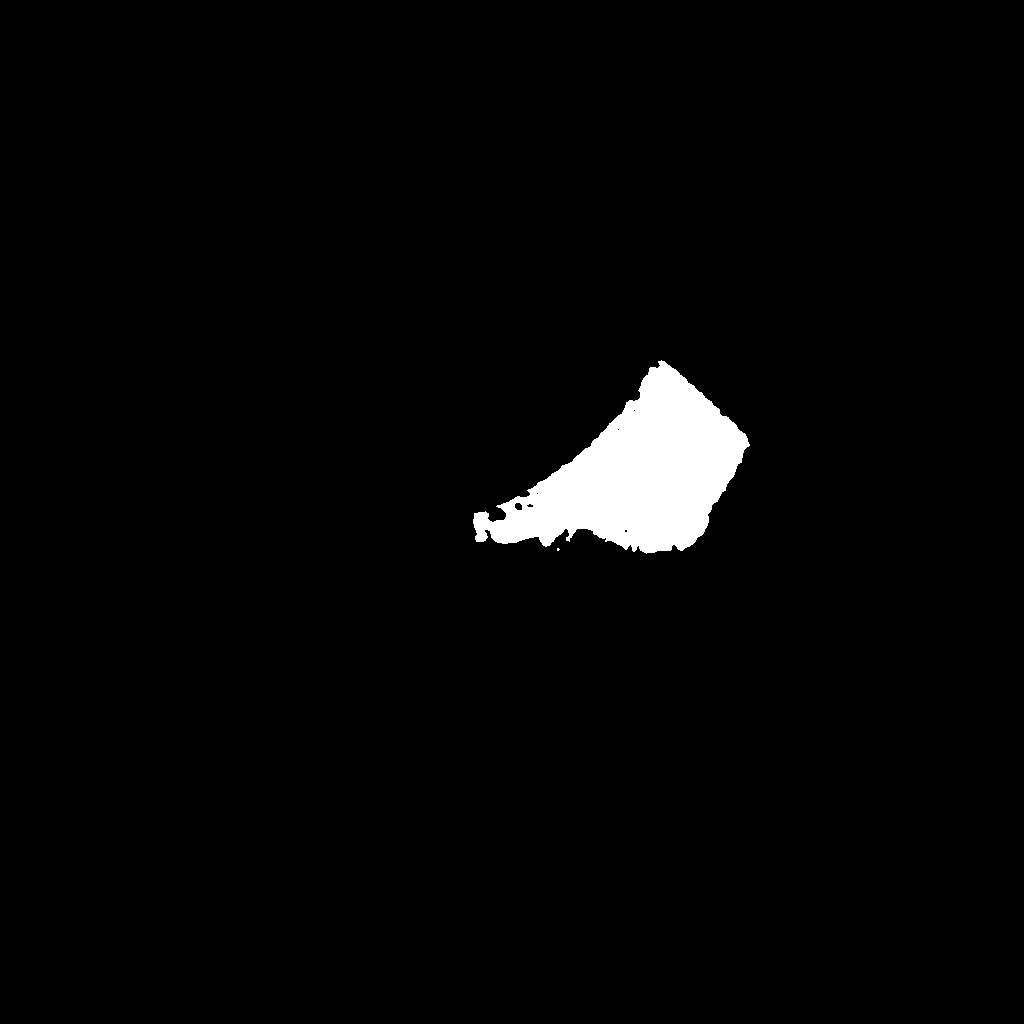}
	\end{minipage}
	\begin{minipage}[b]{0.10\textwidth}
		\centering
		\includegraphics[width=\textwidth]{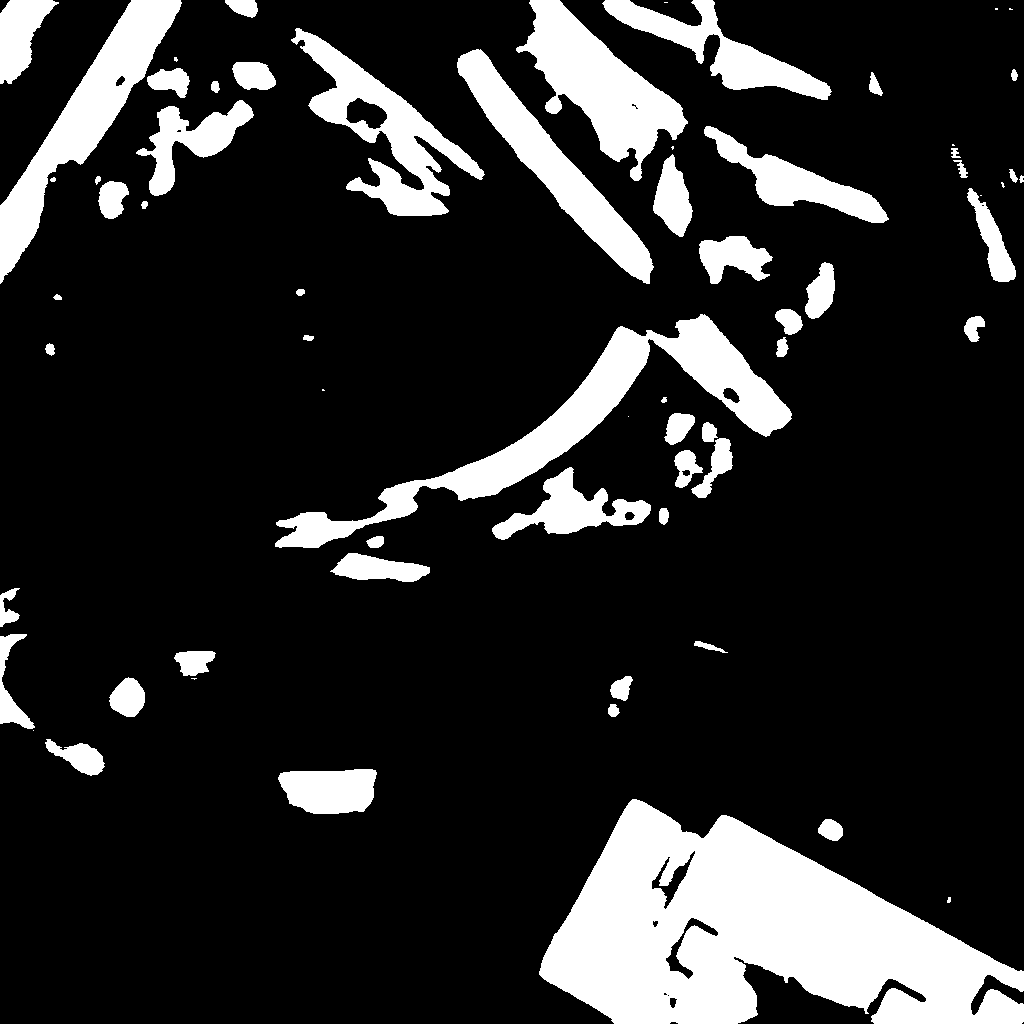}
	\end{minipage}\\[3pt]
	\begin{minipage}[b]{0.10\textwidth}
		\centering
		\includegraphics[width=\textwidth]{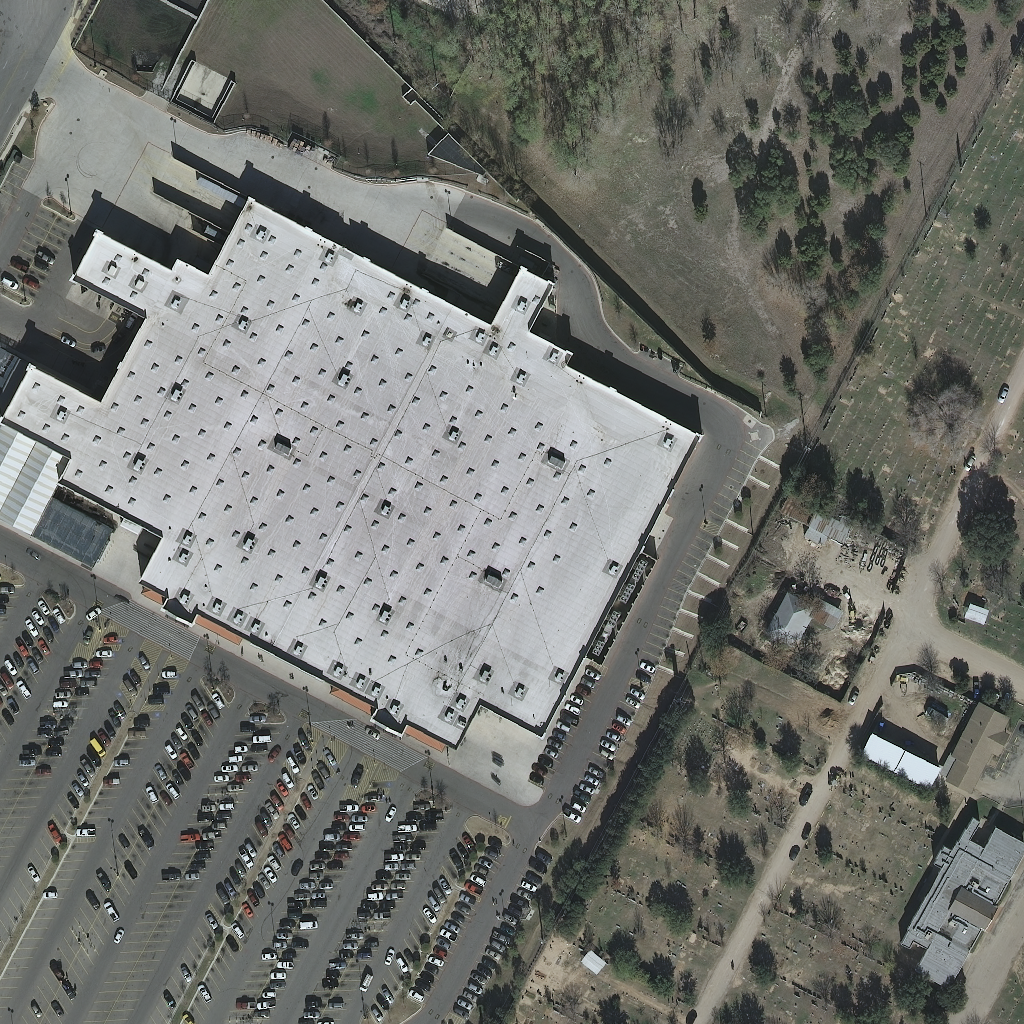}
	\end{minipage}
	\begin{minipage}[b]{0.10\textwidth}
		\centering
		\includegraphics[width=\textwidth]{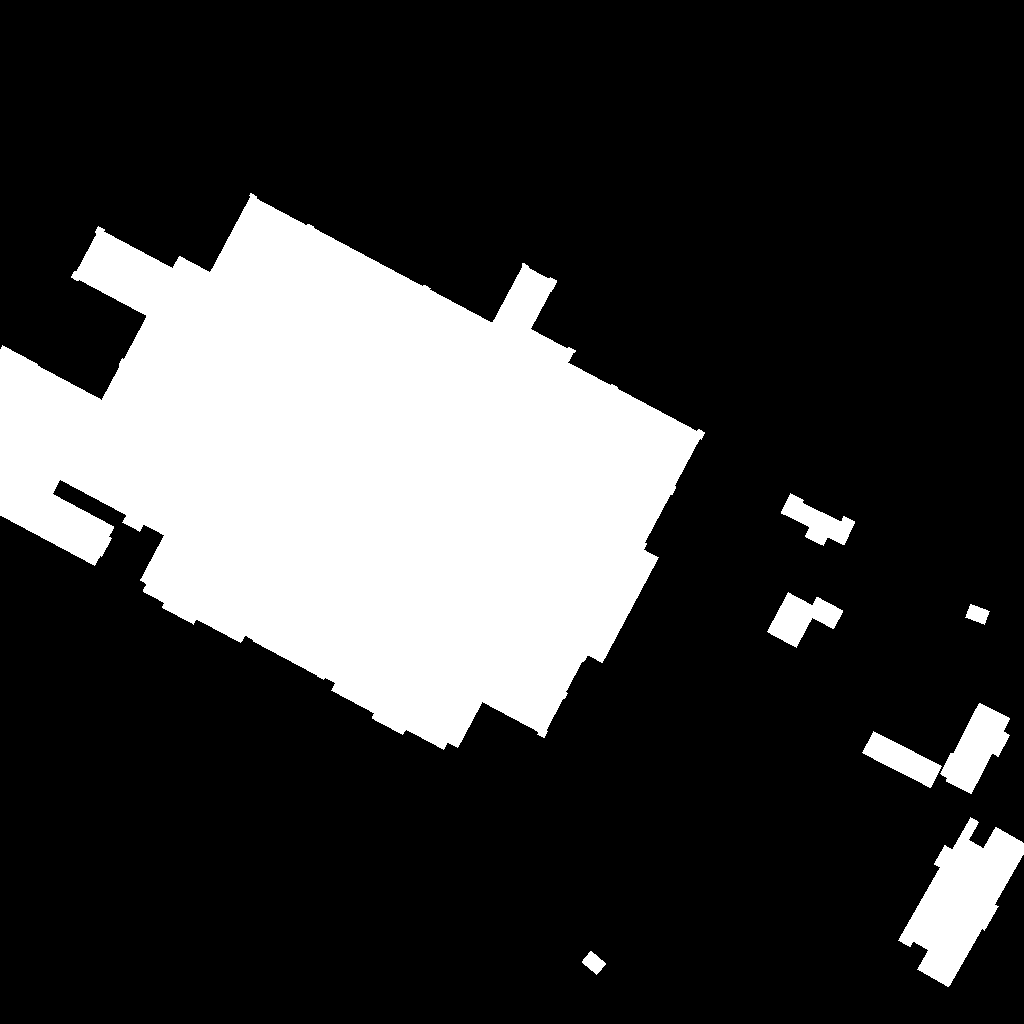}
	\end{minipage}
	\begin{minipage}[b]{0.10\textwidth}
		\centering
		\includegraphics[width=\textwidth]{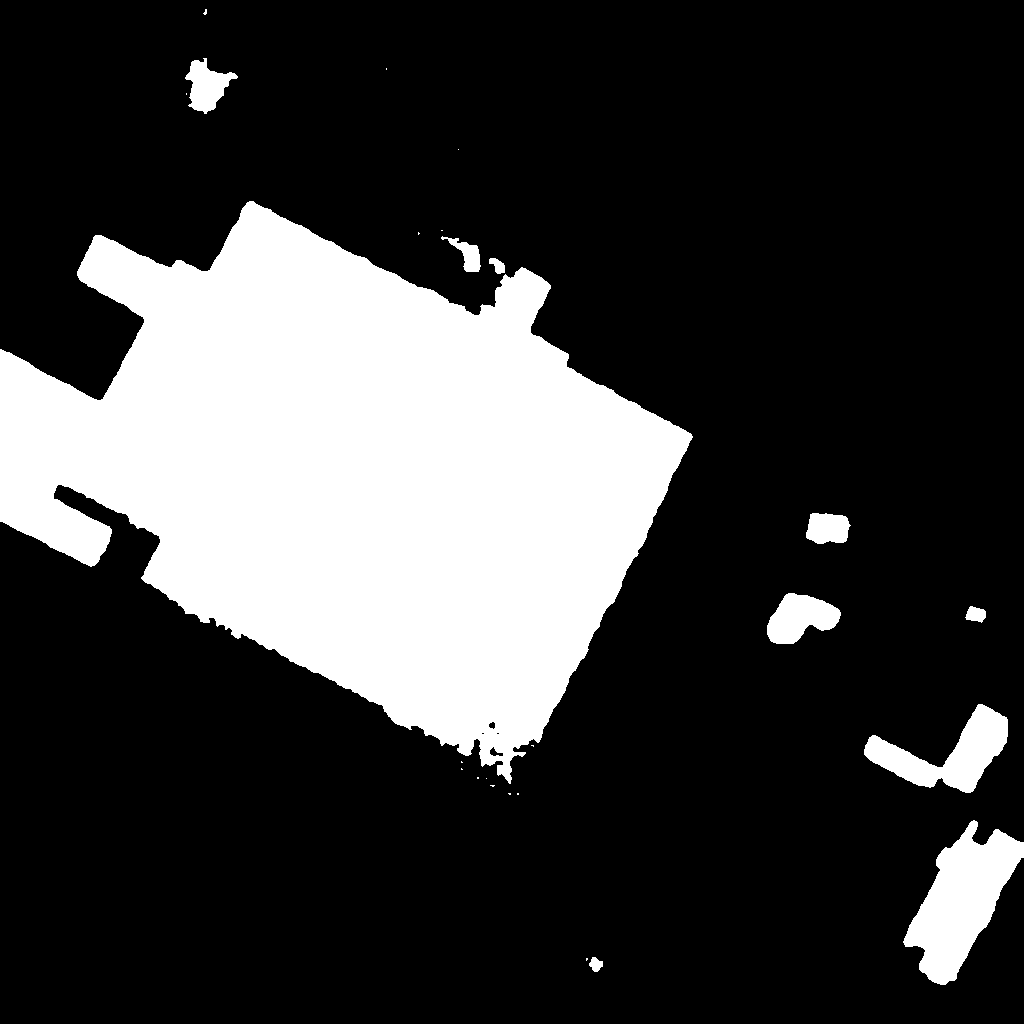}
	\end{minipage}
	\begin{minipage}[b]{0.10\textwidth}
		\centering
		\includegraphics[width=\textwidth]{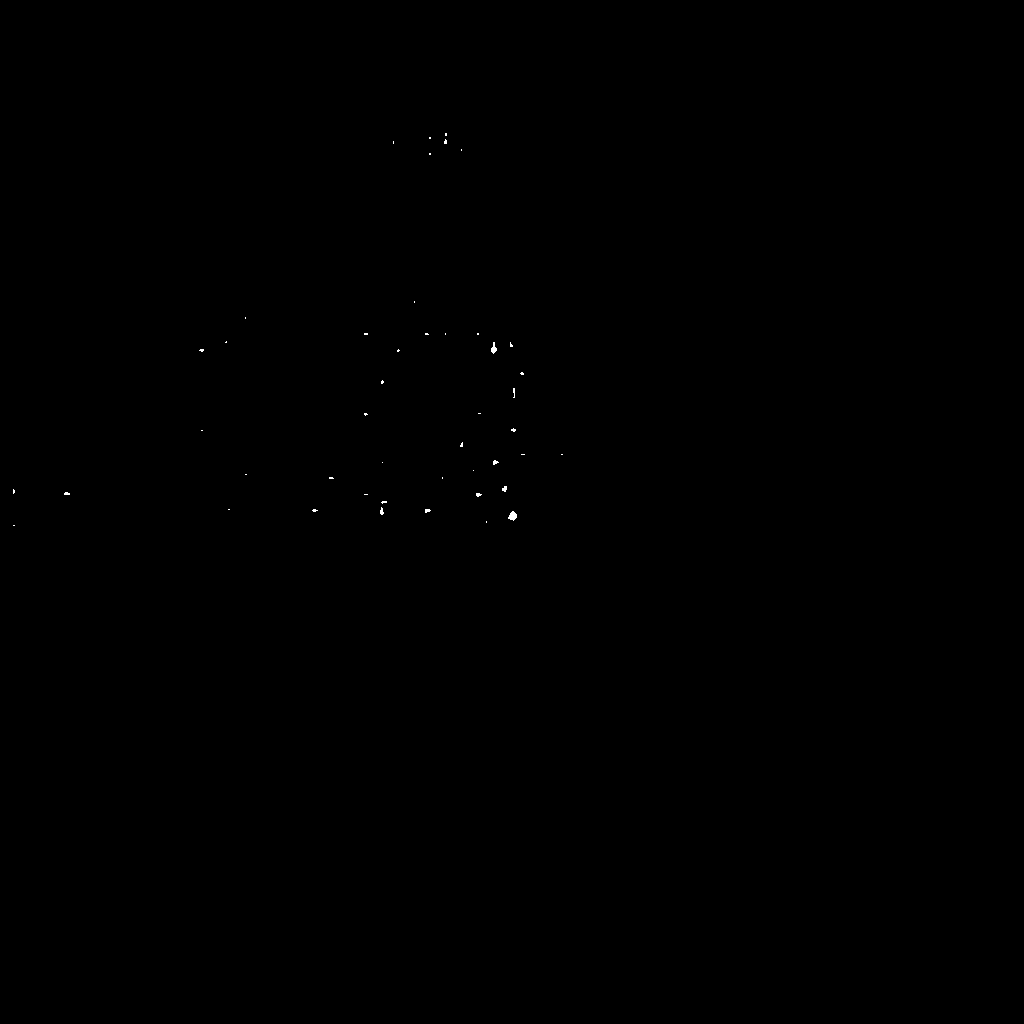}
	\end{minipage}
	\begin{minipage}[b]{0.10\textwidth}
		\centering
		\includegraphics[width=\textwidth]{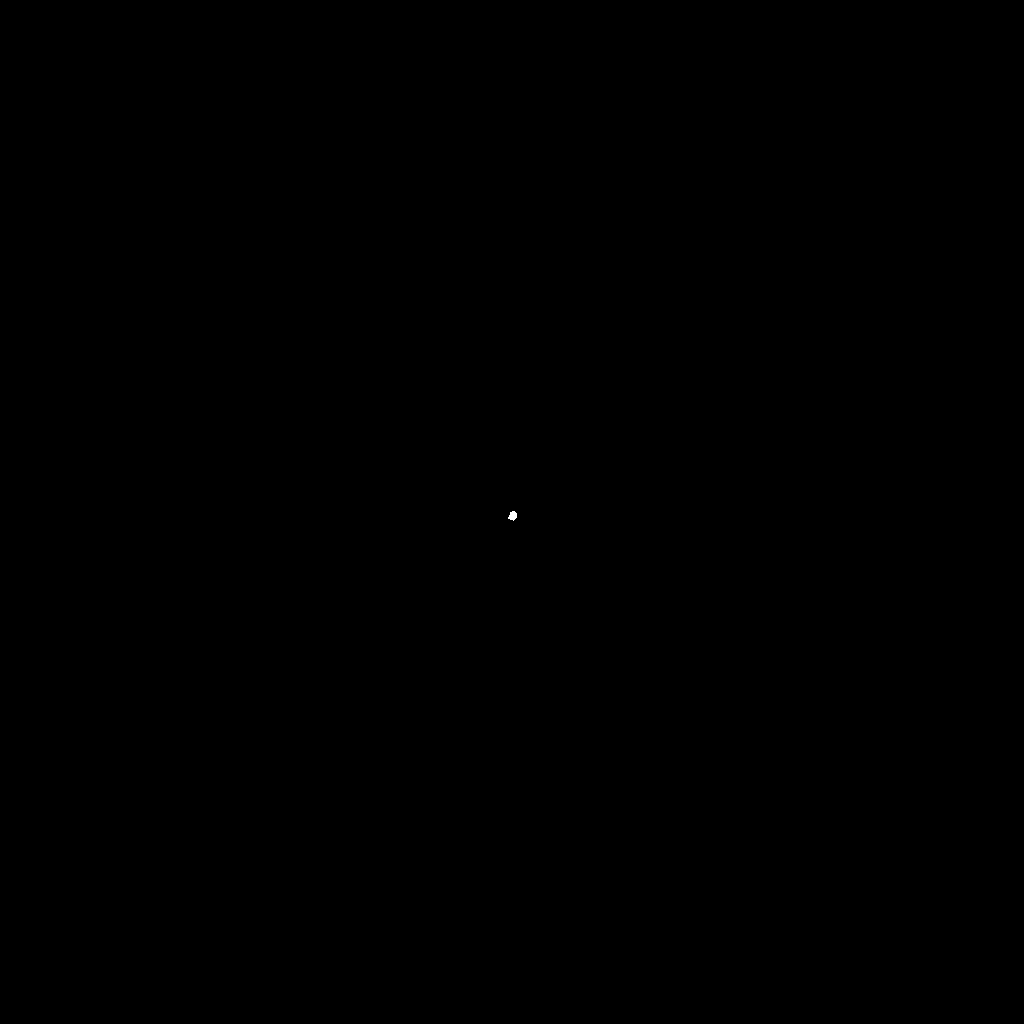}
	\end{minipage}
	\begin{minipage}[b]{0.10\textwidth}
		\centering
		\includegraphics[width=\textwidth]{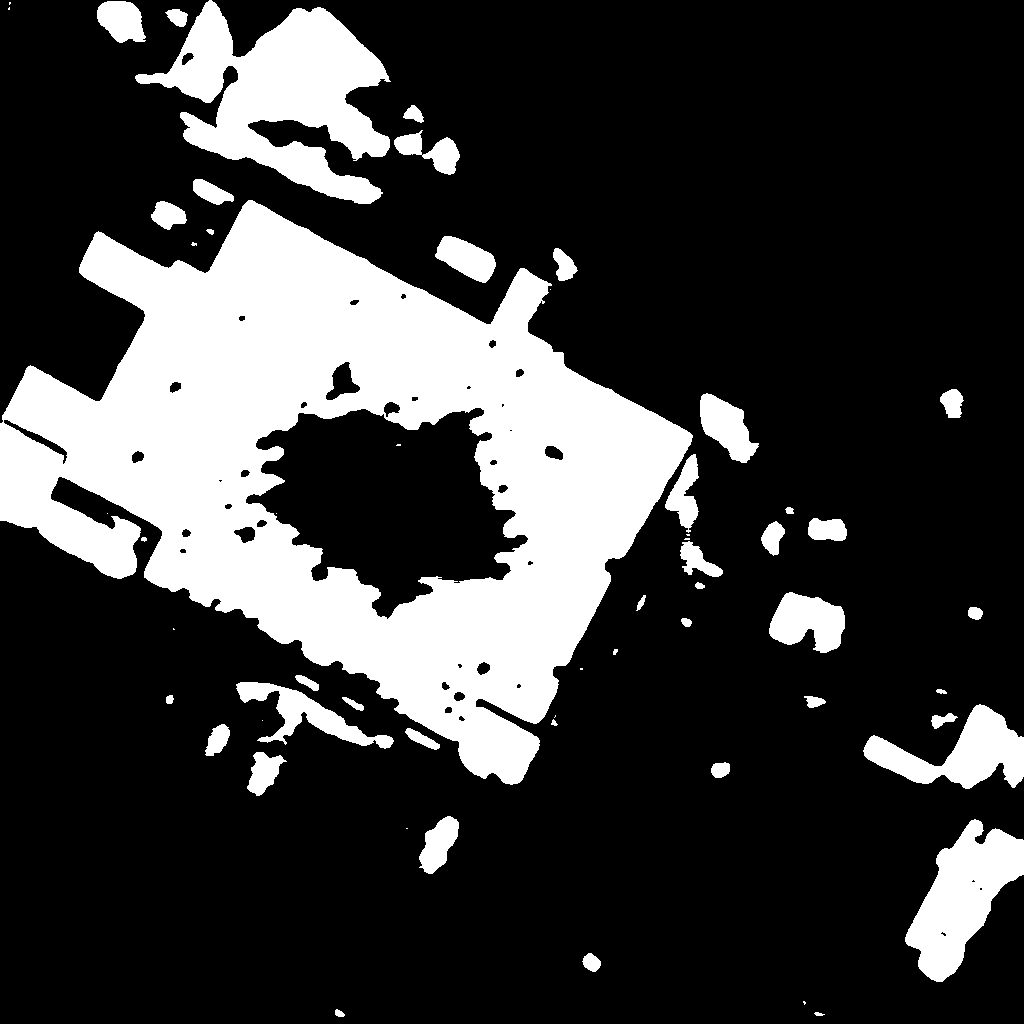}
	\end{minipage}\\[3pt]
	\begin{minipage}[b]{0.10\textwidth}
		\centering
		\includegraphics[width=\textwidth]{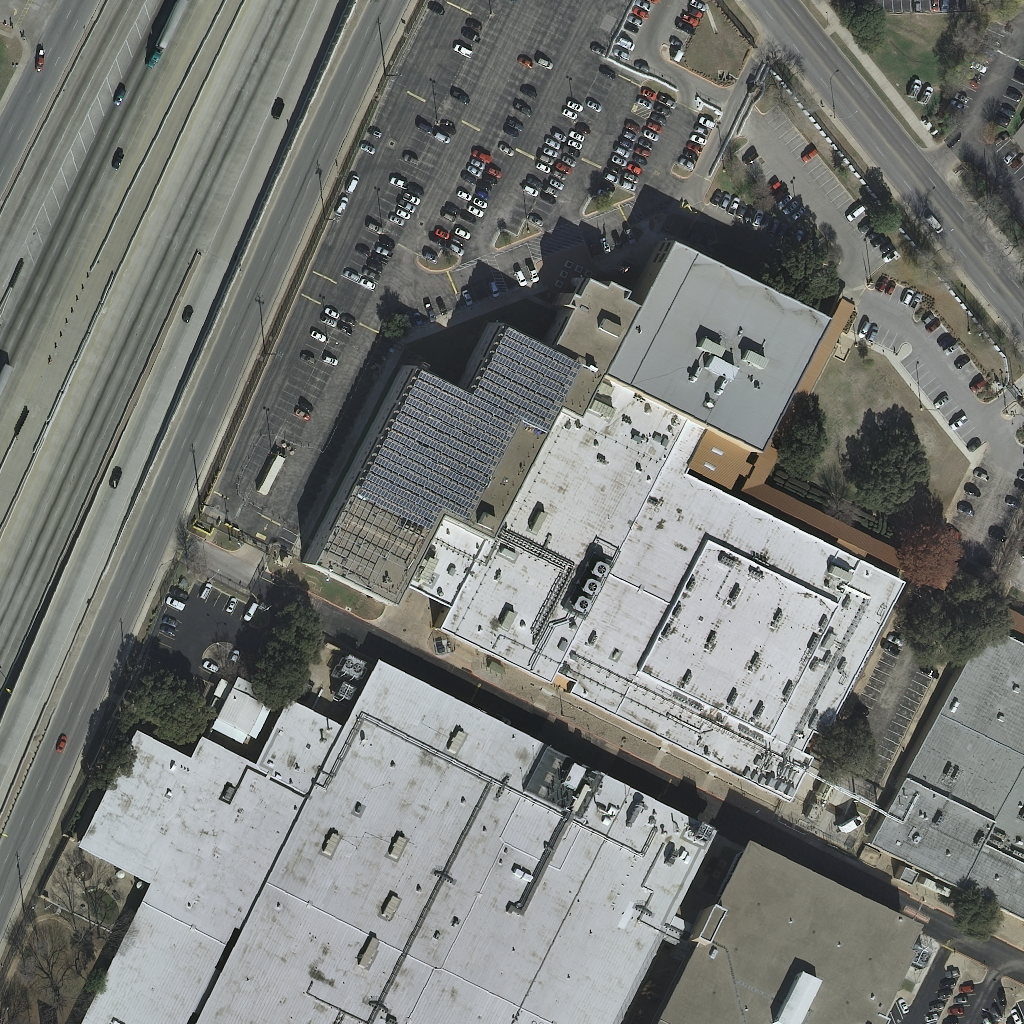}
		\scriptsize{Original Image}
	\end{minipage}
	\begin{minipage}[b]{0.10\textwidth}
		\centering
		\includegraphics[width=\textwidth]{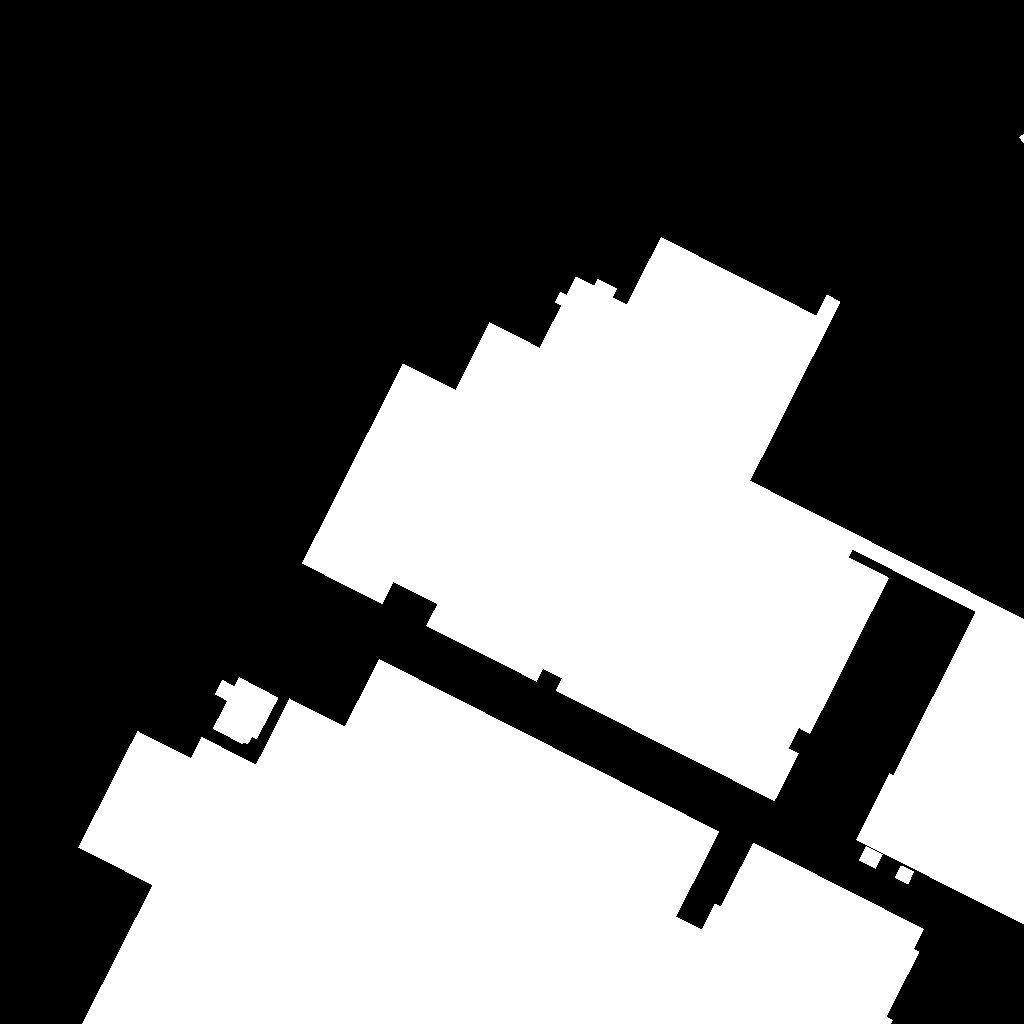}
		\scriptsize{Ground Truth}
	\end{minipage}
	\begin{minipage}[b]{0.10\textwidth}
		\centering
		\includegraphics[width=\textwidth]{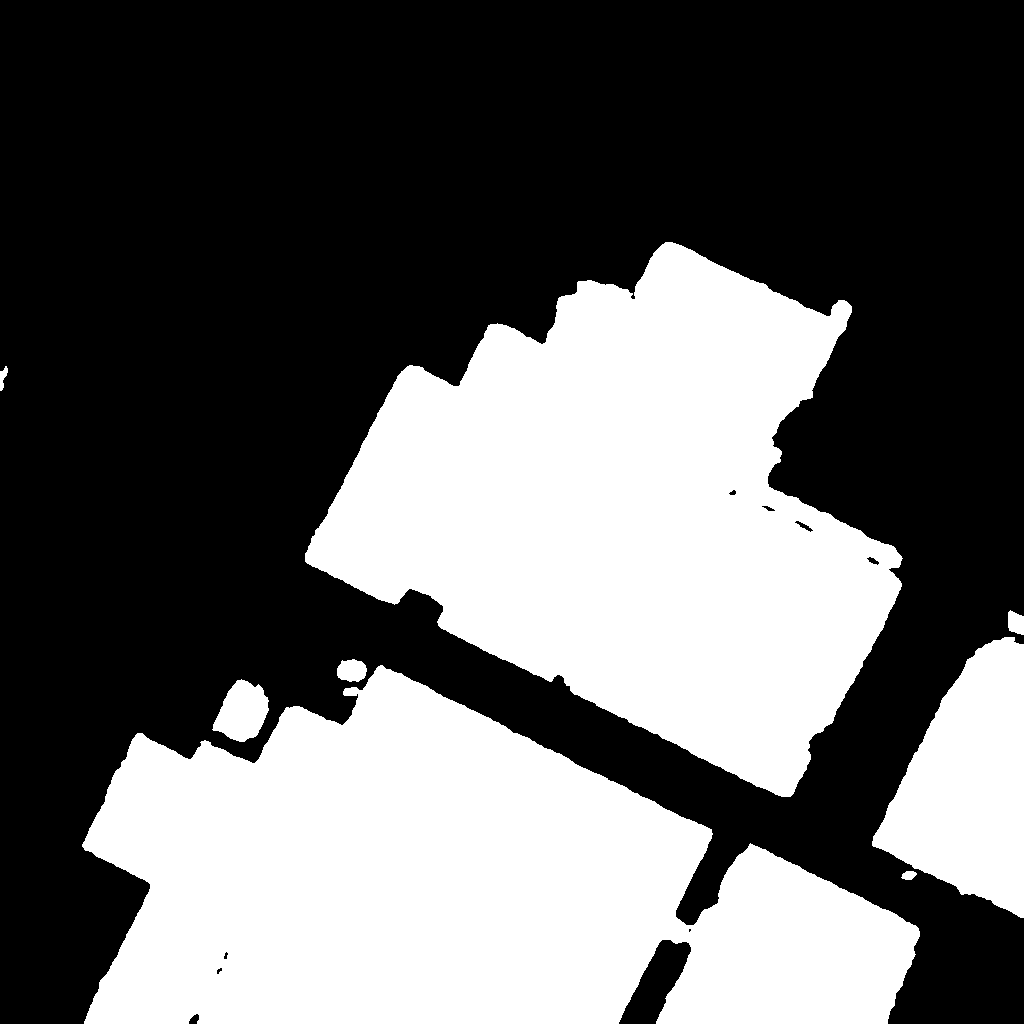}
		\scriptsize{RSAM-Seg}
	\end{minipage}
	\begin{minipage}[b]{0.10\textwidth}
		\centering
		\includegraphics[width=\textwidth]{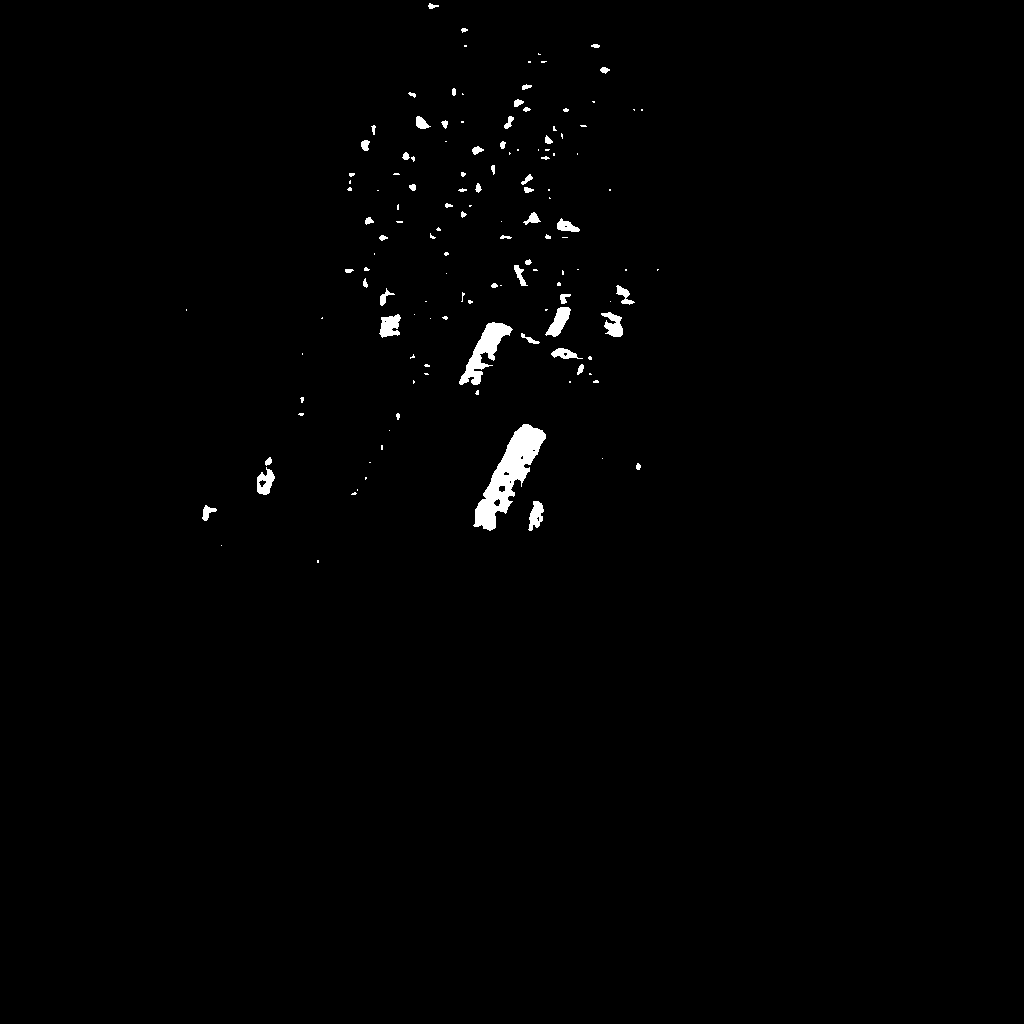}
		\scriptsize{SAM (center -)}
	\end{minipage}
	\begin{minipage}[b]{0.10\textwidth}
		\centering
		\includegraphics[width=\textwidth]{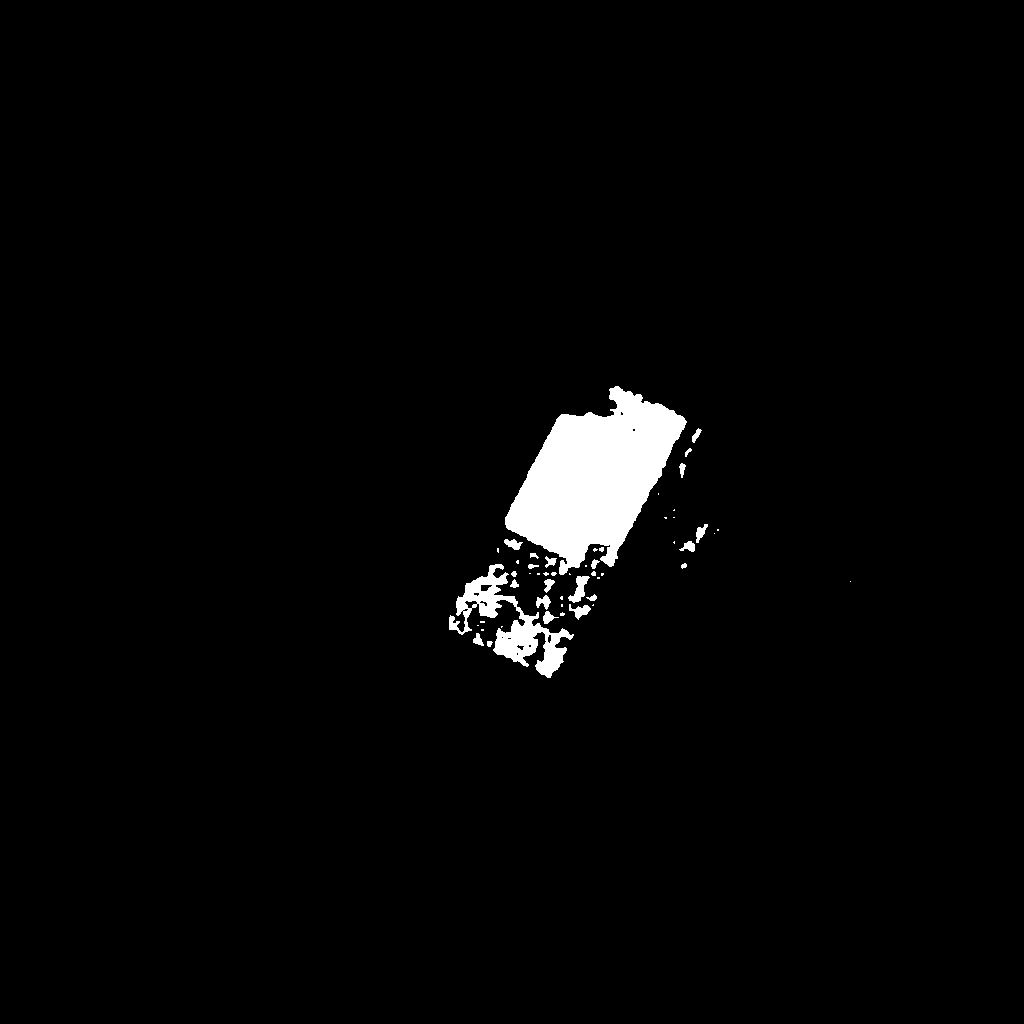}
		\scriptsize{SAM (center +)}
	\end{minipage}
	\begin{minipage}[b]{0.10\textwidth}
		\centering
		\includegraphics[width=\textwidth]{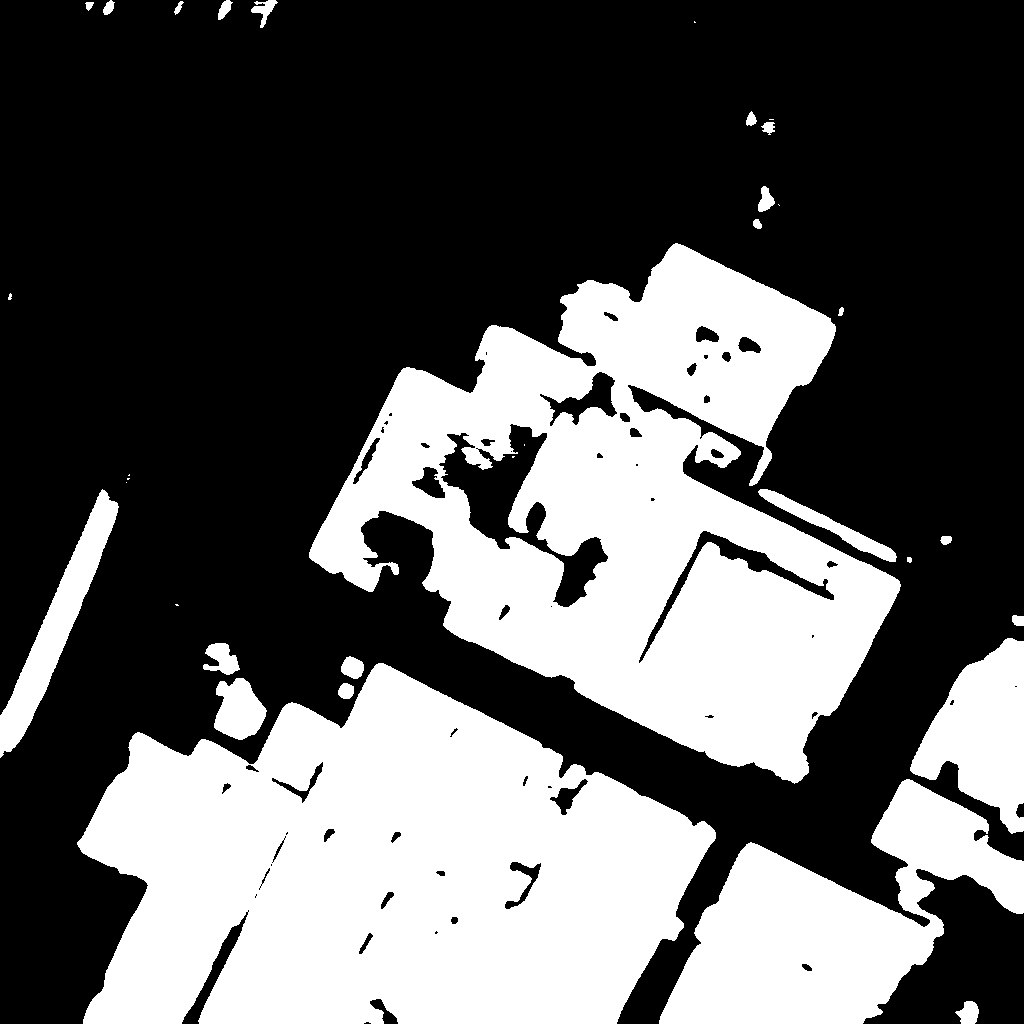}
		\scriptsize{U-Net}
	\end{minipage}
	\caption{Comparison of building segmentation results on Inria dataset with RSAM-Seg, SAM and U-Net.}
	\label{fig:inriaQuality}
	\end{figure*}

	The quantitative results in the building scenario are summarized in Table \ref{tab.RSAMRES} and the visualization results are listed in Figure \ref{fig:inriaQuality}. 
	
	Examining the "Building" row in Table \ref{tab.RSAMRES}, it's observable that the results surpass SAM across multiple evaluation metrics, while slightly exceeding the baseline by 5\% and achieves a substantial average accuracy improvement of 42.71\% under both operational modes of SAM in overall accuracy.
	
	Upon scrutinizing the images in the central row of Figure \ref{fig:inriaQuality}, RSAM-Seg accurately distinguishes both the overall structures and scattered buildings. Furthermore, from the top-left images, RSAM-Seg effectively avoids interference from similar elevated structures within the scene. In contrast, SAM is limited by its dependence on prompts, resulting in only segmenting the area around the point prompt. Meanwhile, U-Net struggles with the segmentation when facing highway structures, leading to misclassification in certain scenarios. RSAM-Seg performs well in complex urban environments, highlighting its potential as a valuable tool for urban planning and management.
	
		\begin{figure*}[!htb]
		\centering
		\begin{minipage}[b]{0.10\textwidth}
			\centering
			\includegraphics[width=\textwidth]{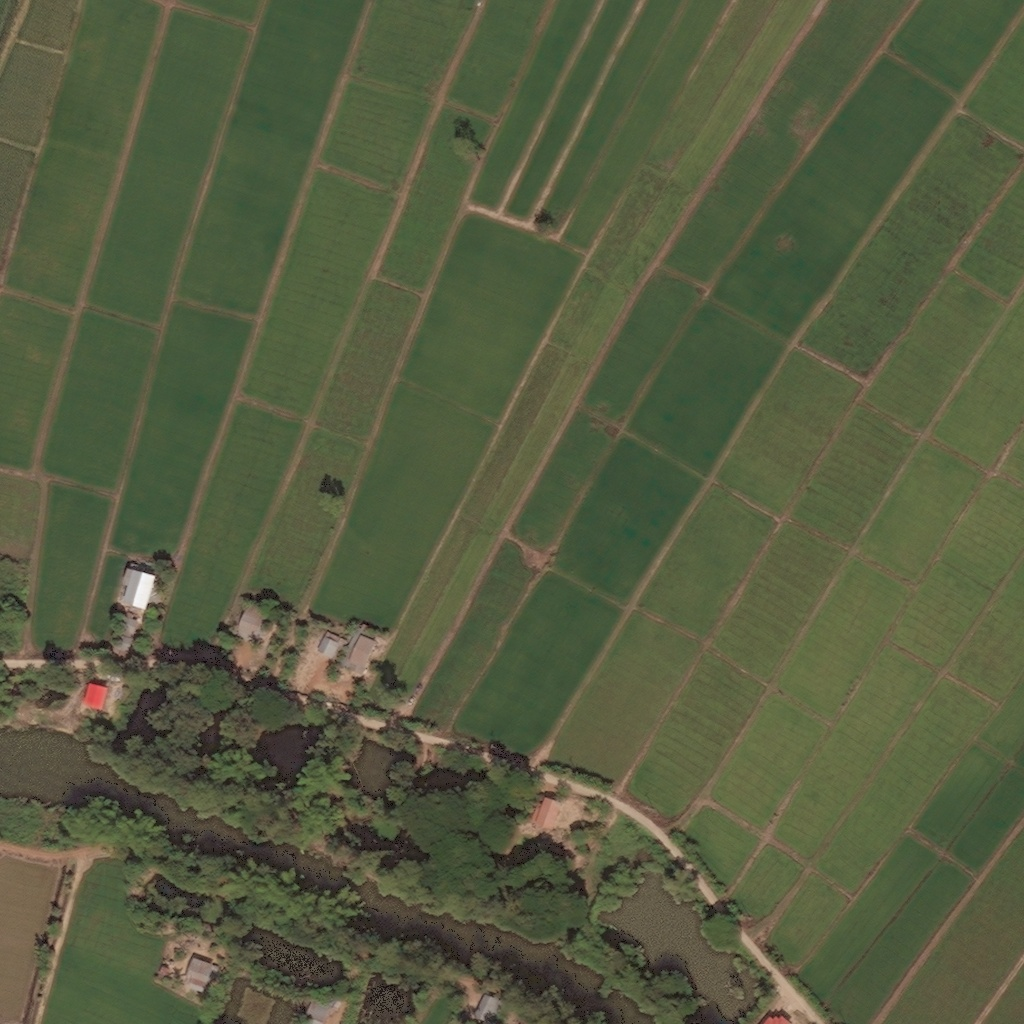}
		\end{minipage}
		\begin{minipage}[b]{0.10\textwidth}
			\centering
			\includegraphics[width=\textwidth]{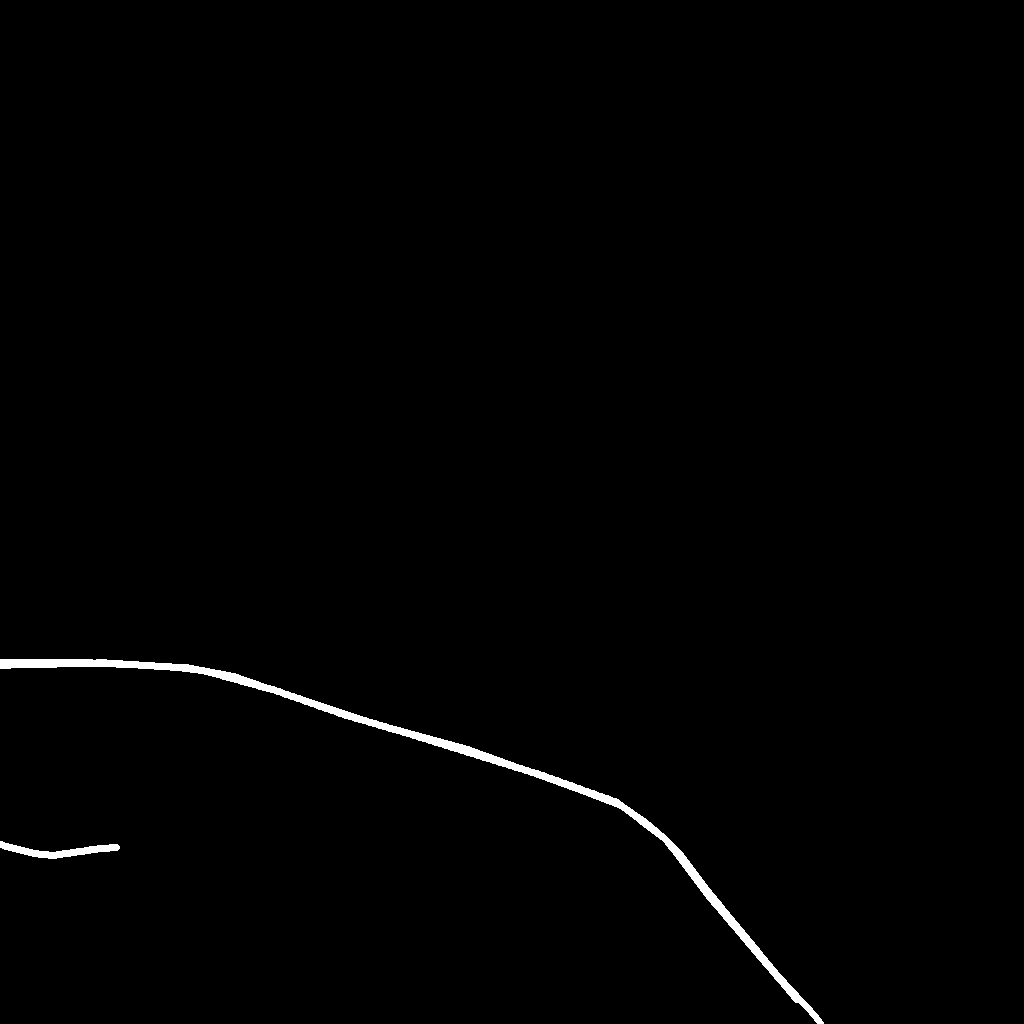}
		\end{minipage}
		\begin{minipage}[b]{0.10\textwidth}
			\centering
			\includegraphics[width=\textwidth]{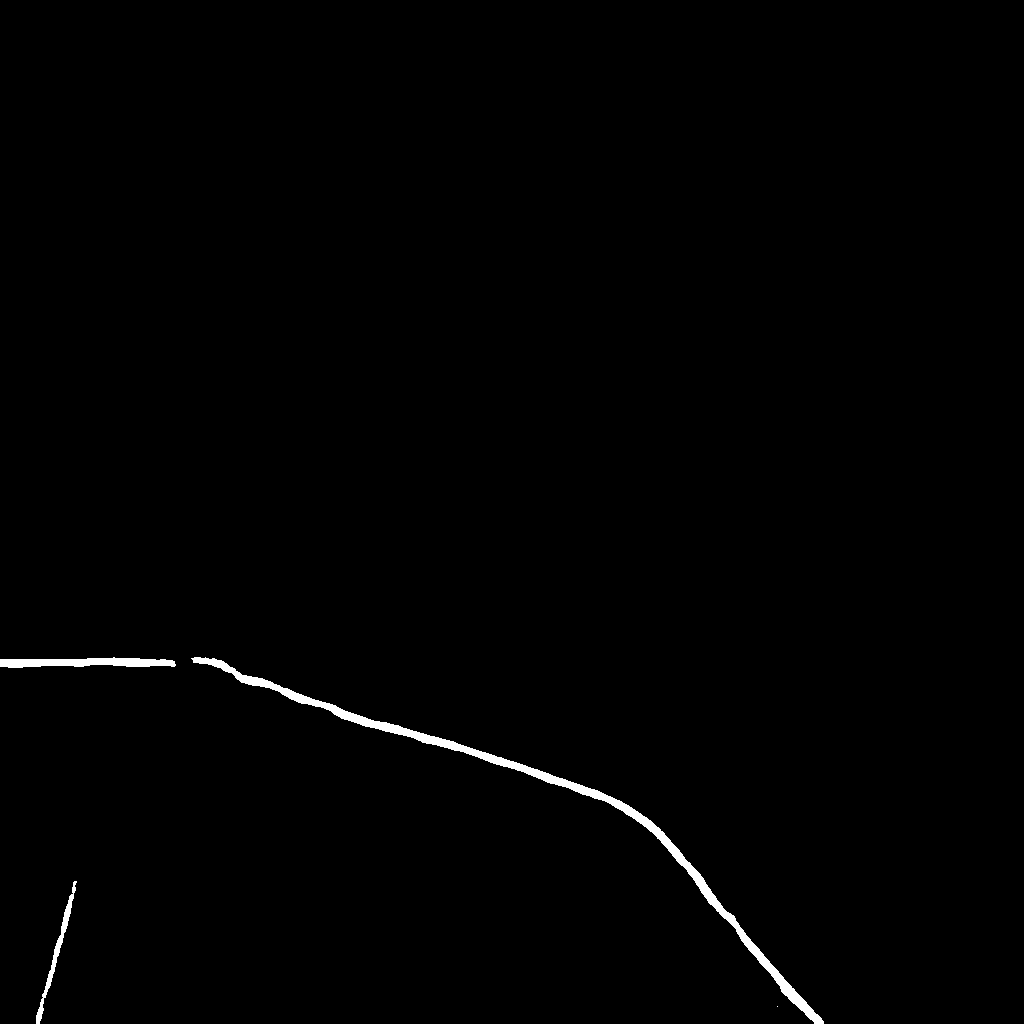}
		\end{minipage}
		\begin{minipage}[b]{0.10\textwidth}
			\centering
			\includegraphics[width=\textwidth]{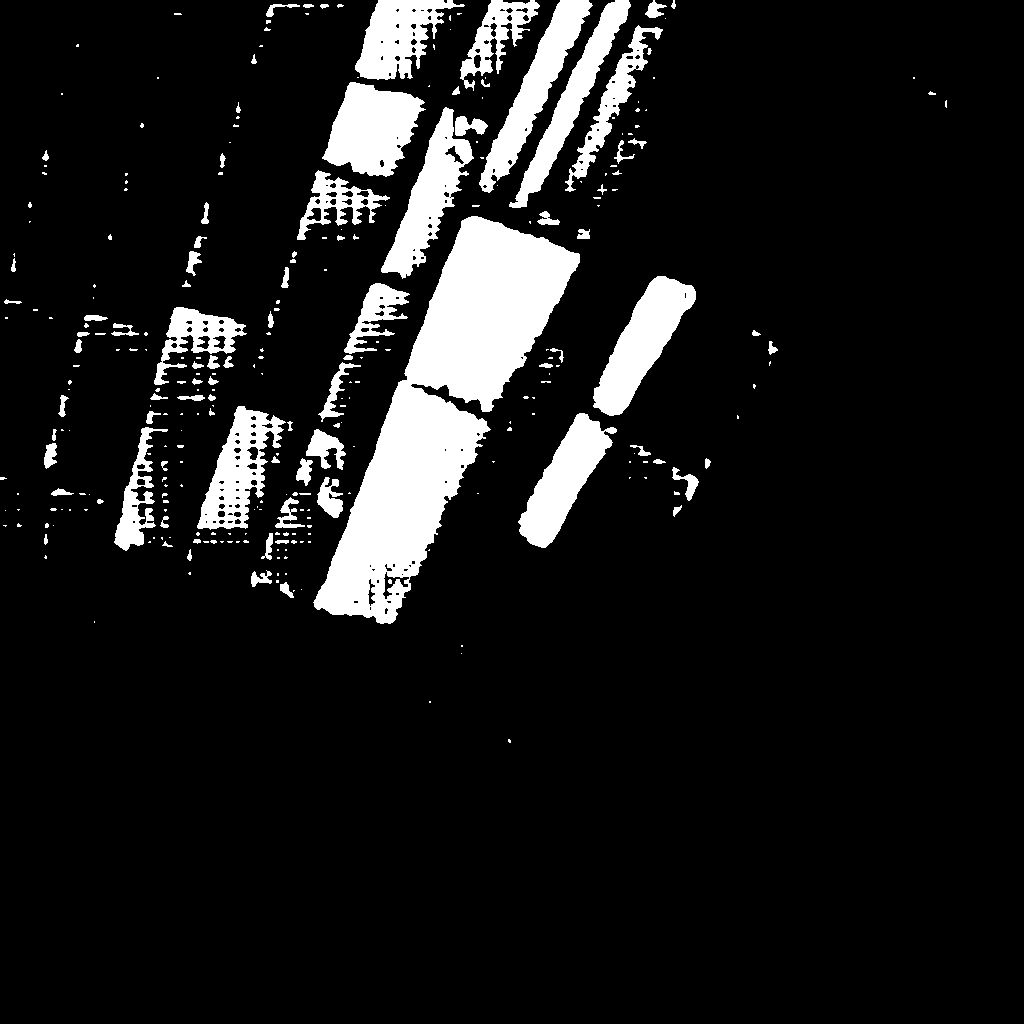}
		\end{minipage}
		\begin{minipage}[b]{0.10\textwidth}
			\centering
			\includegraphics[width=\textwidth]{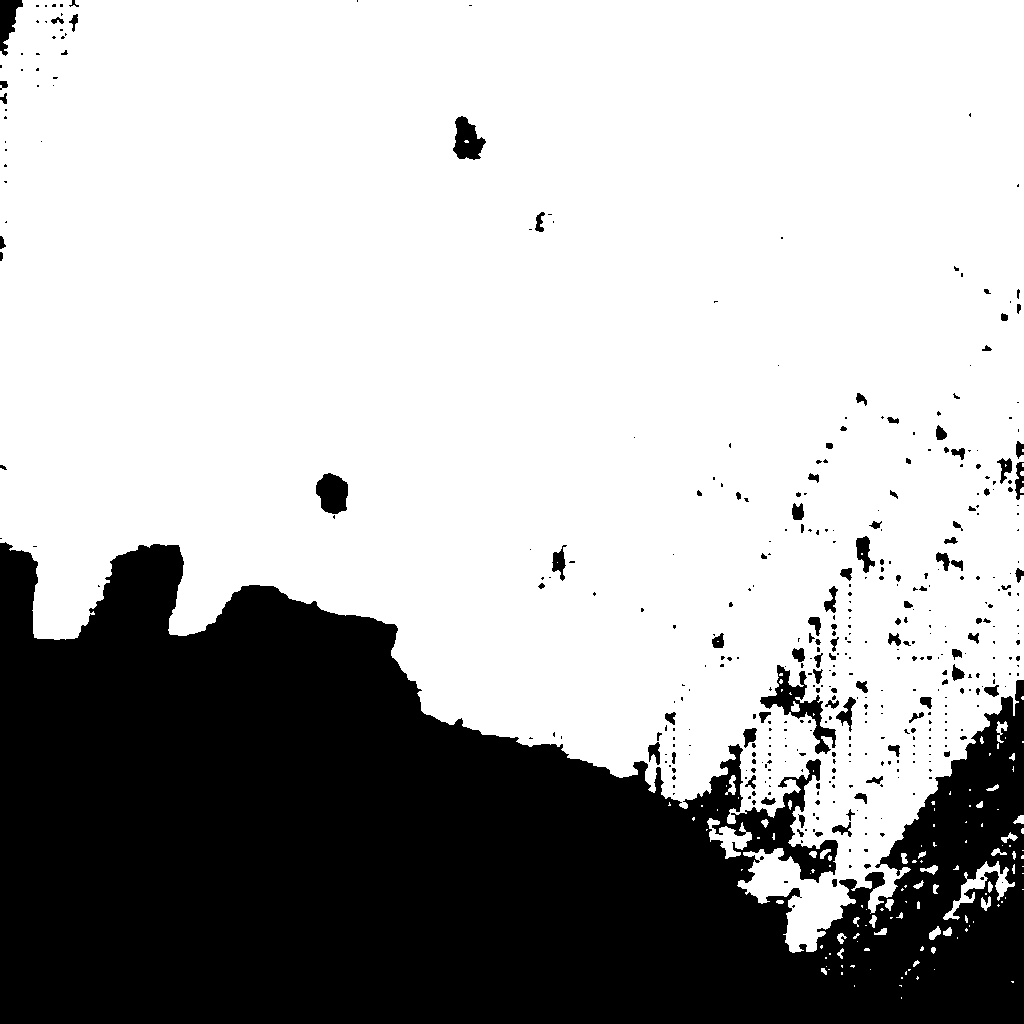}
		\end{minipage}
		\begin{minipage}[b]{0.10\textwidth}
			\centering
			\includegraphics[width=\textwidth]{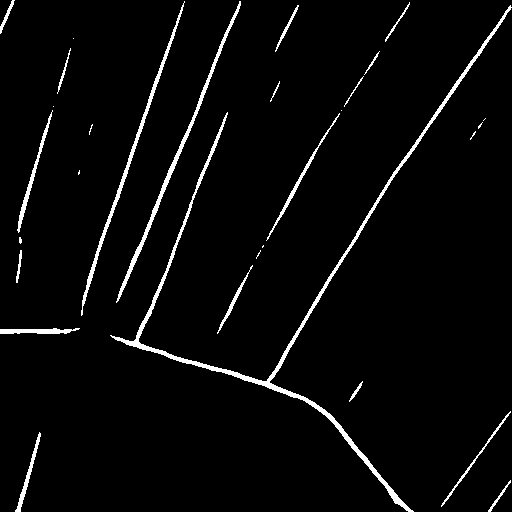}
		\end{minipage}\\[3pt]
		\begin{minipage}[b]{0.10\textwidth}
			\centering
			\includegraphics[width=\textwidth]{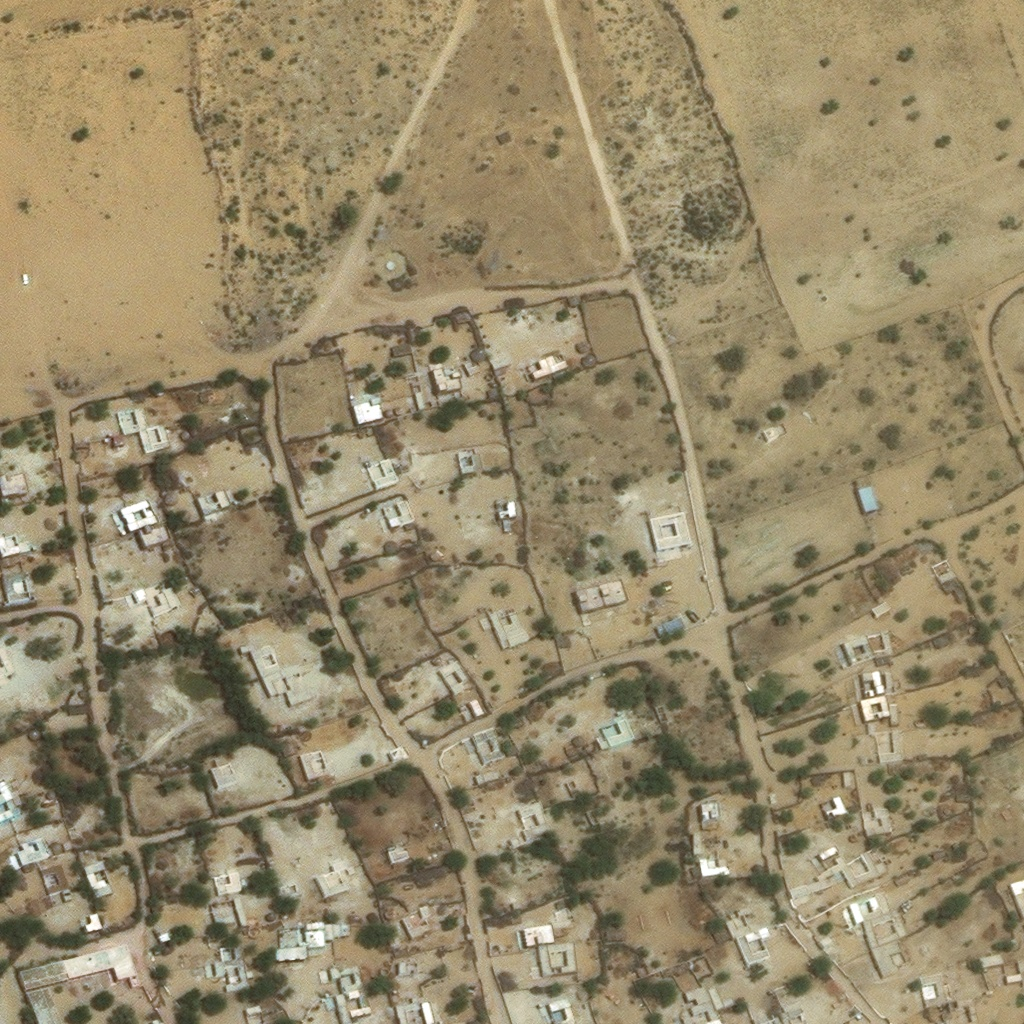}
		\end{minipage}
		\begin{minipage}[b]{0.10\textwidth}
			\centering
			\includegraphics[width=\textwidth]{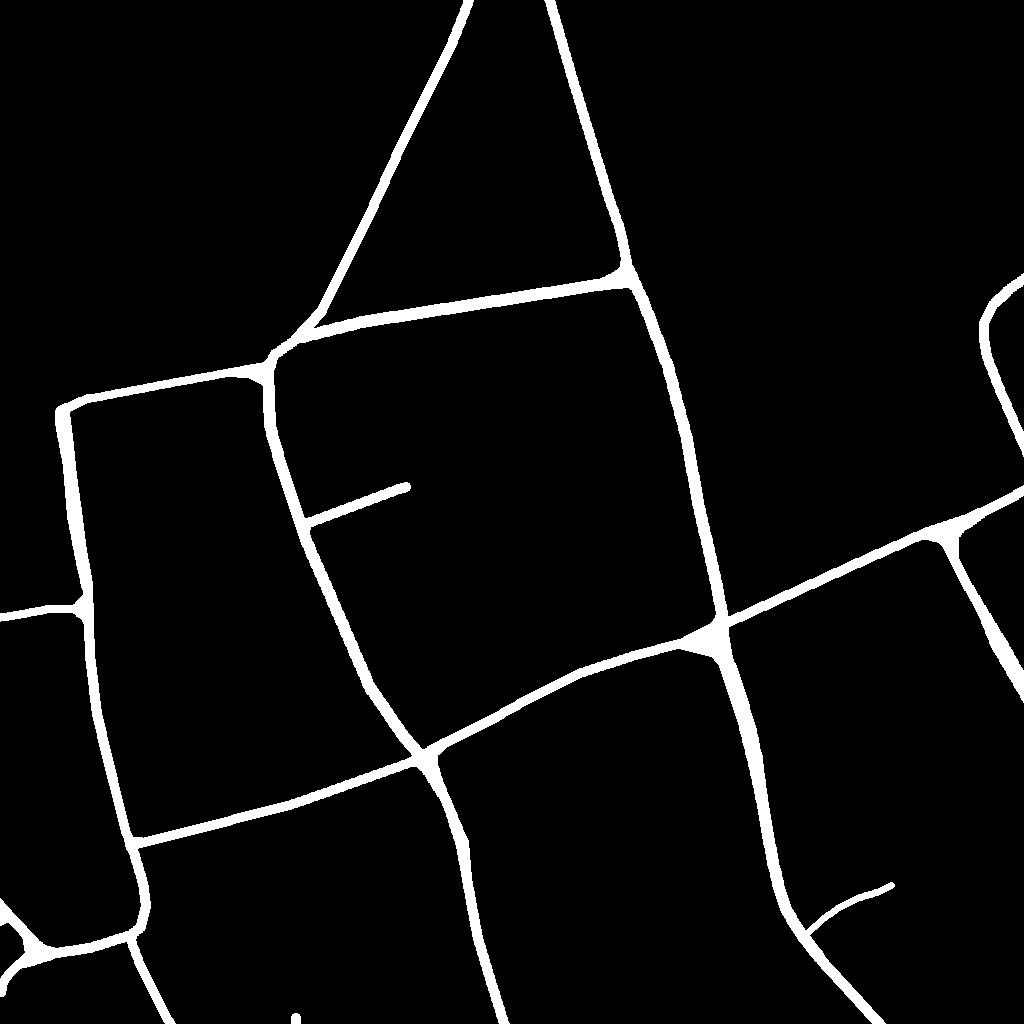}
		\end{minipage}
		\begin{minipage}[b]{0.10\textwidth}
			\centering
			\includegraphics[width=\textwidth]{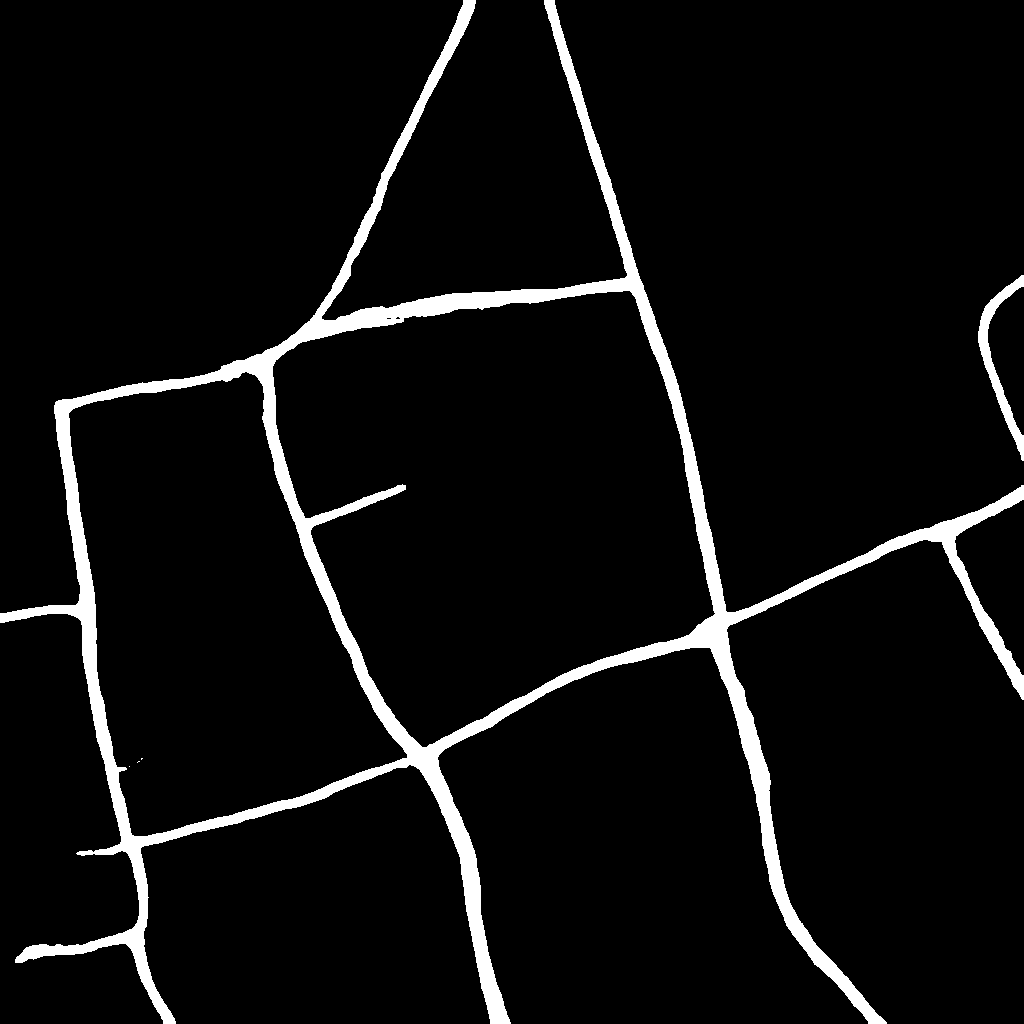}
		\end{minipage}
		\begin{minipage}[b]{0.10\textwidth}
			\centering
			\includegraphics[width=\textwidth]{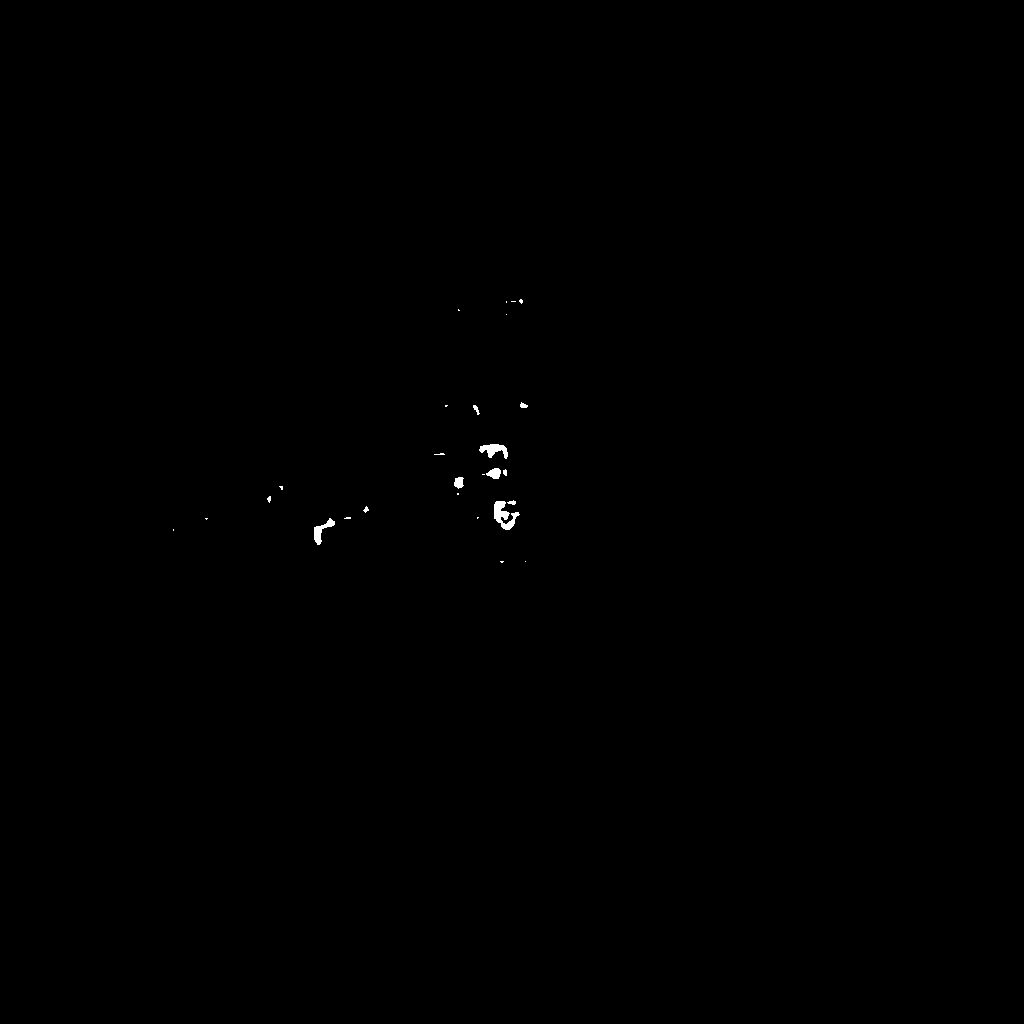}
		\end{minipage}
		\begin{minipage}[b]{0.10\textwidth}
			\centering
			\includegraphics[width=\textwidth]{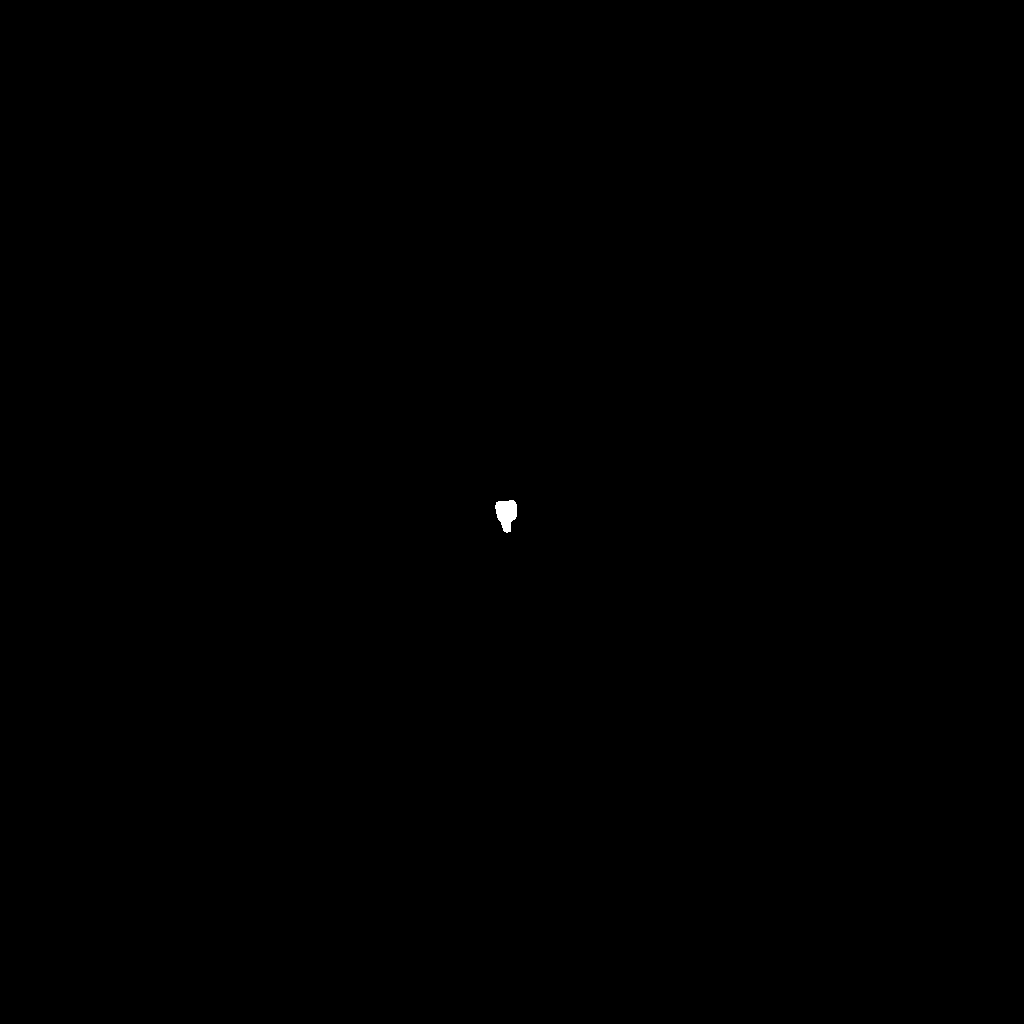}
		\end{minipage}
		\begin{minipage}[b]{0.10\textwidth}
			\centering
			\includegraphics[width=\textwidth]{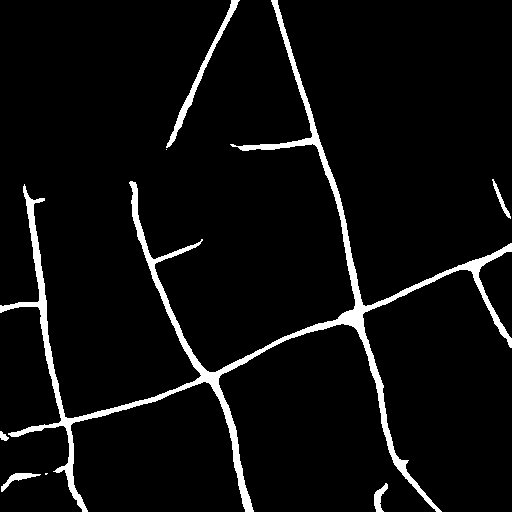}
		\end{minipage}\\[3pt]
		\begin{minipage}[b]{0.10\textwidth}
			\centering
			\includegraphics[width=\textwidth]{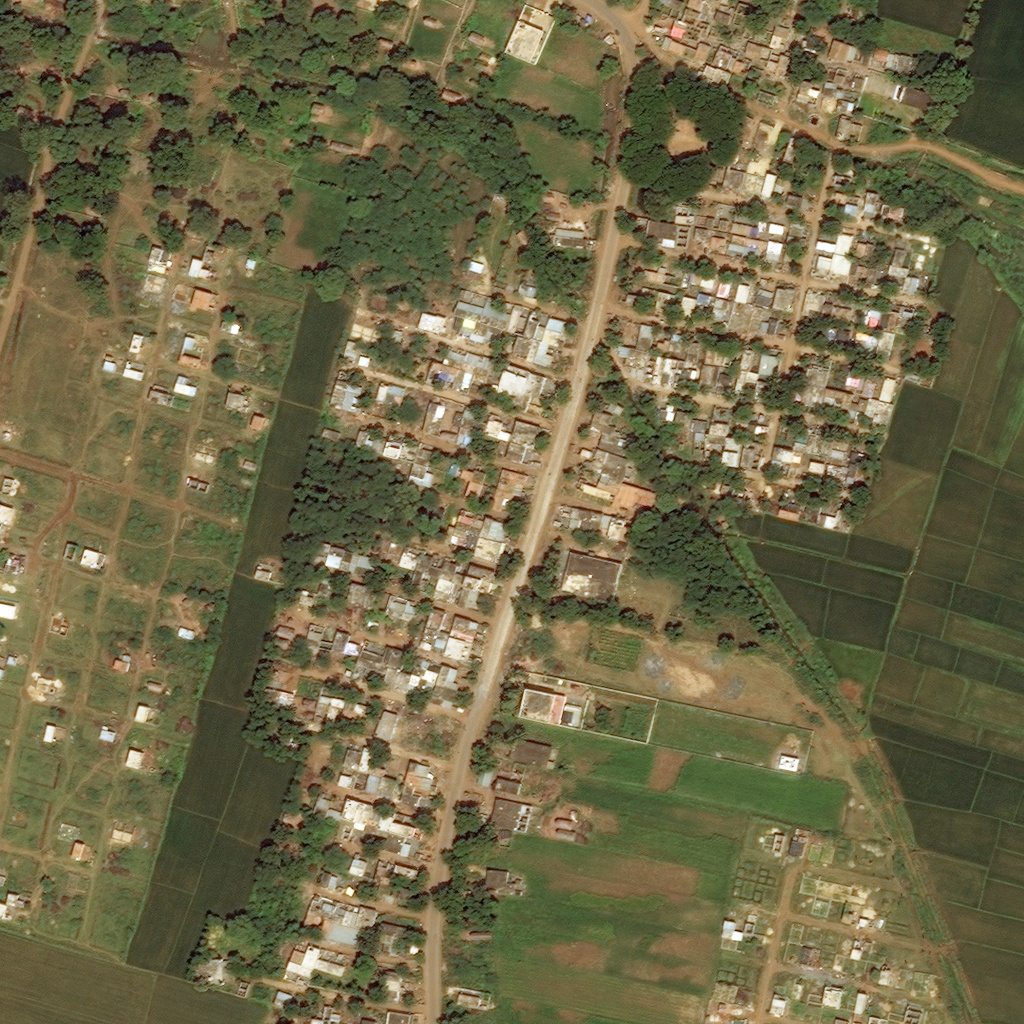}
			\scriptsize{Original Image}
		\end{minipage}
		\begin{minipage}[b]{0.10\textwidth}
			\centering
			\includegraphics[width=\textwidth]{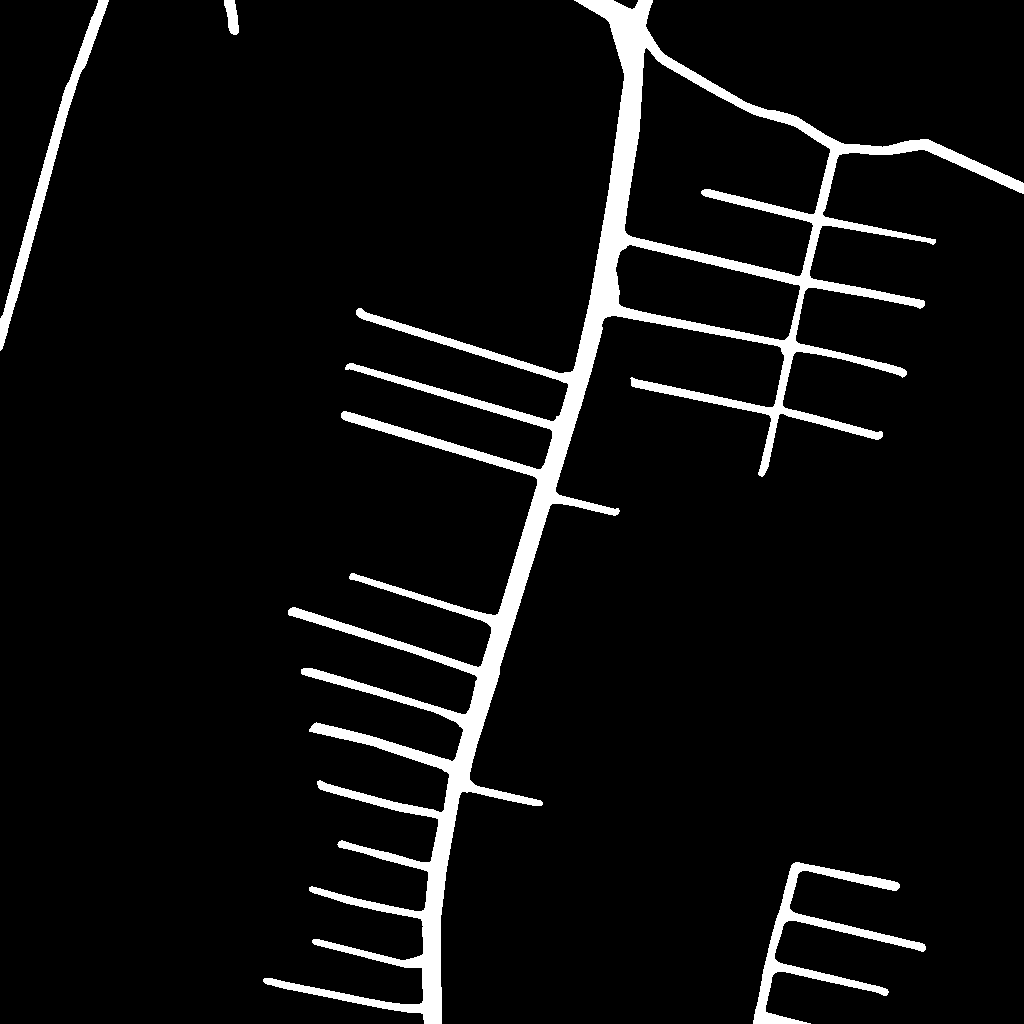}
			\scriptsize{Ground Truth}
		\end{minipage}
		\begin{minipage}[b]{0.10\textwidth}
			\centering
			\includegraphics[width=\textwidth]{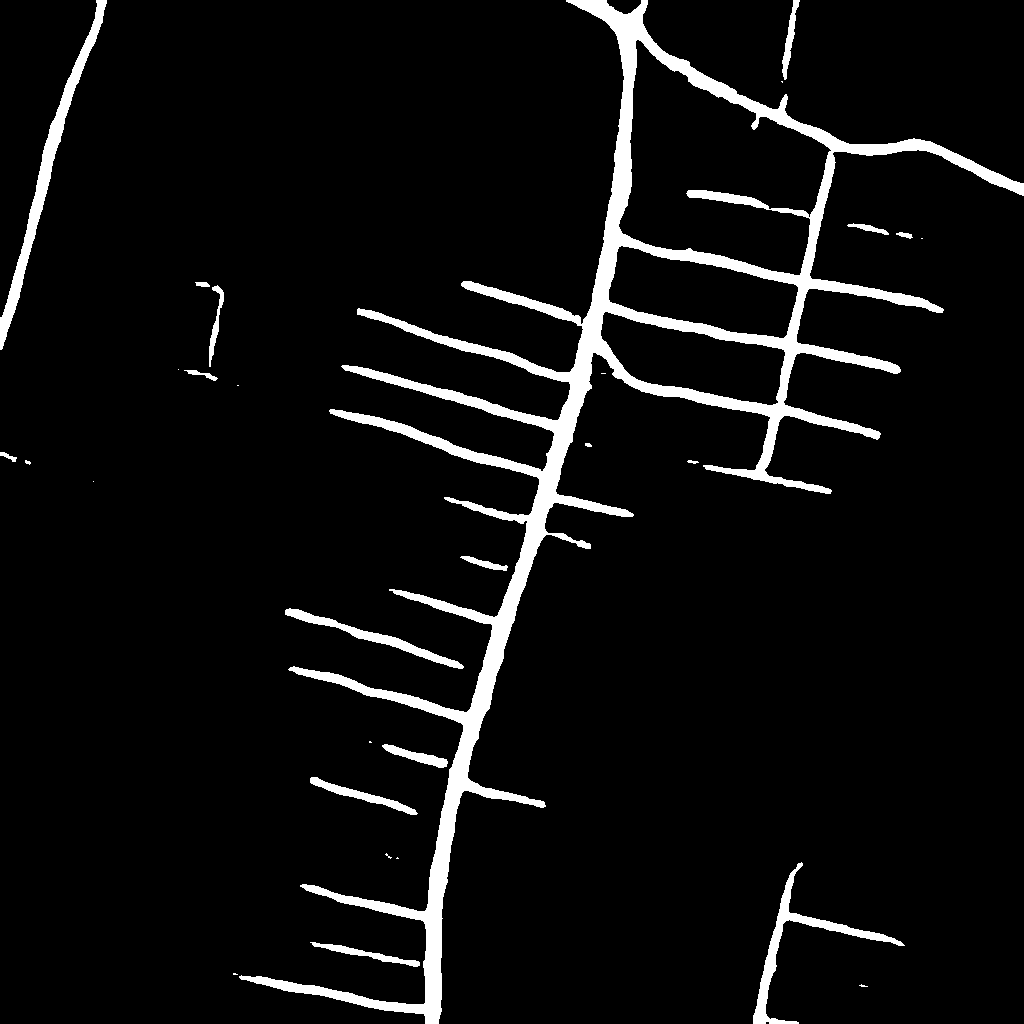}
			\scriptsize{RSAM-Seg}
		\end{minipage}
		\begin{minipage}[b]{0.10\textwidth}
			\centering
			\includegraphics[width=\textwidth]{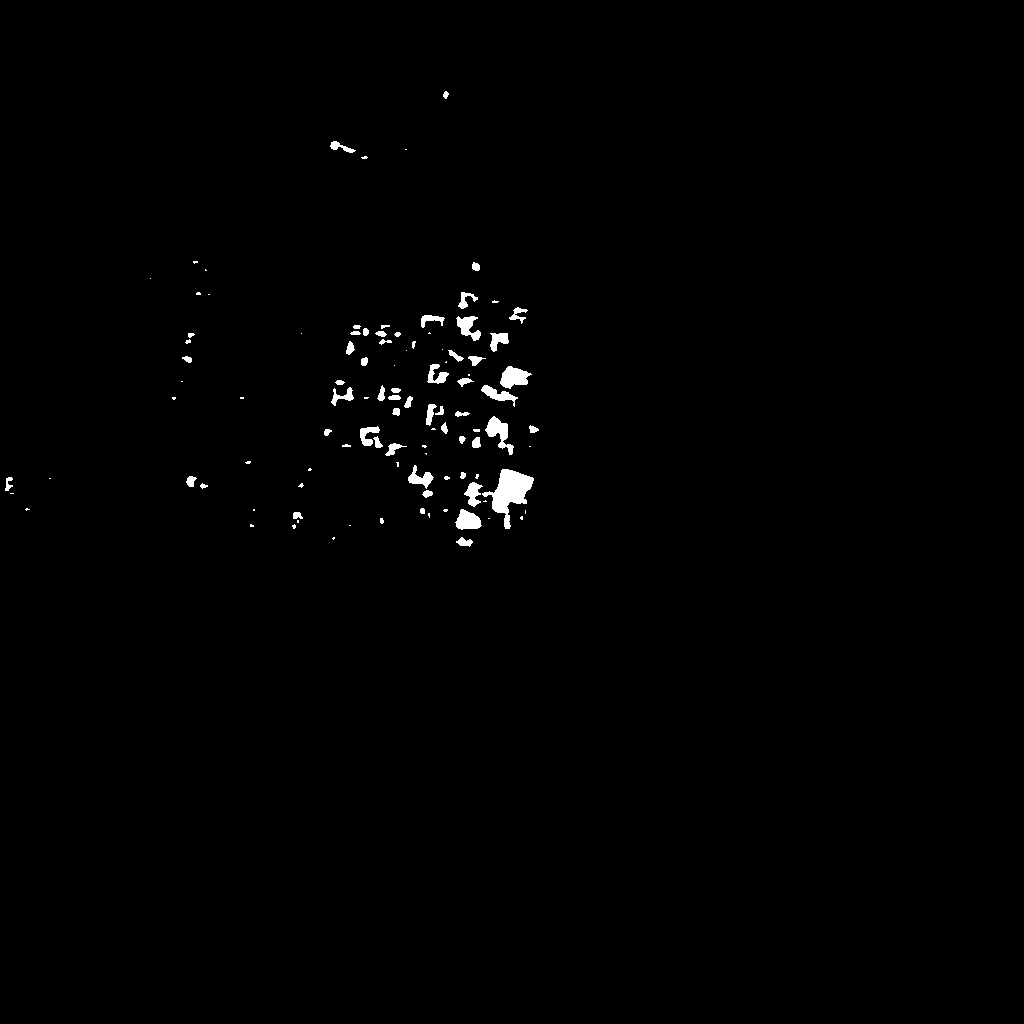}
			\scriptsize{SAM (center -)}
		\end{minipage}
		\begin{minipage}[b]{0.10\textwidth}
			\centering
			\includegraphics[width=\textwidth]{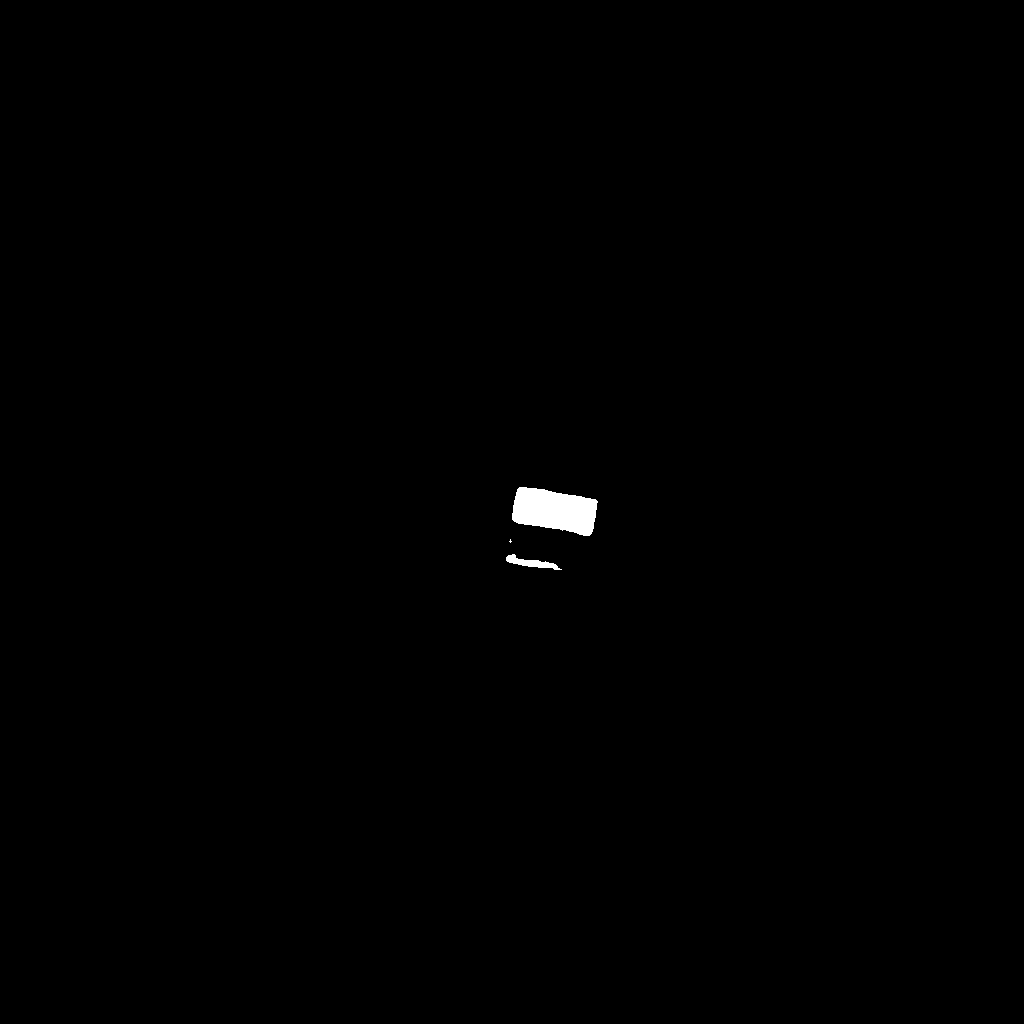}
			\scriptsize{SAM (center +)}
		\end{minipage}
		\begin{minipage}[b]{0.10\textwidth}
			\centering
			\includegraphics[width=\textwidth]{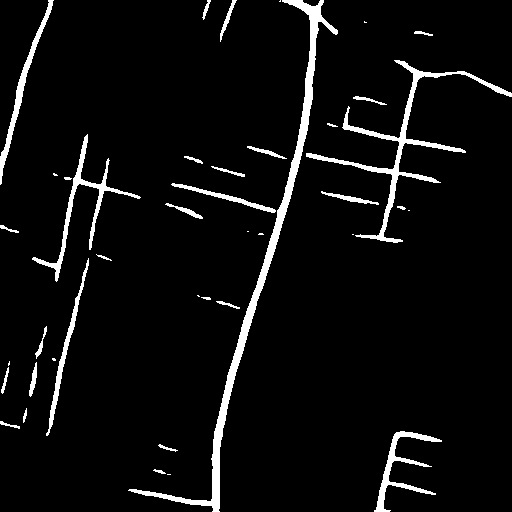}
			\scriptsize{U-Net}
		\end{minipage}
		\caption{Comparison of road Segmentation results on DG-Road dataset with RSAM-Seg, SAM and U-Net.}
		\label{fig:DGRoadQuality}
	\end{figure*}
	
	\begin{figure*}[!htb]
		\centering
		\begin{minipage}[b]{0.10\textwidth}
			\centering
			\includegraphics[width=\textwidth]{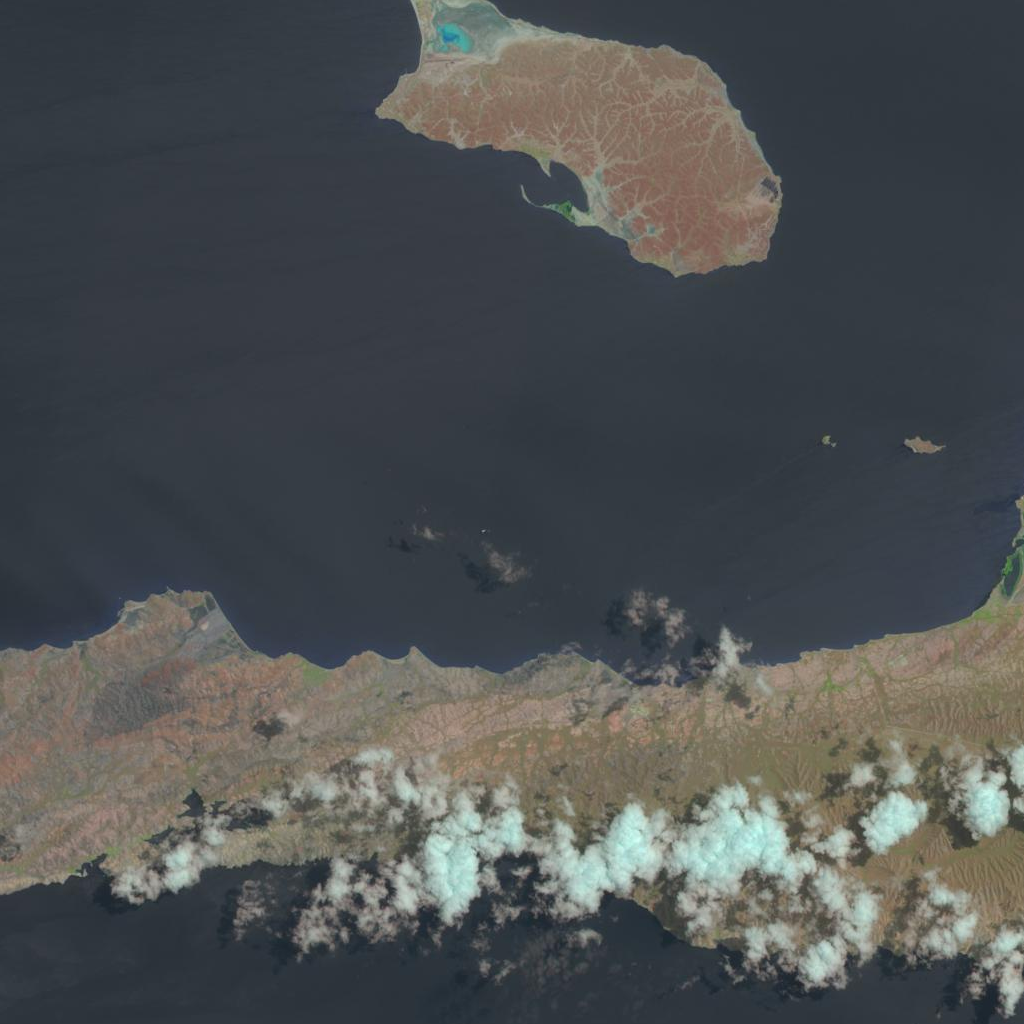}
		\end{minipage}
		\begin{minipage}[b]{0.10\textwidth}
			\centering
			\includegraphics[width=\textwidth]{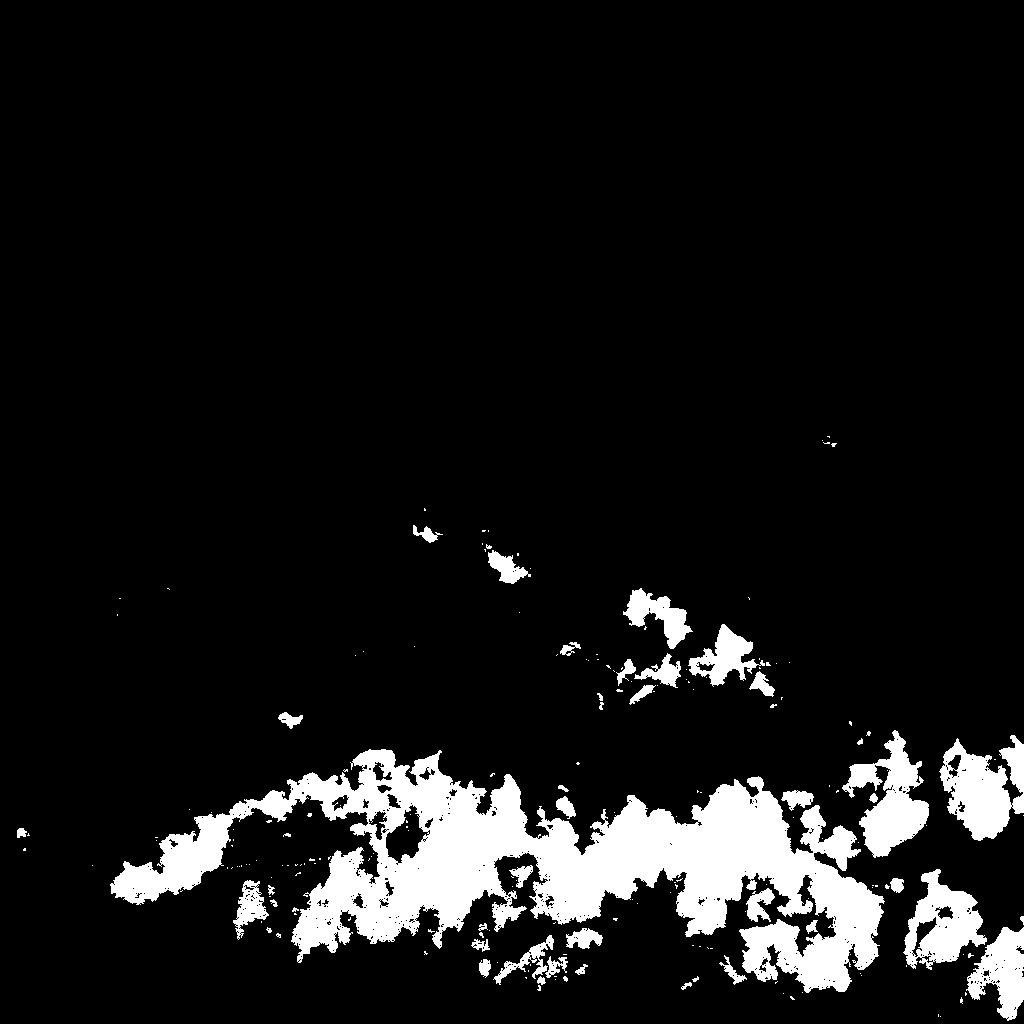}
		\end{minipage}
		\begin{minipage}[b]{0.10\textwidth}
			\centering
			\includegraphics[width=\textwidth]{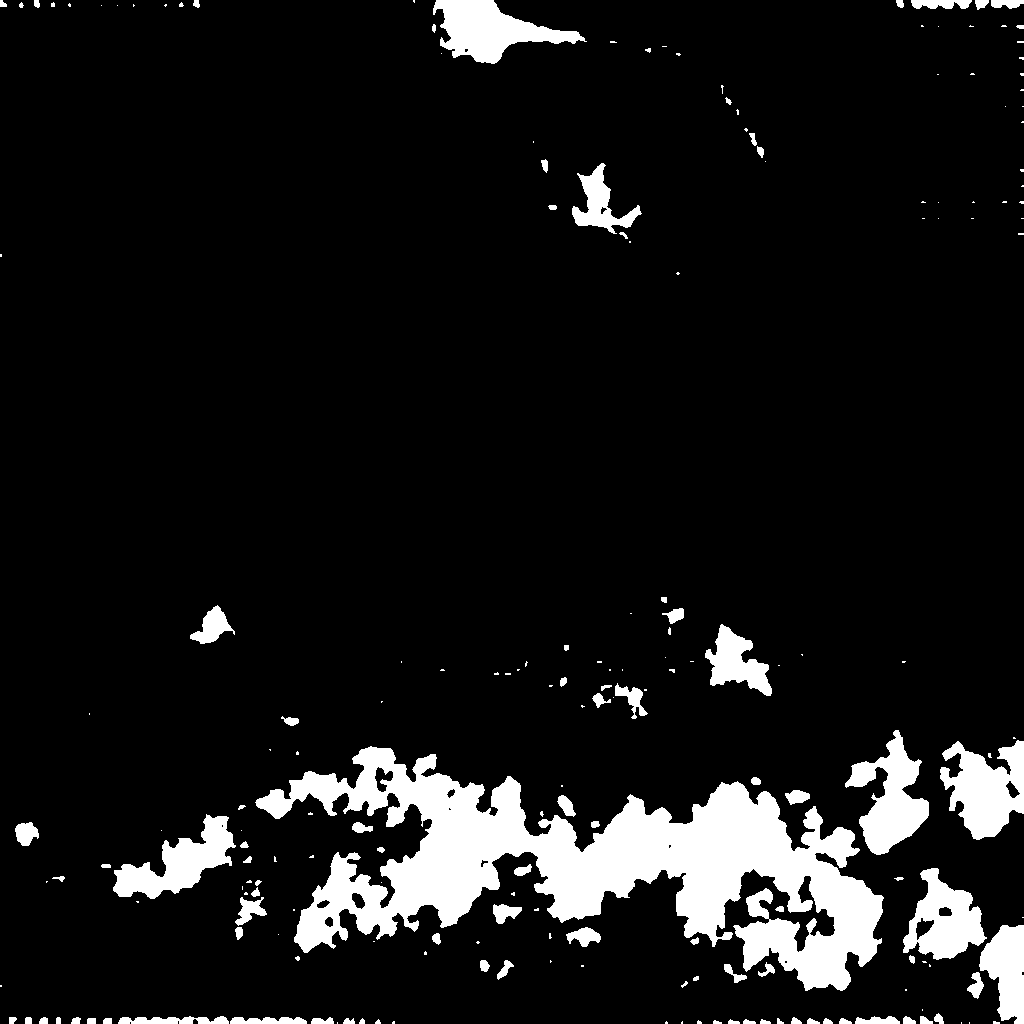}
		\end{minipage}
		\begin{minipage}[b]{0.10\textwidth}
			\centering
			\includegraphics[width=\textwidth]{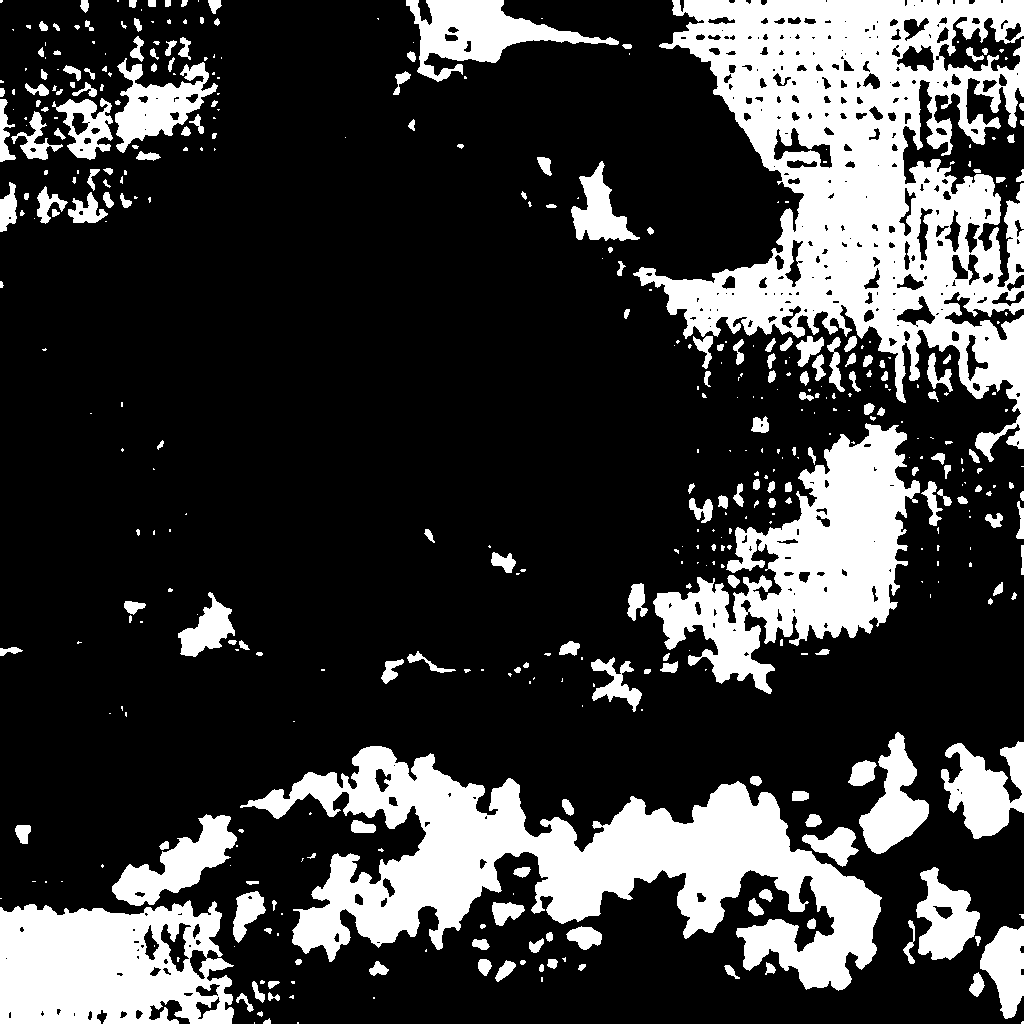}
		\end{minipage}
		\begin{minipage}[b]{0.10\textwidth}
			\centering
			\includegraphics[width=\textwidth]{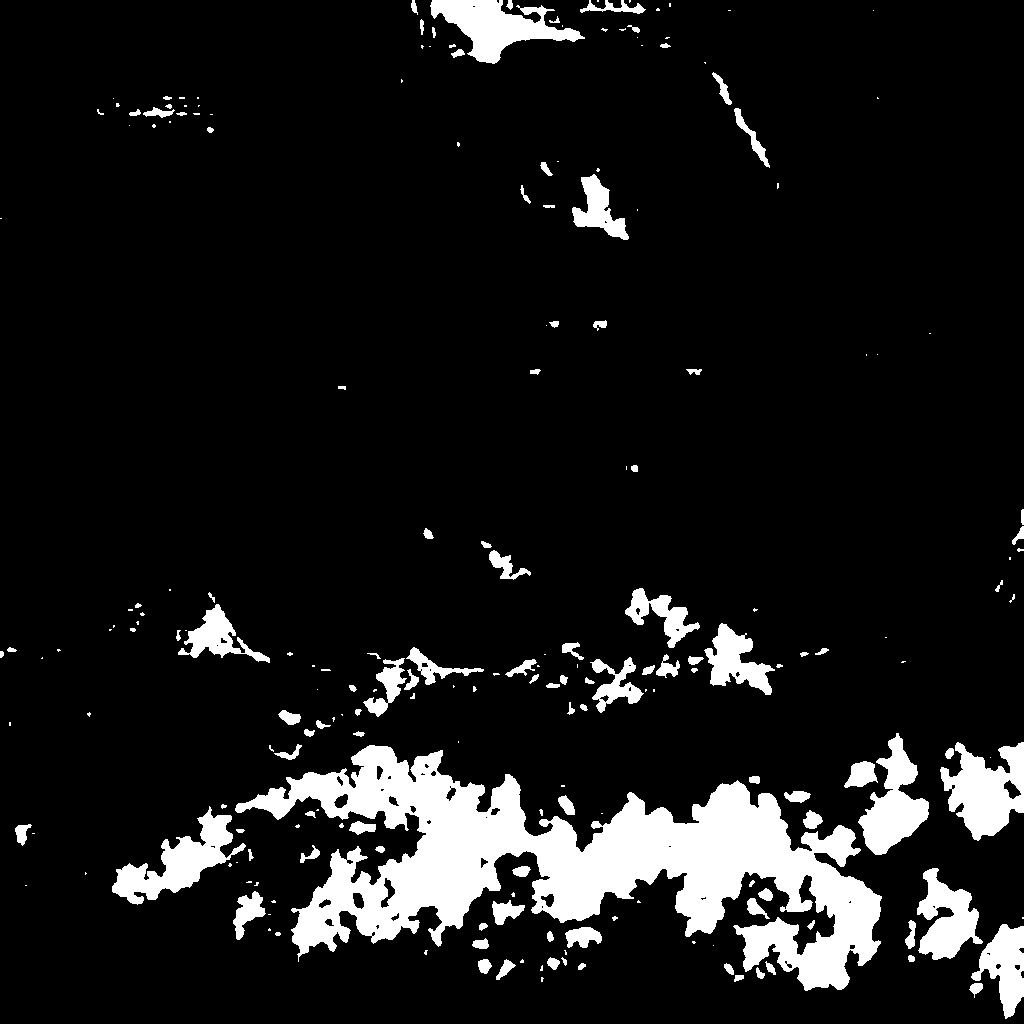}
		\end{minipage}
		\begin{minipage}[b]{0.10\textwidth}
			\centering
			\includegraphics[width=\textwidth]{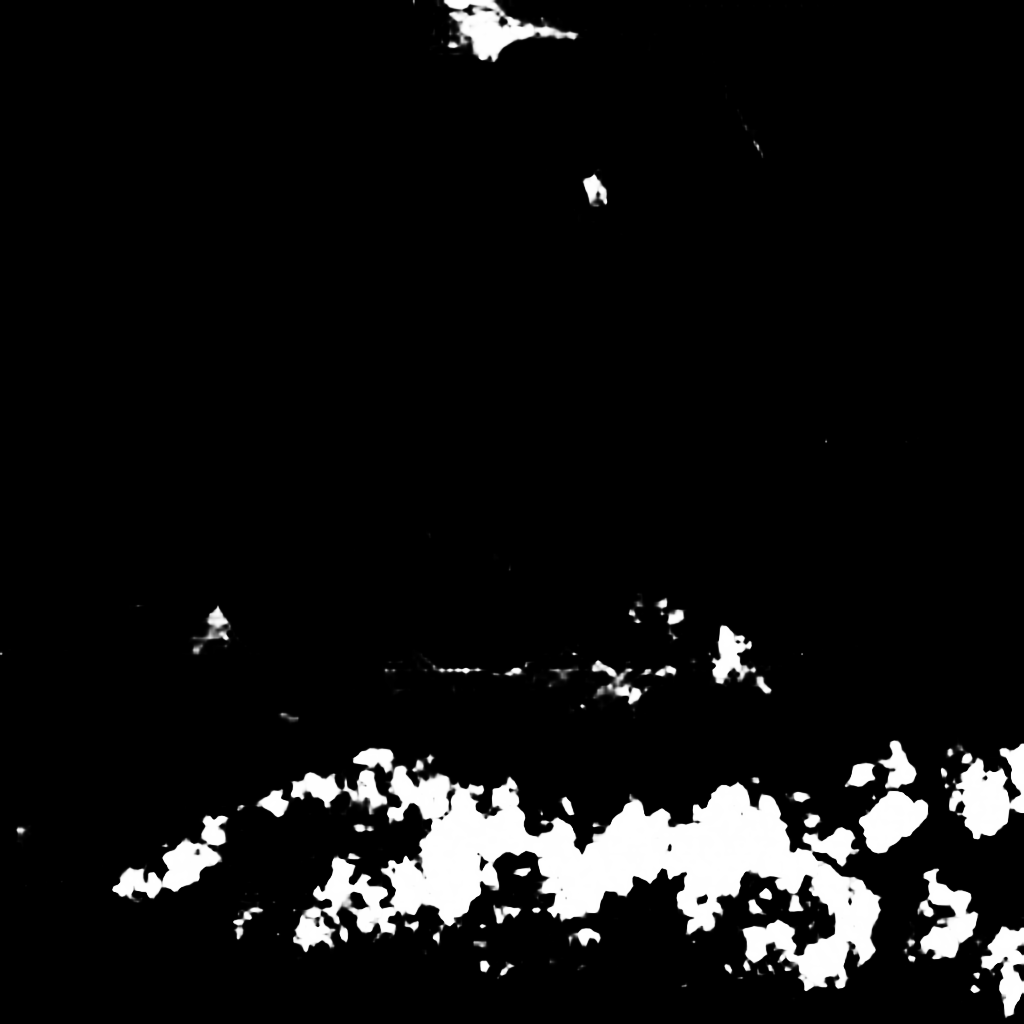}
		\end{minipage}\\[3pt]
		\begin{minipage}[b]{0.10\textwidth}
			\centering
			\includegraphics[width=\textwidth]{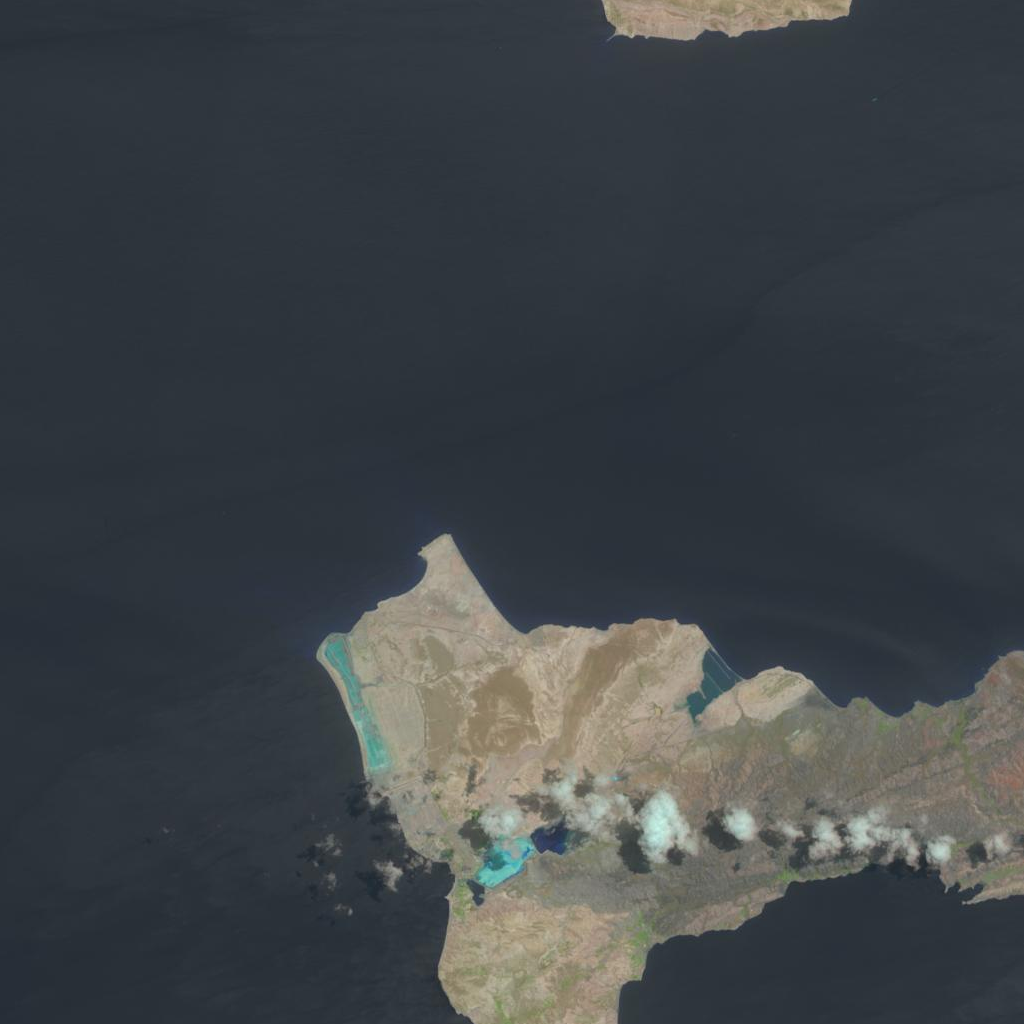}
		\end{minipage}
		\begin{minipage}[b]{0.10\textwidth}
			\centering
			\includegraphics[width=\textwidth]{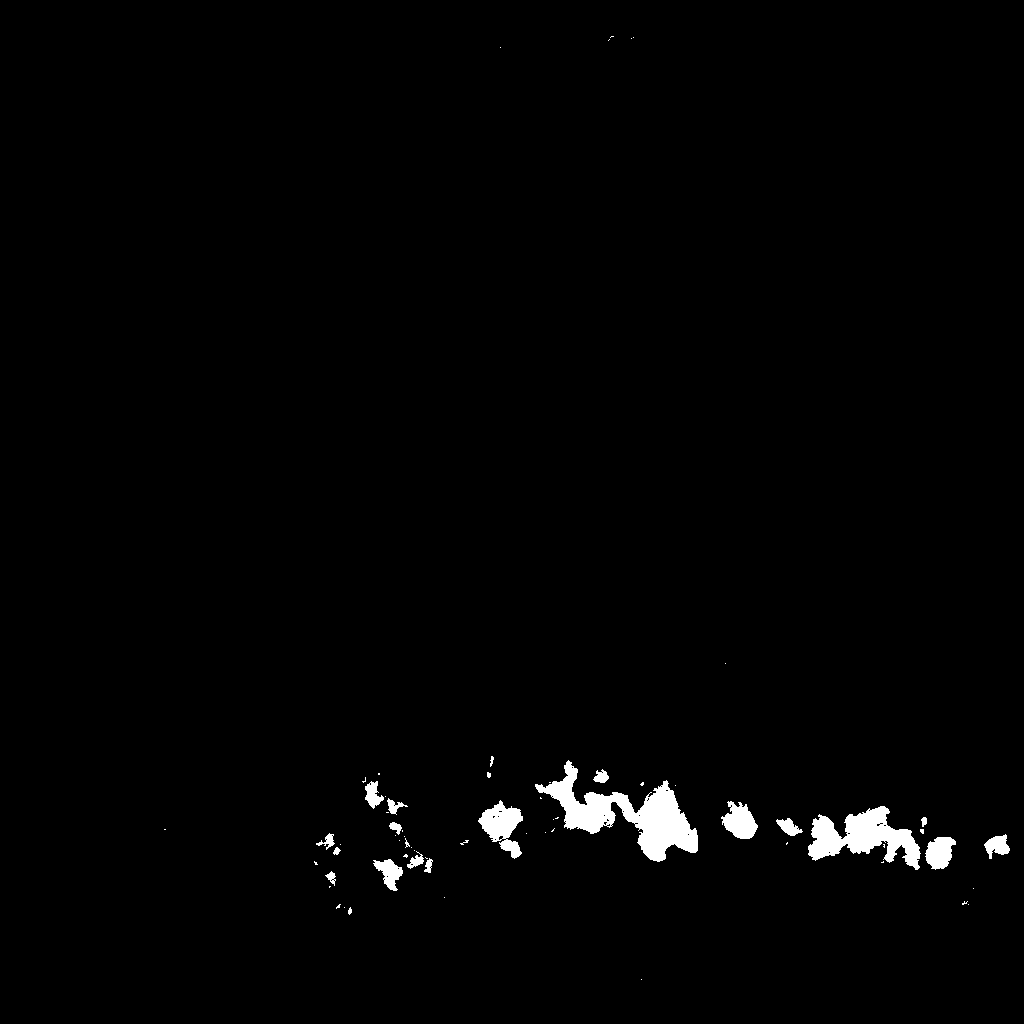}
		\end{minipage}
		\begin{minipage}[b]{0.10\textwidth}
			\centering
			\includegraphics[width=\textwidth]{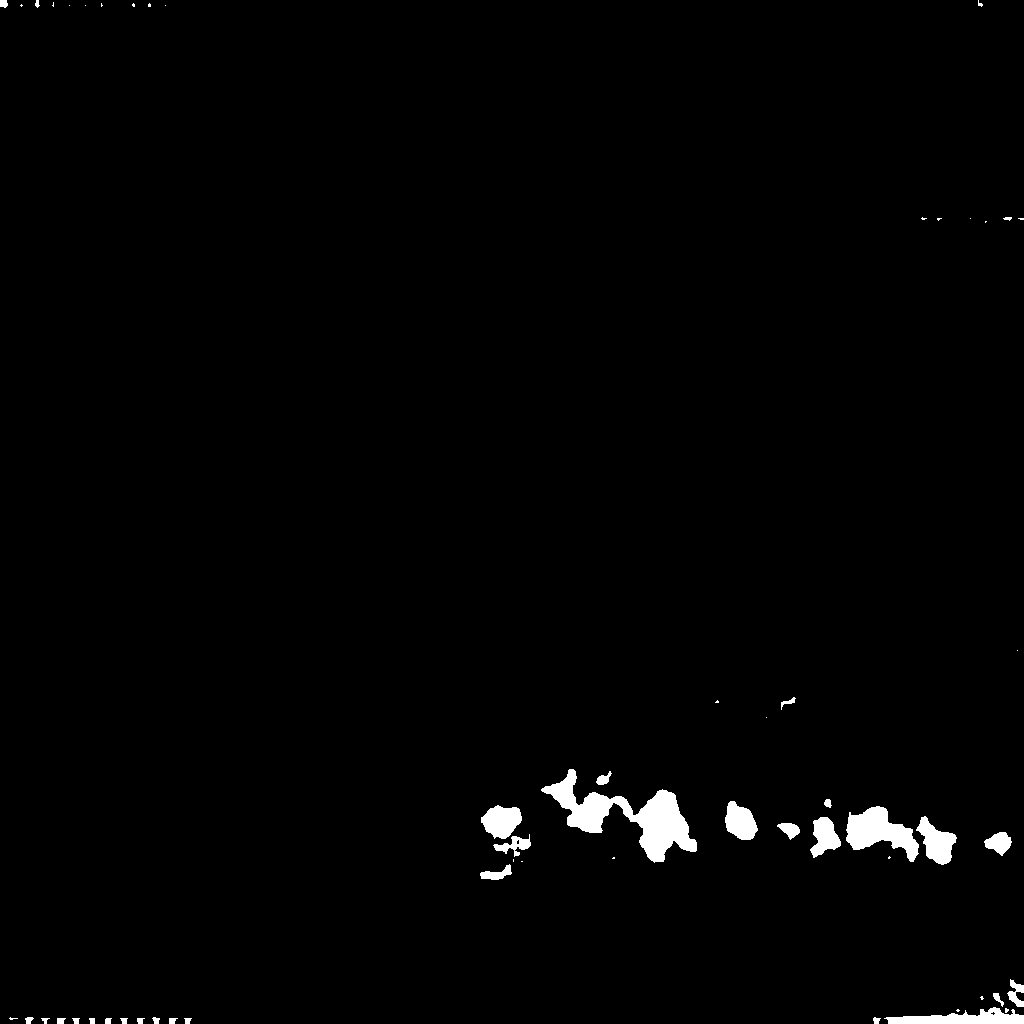}
		\end{minipage}
		\begin{minipage}[b]{0.10\textwidth}
			\centering
			\includegraphics[width=\textwidth]{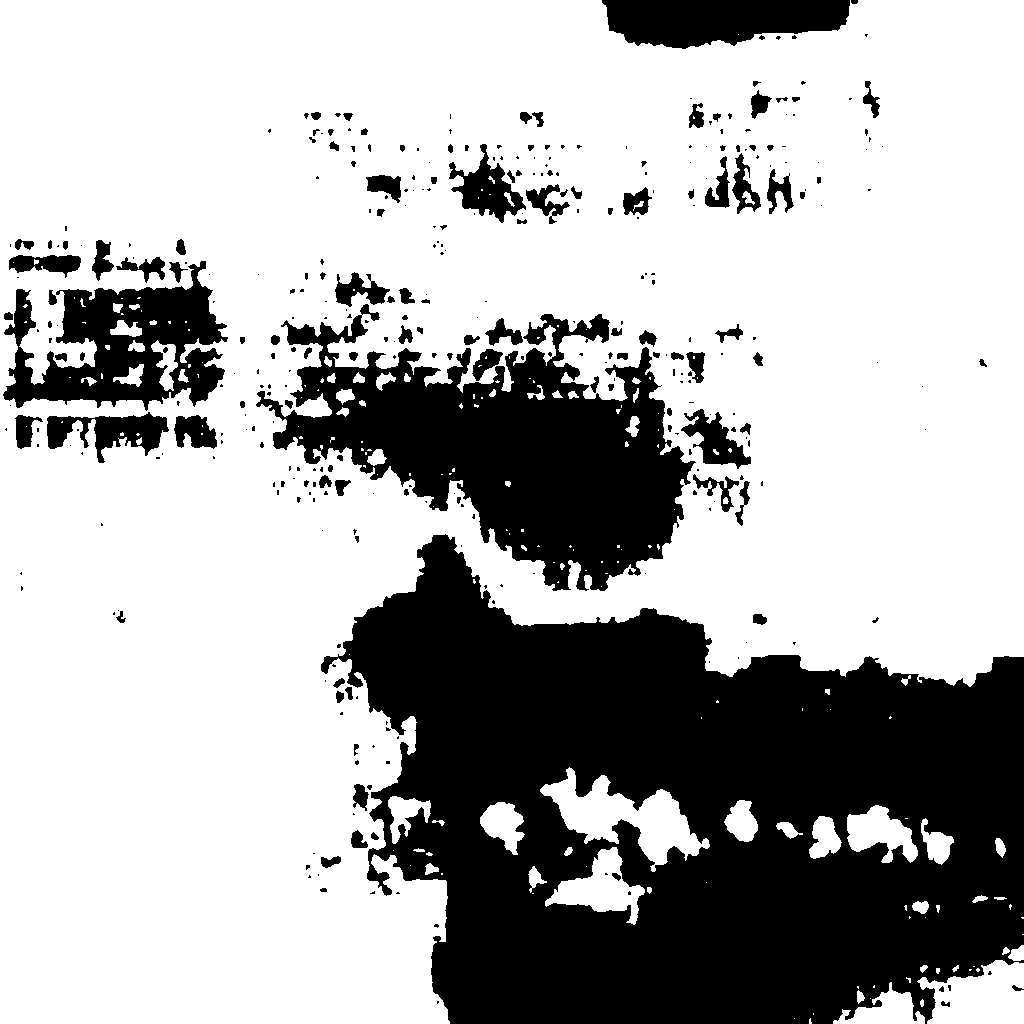}
		\end{minipage}
		\begin{minipage}[b]{0.10\textwidth}
			\centering
			\includegraphics[width=\textwidth]{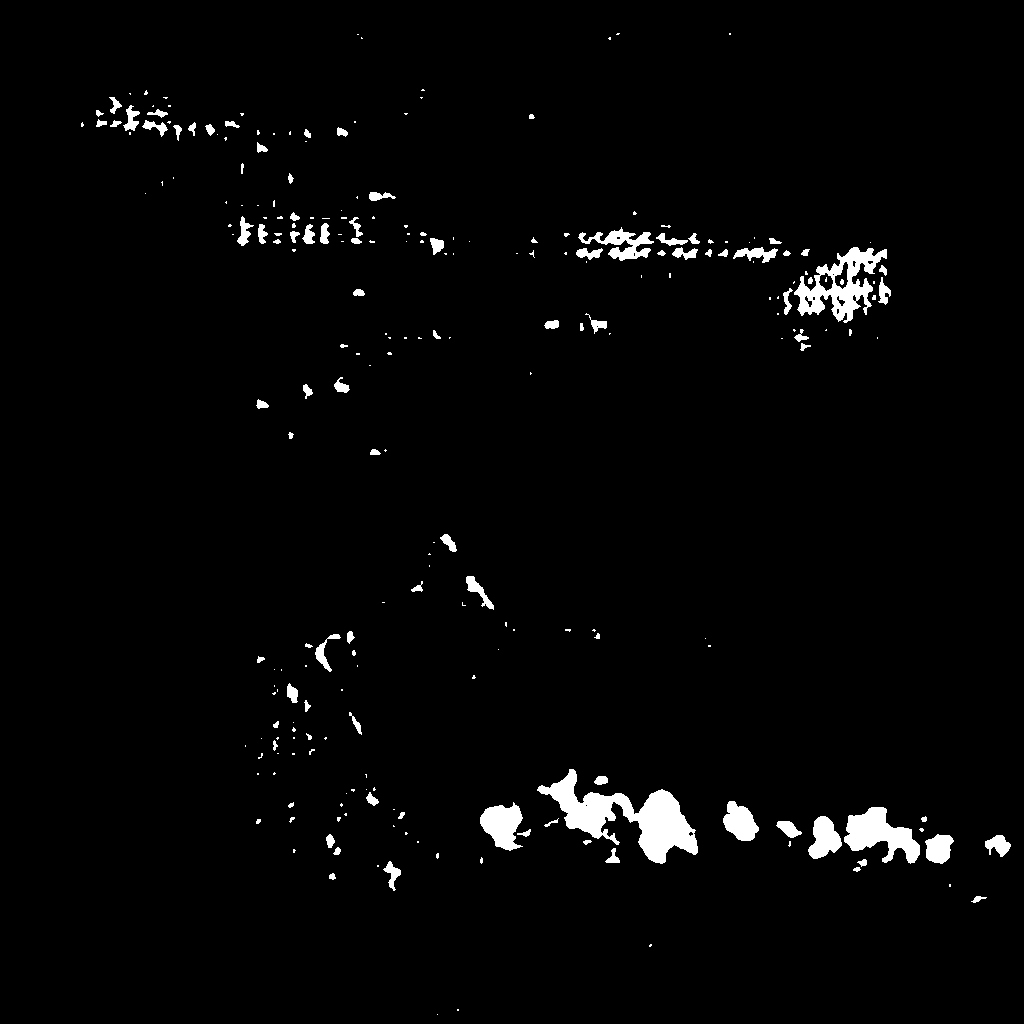}
		\end{minipage}
		\begin{minipage}[b]{0.10\textwidth}
			\centering
			\includegraphics[width=\textwidth]{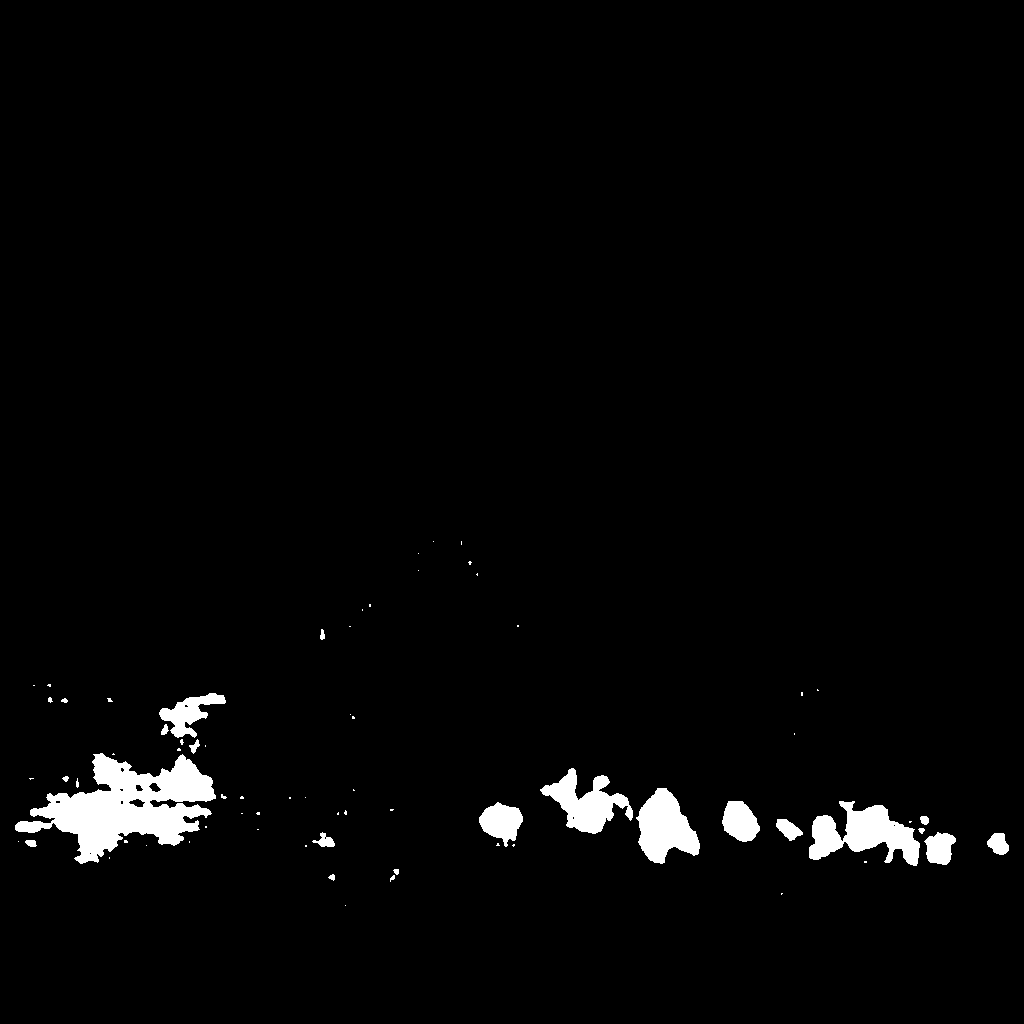}
		\end{minipage}\\[3pt]
		\begin{minipage}[b]{0.10\textwidth}
			\centering
			\includegraphics[width=\textwidth]{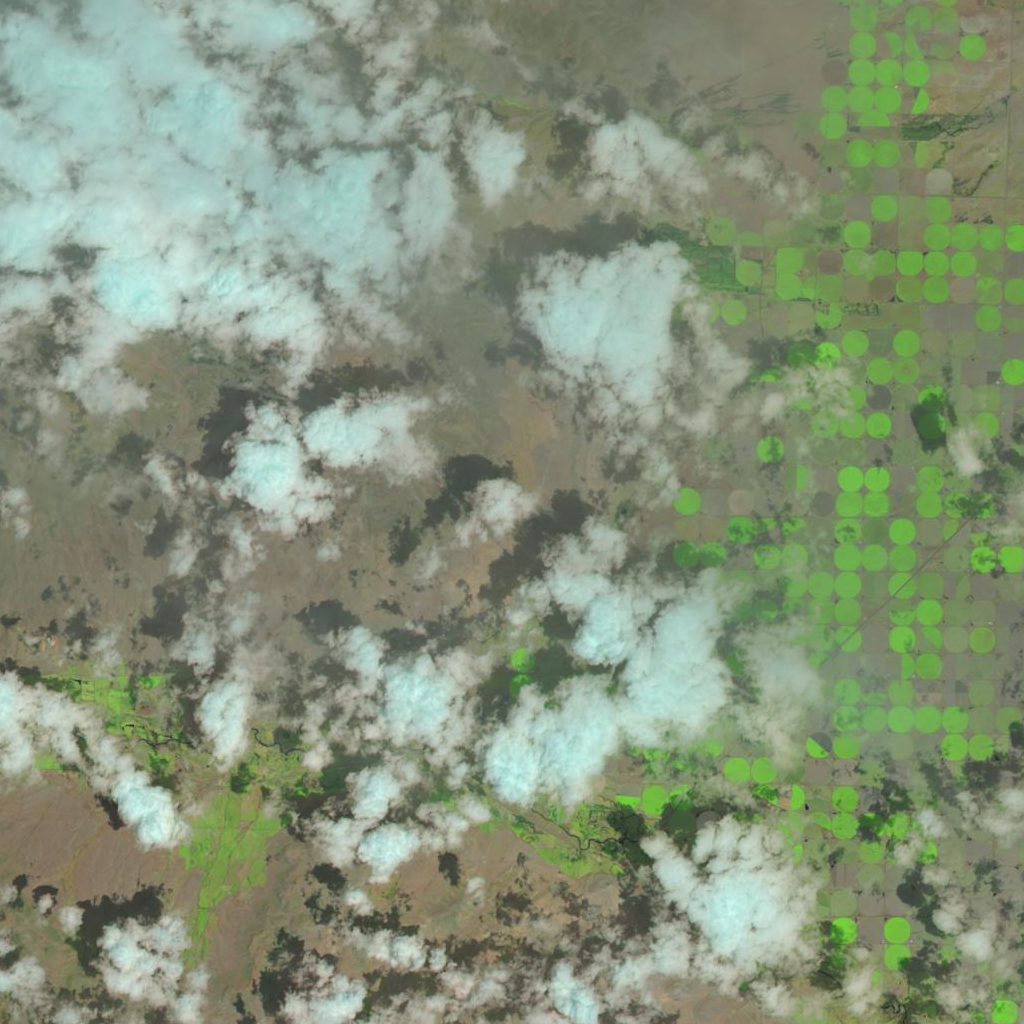}
			\scriptsize{Original Image}
		\end{minipage}
		\begin{minipage}[b]{0.10\textwidth}
			\centering
			\includegraphics[width=\textwidth]{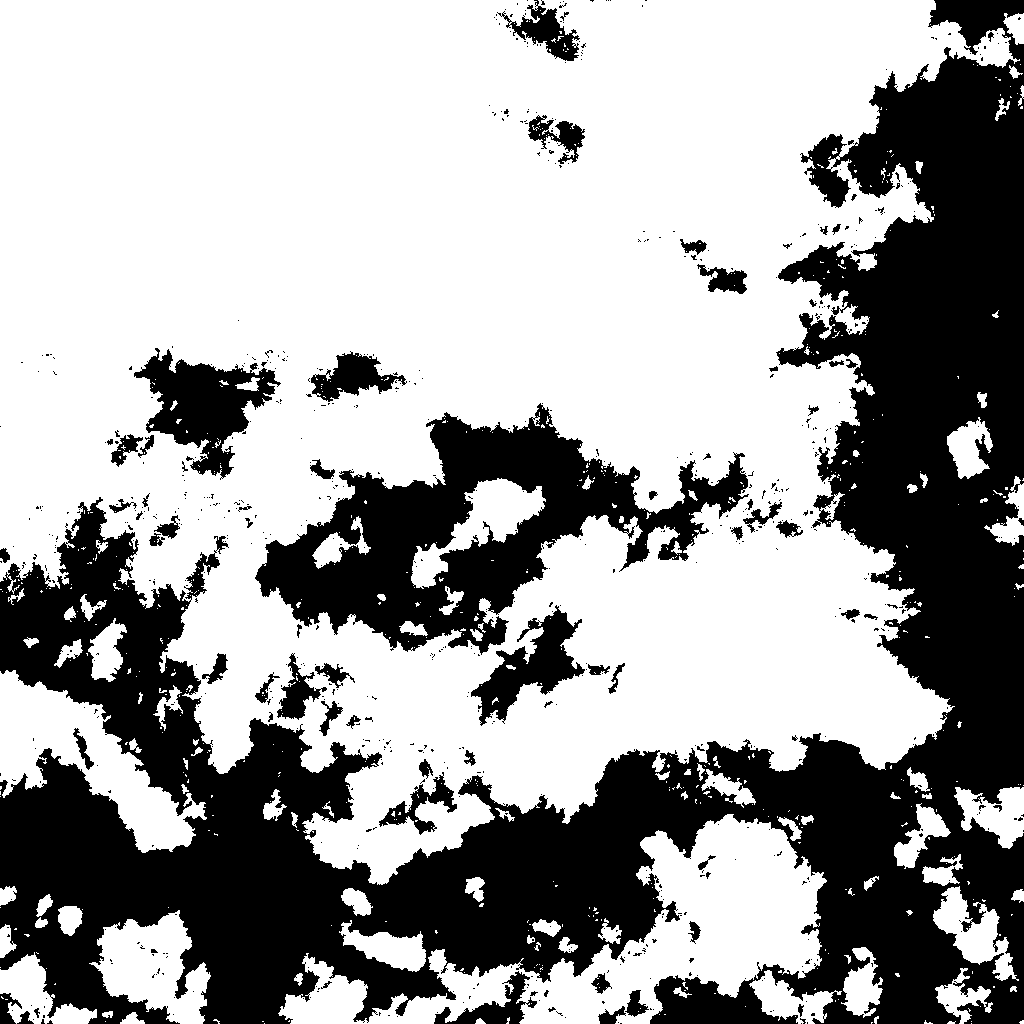}
			\scriptsize{Ground Truth}
		\end{minipage}
		\begin{minipage}[b]{0.10\textwidth}
			\centering
			\includegraphics[width=\textwidth]{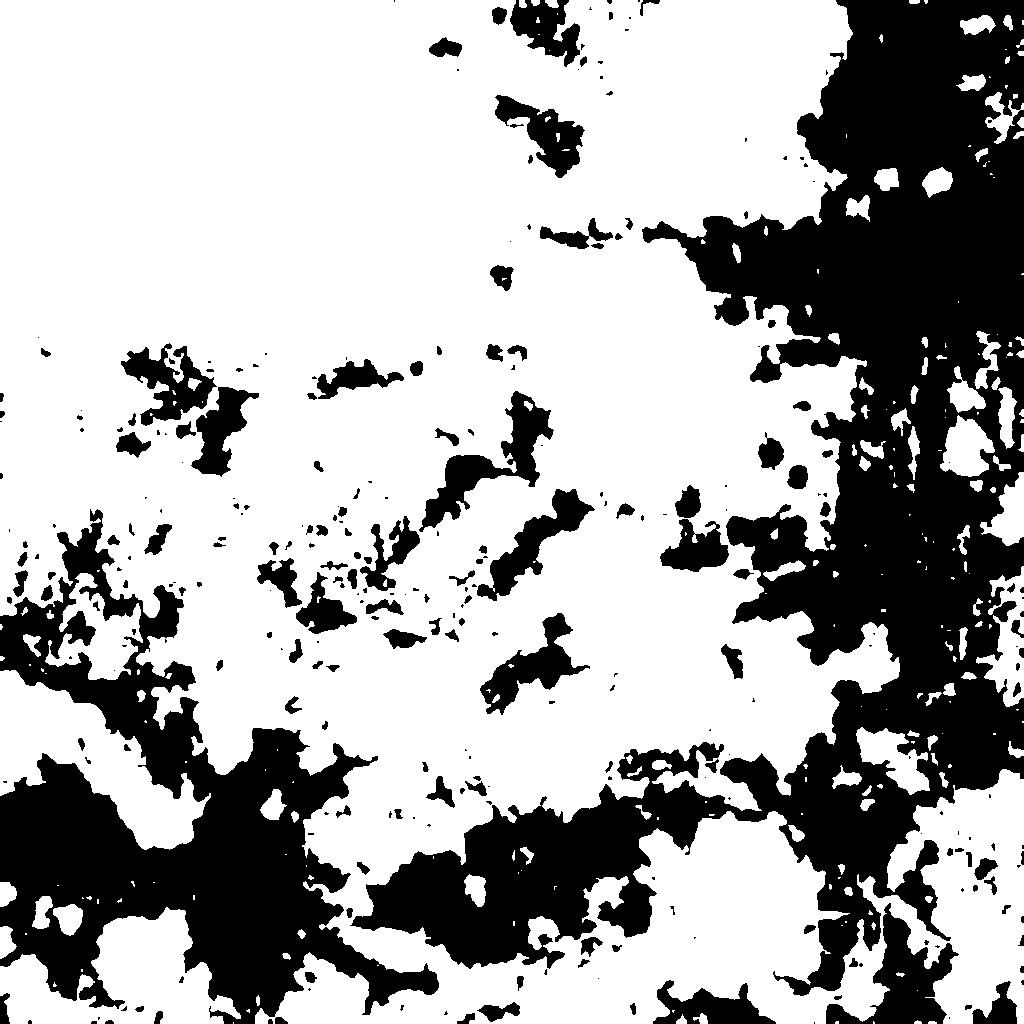}
			\scriptsize{RSAM-Seg}
		\end{minipage}
		\begin{minipage}[b]{0.10\textwidth}
			\centering
			\includegraphics[width=\textwidth]{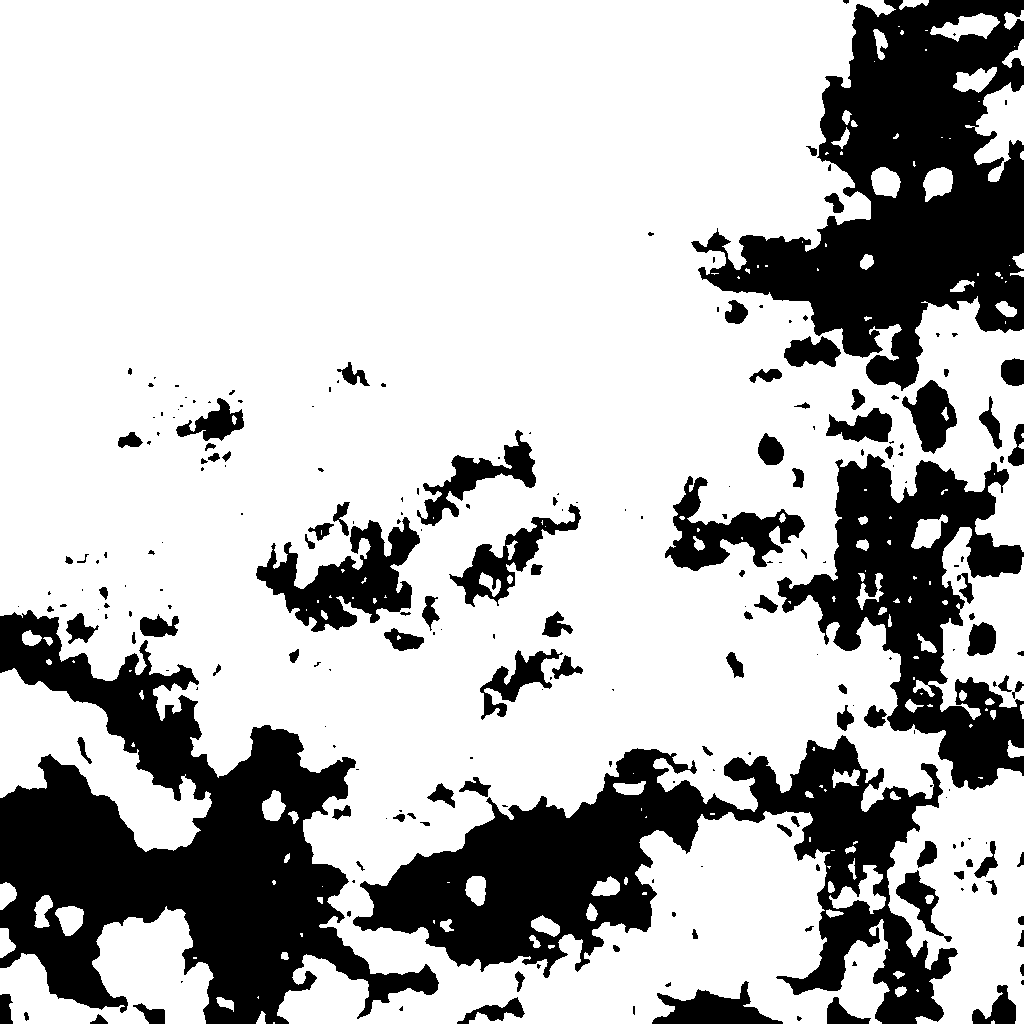}
			\scriptsize{w/o $\mathrm{F}_{\mathrm{hfc}}$}
		\end{minipage}
		\begin{minipage}[b]{0.10\textwidth}
			\centering
			\includegraphics[width=\textwidth]{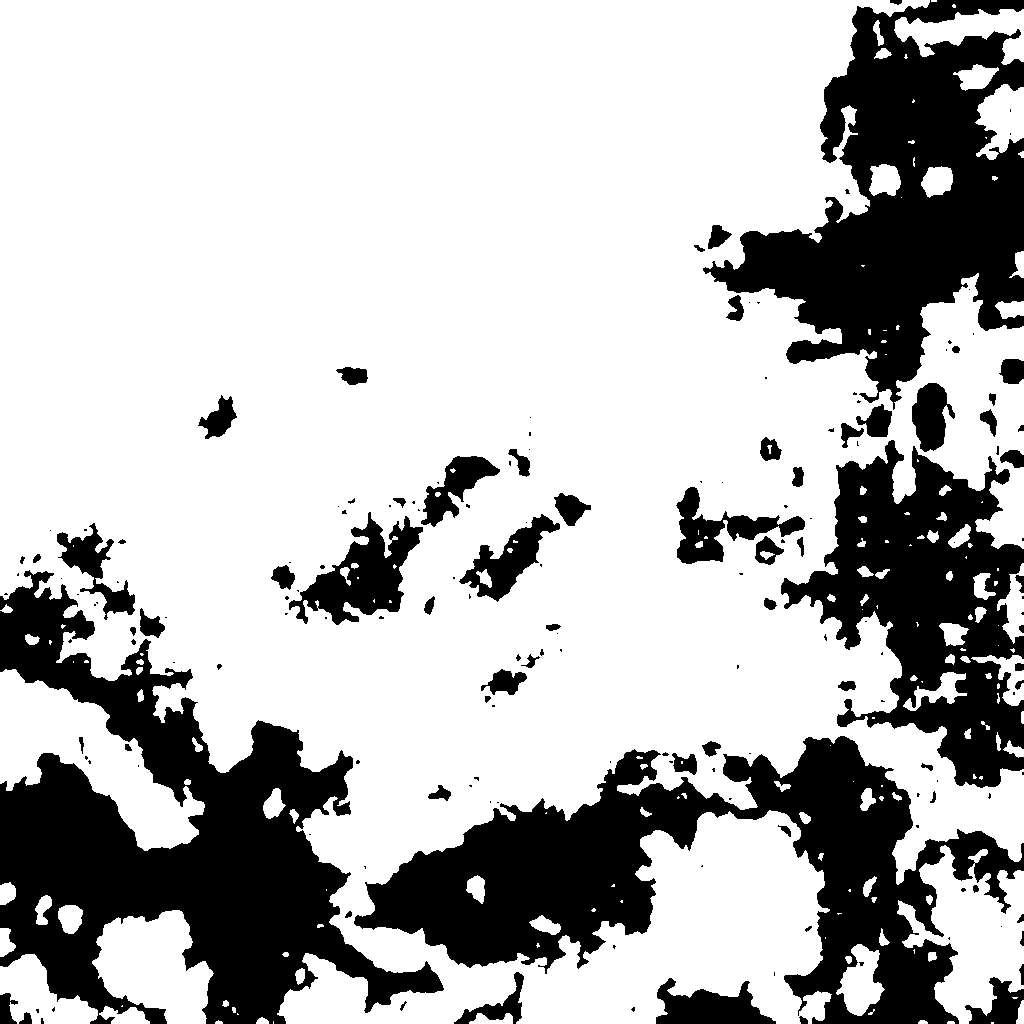}
			\scriptsize{w/o $\mathrm{F}_{\mathrm{pe}}$}
		\end{minipage}
		\begin{minipage}[b]{0.10\textwidth}
			\centering
			\includegraphics[width=\textwidth]{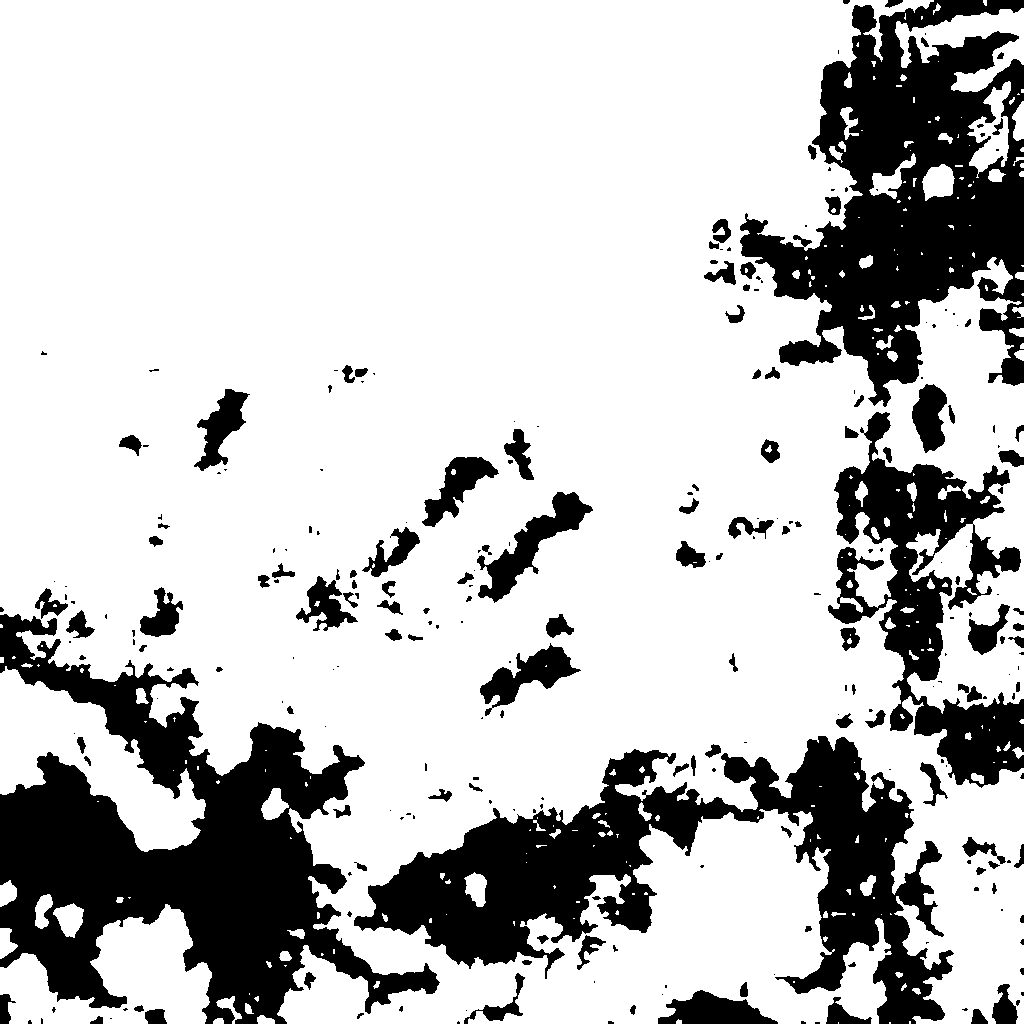}
			\tiny{w/o Adapter-Scale}
		\end{minipage}
		\caption{Visualization results on the ablation study for RSAM-Seg on 38-Cloud dataset.}
		\label{fig:AblationQuality}
	\end{figure*}
	
	\subsubsection{Results in the road scenario}
	The quantitative results in the road scenario are summarized in Table \ref{tab.RSAMRES} and the visualization results are listed in Figure \ref{fig:DGRoadQuality}. 
	
	The "Road" row in Table \ref{tab.RSAMRES} reveals RSAM-Seg outperforms the baseline with an 11\% improvement in F1 score and SAM exhibits suboptimal performance around 45\% in mIoU, which possibly attributed to the narrowness of the road and indistinct demarcation from the surrounding environment.
	
	Considering the images in the third row of Figure \ref{fig:DGRoadQuality}, RSAM-Seg demonstrates the ability to distinguish densely roads, presenting well-defined and more complete road segments. SAM cannot effectively distinguish roads and is easily disrupted by surrounding environments, such as farmland, as seen in the experiments. Additionally, the U-Net is susceptible to misclassification of similar road-like areas, such as gaps between farmland or buildings, when segmenting roads.
	
	After conducting experiments on datasets from various remote sensing domains, it can be observed that SAM has the limitations that rely heavily on human annotations or prompts. However RSAM-Seg can perform significantly better than the original SAM approach and automate the segmentation process without the need for manually annotated data or prompts in specific remote sensing scenes. This modification enables SAM to better adapt to segmentation tasks in remote sensing imagery, making it a valuable tool for a wide range of remote sensing applications.
	
	\subsection{Ablation study}
	In order to systematically assess the contribution of different components in our proposed approach, an ablation study is conducted on 38-Cloud dataset and results are presented in Table \ref{tab.ablation}.
	
	\begin{figure*}[!htb]
		\centering
		\begin{minipage}[b]{0.10\textwidth}
			\centering
			\includegraphics[width=\textwidth]{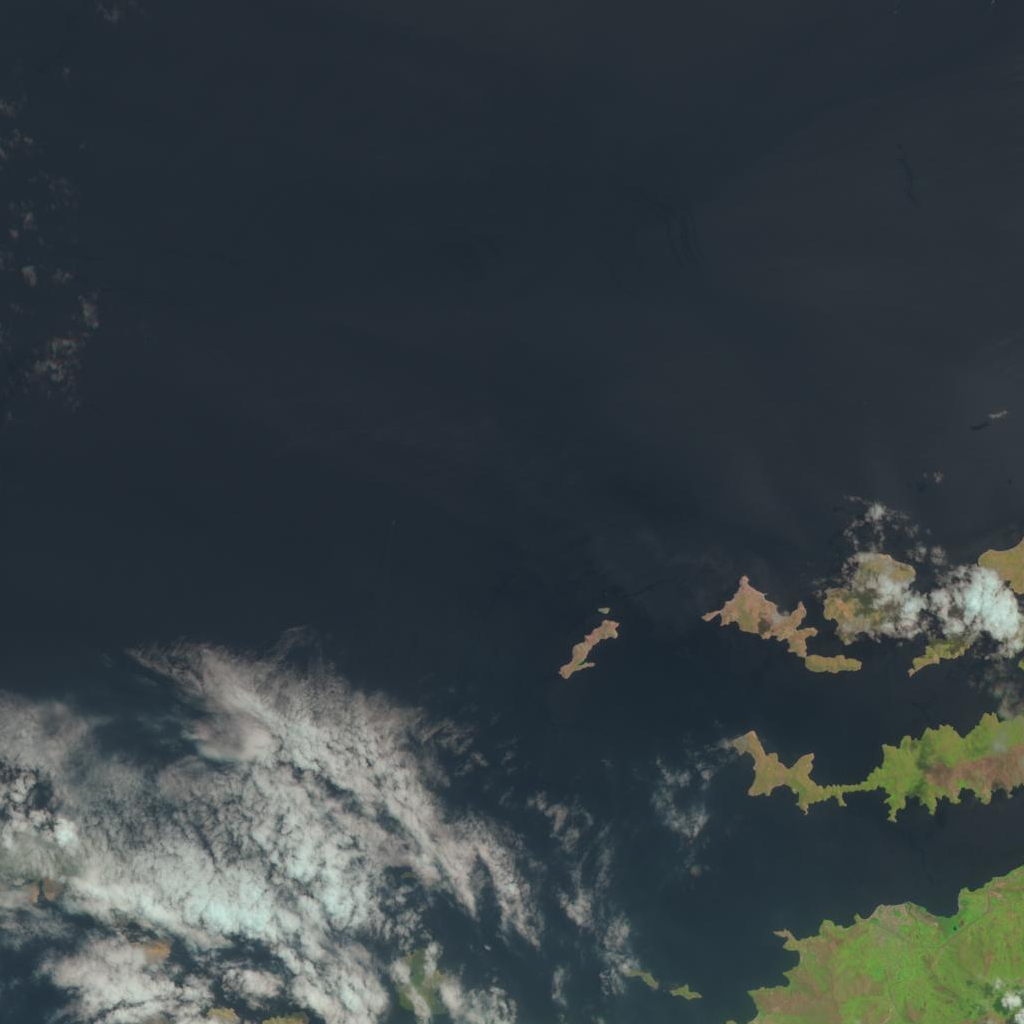}
		\end{minipage}
		\begin{minipage}[b]{0.10\textwidth}
			\centering
			\includegraphics[width=\textwidth]{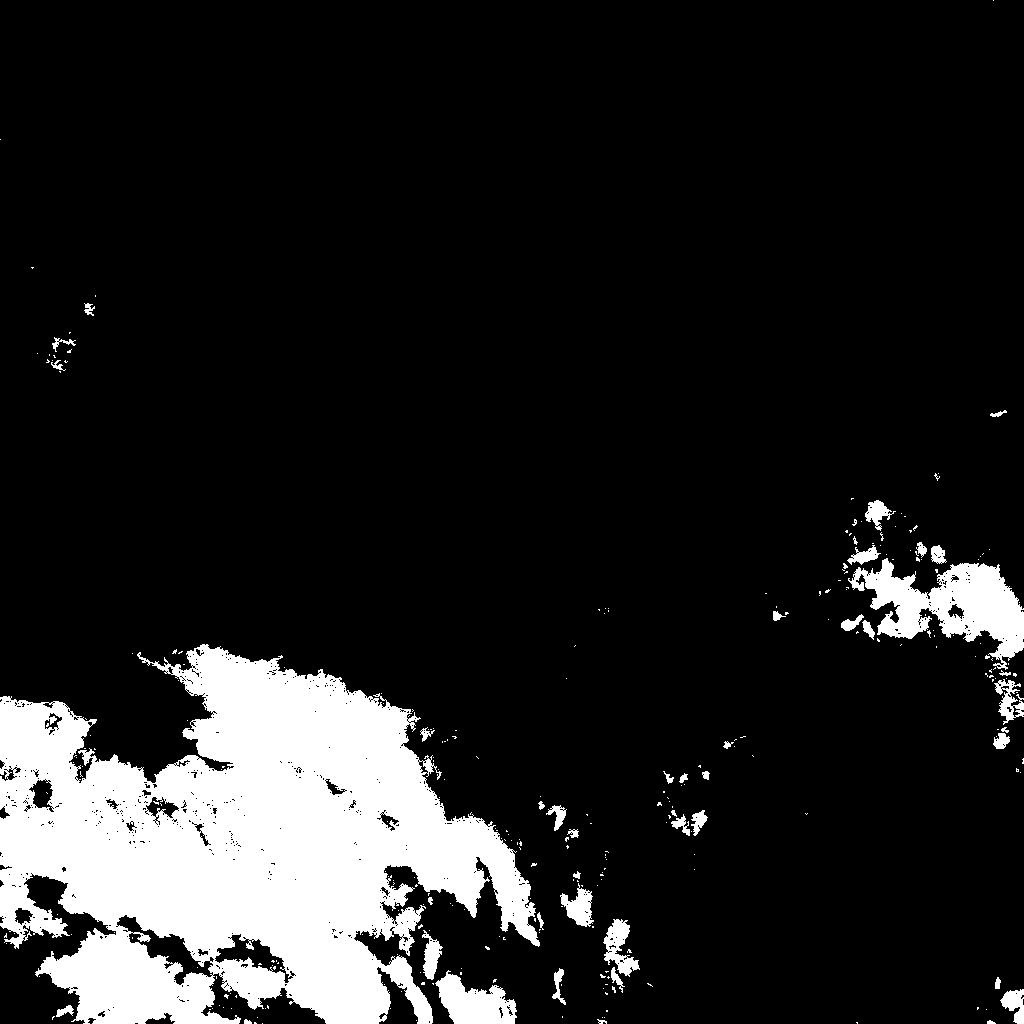}
		\end{minipage}
		\begin{minipage}[b]{0.10\textwidth}
			\centering
			\includegraphics[width=\textwidth]{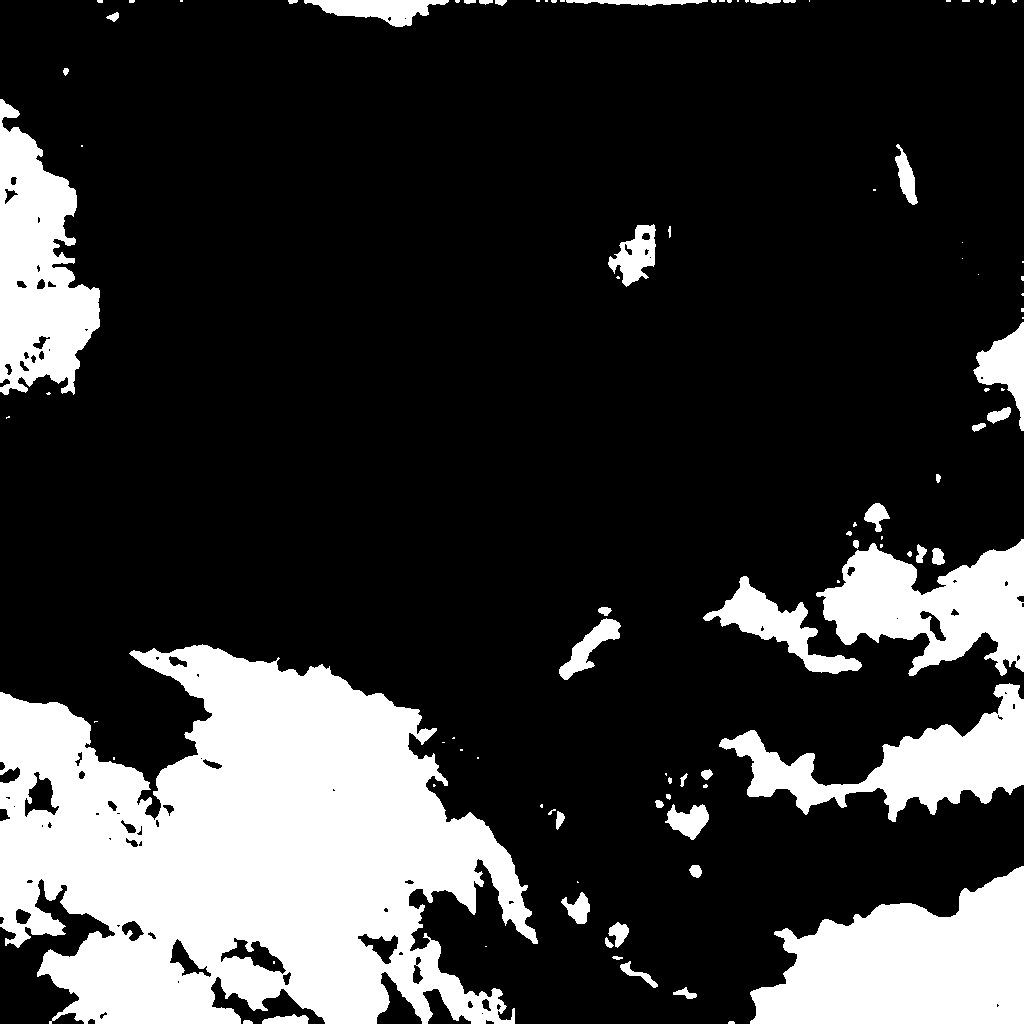}
		\end{minipage}
		\begin{minipage}[b]{0.10\textwidth}
			\centering
			\includegraphics[width=\textwidth]{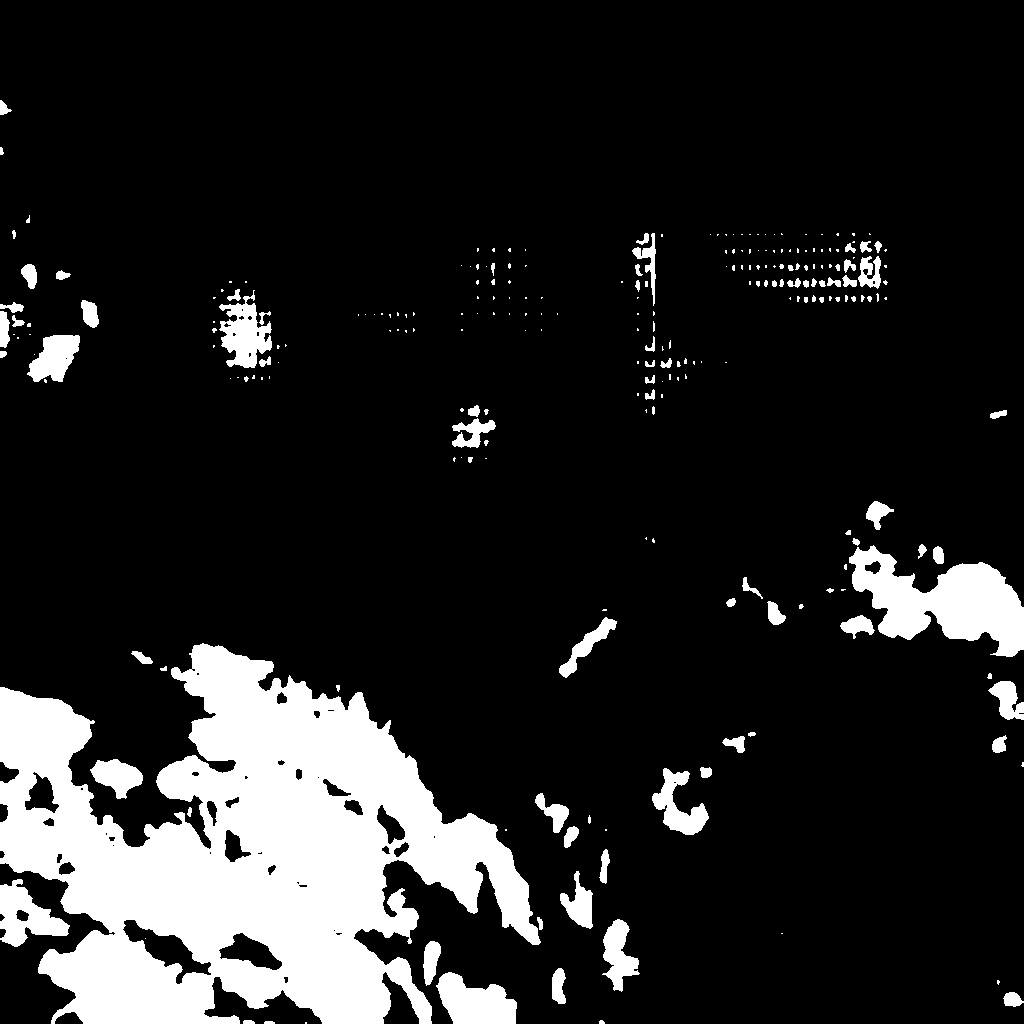}
		\end{minipage}
		\begin{minipage}[b]{0.10\textwidth}
			\centering
			\includegraphics[width=\textwidth]{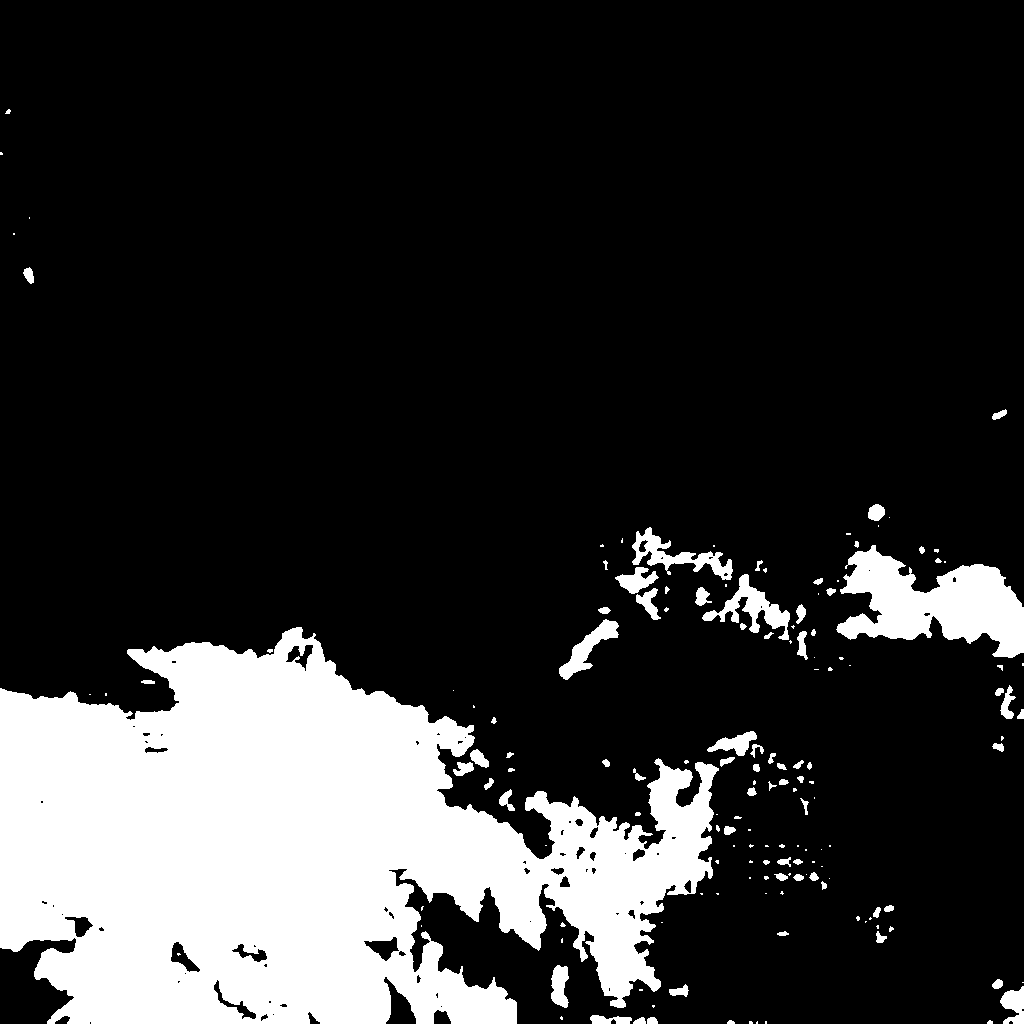}
		\end{minipage}
		\begin{minipage}[b]{0.10\textwidth}
			\centering
			\includegraphics[width=\textwidth]{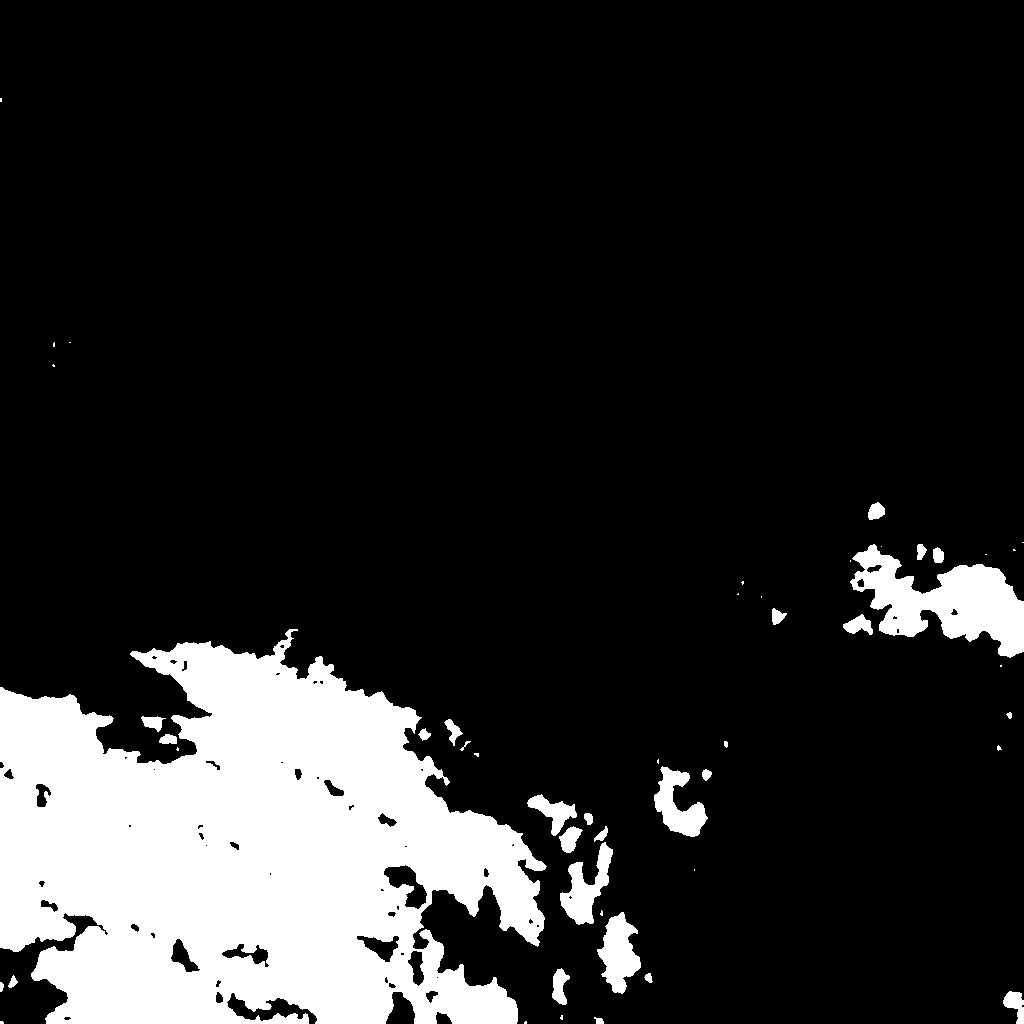}
		\end{minipage}\\[3pt]
		\begin{minipage}[b]{0.10\textwidth}
			\centering
			\includegraphics[width=\textwidth]{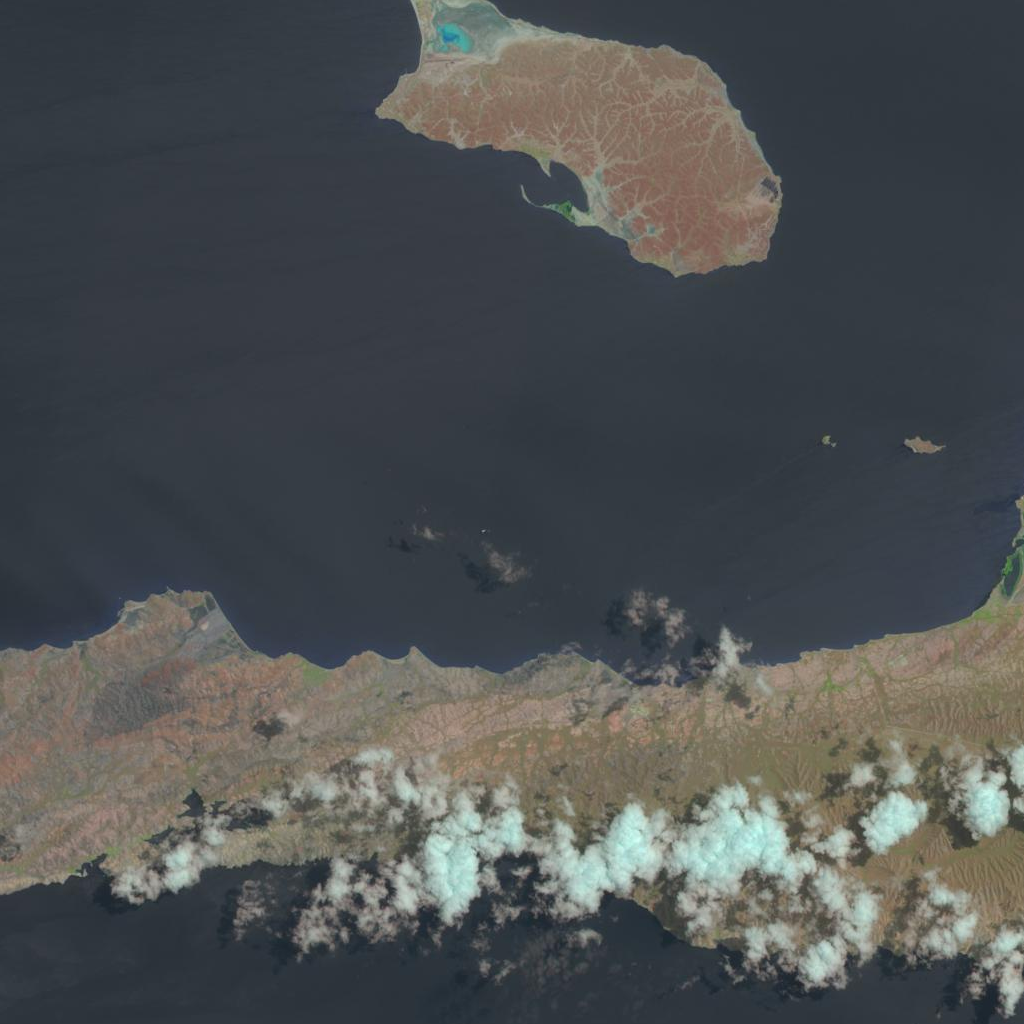}
		\end{minipage}
		\begin{minipage}[b]{0.10\textwidth}
			\centering
			\includegraphics[width=\textwidth]{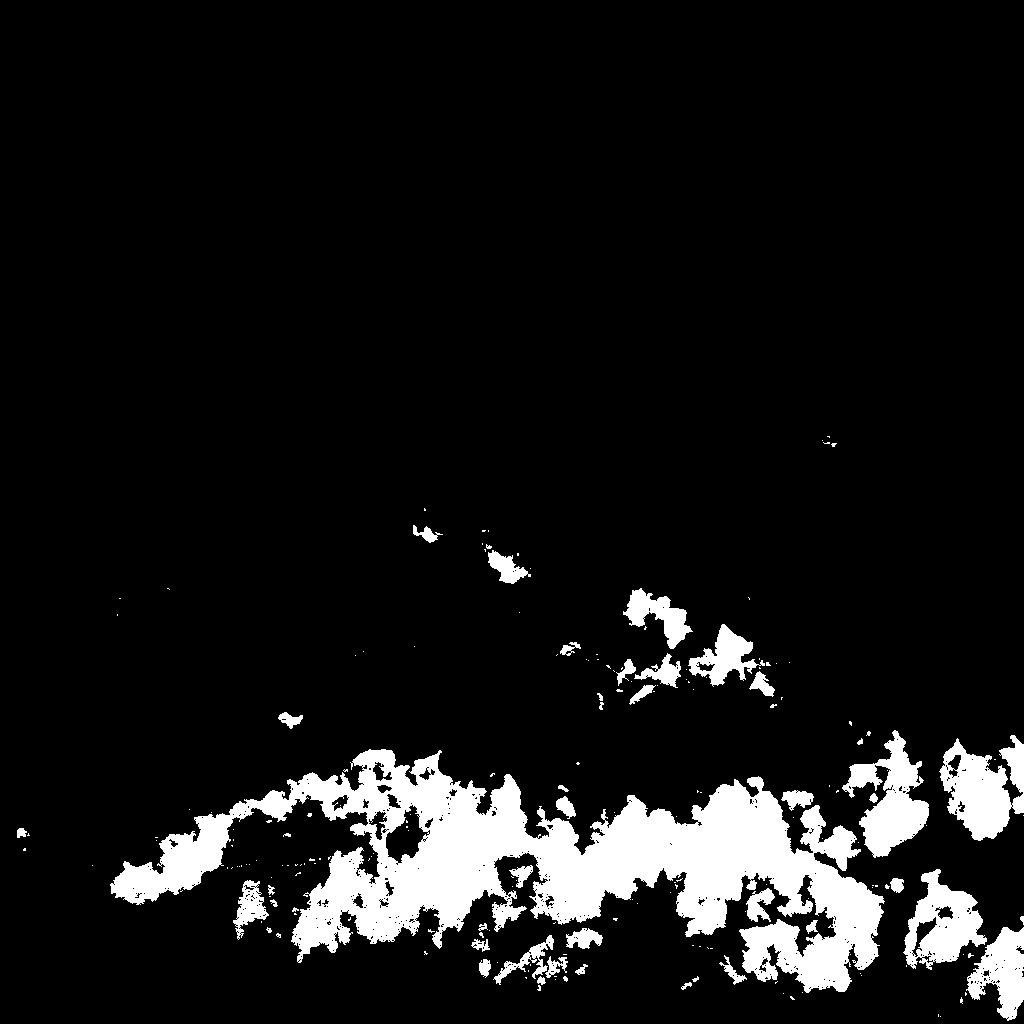}
		\end{minipage}
		\begin{minipage}[b]{0.10\textwidth}
			\centering
			\includegraphics[width=\textwidth]{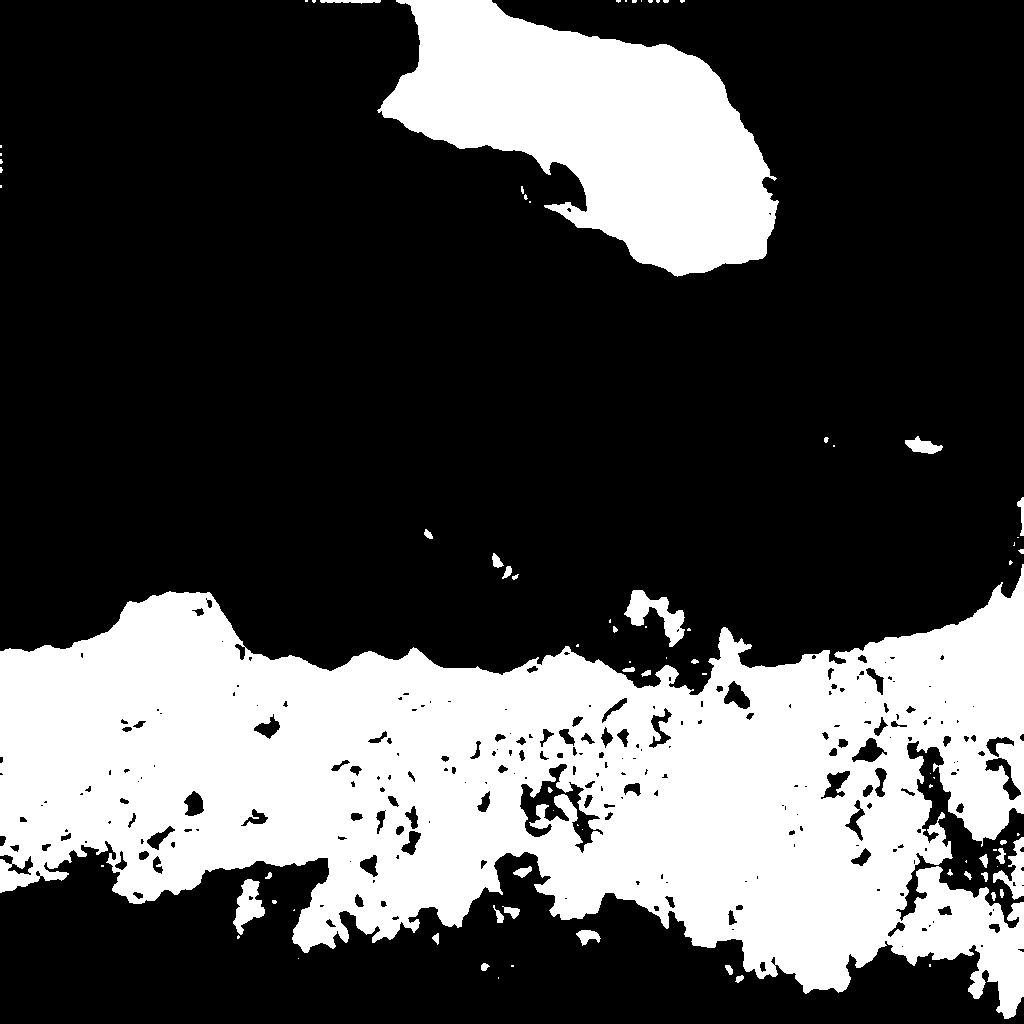}
		\end{minipage}
		\begin{minipage}[b]{0.10\textwidth}
			\centering
			\includegraphics[width=\textwidth]{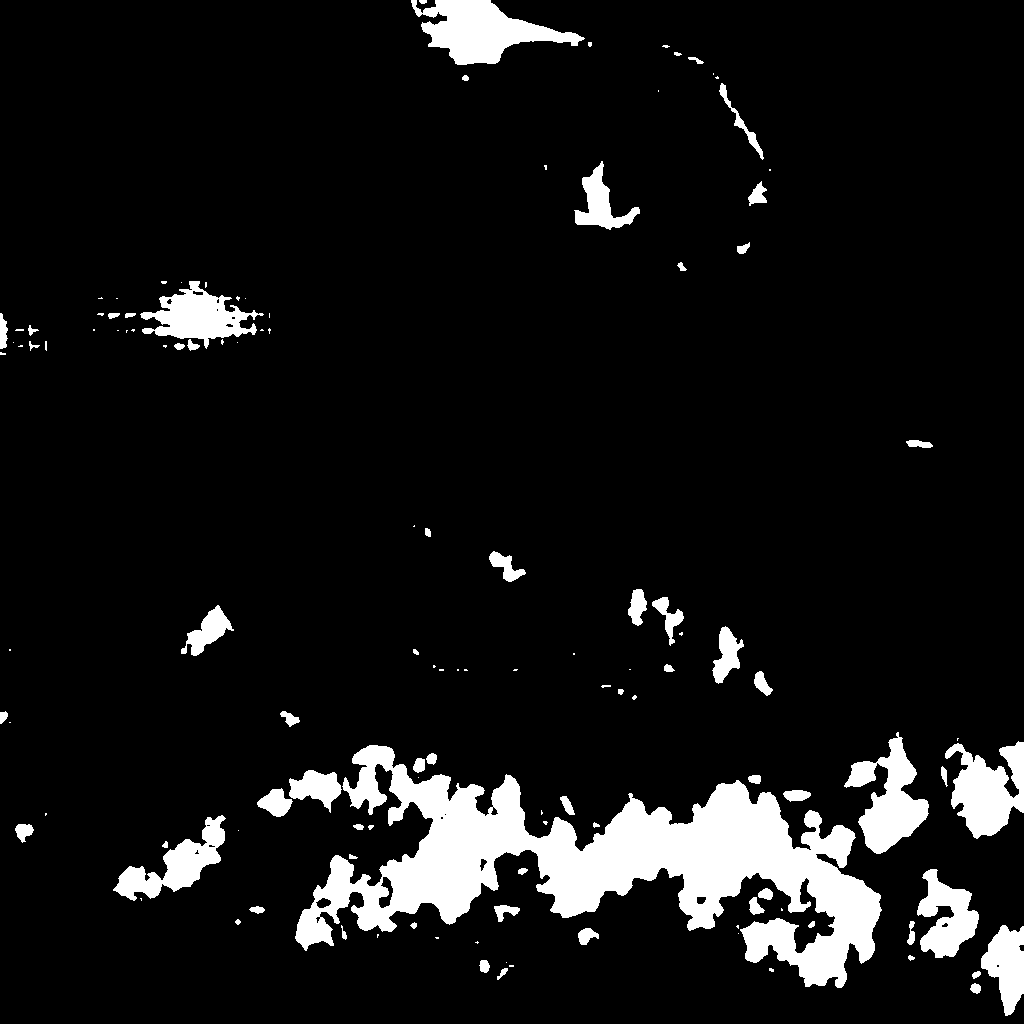}
		\end{minipage}
		\begin{minipage}[b]{0.10\textwidth}
			\centering
			\includegraphics[width=\textwidth]{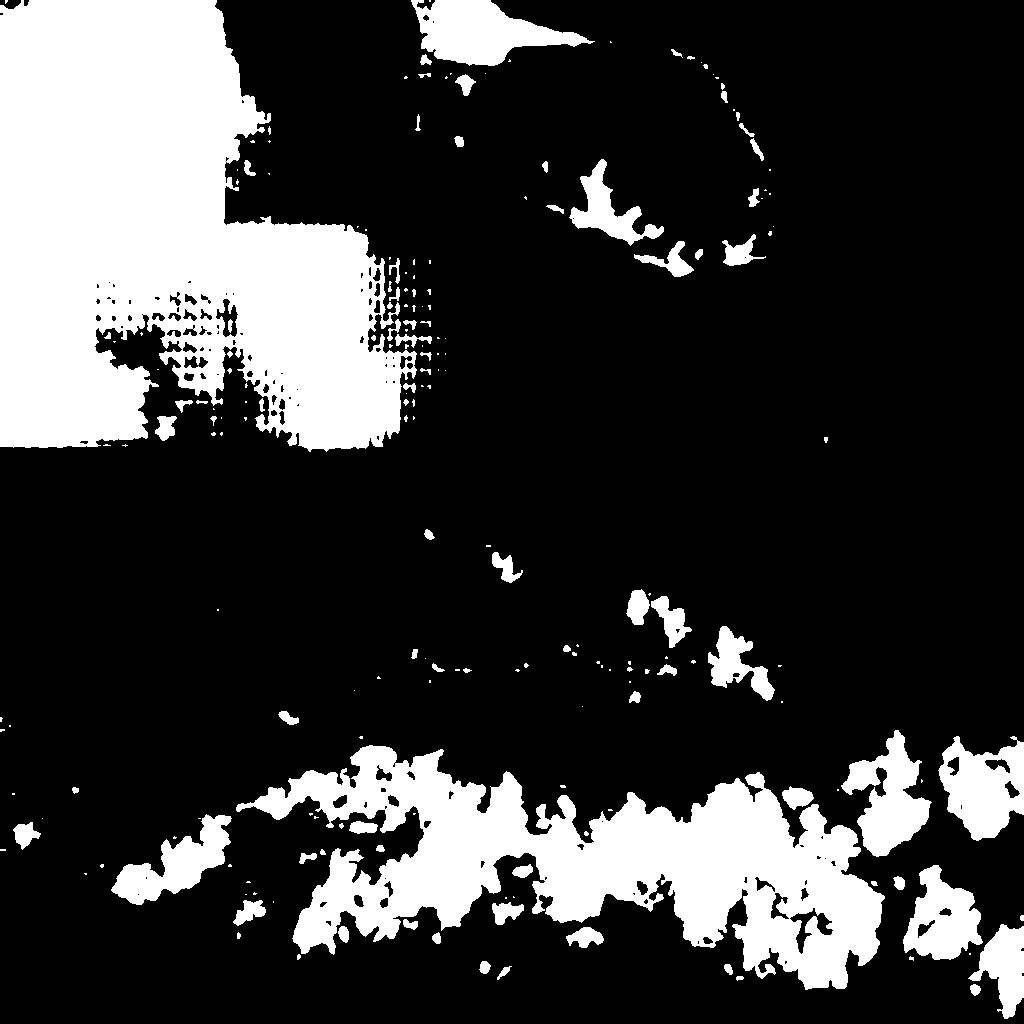}
		\end{minipage}
		\begin{minipage}[b]{0.10\textwidth}
			\centering
			\includegraphics[width=\textwidth]{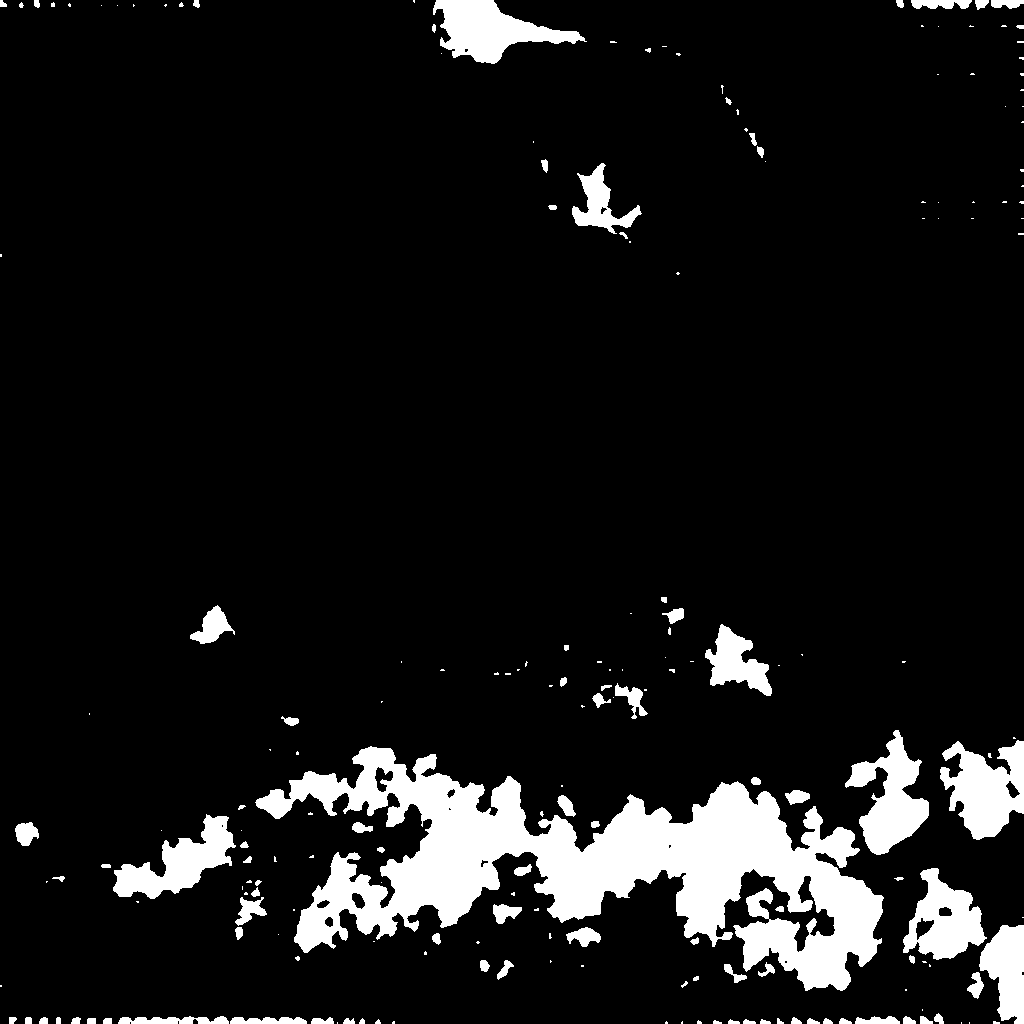}
		\end{minipage}\\[3pt]
		\begin{minipage}[b]{0.10\textwidth}
			\centering
			\includegraphics[width=\textwidth]{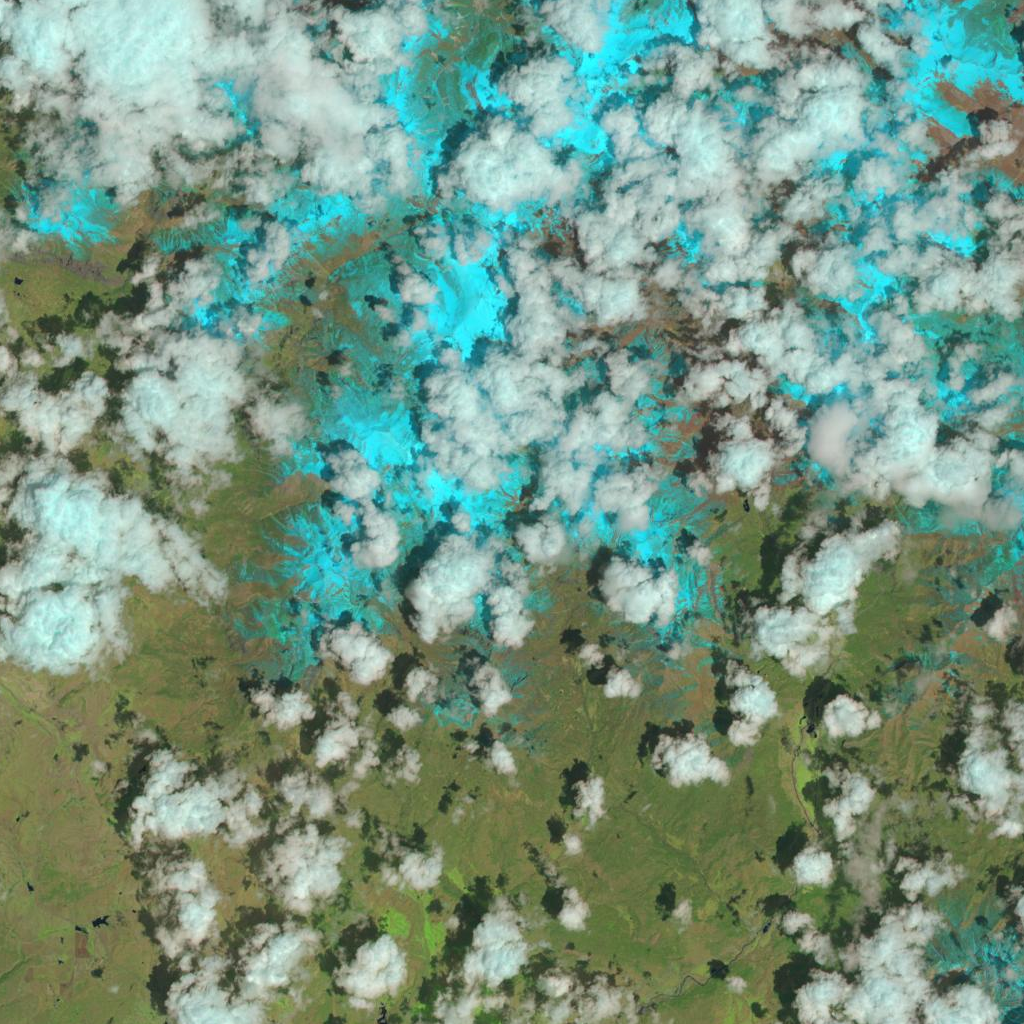}
			\scriptsize{Original Image}
		\end{minipage}
		\begin{minipage}[b]{0.10\textwidth}
			\centering
			\includegraphics[width=\textwidth]{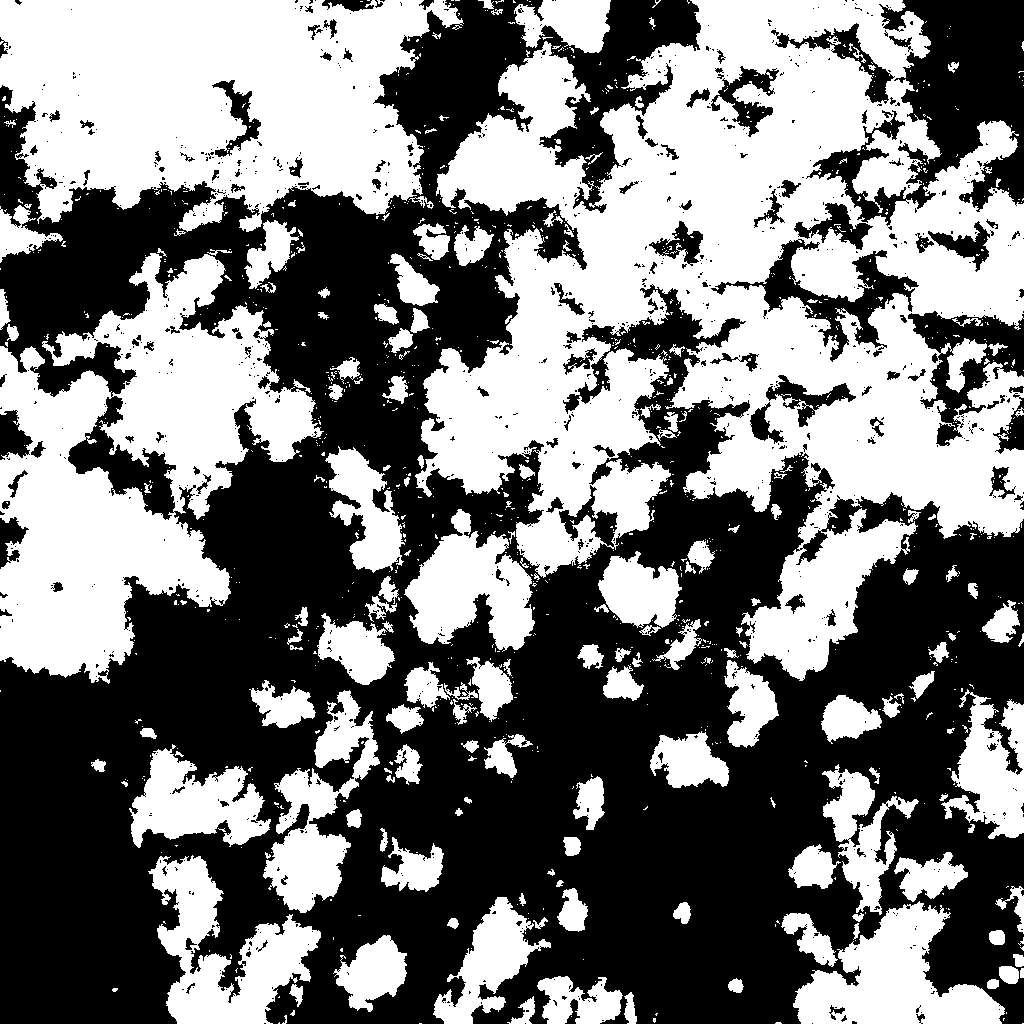}
			\scriptsize{Ground Truth}
		\end{minipage}
		\begin{minipage}[b]{0.10\textwidth}
			\centering
			\includegraphics[width=\textwidth]{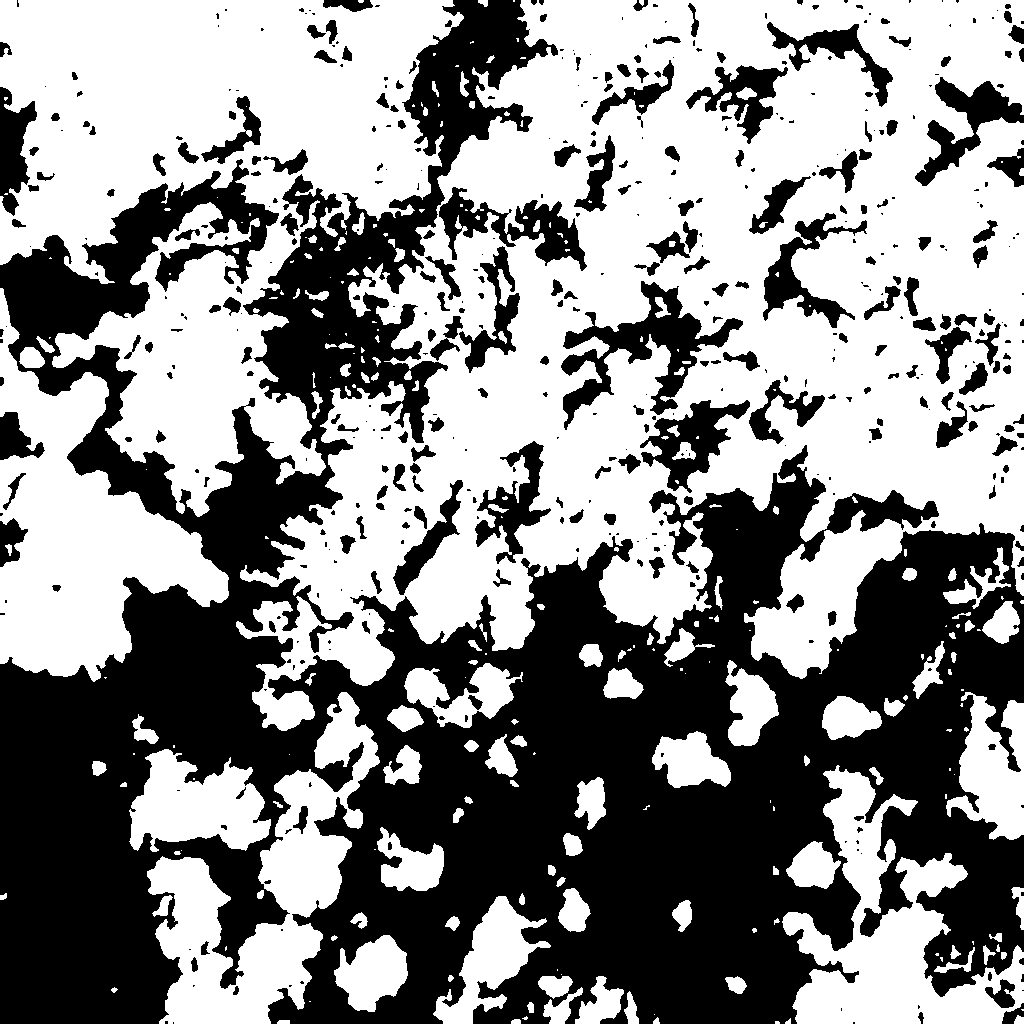}
			\scriptsize{1\% training set}
		\end{minipage}
		\begin{minipage}[b]{0.10\textwidth}
			\centering
			\includegraphics[width=\textwidth]{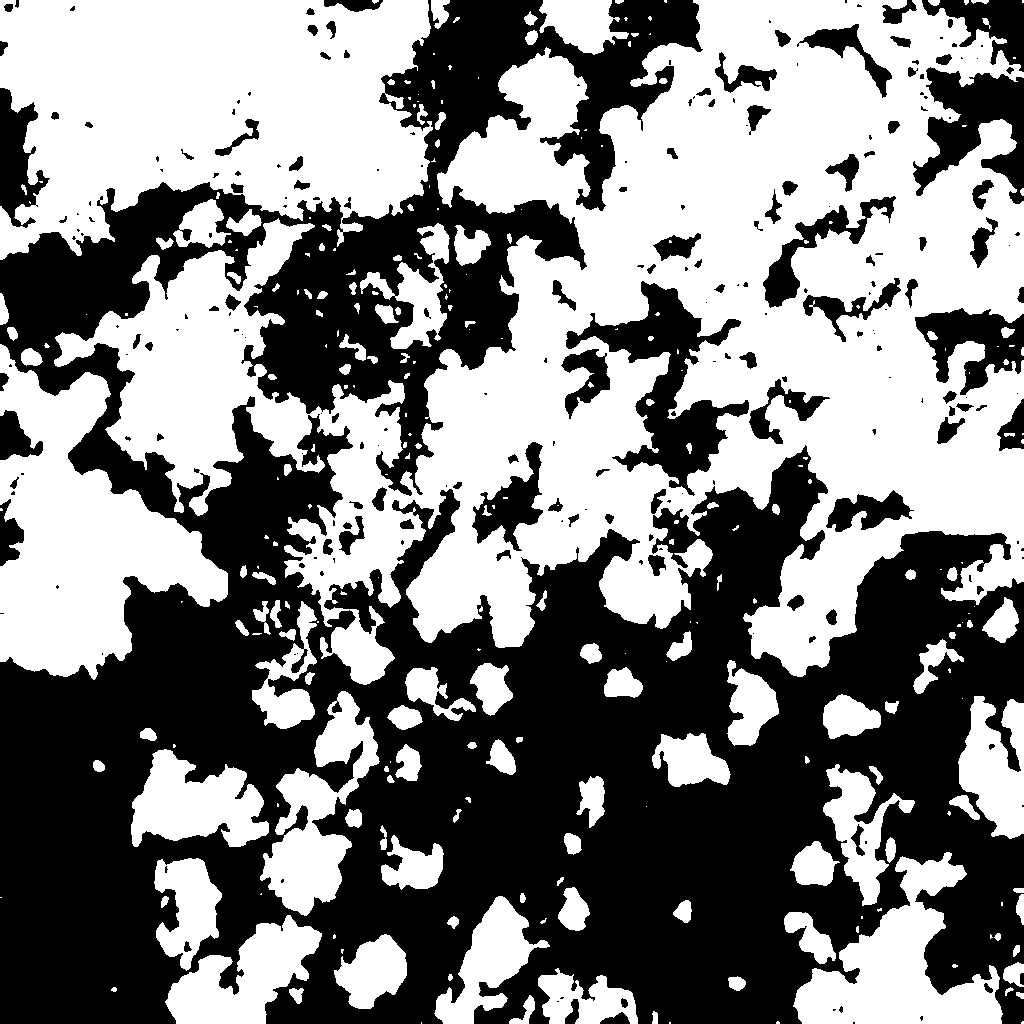}
			\scriptsize{10\% training set}
		\end{minipage}
		\begin{minipage}[b]{0.10\textwidth}
			\centering
			\includegraphics[width=\textwidth]{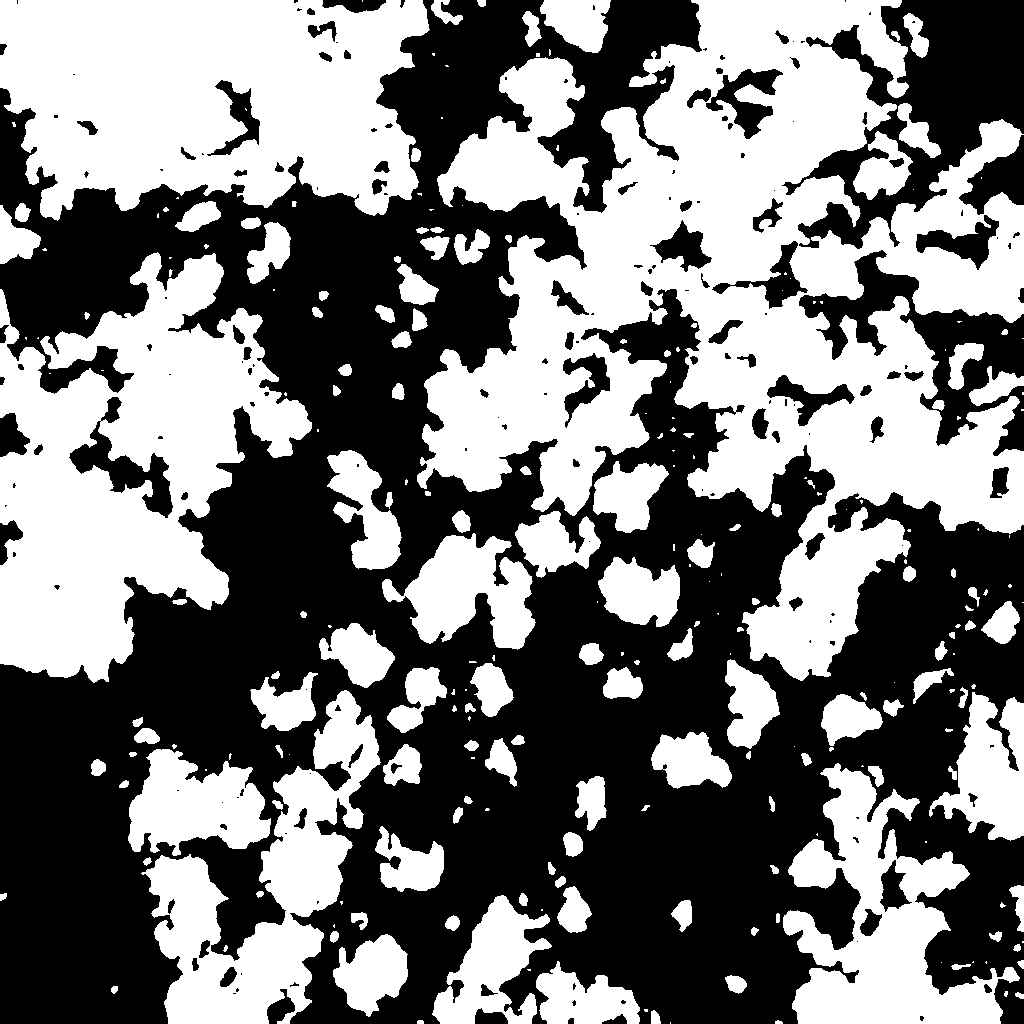}
			\scriptsize{30\% training set}
		\end{minipage}
		\begin{minipage}[b]{0.10\textwidth}
			\centering
			\includegraphics[width=\textwidth]{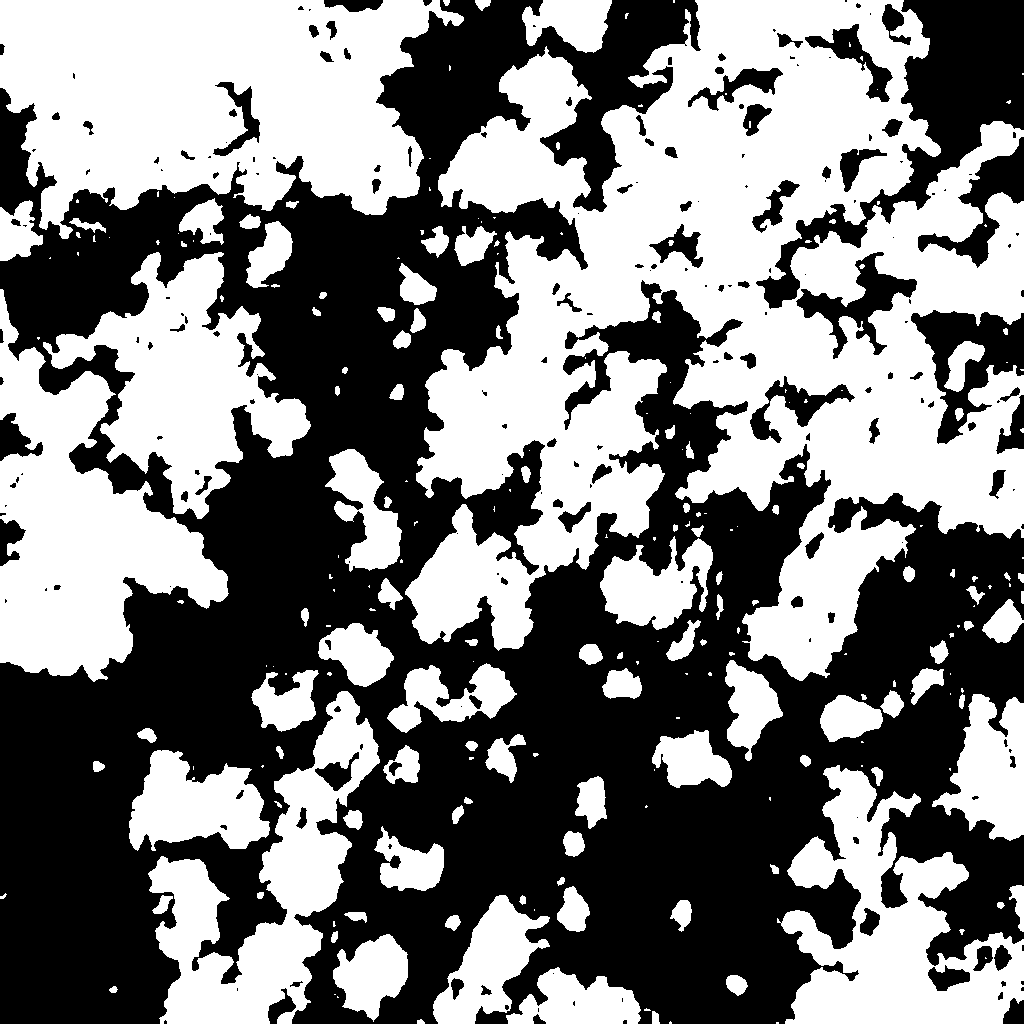}
			\scriptsize{70\% training set}
		\end{minipage}
		\caption{Examples of few-shot segmentation results on 38-Cloud dataset.}
		\label{fig:discussionQuality}
	\end{figure*}
	
	\begin{figure*}[!htb]
		\centering
		\begin{minipage}[b]{0.10\textwidth}
			\centering
			\includegraphics[width=\textwidth]{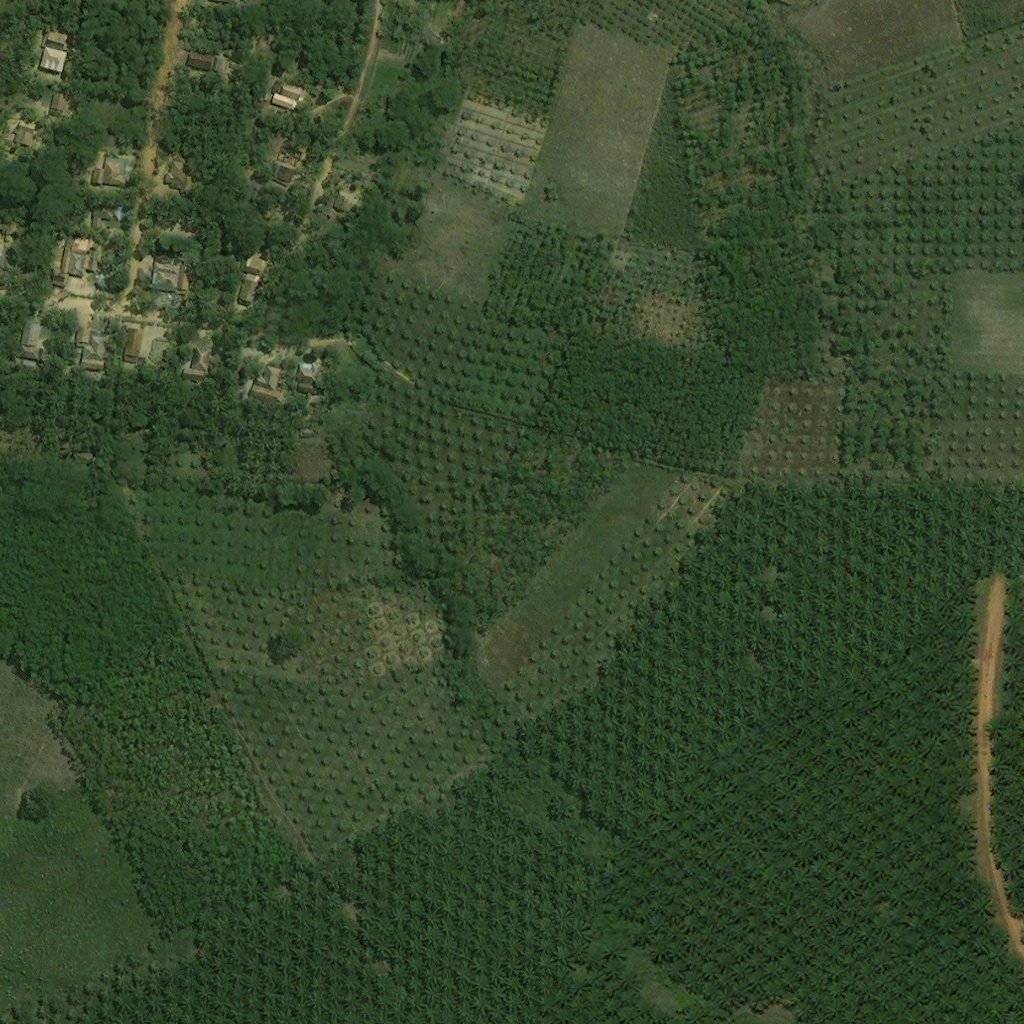}
		\end{minipage}
		\begin{minipage}[b]{0.10\textwidth}
			\centering
			\includegraphics[width=\textwidth]{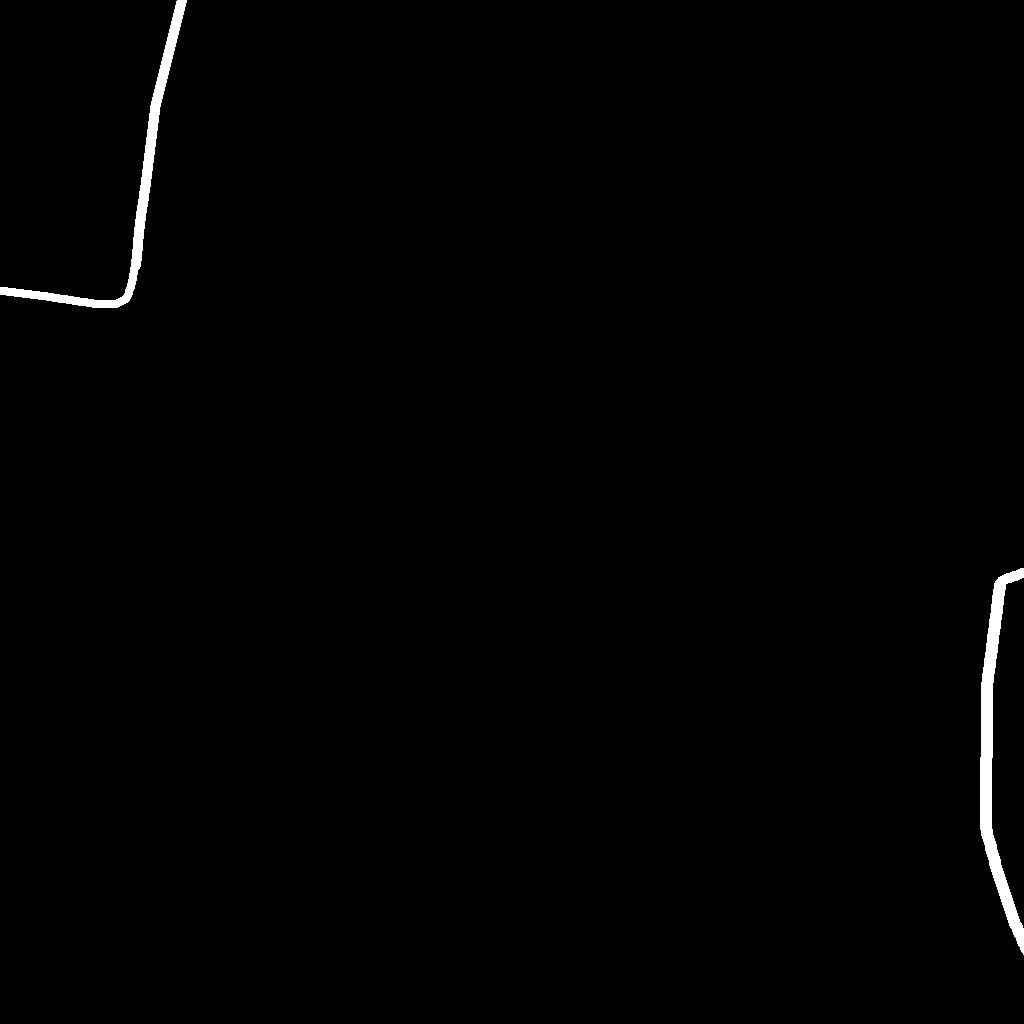}
		\end{minipage}
		\begin{minipage}[b]{0.10\textwidth}
			\centering
			\includegraphics[width=\textwidth]{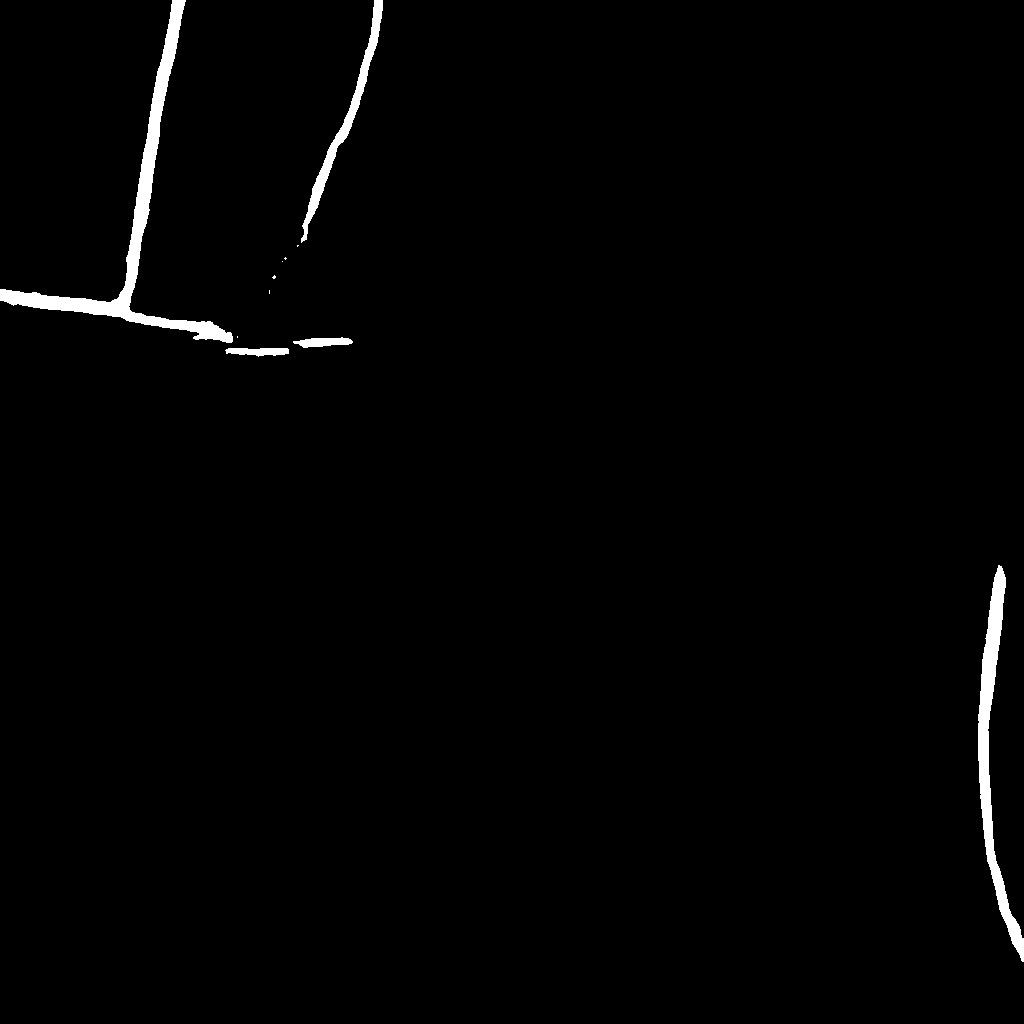}
		\end{minipage}
		\begin{minipage}[b]{0.10\textwidth}
			\centering
			\includegraphics[width=\textwidth]{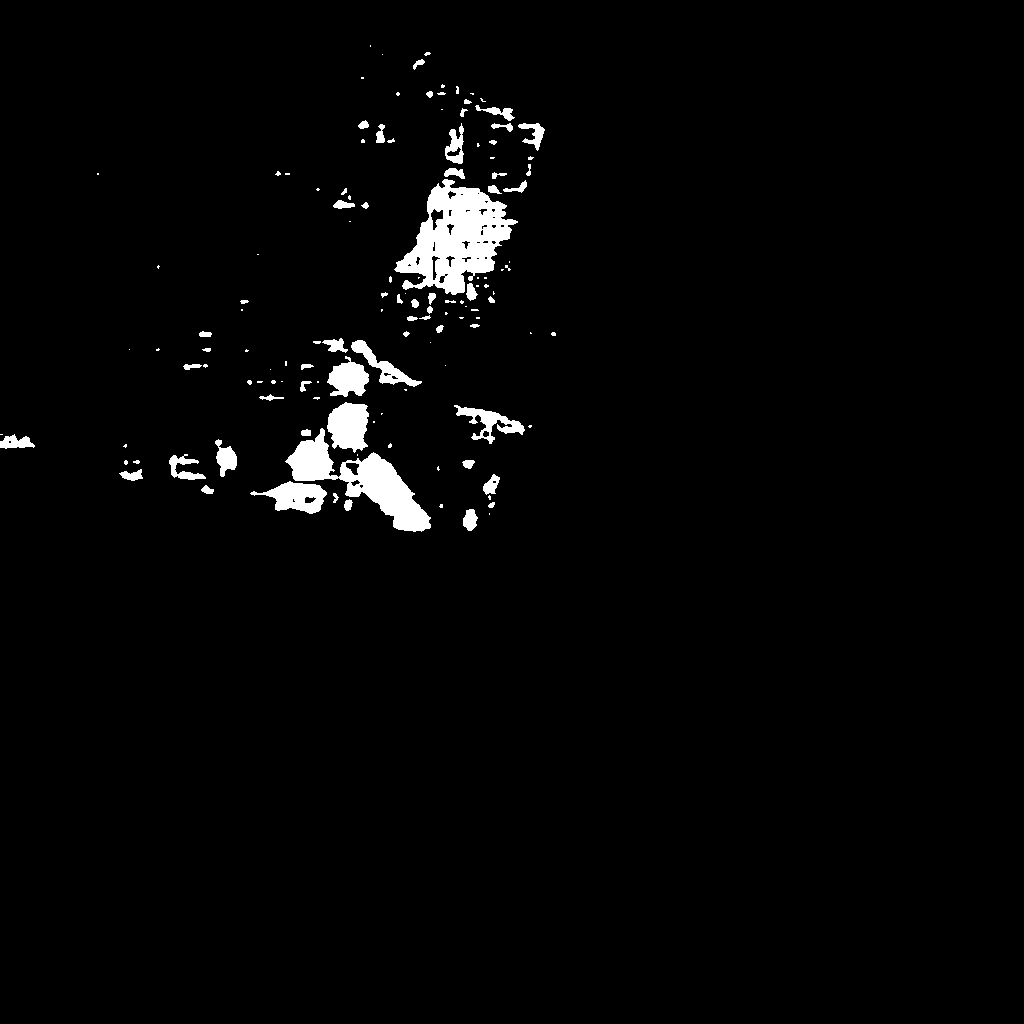}
		\end{minipage}
		\begin{minipage}[b]{0.10\textwidth}
			\centering
			\includegraphics[width=\textwidth]{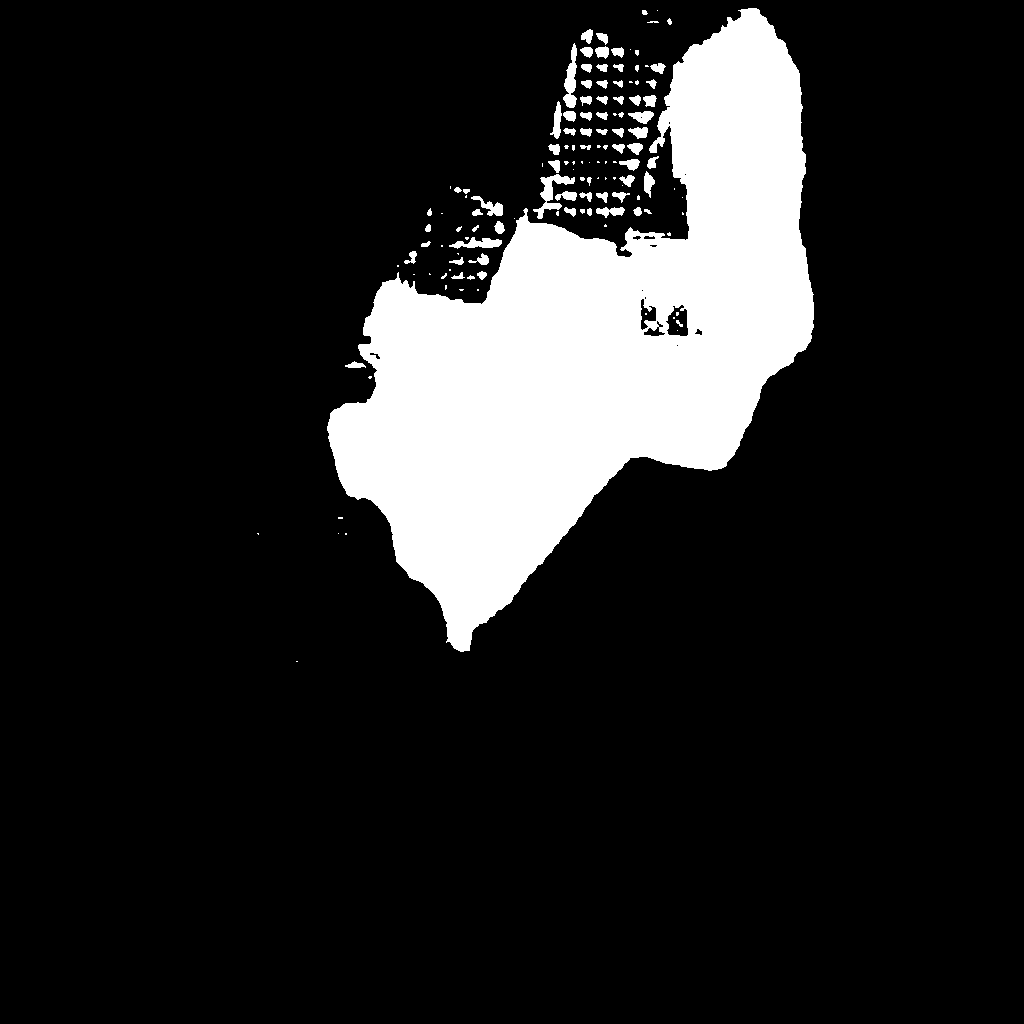}
		\end{minipage}
		\begin{minipage}[b]{0.10\textwidth}
			\centering
			\includegraphics[width=\textwidth]{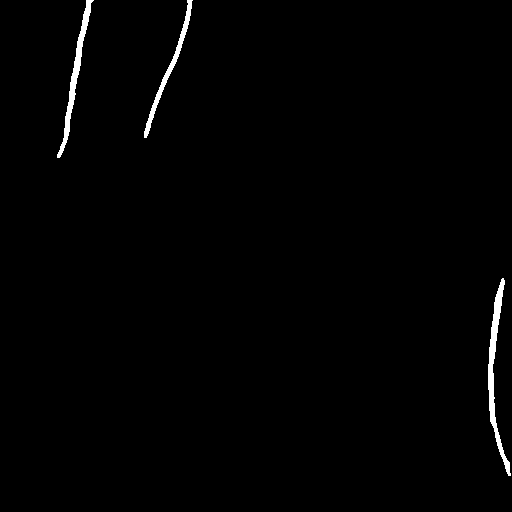}
		\end{minipage}\\[3pt]
		\begin{minipage}[b]{0.10\textwidth}
			\centering
			\includegraphics[width=\textwidth]{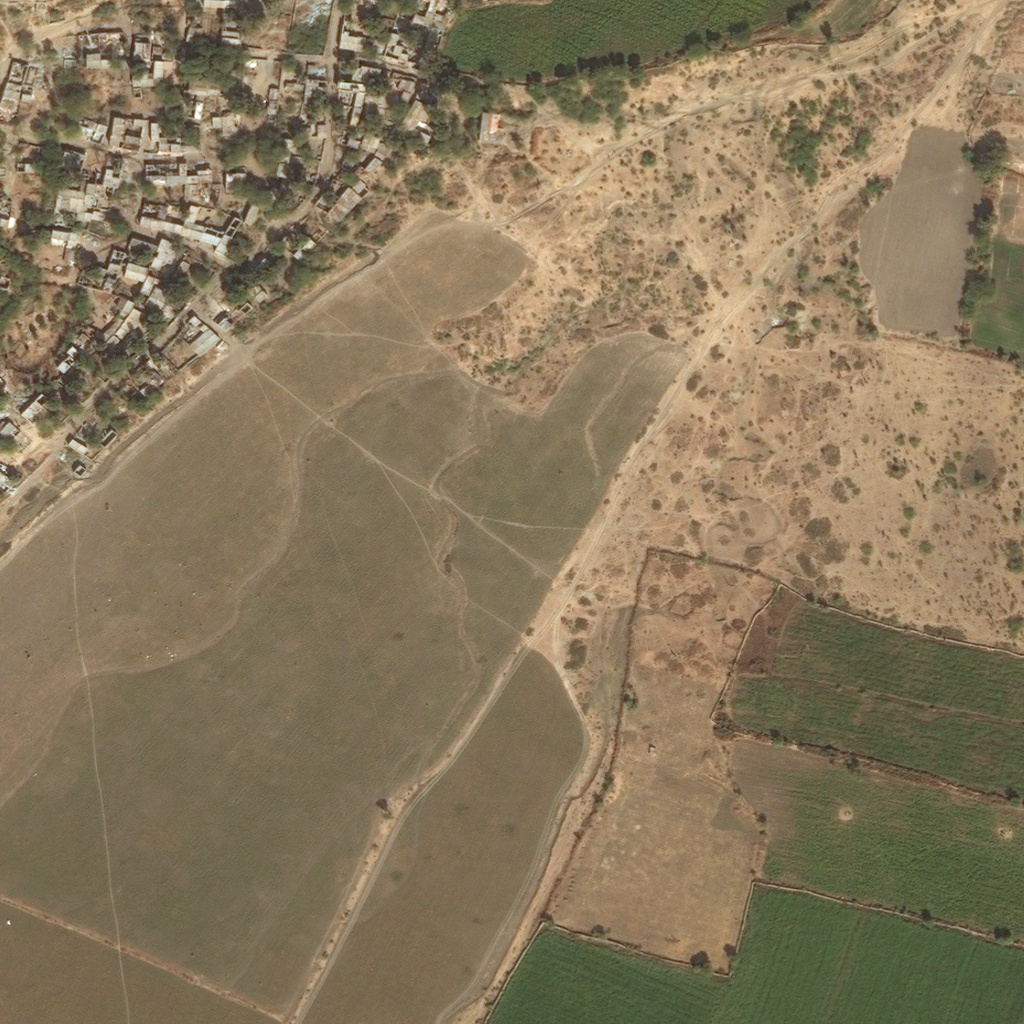}
		\end{minipage}
		\begin{minipage}[b]{0.10\textwidth}
			\centering
			\includegraphics[width=\textwidth]{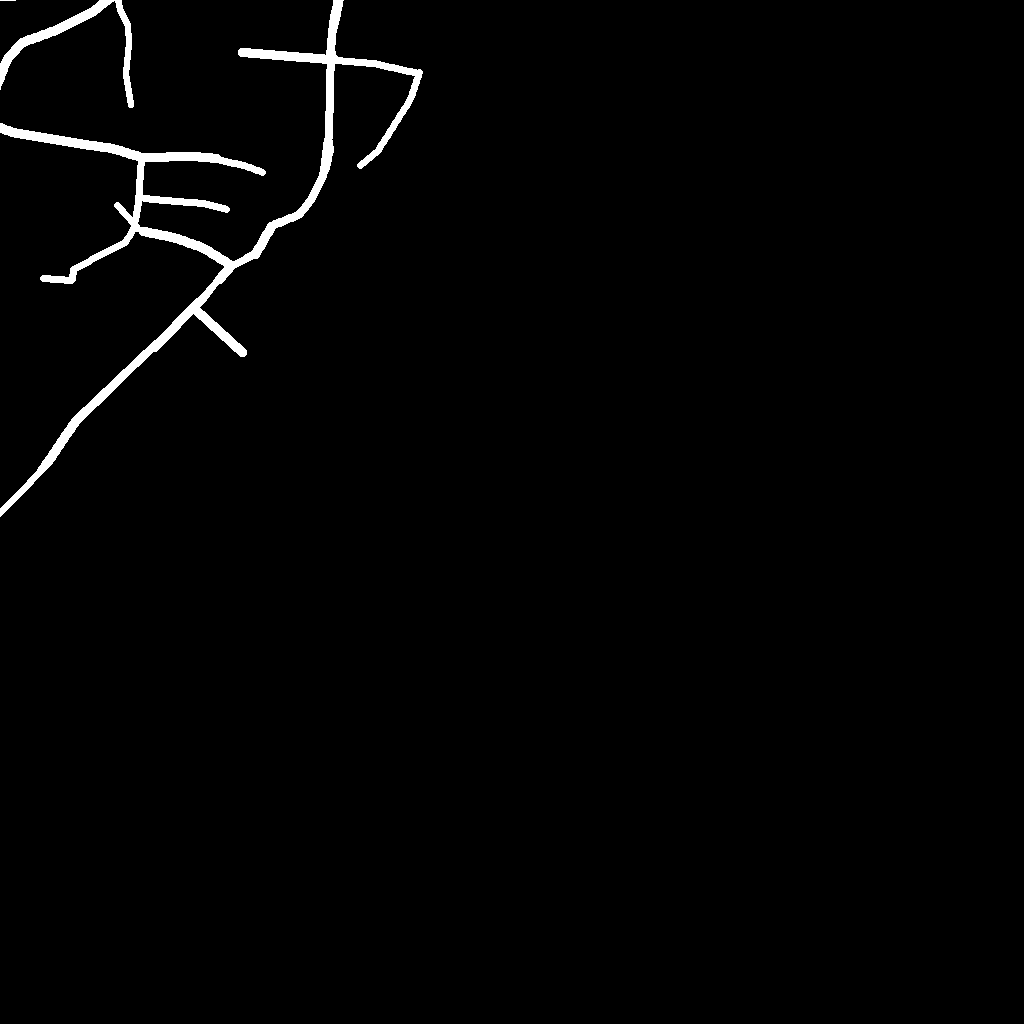}
		\end{minipage}
		\begin{minipage}[b]{0.10\textwidth}
			\centering
			\includegraphics[width=\textwidth]{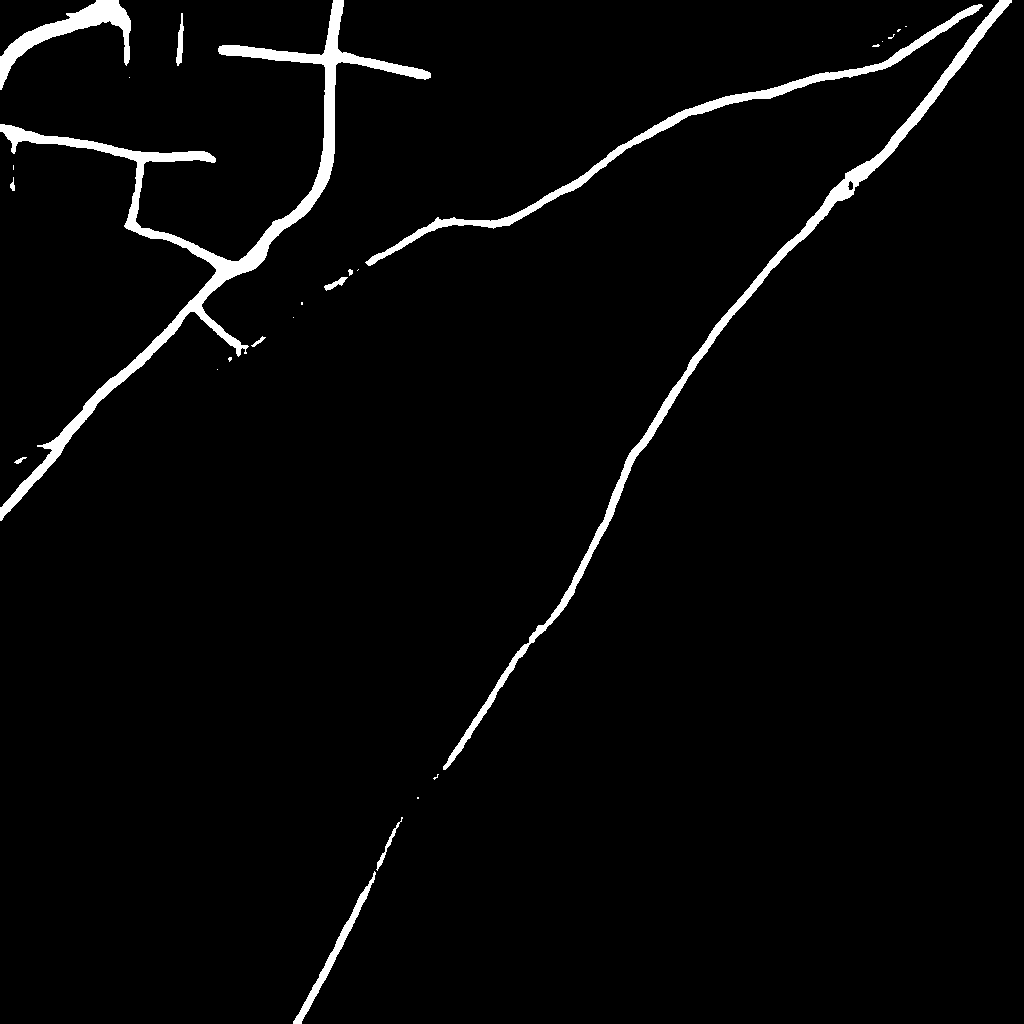}
		\end{minipage}
			\begin{minipage}[b]{0.10\textwidth}
			\centering
			\includegraphics[width=\textwidth]{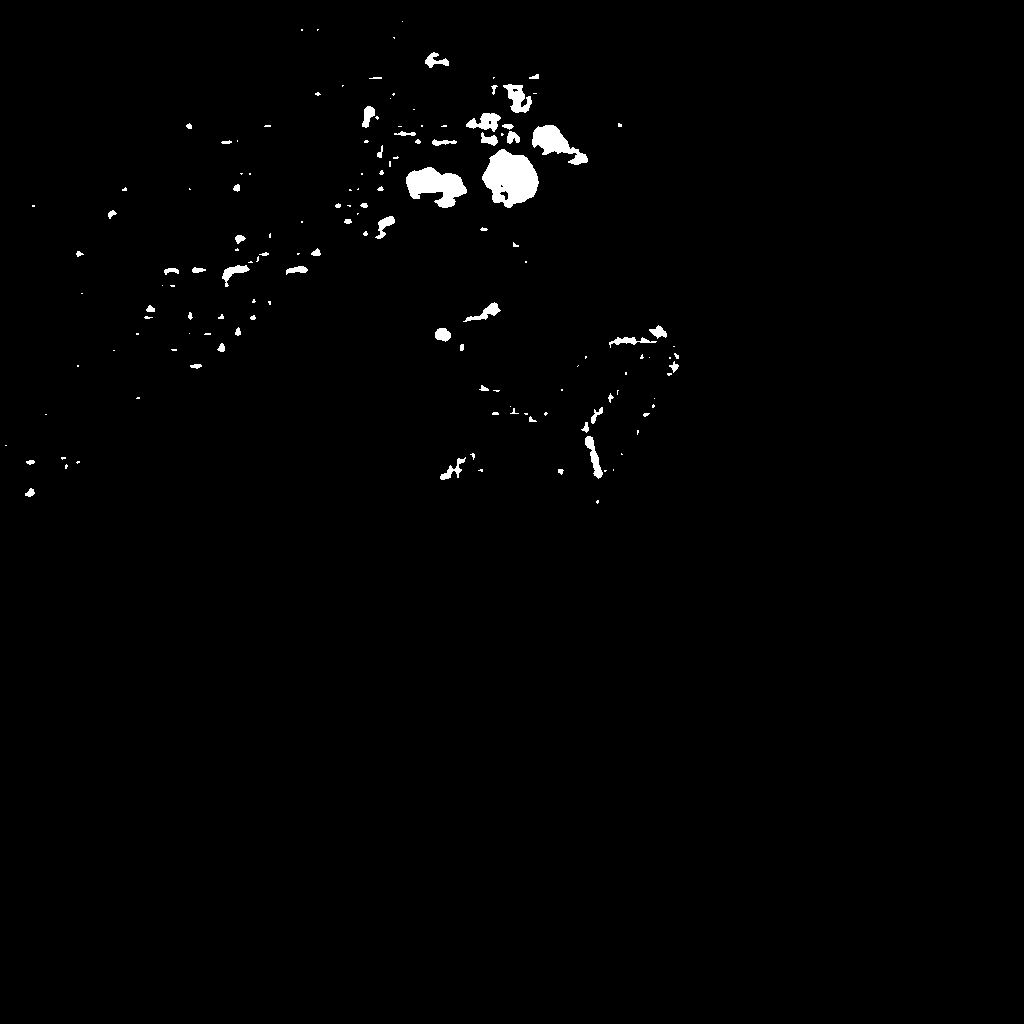}
		\end{minipage}
		\begin{minipage}[b]{0.10\textwidth}
			\centering
			\includegraphics[width=\textwidth]{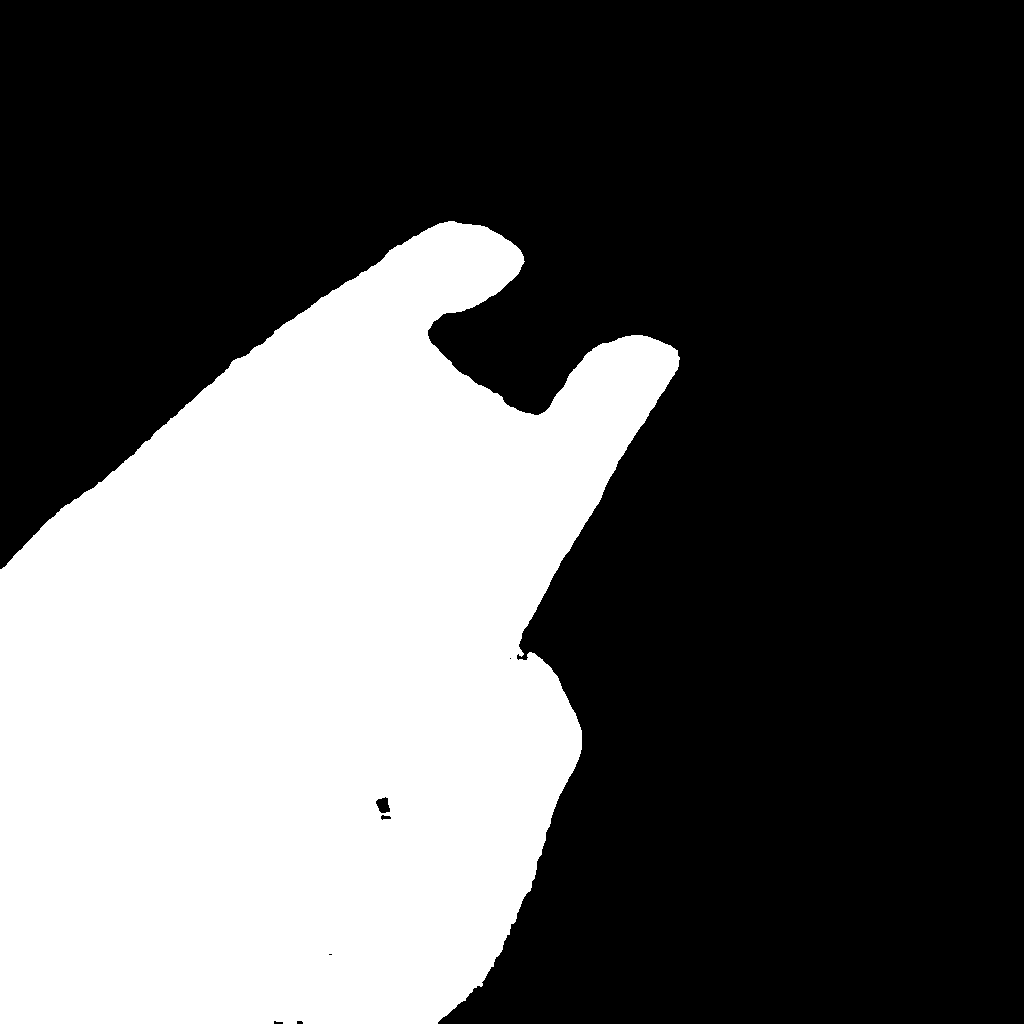}
		\end{minipage}
		\begin{minipage}[b]{0.10\textwidth}
			\centering
			\includegraphics[width=\textwidth]{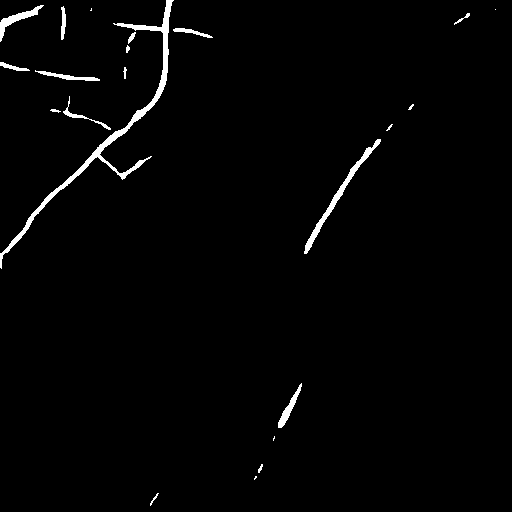}
		\end{minipage}\\[3pt]
		\begin{minipage}[b]{0.10\textwidth}
			\centering
			\includegraphics[width=\textwidth]{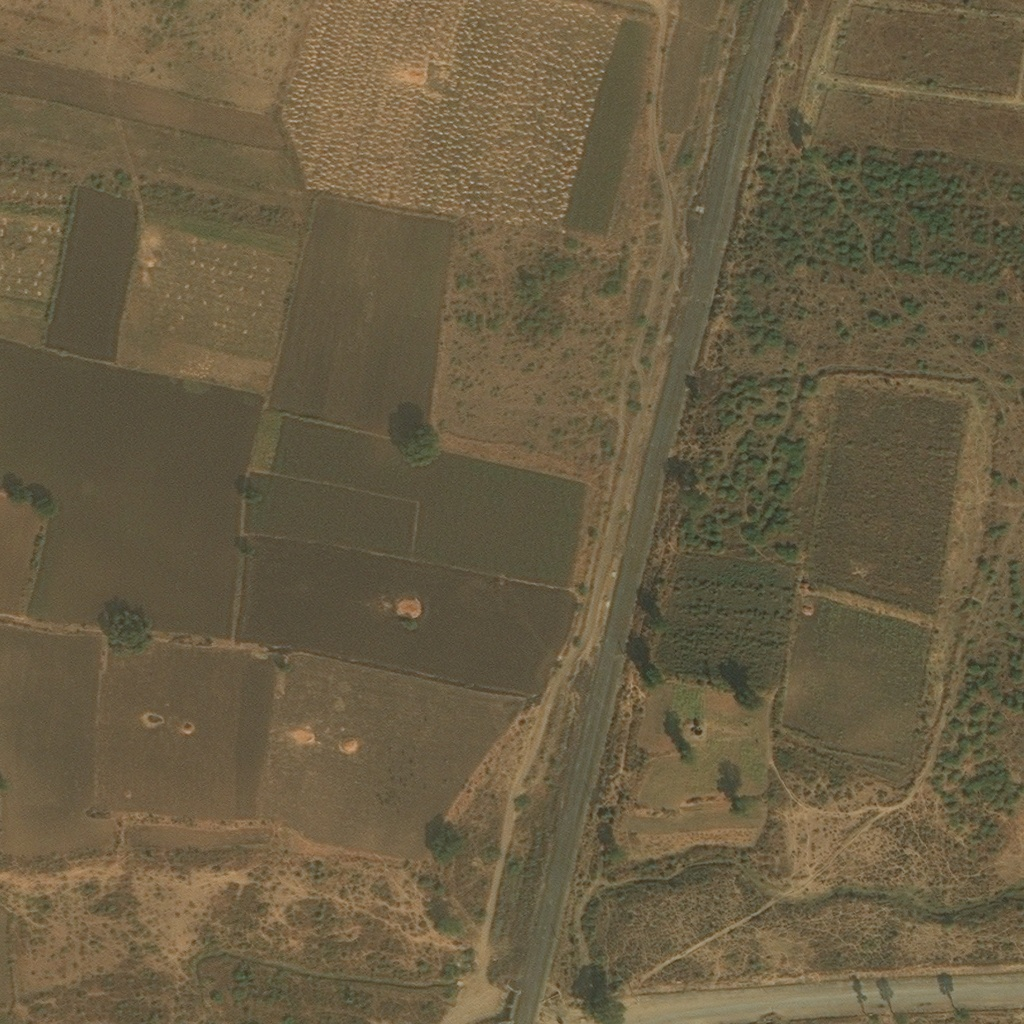}
			\scriptsize{Original Image}
		\end{minipage}
		\begin{minipage}[b]{0.10\textwidth}
			\centering
			\includegraphics[width=\textwidth]{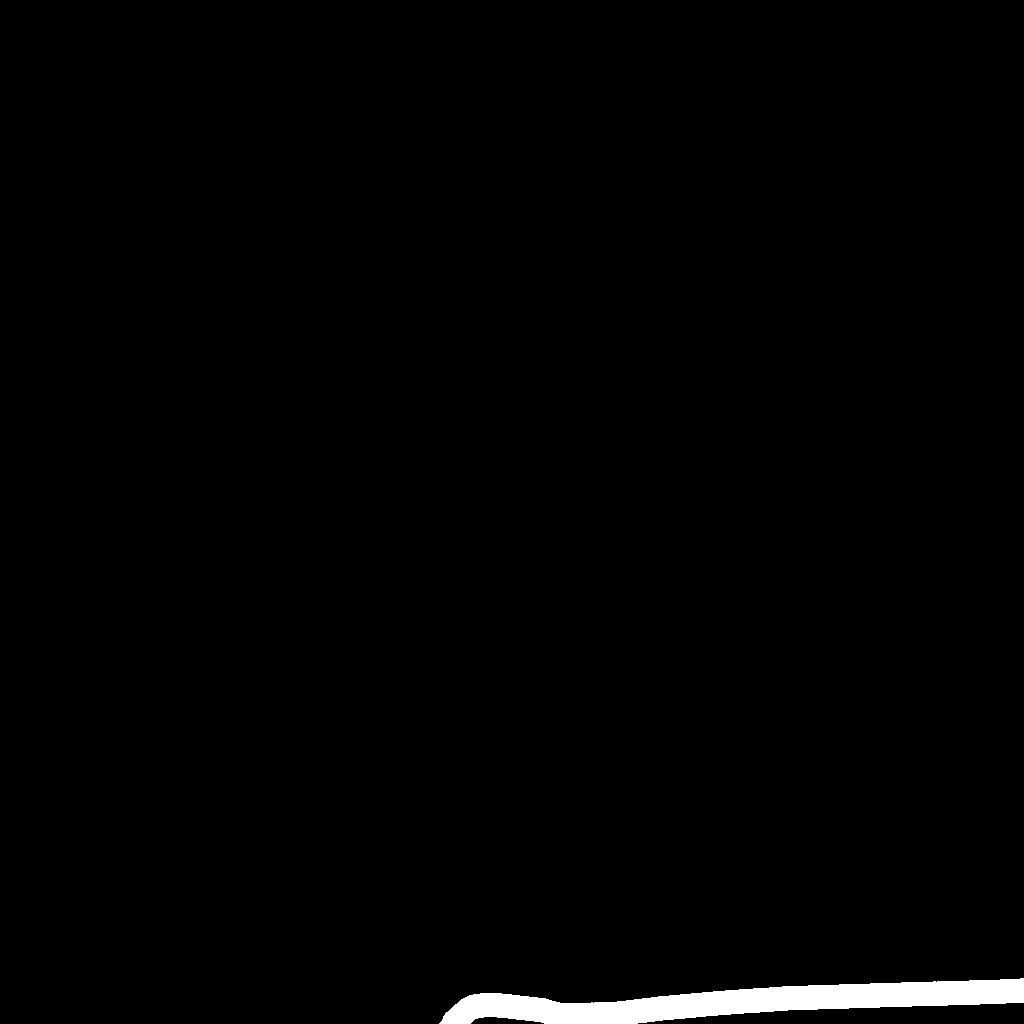}
			\scriptsize{Ground Truth}
		\end{minipage}
		\begin{minipage}[b]{0.10\textwidth}
			\centering
			\includegraphics[width=\textwidth]{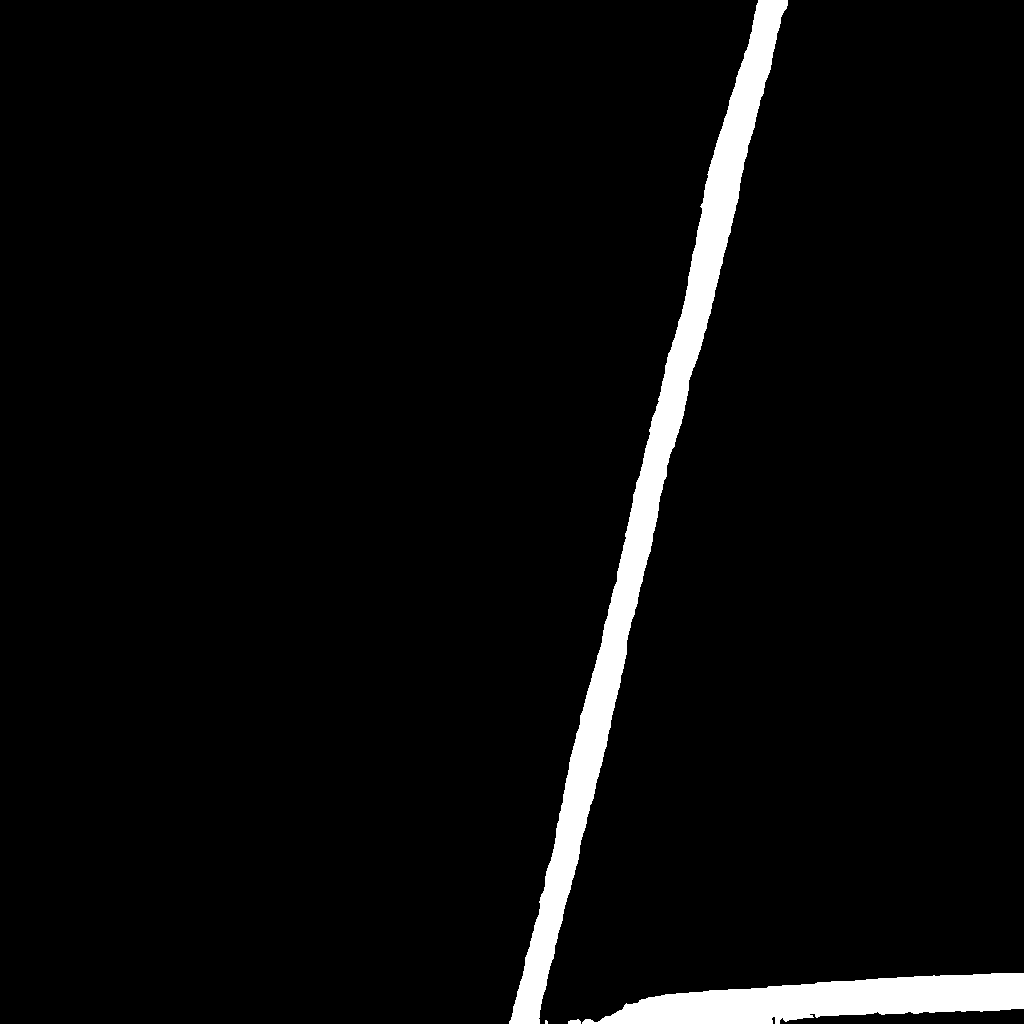}
			\scriptsize{RSAM-Seg}
		\end{minipage}
			\begin{minipage}[b]{0.10\textwidth}
			\centering
			\includegraphics[width=\textwidth]{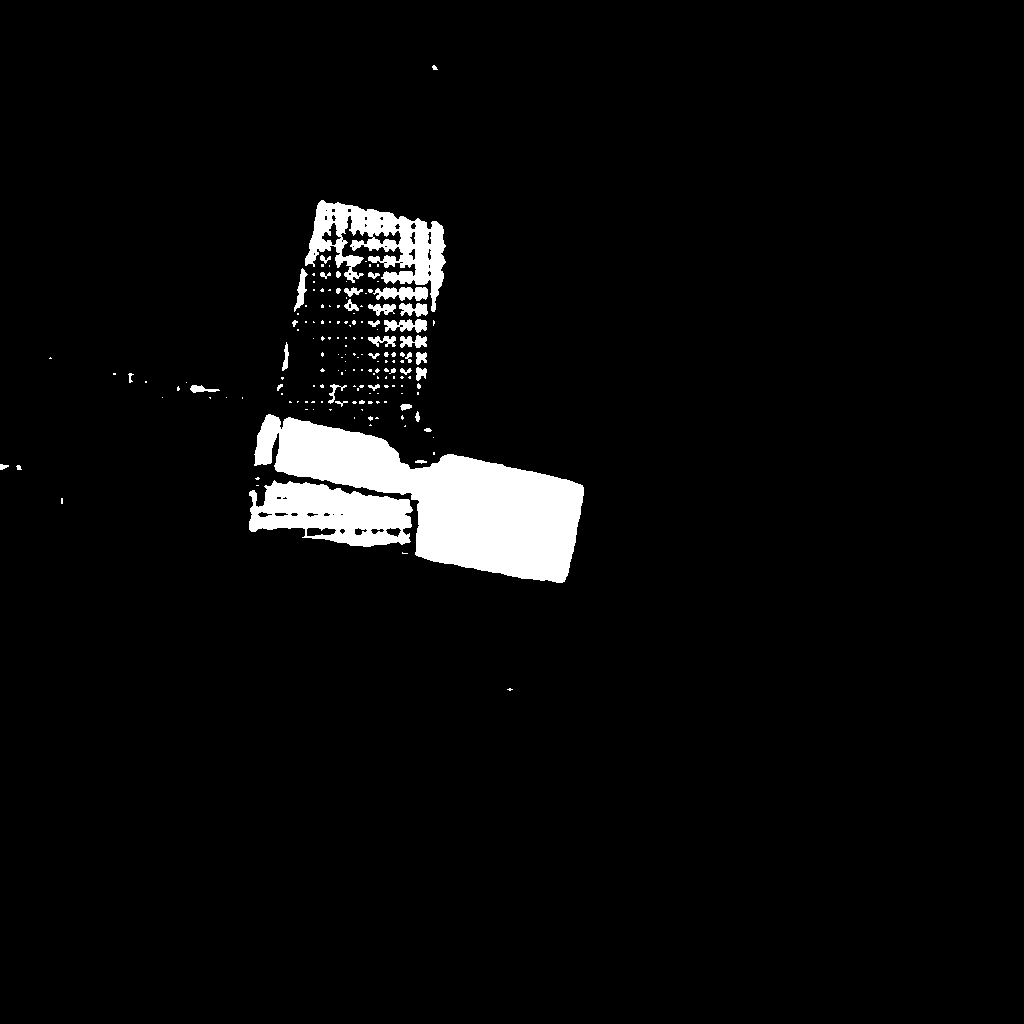}
			\scriptsize{SAM (center -)}
		\end{minipage}
		\begin{minipage}[b]{0.10\textwidth}
			\centering
			\includegraphics[width=\textwidth]{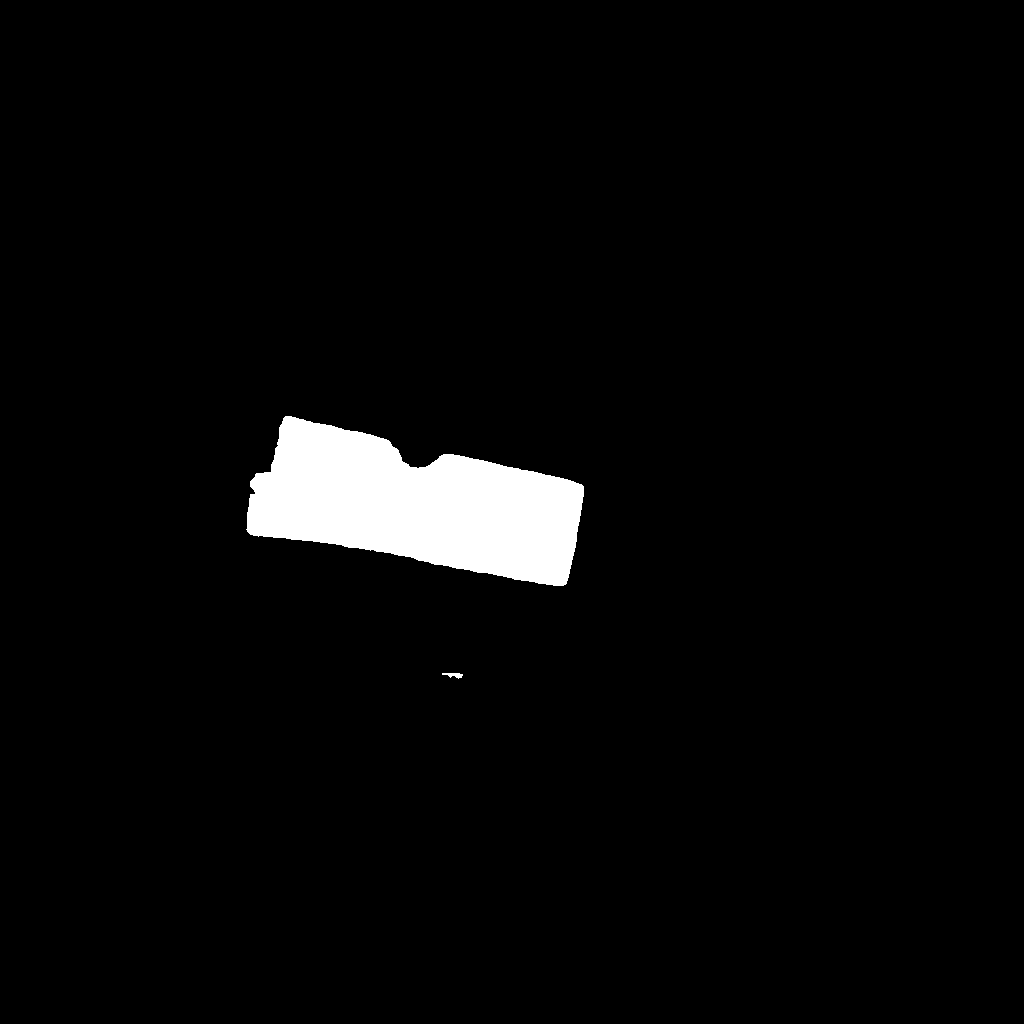}
			\scriptsize{SAM (center +)}
		\end{minipage}
		\begin{minipage}[b]{0.10\textwidth}
			\centering
			\includegraphics[width=\textwidth]{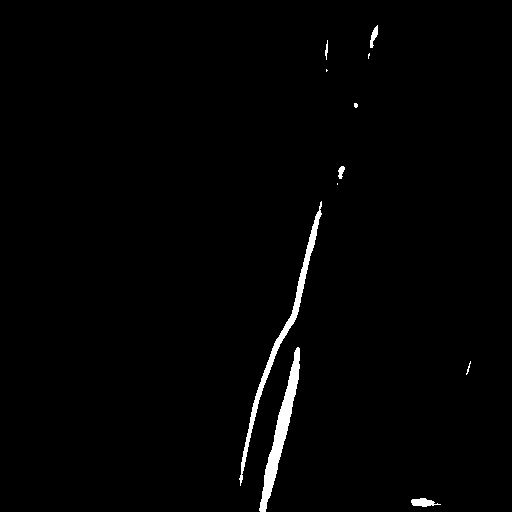}
			\scriptsize{U-Net}
		\end{minipage}
		\caption{Examples of completion results on the DG-Road dataset.}
		\label{fig:discussionBGuality}
	\end{figure*}
	
		\begin{table}[h]\normalsize
		\caption{Ablation study for RSAM-Seg, showing the impact of Adapter-Scale and $\mathrm{F}_{\mathrm{pe}}$,  $\mathrm{F}_{\mathrm{hfc}}$ within Adapter-Feature on segmentation performance.}
		\centering
		\resizebox{\columnwidth}{!}{%
		\begin{tabular}{|cc|c|c|c|c|c|c|c|}
			\toprule[2pt]
			\multicolumn{2}{|c|}{Method} & Jaccard & Precision & Recall & Specificity & F1 score & Overall accu & mIoU \\ \hline
			\multicolumn{2}{|c|}{RSAM-Seg} & $\mathbf{0.731}$ & $\mathbf{0.8301}$ & 0.8396 & $\mathbf{0.8301}$ &$\mathbf{0.8152}$ & $\mathbf{0.9197}$ & $\mathbf{0.7646}$ \\ \hline
			\multicolumn{1}{|c|}{\multirow{2}{*}{Adapter-Feature}}& w/o $\mathrm{F}_{\mathrm{pe}}$ & 0.723 & 0.8146 & 0.8469 & 0.8146 & 0.8049 & 0.9131 & 0.7597 \\ \cline{2-9} 
			\multicolumn{1}{|c|}{} & w/o $\mathrm{F}_{\mathrm{hfc}}$& 0.7118 & 0.7888 & $\mathbf{0.8563}$ & 0.7888 & 0.7885 & 0.892 & 0.7364 \\ \hline
			\multicolumn{2}{|c|}{RSAM-Seg w/o Adapter-Scale} & 0.7287 & 0.8173 & 0.8562 & 0.8173 & 0.8114 & 0.9056 & 0.7608 \\ 	\toprule[2pt]
		\end{tabular}%
	}
	\label{tab.ablation}
	\end{table}
	
	Table \ref{tab.ablation} presents the results on the 38-Cloud dataset, where the impact of removing $\mathrm{F}_{\mathrm{pe}}$ and $\mathrm{F}_{\mathrm{hfc}}$ within the Adapter-Feature, as well as removing Adapter-Scale from RSAM-Seg is evaluated. The experiments show that the $\mathrm{F}_{\mathrm{hfc}}$ significantly improves the performance of RSAM-Seg on several evaluation metrics, indicating that it introduces high-frequency information into the model that is crucial for accurate image segmentation in remote sensing applications. Additionally, both the $\mathrm{F}_{\mathrm{pe}}$ and Adapter-Scale modules contribute to the overall performance of RSAM-Seg in processing remote sensing imagery. 

	The visualization results are listed in Figure \ref{fig:AblationQuality}, which can be observed that the $\mathrm{F}_{\mathrm{hfc}}$ effectively reduces the interference from the surrounding environment of the clouds on the classification results. Furthermore, the combination of Adapter-Scale and $\mathrm{F}_{\mathrm{pe}}$ further enhances the segmentation performance. 
	
	By analyzing both the quantitative and visualization results, the critical role of each component in the proposed method can be observed. These findings not only validate the effectiveness of RSAM-Seg but also provide valuable insights for future research and development in this field.
	\section{Discussion}
	\subsection{Few-shot scenario}
	In the experiment, RSAM-Seg exhibits a commendable degree of accuracy even in the challenging few-shot scenarios. Moreover, as the sample size expands, the performance of the model demonstrates a notable enhancement in terms of precision and predictive capability.
	
	\begin{table}[!htb]\normalsize
			\caption{The impact of dataset size on few-shot results}
		\resizebox{\columnwidth}{!}{%
			\begin{tabular}{|c|c|c|c|c|c|c|c|c|c|}
				\toprule[2pt]
				Dataset & Jaccard & Precision & Recall & Specificity & F1 score & Overall accu & mIoU \\ \hline
				1\% 38-Cloud & 0.5552 & 0.7389 & 0.6777 & 0.7389 & 0.6561 & 0.7919 & 0.6412 \\ \hline
				10\% 38-Cloud & 0.6733 & 0.7984 & 0.8032 & 0.7984 & 0.7587 & 0.8738 & 0.7172 \\ \hline
				30\% 38-Cloud &0.6940 & 0.7723 & 0.8597 & 0.7723 & 0.7770 & 0.8797 & 0.7234 \\
				\hline
				70\% 38-Cloud &$\mathbf{0.731}$ & $\mathbf{0.8301}$ & $\mathbf{0.8396}$ & $\mathbf{0.8301}$ &$\mathbf{0.8152}$ & $\mathbf{0.9197}$ & $\mathbf{0.7646}$ \\
				\toprule[2pt]
			\end{tabular}%
		}
		\label{tab.fewshot}
	\end{table}
	
	The performance in the context of few-shot learning of RSAM-Seg is assessed using the 38-Cloud dataset. Table \ref{tab.fewshot} reflects the results under the condition where the original test set remains unchanged, 1\% , 10\%, 30\% and 70\% of images from the training set are randomly selected as new training subsets. Compared to the results of U-Net in Table \ref{tab.RSAMRES}, RSAM-Seg demonstrates comparable efficacy to U-Net when only utilizing 10\% of the dataset.
	
	The visualization results are listed in the Figure \ref{fig:discussionQuality}, which reveals the potential of the methodology in the domain of remote sensing, particularly for few-shot image segmentation tasks.
	
	\subsection{Beyond Ground Truth}
	The experimental findings reveal a observation that our method surpasses the ground truth annotations of the dataset in certain scenarios, yielding segmentation results that exhibit superior accuracy and fidelity. 
	
	The Figure \ref{fig:discussionBGuality} shows the segmentation results in the road segmentation scenario. The snippets in the second row clearly demonstrate the capability of RSAM-Seg to segment roads that ground truth failed to identify, showcasing the completion capability of RSAM-Seg in delineating road regions from remote sensing imagery.
	
	The segmentation results also indicate that the integration of domain-specific prior knowledge from remote sensing scenes into the SAM holds substantial promise for enhancing the construction of remote sensing datasets. RSAM-Seg exhibits generalizability, positioning it as a formidable tool for auxiliary annotation purposes, thereby mitigating the burdensome costs associated with manual annotation and concurrently amplifying the overall efficiency of the process.
	
	\section{Conclusion}
	We propose RSAM-Seg by incorporating specific prior information from the remote sensing domain and combined the high-frequency information of the images with their intrinsic features as prompts without manual prompt. Adapter-Feature and Adapter-Scale are integrated to enhance performance in semantic segmentation tasks involving remote sensing imagery. To evaluate the proposed methodology, comprehensive experiments are conducted in cloud, buildings, fields and roads scenarios. A meticulous comparative analysis is also conducted, benchmarking RSAM-Seg against the original architecture as well as the widely adopted U-Net model in the general semantic segmentation domain. The findings suggest that leveraging the incorporation of prior information, RSAM-Seg demonstrates promising capabilities in few-shot learning scenarios. Furthermore, RSAM-Seg holds potential as an auxiliary annotation tool, offering a novel approach to facilitate dataset creation while mitigating associated costs.
	
	In the future, the primary focus will be on multi-object segmentation in few-shot scenarios, emphasizing the improvement of segmentation accuracy. Concurrently, there will be exploration into the optimization of efficiency and model compactness.

	\ifCLASSOPTIONcaptionsoff
	\newpage
	\fi

	
	

	\bibliographystyle{IEEEtran}
	%
	\bibliography{doc}
	
	%
	
	
	\begin{IEEEbiography}[{\includegraphics[width=1in,height=1.25in,clip,keepaspectratio]{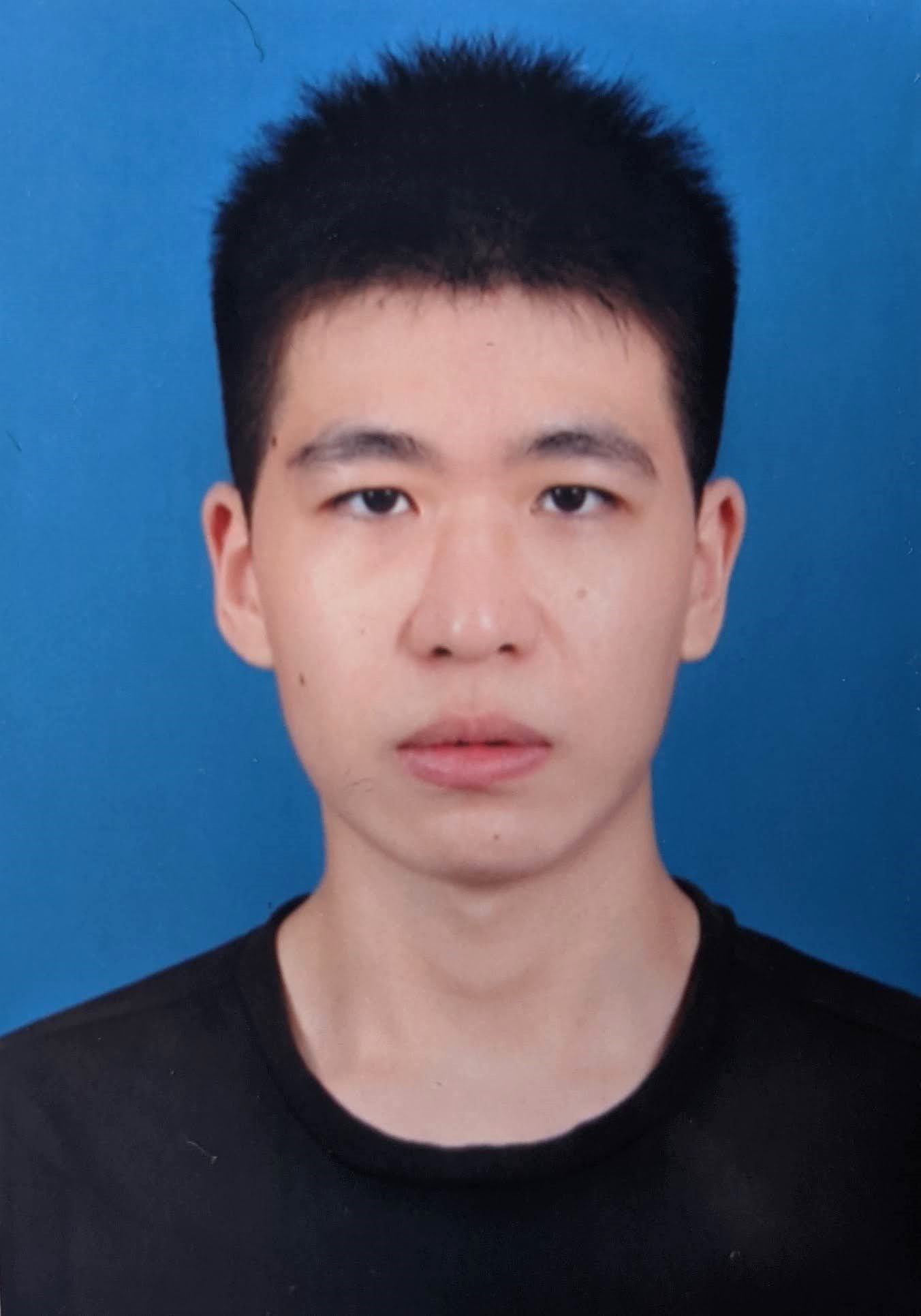}}]{Jie Zhang}
		 is currently pursuing the M.S. degree in computer science from Nanjing Forestry University, Nanjing, China. His research interests include computer vision and remote sensing image processing.
	\end{IEEEbiography}
	\begin{IEEEbiography}[{\includegraphics[width=1in,height=1.25in,clip,keepaspectratio]{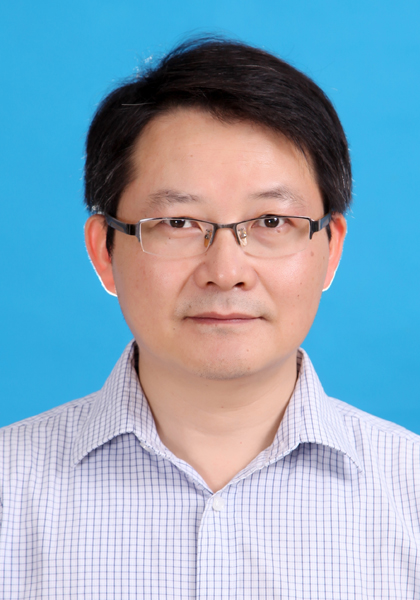}}]{XUBING YANG}
		received his B.S. degree in mathematics from Anhui University in 1997. He completed his M.S. and Ph.D. degrees in computer applications at Nanjing University of Aeronautics \& Astronautics (NUAA) in 2004 and 2008, respectively. Since 2008, he joined Nanjing Forestry University and now worked as an associate professor at computer science and engineering department at NFU. His research interests include pattern recognition, machine learning, and neural computing. 
	\end{IEEEbiography}
	\begin{IEEEbiography}[{\includegraphics[width=1in,height=1.25in,clip,keepaspectratio]{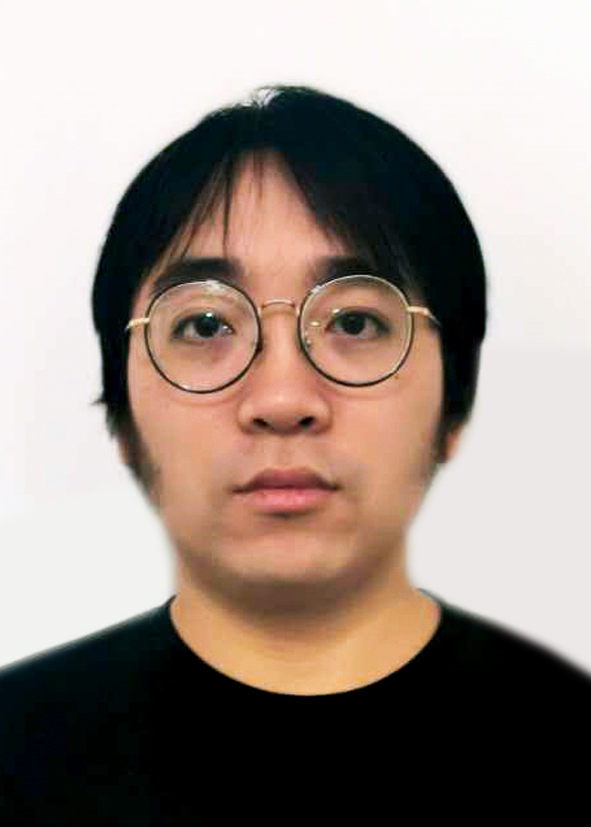}}]{RUI JIANG}
	received the B.S. and Ph.D. degrees in electronic engineering from the Nanjing University of Aeronautics and Astronautics (NUAA), Nanjing, China, in 2007 and 2013, respectively.  In 2013 he joined the Department of Telecommunications and Information Engineering, Nanjing University of Posts and Telecommunications (NUPT), Nanjing, China, where He is now an associate professor. His research interests include machine learning and Internet of things (IoT).
	\end{IEEEbiography}
	\begin{IEEEbiography}[{\includegraphics[width=1in,height=1.25in,clip,keepaspectratio]{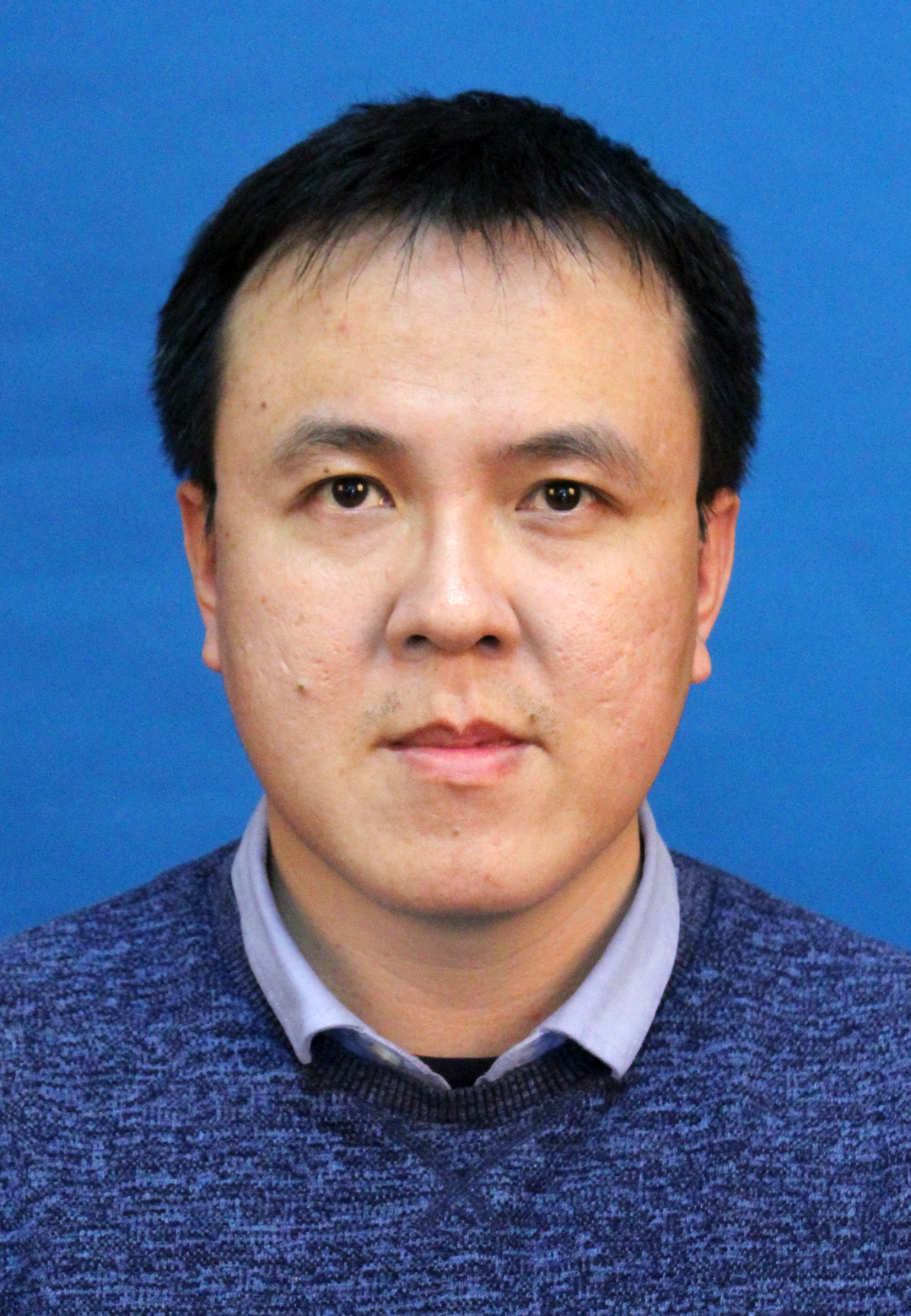}}]{WEI SHAO}
	received the B.Sc. and M.Sc. degrees in information and computing science from Nanjing University of Technology, China in 2009 and 2012, respectively, and the Ph.D. degree in software engineering from Nanjing University of Aeronautics and Astronautics, China in 2018. He is currently an associate professor with College of Computer Science and Technology, Nanjing University of Aeronautics and Astronautics, China. His research interests include machine learning and bioinformatics.
	\end{IEEEbiography}
	\begin{IEEEbiography}[{\includegraphics[width=1in,height=1.25in,clip,keepaspectratio]{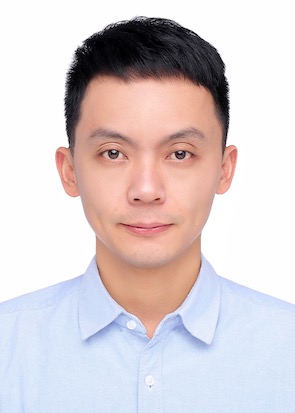}}]{LI ZHANG}
		received the B.S. degree in computer science from Changsha University of Science \& Technology, the M.S. and Ph.D degree in computer science from Nanjing University of Aeronautics and Astronautics (NUAA) in 2007, 2010 and 2015 respectively.  He is currently an associate professor in the College of Information Science and Technology, Nanjing Forestry University.  His research interests include machine learning, remote sensing and medical imaging analysis.
	\end{IEEEbiography}	

	
	
	
	
	

	
	
\end{document}